\documentclass{article}
\usepackage{custom_tex}

\title{Minimax Optimal Fair Classification \\
with Bounded Demographic Disparity}

\author
{Xianli Zeng\footnote{
University of Pennsylvania.	\texttt{zengxl19911214@gmail.com}.},\,\,
Guang Cheng\footnote{University of California, Los Angeles. 
\texttt{guangcheng@ucla.edu}.}\,\,
and
Edgar Dobriban\footnote{
University of Pennsylvania.	\texttt{dobriban@wharton.upenn.edu}.}}

\allowdisplaybreaks[4]

\date{\today}

\begin{document}
 
\begin{sloppypar}
\maketitle

\begin{abstract}
Mitigating the disparate impact of statistical machine learning methods is crucial for ensuring fairness. While extensive research aims to reduce disparity, the effect of using a \emph{finite dataset}---as opposed to the entire population---remains unclear. This paper explores the statistical foundations of fair binary classification with two protected groups, focusing on controlling demographic disparity, defined as the difference in acceptance rates between the groups.
Although fairness may come at the cost of accuracy even with infinite data, we show that using a finite sample incurs additional costs due to the need to estimate group-specific acceptance thresholds. 
We study the minimax optimal classification error while constraining demographic disparity to a user-specified threshold. 
To quantify the impact of fairness constraints, we introduce a novel measure called \emph{fairness-aware excess risk} and derive a minimax lower bound on this measure that all classifiers must satisfy.
Furthermore, we propose FairBayes-DDP+, a group-wise thresholding method with an offset that we show attains the minimax lower bound. Our lower bound proofs involve several innovations. Experiments support that 
FairBayes-DDP+
controls disparity at the user-specified level, while being faster and having a more favorable fairness-accuracy tradeoff than several baselines.
\end{abstract}

\tableofcontents

\section{Introduction}
\label{introduction}
Fairness, a concept closely related to justice, has been studied for thousands of years,
dating back at least to Plato's Republic \citep{Plato,Rawlsjus,Rawlsfair}. 
Many laws and provisions aim to ensure fairness and protect the rights and interests of individuals,
especially those of vulnerable groups.
Recently, the fairness of automated decision-making systems enabled by statistical machine
learning has come into question.
Due to their ever-improving performance, advanced machine learning methods 
are increasingly being utilized in high-stakes
sectors—ranging from credit lending \citep{MA201824} and criminal recidivism forecasting
\citep{JJSL2016} to medical diagnoses \citep{MQ2017}—where their decisions profoundly affect
individual lives.

\begin{figure}[t]
\begin{minipage}{.46\linewidth}
\begin{center}
\centerline{\includegraphics[width=0.95\columnwidth]{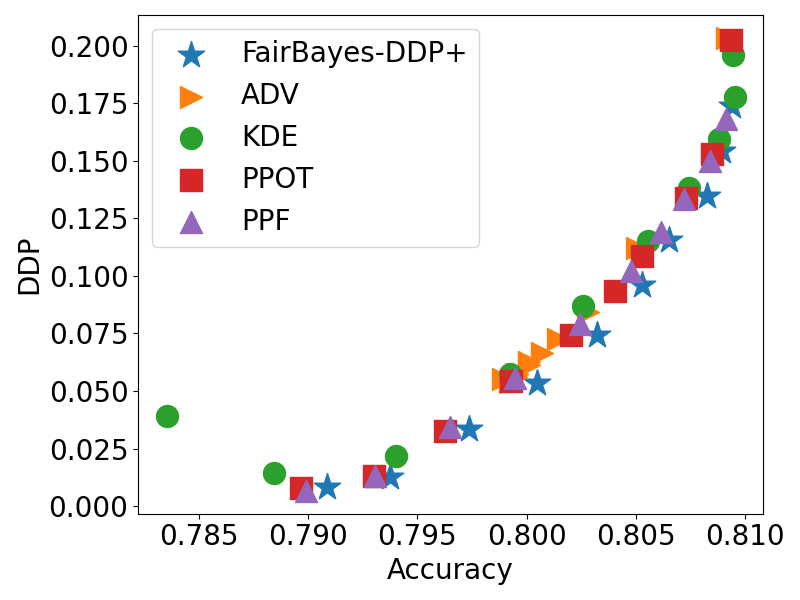}}
\caption{\small 
Our FairBayes-DDP+  method achieves a better fairness-accuracy tradeoff than other baselines on the "Adult" dataset.
Here DDP is the demographic disparity, i.e., the difference in the probabilities of a positive classification among the two protected groups; see \Cref{sec:eda} for details. } \label{fig::real_data_two_dimen} \end{center} \end{minipage} \hfill 
\begin{minipage}{.46\linewidth} \captionof{table}{\small Performance of our FairBayes-DDP+ method with various pre-specified disparity levels on the ``Adult'' dataset; see \Cref{sec:eda}. Our method controls the demographic disparity (DDP) at a user-specified value $\delta$, while achieving a high accuracy (ACC); while having a small standard deviation (SD) over 1000 repetitions.}
\label{tab:real_data_two_dimen}
\begin{center}
\centering
{\footnotesize
\begin{tabular}{c|cc}\hline
$\delta$ & DDP (SD) & ACC (SD) \\\hline
0.00 & 0.008 (0.003) & 0.791 (0.001)\\
0.02 & 0.013 (0.003) & 0.794 (0.001)\\
0.04 & 0.033 (0.003) & 0.797 (0.001)\\
0.06 & 0.054 (0.003) & 0.800 (0.001)\\
0.08 & 0.074 (0.003) & 0.803 (0.000)\\
0.10 & 0.096 (0.003) & 0.805 (0.000)\\
0.12 & 0.115 (0.002) & 0.807 (0.000)\\
0.14 & 0.135 (0.002) & 0.808 (0.000)\\
\hline \end{tabular}
}
\end{center}
\end{minipage}
\end{figure}
Concurrently, these powerful predictive
models risk making discriminative decisions against certain protected groups,
such as those defined by race, gender, and other characteristics \citep[e.g.,][]
{JJSL2016,FBC2016,corbett2018measure}. This has motivated a growing body of work literature on the algorithmic aspects of achieving fairness (see Section
\ref{Literature}).
However, statistical considerations---such as the effect of having a finite dataset on fairness and accuracy in the entire population, and the optimal use of data---are much less studied.

To shed light on the statistical aspects of fairness, 
we study fair classification,  
where various population-level fairness
criteria exist, see \Cref{Literature}.
Classifiers that conform to such fairness constraints and are most accurate---Bayes-optimal---in the population have been identified
\citep{CSFG2017, MW2018, CCH02019, SC2021, Weietal20a,zeng2022fair,zeng2024bayesoptimal}. 
From these works, it is known that there can be fundamental trade-offs between accuracy and fairness even if the entire population is known and available to determine a classifier.

However, it is not known how much additional cost using  a \emph{finite dataset} induces.
What is the best possible---minimax optimal---accuracy and
fairness that we can achieve with a finite sample? 
For a different problem, fair regression, this has been studied by
 \cite{chzhen2022minimax} and
\cite{fukuchi2022minimax}; 
but even the definitions of fairness are unrelated.

To study fairness in classification, we consider the most commonly discussed fairness metric, demographic parity.
Since the accuracy of unfair classifiers can be higher than that of fair classifiers,
we introduce a novel notion of \emph{fairness-aware excess risk} to measure performance (See Definition \ref{def:fairawarerisk}).
This metric coincides with the conventional excess risk when the
classifier is fair, and appropriately penalizes unfairness otherwise.
Further, it is minimized
by the most accurate---Bayes-optimal---fair classifier.

Since the properties of the data distribution affect performance, 
we quantify the behavior of the data near the decision boundary 
via the margin condition studied in
non-parametric classification \citep{TsybakovBayes2004,tsy2007,Was2013}.
In fair binary classification with a binary protected attribute,
we derive a
minimax lower bound for the error
when the group-wise probabilities of the positive class---or, the \emph{regression functions}---are H\"older-smooth
and the group-wise density functions of features
satisfy a so-called strong density condition \citep{tsy2007},

When the disparity constraint is
sufficiently stringent, 
the group-wise acceptance thresholds need to be adjusted to satisfy the fairness constraint.
Estimating these thresholds incurs an additional error, and the minimax lower bound is
determined by the maximum of this and the error in estimating the  regression functions for each protected group. 
Deriving the additional term in the lower bound requires an innovative argument, 
by proposing an intricate
novel construction of two similar distributions 
with distinct decision thresholds
in 
Le Cam's two point lower bound method
(see Part II of \Cref{pf:thm:lower_bound} for details).

After deriving the lower bound, we complete the minimax analysis by
proposing FairBayes-DDP+, a method that we show is minimax rate optimal.
Our method 
is a group-wise thresholding algorithm, and  
improves previous estimators
of fair Bayes-optimal classifiers \citep{MW2018,zeng2022fair,zeng2024bayesoptimal}
in two key components: (1)
it identifies and adapts to possible jump discontinuities of the disparity as a function of the group-wise threshold (see \Cref{sec:discont}) and 
(2) it introduces offsets to
handle the case where the decision boundary has a positive probability (see \Cref{sec:offsets}). 
We prove that FairBayes-DDP+ is minimax optimal and asymptotically controls disparity.

We summarize our contributions as follows.
\begin{itemize}
    
    \item \textbf{Minimax lower bound in binary classification with a bounded demographic disparity:}  
   We study classification problems with a constraint on demographic disparity. 
   We introduce the notion of fairness-aware excess risk (Definition \ref{def:fairawarerisk}) to measure the performance of classifiers given fairness constraints. 
   When the data distribution
   satisfies appropriate versions of 
   Hölder-smoothness condition and Tsybakov noise condition \citep{TsybakovBayes2004}, we derive the a minimax lower bound for fair classification with a bounded demographic disparity. 
   We find that, in addition to a population-level effect, fairness may or may not have a significant effect on accuracy in a finite sample. 
 Our analysis requires  a novel construction of distributions in the lower bound, in the case  where the group-wise decision thresholds need to be adjusted to satisfy the fairness constraint.

    \item \textbf{Minimax optimal classifier:}
We introduce FairBayes-DDP+, an algorithm for binary fair classification, improving 
on previous methods in two key ways:
(1)
by adapting to possible jump discontinuities of the disparity as a function of the group-wise threshold (see \Cref{sec:discont}) and 
(2) by introducing offsets to
handle a decision boundary with a positive measure (see \Cref{sec:offsets}). 
We further prove that FairBayes-DDP+ attains minimax optimality. 
In experiments, we compare it with several baselines and show that it has a competitive performance. 
FairBayes-DDP+ controls disparity at the user-specified level, and attains a better tradeoff between fairness and accuracy in a finite sample than baselines. See \Cref{fig::real_data_two_dimen} and \Cref{tab:real_data_two_dimen} for a brief example, and see \Cref{sim} for details. Our numerical results can be reproduced with the code provided at {\url{https://github.com/XianliZeng/FairBayes-DDP-Plus}.}

\end{itemize}
\section{Related Literature}\label{Literature}
There is a great deal of related work, and we can only discuss the most closely related papers.

{\bf Definitions of Fairness.}
Many fairness metrics have been developed. 
Group fairness \citep[e.g.,][etc]{CFM2009,DHPT2012,HPS2016} targets parity across
 protected groups, while individual fairness \citep[e.g.,][etc]{JKMR2016,PKG2019,RBMF2020} aims to provide nondiscriminatory predictions
for similar individuals.  

{\bf Algorithms Aiming for Fairness.}
There is a large literature on fair machine learning algorithms,
broadly categorized  into three types: pre-processing \citep[e.g.,][etc]{FFMS2015,KJ2016,JL2019,CFDV2017}, in-processing \citep[e.g.,][etc]{GCAM2016,ZBR2017,NH2018,CLKVN2019,CJG2019,CHS2020}, and post-processing \citep[e.g.,][etc]{BJA2016,CSFG2017,VSG2018,MW2018,CCH02019,I2020,SC2021,JSW2022}, see \cite{SC2020} for a review.

Our method is a post-processing algorithm, 
aiming to mitigate disparities in the output 
of a classifier. 
Specifically, it is a group-wise thresholding rule \citep[e.g.,][etc]{BJA2016,CSFG2017,VSG2018,MW2018,CCH02019,I2020,SC2021,JSW2022}, estimating the probability of a positive label given the features for each protected group, 
and assigning thresholds to protected groups
aiming for parity. 
\cite{MW2018,zeng2024bayesoptimal} propose post-processing algorithms aiming to estimate the Bayes-optimal classifier, but do not study the finite-sample performance of their methods.
We refine their method with an offset and show that  
it achieves the minimax optimal rate.

{\bf Nonparametric Classification and Minimax Optimal Rate.}
For a binary classification problem
where the goal is to predict a label $Y \in \{0,1\}$ based on
observed $d$-dimensional features $x \in \X:= \R^d$, 
a probabilistic classifier $f$ is a function\footnote{All functions considered will be measurable with respect to the Borel sigma algebras on the input and output spaces; 
this will not be mentioned further.} $\X \to [0,1]$ that specifies the probability of predicting $\widehat{Y}_f=1$ given $X=x$, i.e., $f(x)=\P(\widehat{Y}_f=1\mid X=x)$ for all $x\in\X$. 
Classification methods include plug-in rules, 
which estimate the regression function
$\eta:x\mapsto \P(Y=1\mid X=x)$ 
and makes decisions by thresholding it, 
and empirical risk minimizers (ERM). 
The convergence rates and minimax optimality of both methods have been studied \citep[e.g.,][etc]{MammenTsybakov1999,Yangminimax1999}.

When $x\mapsto \P(Y=1\mid X=x)$ is 
$\beta$-H\"older-smooth, 
$n$ is the sample size,
and
$d$ is the dimensionality,
\cite{Yangminimax1999} showed that the convergence rate of 
a plug-in classifier is $n^{-\beta/(2\beta+d)}$,
the same as the convergence rate of the estimated regression function.
Moreover, that work proved that the rate is minimax optimal. 
When the regression function is well-behaved near the decision boundary, the convergence rate is faster. 
By considering boundary fragments with $\beta$-smooth boundaries and noise satisfying the $\gamma$-exponent condition, \cite{MammenTsybakov1999} and \cite{TsybakovBayes2004} proved that the minimax convergence rate is $n^{-\beta(\gamma+1)/[\beta(\gamma+2)+(d-1)\gamma]}$, which can be achieved by ERM rules.
\cite{tsy2007}
showed that a plug-in rule with a local polynomial regression estimate is minimax optimal
under the $\gamma$-exponent condition and 
for $\beta$-smooth regression functions, 
with a rate $n^{-\beta(\gamma+1)/(2\beta+d)}$.

\section{Classification with a Bounded Demographic Parity}\label{Pre_and _Not}

In fair binary classification problems 
 with labels in $\mathcal{Y}=\{0,1\}$, 
 two types of features are observed: the usual features $X\in\X$, and the binary protected (or, sensitive)  features $A\in\A=\{0,1\}$\footnote{Conventionally,  we consider $A=0$ to represent the underprivileged group that could potentially face discrimination.}, with respect to which we aim to be fair. 
For example, in a credit lending setting, 
$X$ could refer to 
 education level and income, $A$ could indicate the race or gender of the individual, and $Y$ could correspond to the status of repayment or defaulting on a loan.
Here and below, for all $a\in \A$ and $y\in\mathcal{Y}=\{0,1\}$, we denote by $\P_X$, $\P_{X|A=a}$ and $\P_{X|A=a,Y=y}$ the marginal distribution function of $X$, the conditional distribution function of $X$ given $A=a$, and the conditional distribution of $X$ given $A=a,Y=y$, respectively. 

To evaluate the fairness of a classifier, 
we consider demographic parity, 
the possibly most popular fairness metric.\footnote{In future work, we expect that our insights can seamlessly be extrapolated to other group fairness metrics, including equality of opportunity \citep{HPS2016} and predictive equality \citep{CSFG2017}.}
A
probabilistic classifier $f:\mX\times\mA\to [0,1]$ 
specifies the probability of predicting $\widehat{Y}_f=1$ given $X=x$ and $A=a$, i.e., $f(x)=\P(\widehat{Y}_f=1\mid X=x, A=a)$ for $(x,a)\in \mX\times \mA$. 
The classifier $f$
satisfies \emph{demographic parity} if its prediction $\widehat{Y}_f$
   is probabilistically independent of the protected attribute $A$: $\widehat{Y}_f\indep A$, so that
   $\P_{X|A=1}\lsb \widehat{Y}_f  = 1\rsb  =\P_{X|A=0}\lsb \widehat{Y}_f  = 1\rsb .$
However, demographic parity may be too stringent in certain cases, and it is desired to have more flexible metrics controlling disparate impact. 
To measure the disparate impact of a classifier,
 we use the \emph{demographic disparity} or DDP \citep{CHS2020}, i.e., the difference in the probabilities of predicting $\widehat{Y}_f  = 1$ across groups:
\begin{align}\label{eq:disparity level}
      \textup{DDP}(f)&= \P_{X|A = 1}(\widehat{Y}_f = 1) -\P_{X|A = 0}\lsb \widehat{Y}_f  = 1\rsb. 
\end{align}
 We  denote by $\mathcal{F}_{\delta}$ the set of  functions satisfying the \emph{$\delta$-parity constraint} $|\textup{DDP}(f)|\le \delta$, so that
$$\mathcal{F}_{\delta}=\{f: \mX\times\mA\to [0,1]\,:\, |\textup{DDP}(f)|\le \delta\}.$$
Subject to this $\delta$-parity constraint, we aim to minimize the misclassification error. 
This is achieved by \emph{$\delta$-fair Bayes-optimal classifiers}, defined as
\begin{equation}\label{boc}
f_{\delta}^\star\in \underset{f\in\mathcal{F}_{\delta}}{\argmin}\,\,  \P\lsb Y\neq \widehat{Y}_f\rsb.    
\end{equation}

\subsection{Fair Bayes-Optimal Classifier under Demographic Parity}
\label{sec:Bayes}
The classification thresholds
of fair Bayes-optimal classifiers need to be adjusted for each group,
see  \cite{CSFG2017, MW2018, CCH02019, SC2021, Weietal20a,zeng2022fair,zeng2024bayesoptimal} and Proposition \ref{thm-opt-dp} for details.
To leverage these results, we need some additional notation. 

Intuitively, to minimize the error, we should output $\widehat{Y}=1$ if the probability of $Y=1$ given $X=x$ and $A=a$ is large.
Therefore, the \emph{group-conditional probabilities}---or, regression functions---$\eta_{a}$, $a\in \A$ defined for all $x\in \X$ via $\eta_{a}(x):=\P(Y=1\mid A=a,X=x)$,  play a crucial role.
All optimal classifiers $f$ will aim to output $\widehat{Y}_f=1$ if $\eta_{a}(x)$ is large.

To explain this in detail,
for $a\in \A$,
we denote $p_a:=\P(A=a)$, 
and let $\mathcal{T}=[-\min(p_1,p_0),\min(p_1,p_0)]$.
For $a\in\{0,1\}$
and $t\in \mathcal{T}$, define \emph{group-wise thresholds} via the formula $T_a(t)={1}/{2}+{(2a-1)t}/({2p_a})$.
Let $\mathcal{F}^t$ be the class of 
\emph{group-wise thresholding rules} 
with thresholds $T_a(t)$ for each $a\in \A$ as follows;
where $I(\cdot)$ is the indicator function that equals unity if its argument is true, and zero otherwise: 
$$\mathcal{F}^t:=\lbb  f: \X\times\A\to[0,1],\, f(x,a)=I\lsb\eta_a(x)>T_a(t)\rsb+\tau_a I\lsb\eta_a(x)=T_a(t)\rsb, \ (\tau_1,\tau_0)\in [0,1]^2\rbb.$$
These classifiers output $\widehat{Y}_f=1$ if $\eta_{a}(x)$ is large, and the thresholds $T_a(t)$ are allowed to depend on the group $a\in \A$.
It turns out that parametrizing  the thresholds of acceptance via $t\mapsto T_a(t)$ for $t\in\T$ suffices to obtain Bayes-optimal classifiers.

Further,
define 
the \emph{disparity functions}
$D_{-}:\mathcal{T}\to \R$ and $D_{+}:\mathcal{T}\to \R$
such that for all $t\in \mathcal{T}$,
\begin{equation}\label{eq:Disfun1}
D_{-}(t)=
\Po\lsb\eta_1(X)>\frac12+\frac{t}{2p_1}\rsb-\Pz\lsb\eta_0(X)\ge\frac12-\frac{t}{2p_0}\rsb;
\end{equation}
\begin{equation}\label{eq:Disfun2}
D_{+}(t)=\Po\lsb\eta_1(X)\ge\frac12+\frac{t}{2p_1}\rsb-\Pz\lsb\eta_0(X)>\frac12-\frac{t}{2p_0}\rsb.
\end{equation}
By inspection, both functions are non-increasing, and
for any $t\in \R$, $D_{-}(t) \le D_{+}(t)$.
Moreover $D_-$ is right-continuous and $D_+$ is left-continuous.
It is not hard to see, and it is shown in \cite{zeng2024bayesoptimal}, that for all $t\in \mathcal{T}$,
the DDP of group-wise thresholding rules ranges between $D_{-}(t)$ and $D_{-}(t)$; specifically
$$D_{-}(t)=\inf_{f\in\mathcal{F}^t} \textup{DDP}(f) \,\text{ and }\, D_{+}(t)=\sup_{f\in\mathcal{F}^t} \textup{DDP}(f).
$$
In particular, $D_{-} (0)$ and $D_{+} (0)$ are, respectively, 
the infimum and supremum of the DDP 
over all unconstrained Bayes-optimal classifiers from  \eqref{boc} with $\delta=\infty$.

We will focus on the setting where the group-wise thresholds are uniquely defined, which holds if there is enough probability mass near the decision boundaries (and is ensured by our formal conditions to follow). 
In this case, 
a
$\delta$-fair Bayes-optimal classifier
$f^\star_{\delta}$ has the following form.
Define 
the following ``inverse" of the functions $D_-, D_+$ on $\mathbb{R}_{\ge0}$, for $\delta\ge 0$, 
 \begin{equation}\label{eq:t-star-dp}
t^\star_\delta=  \inf_{t\in\mathcal{T}}\{D_{-}(t)<\delta\}= \inf_{t\in\mathcal{T}}\{D_{+}(t)<\delta\}.
\end{equation}
Since $D_-$ is non-increasing and right-continuous,
we have $D_-(t^\star_\delta)=\delta$
if $D_-$ is continuous at $t^\star_\delta$, 
and $D_-(t^\star_\delta)<\delta$ while 
$\lim_{t\to (t^\star_\delta)^-}D_-(t)>\delta$
if $D_-$ has a jump discontinuity at $t^\star_\delta$, 
A similar statement applies to  $D_+$.

Define the \emph{group-wise thresholds of the two groups}
\begin{equation}\label{Tdel}
\Tsdo = \frac12 + \frac{t^\star_\delta}{2p_1} \textnormal{ and }\Tsdz = \frac12 - \frac{t^\star_\delta}{2p_0}.
\end{equation}
Then, for some $\taudso,\taudsz \in[0,1]$---specified later in \eqref{eq:Taustar}---there is a
$\delta$-fair Bayes-optimal classifier
$f^\star_{\delta}$ that is a group-wise thresholding rule of the form, for all  $x,a$,
\beq\label{bao}
f^\star_\delta(x,a)= I\lsb\eta_a(x)>\Tsda\rsb +\taudsa I\lsb\eta_a(x)=\Tsda\rsb.
\eeq
As discussed in Appendix \ref{ha},
the behavior of the fair Bayes-optimal classifiers on the decision boundary $\{x,a: \eta_a(x)=\Tsda\}$ is generally not unique.
For the sake of generality, for instance to deal with discrete-valued data, we will allow the decision boundary to have a positive probability mass. 
However, it will help to have a specific choice of the Bayes-optimal classifier to estimate. 
Our minimax lower bounds and rate of convergence will not depend on the specific choice of the Bayes-optimal classifier.
 
 Moreover, we will 
 assume without loss of generality that $D_+(0)\ge -D_-(0)$.  
 This condition means 
 that among Bayes-optimal classifiers,  
 $\P_{X|A = 1}(\widehat{Y}_f = 1)-\P_{X|A = 0}(\widehat{Y}_f = 1)$ can be larger than its negative.
 In this sense, the group $A=0$ is underprivileged. 
 If this condition does not hold, we can introduce a new variable $\tilde{A}$ defined as $1-A$, which swaps the groups characterized by $A=1$  and $A=0$.
We will construct estimators of $D_+(0)$ and $D_-(0)$, and these may be used to decide which group is underprivileged. 
 
\begin{Remark}[The impact of fairness on Bayes-optimal classifiers]\label{if}
Since by definition $D_{-} (0) \le D_{+} (0)$,
there are three possibilities for $\delta>0$: 
(1) $\delta <  D_{-} (0)$, (2)  $ D_{-} (0)\le\delta <D_{+}(0)$, and (3) $\delta \ge  D_{+} (0)$.
 \begin{enumerate}
 \item {\bf Fairness-impacted case: $\delta <  D_{-} (0)$.}  When $\delta$ is relatively small with $\delta <  D_{-} (0)$, no unconstrained Bayes-optimal classifier satisfies the fairness constraint $|\textup{DDP}(f)|\le \delta$. 
 As a result, we need to estimate the group-wise thresholds, and the fairness constraint has a significant impact on the fair Bayes-optimal classifiers. Thus, we call this case the fairness-impacted case.
 
 \item  {\bf Fair-boundary case: $D_-(0) \le \delta < D_+(0)$.} When $\delta$ is moderately large with
 $D_-(0) \le \delta < D_+(0)$, there 
 is at least one unconstrained classifier $f^\star\in {\argmin_{f\in\mF}}\, \P(Y\neq \widehat{Y}_f)$ satisfying  $|\textup{DDP}(f^\star)|\le \delta$, and the optimal group-wise thresholds  are $1/2$ for both protected groups.  
 In this case,  
   we need to change the classifiers on the decision boundaries $\{x:\eta_a(x)=1/2\}$ to satisfy fairness constraint.
We thus refer to this case as the fair-boundary case. 
Since changes on the decision boundary do not change the accuracy, the misclassification rate of fair Bayes-optimal classifiers 
    equals the unconstrained Bayes error. 
  \item  {\bf Automatically fair case: $\delta\ge D_{+} (0)$.} Finally, we call $\delta\ge D_{+} (0)$ the  automatically fair case, as all unconstrained Bayes-optimal classifiers are $\delta$-fair. 
 \end{enumerate}
As we can see, $t^\star_\delta$ is non-zero only in the fairness-impacted case when $\delta < D_{-} (0)$.
\end{Remark}

\section{Minimax Lower Bound for Fair Classification}\label{sec:prepare} 
The minimax approach from statistical decision theory characterizes fundamental performance limits. 
An estimator is  minimax rate-optimal if its convergence rate matches the minimax lower bound, i.e., 
the best possible rate of convergence over all estimators. In this section, we derive a minimax lower bound for 
the fair classification problem.  
This requires quantifying the performance of classifiers. 
We begin by introducing a proper metric for fair classification problems.

\subsection{Measure of Performance}
Consider first an unconstrained classification problem with 
a
Bayes-optimal classifier $f^\star:=f^\star_{\infty}$
defined in \eqref{boc} with $\delta=\infty$. 
The performance of  a classifier ${f}$
is commonly measured by its excess risk over $f^\star$ \citep[e.g.,][]{hastie2009elements}, defined as 
 \begin{equation}\label{metric2}
 d_R({f},f^\star)
 :=\P(Y\neq \widehat{Y}_{{f}})-\P(Y\neq \widehat{Y}_{f^\star})
 =\sum_{a\in\A}p_a\int \lsb f(x,a)-f^\star(x,a)\rsb
 \lsb1-2\eta_a(x)\rsb d\Pa(x).
 \end{equation}

For fair classification problems, 
a first attempt may be to consider the excess risk of $f$ over a fair Bayes-optimal classifier
$f^\star_\delta$ from \eqref{bao}, 
i.e., $d_R\lsb f,f^\star_\delta\rsb$. 
However, in the fairness-impacted case,  $d_R\lsb f,f^\star_\delta\rsb$ can be \emph{negative}, 
as the fair Bayes-optimal classifier does not generally minimize the unconstrained risk, i.e.,
$f^\star_\delta\notin \argmin_{f\in\mF}\,  \P(Y\neq \widehat{Y}_{f})$. As a result, $d_R\lsb f,f^\star_\delta\rsb$ 
is not directly suitable for measuring the cost of fairness.

As an alternative, 
we define the following {\it fairness-aware excess risk} to quantify the performance of a classifier within the context of fair classification.
Its functional form is analogous to \eqref{metric2}, and we provide further justification below.
\begin{definition}[Fairness-aware excess risk]\label{def:fairawarerisk}
Let $\delta\ge 0$ and $f^\star_\delta$ 
be a $\delta$-fair Bayes-optimal classifier from \eqref{boc}, 
and consider $\Tsda$ from \eqref{Tdel}. 
For any classifier ${f}: \X\times\{0,1\}\to [0,1]$, we define the {\it fairness-aware excess risk} as
\begin{equation}\label{eq:measure}
d_E\lsb{f},f_\delta^\star\rsb=2\sum_{a\in\A}p_a\lmb\int \lsb{f}(x,a)-{f}_\delta^\star(x,a)\rsb\lsb\Tsda-\eta_a(x)\rsb d\P_{X|A=a}(x)\rmb.
\end{equation}
\end{definition}
Observe first that $d_E\lsb{f},f_\delta^\star\rsb\ge 0$, as 
from \eqref{bao},
it follows that
$f(x,a) -f_\delta^\star(x,a) \le 0$ when $ \eta_a(x)
 > \Tsda$ and $f(x,a) -f_\delta^\star(x,a) \ge 0$  when $ \eta_a(x)\le \Tsda.$
Moreover, it follows from Proposition \ref{thm-opt-dp} that
the choice of the $\delta$-fair Bayes-optimal classifier $f^\star_\delta$ does not affect the value of $d_E$.
The following result further elucidates the fairness-aware excess risk $d_E$, connecting it to the classical excess risk $d_R$.
\begin{proposition}[Characterizing fairness-aware excess risk]\label{measure2}
For any classifier ${f}\in\mF$, the fairness-aware excess risk simplifies as follows, in the cases identified in Remark \ref{if}:
\begin{equation*}
d_E\lsb {f},f_\delta^\star\rsb=\left\{\begin{array}{l}
d_R \lsb {f},f_\delta^\star\rsb, \quad\textnormal{ in the automatically fair and fair-boundary cases } \delta\ge 
   D_{-} (0);\\
d_R \lsb {f},f_\delta^\star\rsb+t_\delta^\star\lmb \textup{DDP}(f)-\delta\rmb, \quad\textnormal{ in the fairness-impacted case } \delta<
   D_{-} (0);
\end{array}\right.
\end{equation*}
Moreover, for $\delta$-fair classifiers $f$ with $|\textup{DDP}(f)|\le \delta$, we have
$d_R \lsb {f},f_\delta^\star\rsb\ge d_E({f},f^\star_\delta)$.
\end{proposition}
We will show a \emph{lower bound} on the minimax excess fairness-aware risk $d_E$ over \emph{all classifiers}, and an \emph{upper bound} realized by an \emph{asymptotically $\delta$-fair classifier}. This will ensure that our method is asymptotically optimal with respect to both $d_E$ and $d_R$ among all $\delta$-fair classifiers.

\subsection{Conditions on the Data Distribution}
In this section, we introduce conditions 
on the data distribution that we need in our theoretical analysis, which require
some notations and definitions. 
For a scalar $\beta$,
we denote by $\lbeta_+:=\lceil \beta\rceil-1$ the maximal integer that is strictly less than
$\beta$. 
For an integer $d>0$, 
and a multi-index $s\in \NN^d$, we
denote 
$|s| = s_1+\ldots+s_d.$
Moreover, for $x \in \MR^d$ and $s\in \NN^d$,
we denote
$x^s = x_1^{s_1}\cdot\ldots \cdot x_d^{s_d}$.
The first concept is the smoothness of the 
per-group regression functions $\eta_a$, $a\in\A$.

\begin{definition}[H\"older Smoothness]\label{h}
Consider $\beta>0$
 and any $\lbeta_+$-times continuously differentiable real-valued
function $g$ on $\mathbb{R}^d$. 
For any $x \in \MR^d$,
we denote by $g_x$ the Taylor approximation of degree $\lbeta_+$ of $g$ at
$x$, such that for all $x'\in \R^d$,
$$g_x(x')=\sum_{s\in \NN^d: |s|\le \lbeta_+}\frac{(x'-x)^s}{s!}g^{(s)}(x).$$
For $L_\beta > 0$, 
the $(\beta,L_\beta,\mathbb{R}^d)$-H\"older class of functions, denoted $\Sigma(\beta,L,\mathbb{R}^d)$,
is defined as the set of functions $g :\mathbb{R}^d \to \mathbb{R}$ that are $\lbeta_+$ times continuously
differentiable and satisfy, for any $x,x' \in \mathbb{R}^d$, the inequality
$|g(x') - g_x(x')| \le L_\beta \|x' - x\|^\beta$.
\end{definition}

The next definition is the \emph{margin condition}, which we adapt to the fair classification problem from \cite{TsybakovBayes2004,tsy2007}, \cite{Was2013}, 
and which controls the regularity of the regression function near the decision boundary.
Let $\gamma\ge 0$, 
and let 
$P$ be a distribution defined on $\X\times\A\times \mathcal{Y}$ with conditional probability functions $\eta_a$, 
$a\in \A$. 
For 
$D_{-}$, $D_+$ from \eqref{eq:Disfun1} and \eqref{eq:Disfun2} and
$t^\star_\delta$ from \eqref{eq:t-star-dp}, 
define the boundary probability functions
$g_{\delta,-}, g_{\delta,+}:[0,\infty)\to [0,2]$ 
for the positive (``+") and negative (``-")  sides of  $t^\star_\delta$,
such that for all 
$\ep\ge0$,
\begin{align*}
g_{\delta,-}(\ep)&=D_{-}(t^\star_\delta)-D_{-}(t^\star_\delta+\ep)\\
&=\Po\lsb T^\star_{\delta,1}<\eta_1(X)\le  T_{\delta,1}^\star+\frac\ep{2p_1}\rsb+\Pz\lsb T^\star_{\delta,0}-\frac{\ep}{2p_0}\le \eta_0(X)<  T^\star_{\delta,0}\rsb;\\
g_{\delta,+}(\ep)&=D_{+}(t^\star_\delta-\ep)-D_{+}(t^\star_\delta)\\
&=\Po\lsb T^\star_{\delta,1}-\frac\ep{2p_1}\le \eta_1(X)<  T^\star_{\delta,1}\rsb+\Pz\lsb T^\star_{\delta,0}< \eta_0(X)\le T^\star_{\delta,0}+\frac{\ep}{2p_0}\rsb.
\end{align*}
Clearly, both  $g_{\delta,-}, g_{\delta,+}$ are monotone non-decreasing, 
while $g_{\delta,+}$ is right-continuous and $g_{\delta,-}$  is left-continuous. 
One can verify that all notions introduced so far can be defined 
not just when $\delta\in[0,\infty)$, but also 
when $\delta=\infty$, which corresponds to the unconstrained case.

\begin{definition}[$\gamma$-Margin Condition, Adapted from \cite{TsybakovBayes2004}, \cite{tsy2007}, \cite{Was2013}]\label{m}
Let $\gamma\ge 0$, 
and let 
$P$ be a distribution defined on $\X\times\A\times \mathcal{Y}$ with group-conditional probabilities $\eta_a(x,a)=\P(Y=1\mid A=a,X=x)$, for $x\in\X$ and
$a\in \A$. 
For $\delta\ge 0$,
we say that $(\eta_1,\eta_0)$
satisfies 
the \emph{strong $\gamma$-margin condition} for $\delta \in[0,\infty]$
with respect to $P$ if, first,  there exist constants $\ep_0, U_\gamma>0$ such that, 
\beq\label{ma1}
 \max\{ g_{\delta,-}(\ep),g_{\delta,+}(\ep)\} \le  U_\gamma\ep^\gamma, \text{ for all } 0<\ep<\ep_0;
\eeq
and second, for $j\in\{+,-\}$, if $D_j(t^\star_\delta)=\delta$,\footnote{If $D_-(t^\star_\delta)<\delta<D_+(t^\star_\delta)$, the lower bound is unnecessary.} then
\beq\label{ma2}
 g_{\delta,j}(\ep) \ge  U_\gamma^{-1} \ep^{1/\gamma}, \text{ for all } 0<\ep<\ep_0.
\eeq

\end{definition}
Conditions \eqref{ma1} and \eqref{ma2} provide upper and lower bounds, respectively, on the probability mass of the regression functions near the decision boundaries. 
Condition \eqref{ma1} 
adapts the $\gamma$-exponent condition introduced by \cite{TsybakovBayes2004}, \cite{tsy2007} for characterizing the convergence rate in nonparametric classification to our problem. 
When $\gamma$ is large, the probability mass of the conditional probability function near the decision boundary decays quickly with $\ep\to 0$, 
suggesting 
that the estimating the conditional probability functions $\eta_a$ near the decision boundary is less challenging. 

For conventional classification problems, an upper such as \eqref{ma1} bound is sufficient to characterize problem difficulty.
However, for fair classification, 
a lower bound on the density is also necessary to characterize the estimation error of ${t}^\star_\delta$ when $D_-(t^\star_\delta)=\delta$ or $D_+(t^\star_\delta)=\delta$. 
In the fairness-impacted case from Remark \ref{if}, 
the difficulty of estimating ${t}^\star_\delta>0$ 
is impacted by the behavior 
of $D_{-}$ and $D_{+}$ near ${t}^\star_\delta$. 
This is quantified by $g_{\delta,-}(\epsilon) = D_{-}(t^\star_\delta)-D_{-}(t^\star_\delta+\epsilon)$ and $g_{\delta,+}(\epsilon)=D_{+}(t^\star_\delta-\epsilon)-D_{+}(t^\star_\delta)$ for $\ep>0$. 
Without Condition \eqref{ma2}, $D_{-}$ and $D_{+}$ could potentially be ``flat'', 
making it hard to estimate ${t}^\star_\delta$, as illustrated in case (3) of Figure \ref{fig:estmation_t}. 
In addition, 
if either 
$D_{-}(t^\star_\delta)<\delta$ 
or 
$\delta<D_{+}(t^\star_\delta)$, 
we do not need a lower bound for that side of the distribution around $t^\star_\delta$,
as the gap between $\delta$ and $D_{j}(t^\star_\delta), j\in\{+,-\}$ 
ensures that $t^\star_\delta$ can be estimated accurately;
see case (1) of Figure  \ref{fig:estmation_t}.

Moreover,
we also need 
to ensure that the mass of $X$ is sufficiently ``spread out", as per the following strong density condition introduced by \cite{tsy2007}.
For $x\in\R^d$ and $r\ge 0$,
denote by
$B_{d,2}(x,r)$ the closed $d$-dimensional Euclidean ball centered at $x$ with radius $r$. 

\begin{definition}[Strong Density Condition \citep{tsy2007}]\label{sd}
Fix   a compact set $\widetilde C \subset \mathbb{R}^d$. We say that a distribution of $(X,A)$ with the pair of conditional distributions $(\Po,\Pz)$ satisfies \emph{the strong density condition} if there exist positive constants $c_\mu$, $r_\mu$, $\mu_{\min}$ and $\mu_{\max}$ such that the following  hold.  
For $a\in\{0,1\}$, $\Pa$ is absolutely continuous with respect to the Lebesgue measure $\lambda$ on $\mathbb{R}^d$, and it
is supported on a compact \emph{$(c_\mu,r_\mu)$-regular}
set $\Omega_a \subset \widetilde C$, namely
$$\lambda\left[ B_{d,2}(x,r)\cap \Omega_a\right]\ge  c_\mu \lambda \left[  B_{d,2}(x,r)\cap\widetilde C\right],
\  \textnormal{ for all }
x \in\Omega_a\,
\textnormal{ and }
0 < r \le r_\mu.$$ 
Moreover, 
for $a\in\{0,1\}$,
the density function  $\mu_a$ of $\Pa$ with respect to the Lebesgue measure 
satisfies $\mu_{\min} \le \mu_a(x)\le \mu_{\max}$ for $x \in \Omega_a$ and $\mu_a(x) = 0$ otherwise.
\end{definition} 

Letting $\delta>0$ be the chosen disparity level, with the above definitions, our parameter space $\mathcal{P}_\Sigma(\delta,\beta,L_\beta,\gamma)$---or, $\mathcal{P}_\Sigma$ for short---is defined as:
\begin{definition}[Parameter space]\label{ps}
 For $\delta\ge0$, ${\beta}, L_\beta > 0$ and $\gamma_\delta, \gamma_\infty \ge 0$,
we denote by $\mathcal{P}_\Sigma$ the class of all probability distributions $P$ on $\X\times\A\times\mathcal{Y}$ satisfying the following.
\begin{enumerate}
    \item The group-conditional probability functions $\eta_a: x\mapsto\P(Y=1\mid A=a,X=x)$, for $x\in\X$ and
$a\in \A$, satisfy
    $\eta_1,\eta_0\in\Sigma(\beta,L_{\beta},\R^d)$, for the H\"older parameter space $\Sigma$ from Definition \ref{h}.
    \item Further, $(\eta_1,\eta_0)$ satisfies the margin condition 
    from Definition \ref{m} at $\delta$ for $\gamma_\delta$ and at $\infty$ for $\gamma_\infty$.
    \item  The pair of distributions
    $(\Po,\Pz)$ satisfies  the strong density condition from Definition \ref{sd}. 
\end{enumerate} 
\end{definition}

Without loss of generality, we can assume that $\eta_1$ and $\eta_0$ share the same smoothness parameter $\beta$ and satisfy the margin condition with the same parameters $\gamma_\delta$, $\gamma_\infty$. 
When  $\eta_1\in \Sigma(\beta_1,L_{\beta,1},\R^d)$ satisfies the $\gamma_{\delta,1}$-margin condition and $\eta_0\in \Sigma(\beta_0,L_{\beta,0},\R^d)$ satisfies the $\gamma_{\delta,0}$-margin condition, we can set $\beta=\beta_1\wedge \beta_0$ and  $\gamma_\delta=\gamma_{\delta,1}\wedge \gamma_{\delta,0}$.

We will further assume that $\gamma_{\delta}\beta \le d$,  which in particular holds if 
$\gamma_{\delta}$ and $\beta$ are constants; 
this condition is commonly used when deriving minimax lower bounds in nonparametric classification \citep[e.g.,][etc.]{tsy2007,CaiandWei}.

\subsection{Minimax Lower Bound}
We now present our first major result: a minimax lower bound for fair classification. 
The estimation errors of
both the regression functions and the group-wise  thresholds contribute to the overall error in estimating a $\delta$-fair Bayes-optimal classifier. 
Based on the discussion in Remark \ref{if} and after Definition \ref{m}, there are two cases:
(1) The \emph{non-trivial fairness-impacted regime}, 
where $t^\star_\delta>0$ (fairness-impacted regime from Remark \ref{if}) and 
  $D_{-}\lsb t_\delta^\star\rsb=D_{+}\lsb t_\delta^\star\rsb=\delta$ ($D_-$ and $D_+$ are continuous at $t_\delta^\star$), so that
the estimation of $t^\star_\delta$ may affect the minimax lower bound;
(2) The \emph{classical regime}, which is the complement of case (1).
Here, $t^\star_\delta$ does not need to be estimated, or can be estimated at a fast rate.

\begin{figure}[t]
    \centering
    \includegraphics[width = 0.9\textwidth]{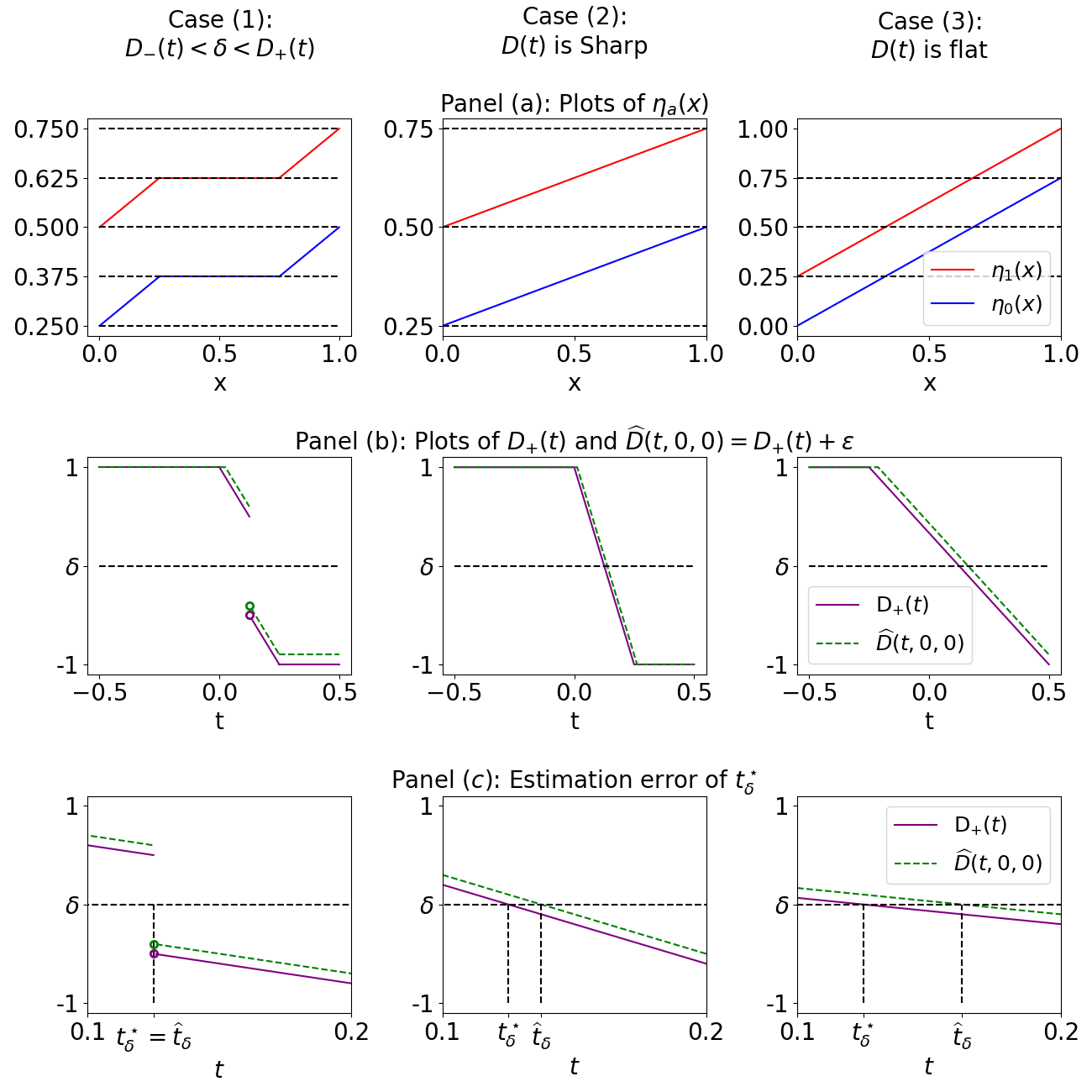}
    \caption{Estimation error of $t^\star_\delta$ in three cases, with $\P(A=1)=1/2$, $X|A=a\sim U(0,1)$ and $\delta = 0$. As we can see, when $D_-(t^\star_\delta) < \delta < D_+(t^\star_\delta)$, $t^\star_\delta$ can be estimated with a fast rate. When $\delta = D_-(t^\star_\delta)$ (or $\delta = D_+(t^\star_\delta)$), the convergence rate depends on the slope of $D_-$ (or $D_+$) near $t^\star_\delta$.}
    \label{fig:estmation_t}
\end{figure}

\begin{theorem}[Minimax lower bound for fair classification]\label{thm:lower_bound}
For a fixed $\delta\ge 0$, let $\beta, \gamma_\delta>0$ be such that $\gamma_\delta\beta\le d$,
and consider the class of distributions $\mathcal{P}_\Sigma=\mathcal{P}_\Sigma(\delta,\beta,L_\beta,\gamma_\delta,\gamma_\infty,c_\mu,r_\mu,\mu_{\min},\mu_{\max})$
from Definition \ref{ps}. 
Let $\delta_{\sgn}=\textup{DDP}(f_\delta^\star)$.\footnote{We have $\delta_{\sgn}=\delta$ in the $\delta$-positive DDP case  when  $D_{-} (0)>\delta$ and $\delta_{\sgn}=-\delta$ in the $\delta$-negative DDP case  when $D_{+} (0)<-\delta$.}
Then, 
there is $C>0$ depending 
only on the problem hyperparameters, 
such that for any $n \ge 1$ and any classifier $\widehat{f}_{\delta,n}$ 
estimating the  $\delta$-fair Bayes-optimal classifier
$f^\star_\delta$ from \eqref{bao},
constructed from a dataset $\mathcal{S}_n=\{(X_i,A_i,Y_i)\}_{i=1}^n$
sampled i.i.d.~from some $P$ from $\mathcal{P}_\Sigma$, we have the following.
\begin{itemize}
  \item[(1). {\bf Non-trivial fairness-impacted regime.}] 
  In the fairness-impacted regime from Remark \ref{if}, if in addition 
  $D_{-}\lsb t_\delta^\star\rsb=D_{+}\lsb t_\delta^\star\rsb=\delta$, we have
\begin{equation}\label{eq:lowbnd:impacted2}
\inf_{\widehat{f}_{\delta,n}}\sup_{\P\in\mathcal{P}}\EN\lmb d_E\lsb\widehat{f}_{\delta,n},f^\star_\delta\rsb\rmb\ge C\lmb n^{-(\gamma_\delta+1)\beta/(2\beta+d)}+ n^{-(\gamma_\delta+1)/(2\gamma_\delta)}\rmb.
\end{equation}

   \item[(2). {\bf{Classical regime.}}] 
  Otherwise, let   $\gamma' = \gamma_\infty$
    in the automatically fair  and  fair-boundary cases from Remark \ref{if}, i.e., 
   for $ 
D_{-} (0)\le\delta$;
 and $\gamma'= \gamma_\delta$
     in the fairness-impacted case from Remark \ref{if}, if further $D_{-}\lsb t_\delta^\star\rsb<\delta$ or $D_{+}\lsb t_\delta^\star\rsb>\delta$.
    Then, we have
\begin{equation}\label{eq:lowbnd:auto}
\inf_{\widehat{f}_{\delta,n}}\sup_{\P\in\mathcal{P}}\EN\lmb d_E\lsb\widehat{f}_{\delta,n},f^\star_\delta\rsb\rmb\ge C n^{-(\gamma'+1)\beta/(2\beta+d)}.
\end{equation}
\end{itemize}

\end{theorem}
We have the following observations:

\begin{enumerate}
  \item In the  fairness-impacted case, 
     there are two sources of error:
     the estimation error of the regression functions near the decision boundaries,
     and the estimation error of the thresholds, i.e., balancing the probability of success in each group.
     The first is well characterized by the  boundary behavior of $\eta_a$, $a\in \{0,1\}$.
      For the second estimation error,
      when $D_{-}\lsb t_\delta^\star\rsb<\delta$ or $D_{+}\lsb t_\delta^\star\rsb>\delta$, $t_\delta^\star$
    can be estimated with a
     rate faster than the regression functions, see Case (1) of Figure \ref{fig:estmation_t}. 
   When $D_{-}\lsb t_\delta^\star\rsb=D_{+}\lsb t_\delta^\star\rsb=\delta$,  and $D_{+}$ or $D_{-}$ is relatively ``steep" near $t^\star_\delta$, 
    one can estimate $t^\star_\delta$ at a faster rate (as shown in Case (2) of Figure  \ref{fig:estmation_t}). 
    In contrast, when $D_{+}$ or $D_{-}$ is relatively ``flat" near $t^\star_\delta$, 
    the error in estimating $t^\star_\delta$ is larger (See Case (3) of Figure \ref{fig:estmation_t}).
    In addition to the population-level
    accuracy loss due to fairness, 
    the constraint
    worsens 
    the minimax lower bound
    when $\gamma_\delta> 1+d/(2\beta)$.
    \item In the automatically fair  and fair-boundary cases, 
    the lower bound  \eqref{eq:lowbnd:auto}
    depends only on the error in estimating the regression functions near the optimal threshold $1/2$; and  
    coincides with the lower bound 
    from conventional non-parametric classification problems \citep{tsy2007}. 
\end{enumerate}

The proof of Theorem \ref{thm:lower_bound}, presented in Section \ref{pf:thm:lower_bound}, consists of two parts,
considering the convergence rate 
of regression functions and of thresholds separately. 
In the first part, 
starting with an approach similar to 
\cite{tsy2007}, 
we construct a set of distributions indexed by the hyper-cube and meticulously verify the distributional assumptions. 
We then leverage Assouad's lemma to derive the lower bound from \eqref{eq:lowbnd:auto}.
In the second step, 
we consider the effect of estimating the thresholds. We depart from the existing proof ideas, 
introducing a novel construction which provides two very similar distributions with different optimal thresholds $t_\delta^\star$. Then, by applying Le Cam's Lemma, we establish the additional term of the lower bound for the fairness-impacted case.

\section{FairBayes-DDP+: Plug-in Thresholding Rule with Offset}\label{sec:FairBayes}
In this section, we complete the picture by proposing an adaptive thresholding estimator that achieves the
minimax lower bounds. 
Together with Theorem \ref{thm:lower_bound},  this establishes the minimax convergence rate in our fair classification problems. 
We first introduce the required estimators.

\subsection{Local Polynomial Estimator of the Regression Function}
In this section, we recall the definition of the local polynomial estimator of the regression function \citep[e.g,][etc]{tsybakov2009nonparametric,tsy2007}. 
For a random variable $(X,Y)$ over $\R^d \times \R$
and 
$n$ i.i.d.~copies $(X_i,Y_i)_{i=1}^n$, 
the local polynomial estimator of
the regression function
$\eta:\R^d\to \R$,
such that for all $x\in \R^d$,
$\eta(x)=\E(Y|X=x)$,  
is defined as follows.
\begin{definition}
For 
a bandwidth
$h > 0$, $x \in \MR^d$, for an integer $k\ge 0$ and a kernel
$K: \MR^d \to [0,\infty)$, denote by $\widehat{\theta}_x$ a polynomial on $\MR^d$ of degree $k$,
whose $k+1$ coefficients minimize\footnote{A minimizer always exists, but may not be unique.} over $\R^{k+1}$
\begin{equation}\label{LPest}
\sum_{i=1}^n\left(Y_i-\widehat\theta_x(X_i-x)\right)^2 K\lsb\frac{X_i-x}{h}\rsb.
\end{equation}
The local polynomial estimator $\widehat{\eta}^{\LP}_n(x)$ of order $k$ of the
value $\eta(x)$ of the regression function at the point $x$ is defined by $\widehat{\eta}^{\LP}_n
(x)=\widehat{\theta}_x (0)$ if $\widehat{\theta}_x$ is a unique minimizer of \eqref{LPest}, and $\widehat\eta^{\LP}_n(x)=0$ if $\widehat{\theta}_x$ is a non-unique minimizer. 
\end{definition}
For a multi-index $s\in \mathbb{N}^d$, and $x\in \R^d$, 
we introduce the vector $U(x)=(x^s)_{|s|\le k}$ and matrix $Q=(Q_{s_1,s_2})_{|s_1|,|s_2|\le k}$
with
$$Q_{s_1,s_2}(x)=\sum_{i=1}^n(X_i-x)^{s_1+s_2}K\lsb \frac{X_i-x}{h}\rsb.$$
\cite{tsy2007} show that if the matrix $Q$ is positive definite, there exists a
unique polynomial on $\R^d$ of degree $k$ minimizing \eqref{LPest}. The corresponding local polynomial regression function
estimator equals, for all $x\in\R^d$,
$$\eta^{\LP}_n(x)=\sum_{i=1}^n Y_i K\lsb\frac{X_i-x}{h}\rsb U^\top  (0)Q^{-1}U(X_i-x).$$
We refer readers to \cite{tsy2007} and \cite{tsybakov2009nonparametric} for more details about local polynomial estimators. 

\subsection{Bandwidth Parameter with Possible Jump Discontinuity}\label{sec:discont}
We saw in Theorem \ref{thm:lower_bound}
that even in the fairness-impacted case, if 
 $D_{-}\lsb t_\delta^\star\rsb<\delta$ or $D_{+}\lsb t_\delta^\star\rsb>\delta$, the fair Bayes-optimal classifier can be estimated with the classical 
 convergence rate for non-parametric regression. 
 This happens if $t_\delta^\star$ is a jump discontinuity point\footnote{For a function $D: \R\to \R$, we say that $t_0$ is a jump discontinuity point of $D$ if both the left limit $\lim_{t\to t_0^-}D(t)$ and right limit $\lim_{t\to t_0^+}D(t)$ of $D$ at $t_0$ are finite, and $\lim_{t\to t_0^+}D(t)\neq\lim_{t\to t_0^-}D(t)$.
 }  
 of $t\mapsto D_-(t)-\delta$ or $\mapsto D_+(t)-\delta$; by definition, $D_-$ and $D_+$ share the same jump discontinuity points.
 Even though the disparity functions $D_-$, $D_+$ 
 are estimated at a non-parametric rate, 
 its jump discontinuities can be estimated with a near-parametric rate. 
 We consider the following strategy to estimate $t^\star_\delta$,
 by taking possible discontinuities into account.

Let $(\Delta_n)_{n\ge 0}$ and $(r_n)_{n\ge 0}$ be two positive sequences that converge to zero slowly (e.g., at  rates on the order of $(\log\log n)^{-1}$  and $(\log n)^{-1}$, respectively).
First, consider the case where $D_-$ and $D_+$ have a jump discontinuity at $t^\star_\delta$. 
If $\delta<D_+(t^\star_\delta)$, for $\Delta_n\to 0_+$, 
we have $D_+(t^\star_\delta)>\delta+\Delta_n>D_-(t^\star_\delta)$. 
This implies that $\inf_{t>0}\{D_+(t)>\delta+\Delta_n\} = t^\star_\delta$. 
Similarly, if $D_-(t^\star_\delta)<\delta$,  we have  $\inf_{t>0}\{D_-(t)>\delta-\Delta_n\} = t^\star_\delta$.
Other other hand, if both $D_-$ and $D_+$ are continuous at $t^\star_\delta$ with $D_-(t^\star_\delta)=D_+(t^\star_\delta)=\delta$, it holds that  $\inf_{t>0}\{D_+(t)>\delta+\Delta_n\}<t_\delta^\star$ and $\inf_{t>0}\{D_-(t)>\delta+\Delta_n\}>t_\delta^\star$.
Motivated by this, 
letting $\widehat{D}_n$ be an estimate of $D_-$ (or $D_+$), we define
$ \htmid, \htmin$ and $\htmax$ as in \eqref{hts} 
and $\widehat{t}_\delta$ as in \eqref{ht} using empirical versions of the above relations.

\subsection{Plug-in Estimators with Offset}\label{sec:offsets}
Next, we consider estimating the group-wise probability functions via
plug-in estimators with an offset \citep{Vert2009}. For a density function 
$g$
defined on $\X$, the $\lambda$-level set of $g$ is 
$\Lambda_g(\lambda)=\{x\in\X, \ g(x)>\lambda\}$.
If $\widehat{g}$ is a consistent estimator of $g$, 
a first thought is to estimate $\Lambda_g(\lambda)$ by the plug-in estimator
$\widehat{\Lambda}_g(\lambda)=\{x\in\X,\ \widehat{g}(x)>\lambda\}$.
However, this can be inconsistent if the boundary set $\{x\in\X, \ g(x) = \lambda\}$ has positive probability. 
Alternatively, \cite{Vert2009} proposed plug-in density estimators with offset $(\ell_n)_{n\ge 1}$:
$$\widetilde{\Lambda}_{g,\ell_n}(\lambda)=\widehat{\Lambda}_{g}(\lambda+\ell_n)=\{x\in\X, \ \widehat{g}(x)>\lambda+\ell_n\},$$
where $\ell_n \ge 0$ 
tends to zero as $n$ tends to infinity; see \Cref{sec:pf-FairBayes} for further discussion.
Similarly, $\{x\in \X, \ g(x)<\lambda\}$ and $\{x\in \X, \ g(x)=\lambda\}$ can be consistently estimated by $\{x\in \X, \ \widehat{g}(x)\le \lambda-\ell_{n}\}$ and $\{x\in \X, \ \lambda-\ell_n< \widehat{g}(x)\le \lambda+\ell_{n}\}$, respectively; and we will adapt such ideas to our problem. 

\subsection{FairBayes-DDP+: Plug-in Estimator with Offset for Fair Classification}\label{sec:plug-in}

\begin{algorithm}[tb]
   \caption{FairBayes-DDP+: Thresholding Rule with Offset for Fair Classification under Demographic Parity}
   \label{alg:FBC1}
\begin{algorithmic}

   \STATE {\bfseries Input:} Disparity level $\delta$. Dataset $\mathcal{S}_n=S_{n,1}\cup S_{n,0}$ with  $\mathcal{S}_n=\{(x_{i},a_i,y_{i})\}_{i=1}^{n}$ and, for $a\in\{0,1\}$, $S_{n,a}=\lbb (\xaj,a,y_{a,j})\rbb_{j=1}^{n_{a}}$.

   \STATE {\textbf{Step 1}:} Estimate $\eta_a$ and $\eta_0$ by local polynomial estimators: 
    \STATE{\quad Let $U(x)=(x^s)_{|s|\le \lbeta_+}$, $Q_a=(Q_{a,s_1,s_2})_{|s_1|,|s_2|\le\lbeta_+}$ with $Q_{a,s_1,s_2}=\sum\limits_{j=1}^{n_{a}} \lsb \xaj-x\rsb^{s_1+s_2}K\lsb \frac{\xaj-x}{h_{n,a}}\rsb,$ \\ \quad and $\bar{B}_a=(\bar{B}_{a,s_1,s_2})_{|s_1|,|s_2|\le\lbeta_+}$ with $\bar{B}_{a,s_1,s_2}={n_a^{-1}h_{n,a}^{-(d+s_1+s_2)}}Q_{a,s_1,s_2}$. \\
    \quad  Define $$\eta^{\LP}_{n,a}(x)=\sum_{j=1}^{n_{a}} Y_{a,j} K\lsb\frac{\xaj-x}{h_{n,a}}\rsb U^\top  (0)Q^{-1}U(X_i-x).$$
    \quad Let
    $$\widehat{\eta}_a(x)=\left\{\begin{array}{lcl}
         0;& \lambda_{\min}(\Bar{B}_a)\le (\log n_{a})^{-1} \text{ or } \eta^{\LP}_{n,a}(x)<0; \\
         1;& \lambda_{\min}(\Bar{B}_a)> (\log n_{a})^{-1} \text{ and } \eta^{\LP}_{n,a}(x)>1; \\ \
         \eta^{\LP}_{n,a}(x), & \text{otherwise}.
    \end{array}\right.$$
    }
       \STATE {\textbf{Step 2}:} Estimate the optimal thresholds:
     \STATE{
\begin{equation}\label{eq:dhatn}
\widehat{D}_{n}(t,\lo,\lz)=
\frac{1}{n_{1}}\sum\limits_{j=1}^{n_{1}}I\lsb\widehat{\eta}_1(\xoj)>\frac12+\frac{nt}{2n_1}+\lo
\rsb-\frac{1}{n_{0}}\sum\limits_{j=1}^{n_{0}}I\lsb\widehat{\eta}_0(\xzj)> \frac12-\frac{nt}{2n_0}+\lz \rsb.
\end{equation}
Set
\begin{equation}\label{hts} 
\left\{\begin{array}{l}
 \htmid = \inf_{t\ge 0}\lbb \widehat{D}_n(t,0,0)<\delta\rbb;\\
\htmin = \inf_{t\ge 0}\lbb \widehat{D}_n(t,0,0)<\delta+\Delta_n\rbb;\\
\htmax = \inf_{t\ge 0}\lbb \widehat{D}_n(t,0,0)<\delta-\Delta_n\rbb,
 \end{array}
 \right.\end{equation}
 \begin{equation}\label{ht}
\widehat{t}_\delta =\left\{
\begin{array}{lcl}
\htmin,    &&  \htmid-\htmin\le r_n;\\
\htmax,    & & \htmid-\htmin> r_n \text{ and } \htmax-\htmid\le r_n;\\
\htmid,    & & \htmid-\htmin> r_n \text{ and } \htmax-\htmid> r_n.
\end{array}
\right.    
\end{equation}
 and
\begin{equation}\label{htau}
\taudho=\rho\lsb\frac{\widehat{\pi}_{n,0,+}-\widehat{\pi}_{n,1,+}+\widehat{\delta}}{\widehat{\pi}_{n,1,=}}\rsb\qquad \text{and}\qquad
  \taudhz=\rho\lsb\frac{ \widehat{\pi}_{n,1,+}-\widehat{\pi}_{n,0,+}-\widehat{\delta}}{ \widehat{\pi}_{n,0,=}}\rsb.
\end{equation}
 with 
\beq\label{hpin}
\pihnap=\frac1{n_{a}}\sum_{j=1}^{n_{a}} I\left(\widehat\eta_a(\xaj)>\Thda+\ela\right),\quad 
\pihnae=\frac1{n_{a}}\sum_{j=1}^{n_{a}} I\left( \Thda-\ela  < \widehat\eta_a (\xaj)\le \Thda+\ela\right),
\eeq
and
\beq\label{hdel}
\widehat{\delta}=\delta\cdot  I\left(\widehat{D}_{n} (0,\lo,-\lz)>\delta\right).
\eeq
}
\STATE {\bfseries Output:}  $\widehat{f}_{\delta,n}$ from \eqref{eq:finalestimate}.
\end{algorithmic}
\end{algorithm}

In this section, we introduce our 
FairBayes-DDP+ method, 
a classifier for minimax  optimal classification under demographic parity. 
This method is based on a two-stage plug-in estimator with offsets, where the first stage estimates the regression functions for each group and the second stage estimates the thresholding adjustment parameter $t_\delta^\star$. 
As elaborated in Appendix \ref{ha},
the behavior of the fair Bayes-optimal classifier on the decision boundary is generally not unique. 
For identifiability, 
we estimate a specific Bayes-optimal classifier, 
chosen such that the probability of $\widehat{Y}=1$
is minimized on the decision boundaries. 
Consider 
the truncation function 
 $\rho: \mathbb{R}\cup\{\infty\}\to [0,1]$ 
defined as
\begin{equation}\label{rho}
\rho\lsb x\rsb=\left\{\begin{array}{lcl}
    0,    &  x\le 0 \\
    1,    &  x \ge 1;\\
    x, &  \text{otherwise.}
    \end{array}\right.
\end{equation}
and the adjustment of $\delta$ 
in the fairness-impacted case given by
\beq\label{dt}
\widetilde{\delta} = \delta I\left(D_{-} (0)>\delta \right)=\left\{\begin{array}{ll}
0,     &  D_-(0)\le \delta; \\
 \delta,    &  D_-(0)>\delta.
\end{array}\right. 
\eeq
We set, 
for $a\in\{0,1\}$, 
\begin{equation}\label{eq:pistar}
\pisap=\Pa(\eta_a(X)>\Tsda) 
\qquad\text{and}\qquad
\pisae=\Pa(\eta_a(X)=\Tsda),
\end{equation} 
and interpreting $x/0 = 0$ for all $x\in \R$ here and in what follows, the randomization probabilities
    \begin{equation}\label{eq:Taustar}
\taudso=\rho\lsb\frac{ \pi^\star_{0,+}-\pi^\star_{1,+}+\widetilde{\delta} }{ \pi^\star_{1,=}}\rsb \qquad\text{and}\qquad
    \taudsz=\rho\lsb\frac{ \pi^\star_{1,+}-\pi^\star_{0,+}-\widetilde{\delta} }{ \pi^\star_{0,=}}\rsb.
\end{equation} 
These choices lead to a specific fair Bayes-optimal classifier with the following properties.
\begin{proposition}[Properties of a specific fair Bayes-optimal classifier]\label{prop:checkcondition}
Consider the fair Bayes-optimal classifier $f^\star_\delta$ from \eqref{bao},
with the thresholds $\Tsda$ from \eqref{Tdel} and 
the randomization probabilities
$\taudsa$ from \eqref{eq:Taustar}. 
\begin{itemize}
    \item In the fairness-impacted case where  $D_-(0)>\delta$, we have that
    $\textup{DDP}(f^\star_\delta)=\delta;$
        \item In the automatically fair and fair-boundary cases where $D_{-} (0)\le\delta,$ we have that
    $$\textup{DDP}(f^\star_\delta)=0 \text{ if } D_-(0)\le 0, \quad
    \text{ and }\quad\textup{DDP}(f^\star_\delta)=D_-(0) \text{ if } 0<D_-(0)\le \delta.$$

\end{itemize}
 \end{proposition}

Now, suppose we have a dataset $\mathcal{S}_n=\{(x_i,a_i,y_i)\}_{i=1}^n$.  
We  separate the data according to the protected information: for $a\in\{0,1\}$, 
we let
$\mathcal{S}_{n,a}= \{(x_i,a_i,y_i)\in \mathcal{S}_{n}, a_i=a \}$ with $n_{a}:=|\mathcal{S}_{n,a}|$. The $j$-th element of $\mathcal{S}_{n,a}$ is denoted as $(\xaj,a,y_{a,j})$, for $j\in[n_a]$.

{\bf Step 1}: 
First, we estimate $\eta_a$ via local polynomial estimation using $\mathcal{S}_{n}$. 
Specifically, consider a kernel $K:\R^d\to \R$ 
satisfying
\begin{equation}\label{Kernel}
\begin{array}{lcl}
(1) \textnormal{ there\, is } c>0 \textnormal{ with } \, K(x)\ge cI(\|x\|\le c), \  \textnormal{ for all } x\in\MR^d; &
(2)\,\int_{\MR^d} K(t)dt=1;    \\
&\\
(3)\,\int_{\MR^d} (1+\|t\|^{4\beta})K^2(t)dt<\infty;&
(4)\,\sup_{t\in\MR^d}(1+\|t\|^{2\beta})K(t)<\infty. 
\end{array}
\end{equation}
One can take, for example, $K$ as the Gaussian kernel.
Let $h_{n,a} > 0$, and consider the matrix $\bar{B}_a= (\bar{B}_
{a,s_1,s_2})_{|s_1|,|s_2|\le \lbeta_+}$, where $$\bar{B}_{a,
s_1,s_2}(x) 
=\frac1{n_{a}h_{n,a}^d}\sum_{j=1}^{n_{a}}\lsb\frac{\xaj-x}{h_{n,a}}\rsb^{s_1+s_2}K\lsb\frac{\xaj-x}{h_{n,a}}\rsb.$$
Define the regression function estimator $\widehat{\eta}_a$ as follows. 
If the smallest eigenvalue of the matrix $\bar{B}_a$ is greater than or equal to $(\log n_{a})^
{-1}$,
for all $x$,
we
set $\widehat\eta_a(x)$ equal to the projection of $\widehat{\eta}_a^{\LP}(x)$ on the interval $[0, 1]$, where $\widehat\eta^{\LP}_a$
is the $LP(\lbeta_+)$ estimator of $\eta_a$ with bandwidth $h_{n,a} > 0$ and kernel $K$ satisfying \eqref{Kernel}. If the smallest eigenvalue of $\bar{B}_a$ is less than $(\log n_{a})^{
-1}$, we set $\widehat\eta_a(x) = 0$.

{\bf  Step 2}: 
In the second step, we start by estimating the acceptance threshold for each protected group, via solving a one-dimensional empirical fairness constraint
and then
determining the prediction on the decision boundaries
when those have strictly positive estimated probability. 
We observe that
the thresholds of the fair Bayes-optimal classifier 
from \eqref{bao}
balance the probability measures of 
the level sets $\{ x\in\X,\eta_1(x)>\Tsdo\}$ and $\{ x\in\X, \eta_0(x)>\Tsdz \}$. 
As a result, we can incorporate 
plug-in estimation with an offset for level set estimation.
Specifically, for $a\in\{0,1\}$, some $\ell_{n,a}>0$, and any $\zeta \in \R$, we 
estimate $\Pa(\eta_a(X)>\zeta)$ and $\Pa(\eta_a(X)\ge \zeta)$
by
$n_{a}^{-1}\sum_{j=1}^{n_{a}}I\lsb\widehat\eta_a\lsb\xaj\rsb>\zeta+\ell_{n,a}\rsb$ and $n_{a}^{-1}\sum_{j=1}^{n_{a}}I\lsb\widehat\eta_a\lsb\xaj\rsb>\zeta-\ell_{n,a}\rsb$, respectively.
With this, 
the probability of the decision boundary can be consistently estimated. 
Based on \eqref{bao}, 
we consider
 the group-wise thresholding rule defined for all $x,a$ by
\begin{equation*}
{f}^{t}_\ell(x,a)=
I\lsb\widehat\eta_{a}(x)> \frac12+\frac{n{t}}{2(2a-1)n_{a}}+\ell_{n,a}\rsb+\tau_aI\lsb\left| \widehat\eta_{a}(x)-  \frac12+\frac{n{t}}{2(2a-1)n_{a}}\right|\le\ela\rsb,
\end{equation*}
where $\eta_a$ and $p_a$ are estimated by plug-in estimators. 

Next, our goal is to construct estimates $\widehat{t}_\delta$ and $\widehat{\tau}_{\delta,a}$ such that the proposed classifier approximately satisfies the fairness constraint. 
With $\widehat{D}_{n}(t,\lo,\lz)$ from \eqref{eq:dhatn},
we consider 
plug-in estimators of $D_{-}(t)$ and $D_{+}(t)$  given by $\widehat{D}_{n}(t,\lo,-\lz)$
and $\widehat{D}_{n}(t,-\lo,\lz)$ with $\lo,\lz>0$,  respectively. 
We estimate $t^\star_\delta$ using the approach introduced in Section \ref{sec:discont}. 
Let $\widehat{t}_\delta$ defined as in \eqref{ht}
and let, for $a\in\{0,1\}$, $\Thda=1/2+(2a-1)n\widehat{t}_\delta/n_{a}$.
We estimate $\taudso$ and $\taudso$ specified in \eqref{eq:Taustar} by $\taudha$  from \eqref{htau},
using 
the plug-in estimates with offsets 
$(\pihnap,\pihnae)$ 
of $(\pisap,\pisap)$ from \eqref{hpin}, respectively. 
Also, $\widetilde{\delta}$ 
from \eqref{dt} is estimated by $\widehat{\delta}$ from \eqref{hdel}.
Our final estimate of the $\delta$-fair Bayes-optimal classifier is 
\begin{equation}\label{eq:finalestimate}
\widehat{f}_{\delta,n}(x,a)=
I\lsb\widehat\eta_{a}(x)>\Thda+\ell_{n,a}\rsb+\widehat{\tau}_{\delta,a}I\lsb|\widehat\eta_{a}(x)-\Thda|\le\ela \rsb.
\end{equation}
Our plug-in method is directly motivated by the fair Bayes-optimal classifier from Theorem \ref{thm-opt-dp}. The offsets $\ell_{n,a}, a\in\{0,1\}$ 
are carefully designed to handle estimation on the boundaries.

\begin{Remark}
 In Step $1$ of our method, we use the local polynomial estimators only for theoretical purposes, as they lead to an upper bound matching the minimax lower bound.
 However, 
 as we show in our experiments,
 in practice we can use other methods, such as support vector machines or deep neural networks, to estimate the regression function for improved performance. 
\end{Remark}

\section{Asymptotic Analysis of FairBayes-DDP+}\label{sec:Asymp_analysis}
In this section, we study the statistical properties of FairBayes-DDP+. 
We first 
derive the convergence rate of our plug-in method, establishing its minimax optimality.
We then show that the constraint $|\textup{DDP}(f)|\le \delta$ is satisfied up to a vanishing error term.

\subsection{Convergence Rate and Minimax Optimality}
In this section, we establish the convergence rate of FairBayes-DDP+. 
The rate depends on the pointwise convergence of $\widehat{\eta}_a$ to $\eta_a$, $a\in \{0,1\}$. 
To quantify this rate, 
the following definition describes a notion of  pointwise convergence of a sequence of estimators of the conditional probability functions $\eta_a$, $a\in \{0,1\}$.  
\begin{definition}[Pointwise convergence]\label{pc}
 Let $\mathcal{P}$ be a class of distributions for $(X,A,Y)$ and fix $U_\eta>0$.
Let $(\phi_{n,1})_{n\ge 1}$ and $(\phi_{n,0})_{n\ge 1}$  
be two positive, monotonically non-increasing sequences.
We say that the estimator sequence $(\widehat{\eta}_{n,1},\widehat\eta_{n,0})_{n\ge 1}$, where $\widehat{\eta}_{n,1},\widehat\eta_{n,0}$ is constructed using a sample of size $n$, converges pointwise at rate $(\phi_{n,1},\phi_{n,0})_{n\ge 1}$ 
uniformly over $\mathcal{P}$
if there are positive constants $c_{1,\eta}$, $c_{2,\eta}$ and $L_\eta$, as well as a set $\Omega\subset \X$,  such that $\P(\Omega)=1$ and, for $a\in\{0,1\}$, 
\begin{equation}\label{eq:regression_function}
\sup_{\P\in\mathcal{P}} \PN\left(\sup_{x\in\Omega}\left|\widehat{\eta}_{n,a}(x) - \eta_a(x)\right| > \ep \right) \le c_{1,\eta} \exp\lsb{-c_{2,\eta}\lsb\ep/\phi_{n,a}\rsb^2}\rsb, \ \  L_\eta \phi_{n,a} < \ep < U_{\eta}.
\end{equation}
\end{definition}

We will usually drop the subscript $n$ and write $\widehat{\eta}_{a} = \widehat{\eta}_{n,a}$.
In the rest of this paper, we let, for 
$\ep>0$, for
$i\in[4]$ and $\iota\in \{t,T,t_1,t_2,r,D,\pi\}$,
and for quantities $c_{i,\iota}>0$, 
\beq\label{psi}
\psi_{n,1,\iota}(\ep)=c_{1,\iota} \exp\lsb-c_{2,\iota} \lsb\ep/[\phi_{n,1}\vee\phi_{n,0}]\rsb^2 \rsb\text{ and }\psi_{n,2,\iota}(\ep) = c_{3,\iota}\exp\lsb-c_{4,\iota}n \ep^2\rsb.
\eeq
With $t^\star_\delta$ from \eqref{eq:t-star-dp}, $D_{-}$ and $D_{+}$ from \eqref{eq:Disfun1} and \eqref{eq:Disfun2},  we denote by $\widetilde{I}^\star(\delta)$ 
the indicator function of the non-trivial fairness-impacted regime introduced in Theorem \ref{thm:lower_bound}, i.e., 
\beq\label{istar}
\widetilde{I}^\star(\delta)= I(\{t^\star_\delta>0\}\cap\{D_-(t^\star_\delta)=D_+(t^\star_\delta)=\delta\}).
\eeq
Moreover, for $\ep>0$ and $r_n>0$, we write
\beq\label{xi}
\omega(\ep,r_n)= {\widetilde{I}^\star(\delta)\cdot\ep}+(1-\widetilde{I}^\star(\delta))\cdot r_n,
\eeq
which selects $\ep$ in the non-trivial fairness-impacted regime, and $r_n$  otherwise.
We derive a a general and abstract convergence rate for $\widehat{t}_\delta$ below, assuming the convergence of $\widehat \eta_a$. Later we will apply this result to our concrete setting.
For two scalars $a,b$, we denote
their maximum by 
$\max\{a,b\}$ or
$a\vee b$, 
and 
their minimum by 
$\min\{a,b\}$ or
$a\wedge b$.
\begin{theorem}[Error bound for estimating $t^\star_\delta$]\label{thm:estimate_of_t}
Let $\mathcal{P}$ be a class of densities on $\X\times \A\times\mathcal{Y}$, 
let $\delta \ge 0$, 
and let $(\phi_{n,1},\phi_{n,0})_{n\ge 1}$ be two positive, monotonically non-increasing sequences such that, for
constants $c_\mu>0$, $\tilde\mu_a \ge 1/2$ for $a\in \{0,1\}$, we have
$\phi_{n,a} \ge c_\mu n^{-\tilde\mu_a}.$
Suppose that   $(\widehat\eta_1,\widehat\eta_0)$ are
$(\phi_{n,1},\phi_{n,0})_{n\ge 1}$-pointwise convergent 
to $(\eta_1,\eta_0)$ as per Definition \ref{pc},  
uniformly over $\mathcal{P}$.
Then,
with $D_{-}$ and $D_{+}$ from \eqref{eq:Disfun1} and \eqref{eq:Disfun2}, 
there are constants $c_{i,t}$, $i\in [4]$, 
$L_{t}$, $U_t$ and $U_{\Delta,t}$ 
such that, if $L_t(\phi_{n,1}\vee\phi_{n,0})<r_n<U_t$, 
$2(g_{\delta,-}(4r_n)\vee g_{\delta,+}(4r_n))<\Delta_n< U_\Delta$ and $L_t(\phi_{n,1}\vee\phi_{n,0})< \ep< r_n$ hold,
then
with $\psi_{n,j,t}$, $j\in\{1,2\}$ from \eqref{psi} and $\omega$ from \eqref{xi}, we have
for any $\ep>0$ that
\begin{equation}\label{eq:tres}
\PN\lsb|\widehat{t}_\delta-{t_{\delta}^\star}| >\ep\rsb
\le 
   \psi_{n,1,t}(\ep)+\sum_{j\in\{-,+\}}I(\delta = D_j(t^\star_\delta))\psi_{n,2,t}\lsb g_{\delta,j}\lsb\omega(\ep,r_n)\rsb\rsb.
\end{equation}

\end{theorem}

The bound in \eqref{eq:tres}
consists of two parts. 
The first is determined by the convergence rates of $\widehat\eta_1$ and $\widehat\eta_0$, as $\widehat{t}_\delta$ is based on them. 
The second depends on the behavior of the conditional probability functions near the decision boundary, for the same reason as explained after \Cref{thm:lower_bound}.
When $D_{-}(t_\delta^\star)<\delta$ or $\delta<D_{-}(t_\delta^\star)$, the second term in \eqref{eq:tres}
disappears, and the convergence rate of $\widehat{t}_\delta$  depends only on the convergence of $\widehat\eta_1$ and $\widehat{\eta_0}$. 

By definition, we have $\Thda=1/2+{(2a-1)\widehat{t}_\delta n}/(2n_{a})$ 
and $\Tsda=1/2+(2a-1)t_\delta^\star/(2p_a)$. 
Moreover, ${n}/{n_{a}}$ is a root-$n$-consistent estimate of $1/{p_a}$ when $p_a>0$. 
Building on these observations, we can show the following corollary, still in an abstract setting:
\begin{corollary}[Error bound for optimal thresholds]\label{constT}
Under the condition of Theorem~\ref{thm:estimate_of_t}, there are constants $c_{i,T}$, $i\in [4]$, $U_{T}$, $L_T$ and $U_\Delta$ 
such that, 
if $L_T(\phi_{n,1}\vee\phi_{n,0})<r_n<U_T$, $2(g_{\delta,-}(4r_n)\vee g_{\delta,+}(4r_n))<\Delta_n< U_{\Delta,T}$ and $L_T(\phi_{n,1}\vee\phi_{n,0})< \ep< r_n$ hold,
then 
with $\psi_{n,j,T}$, $j\in\{1,2\}$, from \eqref{psi} and $\omega$ from \eqref{xi}, we have
for any $\ep>0$ that
    \begin{equation}\label{eq:Tres1}
\PN\lsb|\widehat{T}_{\delta, a}-T^\star_{\delta, a}|>\ep \rsb\le \psi_{n,1,T}(\ep)+\sum_{j\in\{-,+\}}I(\delta = D_j(t^\star_\delta))\psi_{n,2,T}\lsb g_{\delta,j}\lsb\omega(\ep,r_n)\rsb\rsb.
\end{equation}
\end{corollary}
Assuming the convergence rates of the estimated regression functions in \eqref{eq:regression_function},
and with the results on the thresholds from \eqref{eq:tres}, 
we can show that 
FairBayes-DDP+ is asymptotically fair and accurate, again first in an abstract setting.

\begin{theorem}[Fairness-aware excess risk upper bound; abstract version]\label{thm:convergence_rate}
For any $\delta \ge 0$, suppose that the conditions of Theorem~\ref{thm:estimate_of_t} hold. 
Suppose further that the regression functions $(\eta_1, \eta_0)$ satisfy the $\gamma_\delta$-margin condition. Then, the plug-in estimate $\widehat{f}_{\delta,n}$ with offsets $\ela$ from \eqref{eq:finalestimate}, $a\in\{0,1\}$, satisfies, 
with
$\widetilde{I}^\star(\delta)$ from
\eqref{istar},
\begin{equation}\label{eq:upperbnd:impacted1}
\sup_{\P\in\mathcal{P}}\EN\lmb d_E\lsb\widehat{f}_{\delta,n},f^\star_\delta\rsb\rmb\le 
 C \lsb (\phi_{n,1}\vee \phi_{n,0}\vee \lo\vee \lz) +\widetilde{I}^\star(\delta)n^{-1/(2\gamma_\delta)}\rsb^{\gamma_\delta+1}.
\end{equation}
\end{theorem}
The convergence of $d_E(\widehat{f}_{\delta,n},f^\star_\delta)$ remains unaffected by the offsets when $\lo\vee \lz\le C(\phi_{n,1}\vee \phi_{n,0})$. 
Indeed, 
the boundary effects are negligible when considering $d_E$, as the expression $(f(x,a)-f^\star_\delta(x,a))(\Tsda-\eta_a(x))\equiv 0$ holds for any $f$ on the boundary sets. 
However, the offsets are key to ensuring the asymptotic fairness of our method, as demonstrated in the next section. Additionally, they also impact the accuracy through Proposition \ref{measure2}.

 We can make this result concrete
by leveraging 
the point-wise convergence of the local polynomial estimator from \cite{tsy2007},
for the appropriate choice of $h_{n,a}$. 
With this we now show that FairBayes-DDP+ achieves the minimax lower bound derived in Theorem \ref{thm:lower_bound}.

\begin{corollary}[Fairness-aware excess risk upper bound]\label{cor:minimax optimality}
Consider the class of densities $\P_\Sigma$ defined in Definition \ref{ps}. 
For any $\delta>0$, consider the 
FairBayes-DDP+  classifier $\widehat{f}_{\delta,n}$ from \eqref{eq:finalestimate},
where for $a\in\{0,1\}$, $\eta_a$ is estimated by the local polynomial estimators of $(\eta_1,\eta_0)$ with kernel $K$ satisfying 
\eqref{Kernel} and $h_{n,a}\asymp n_a^{
-1/{(2\beta+d)}}$, $\Delta_{n}\asymp (\log\log n)^{-1}$, $r_{n}\asymp (\log n)^{-1}$, and the offsets satisfy $\lo,\lz \asymp n^{-\beta/(2\beta+d)}$.
Then, we have, with
$\widetilde{I}^\star(\delta)$ from
\eqref{istar},
\begin{equation*}\label{eq:upperbnd:impacted2}
\sup_{\P\in\mathcal{P}_\Sigma}\EN\lmb d_E\lsb\widehat{f}_{\delta,n},f^\star_\delta\rsb\rmb\le C\lmb n^{-(\gamma_\delta+1)\beta/(2\beta+d)}+ \widetilde{I}^\star(\delta) n^{-(\gamma_\delta+1)/(2\gamma_\delta)}\rmb.
\end{equation*}
\end{corollary}
The convergence rate stated in equation \eqref{eq:lowbnd:impacted2} matches the minimax lower bound specified in Theorem \ref{thm:lower_bound}. Specifically, 
\begin{itemize}
  \item[(1)] In the non-trivial fairness-impacted regime,  i.e., $t^\star_\delta>0$ and 
  $D_{-}\lsb t_\delta^\star\rsb=D_{+}\lsb t_\delta^\star\rsb=\delta$, we have  $\widetilde{I}^\star(\delta)=1$ and
\begin{equation*} 
\sup_{\P\in\mathcal{P}}\EN\lmb d_E\lsb\widehat{f}_{\delta,n},f^\star_\delta\rsb\rmb\le C\lmb n^{-(\gamma_\delta+1)\beta/(2\beta+d)}+ n^{-(\gamma_\delta+1)/(2\gamma_\delta)}\rmb.
\end{equation*}
   \item[(2)] In the classical regime,  i.e., (2.1) $t^\star_\delta>0$ or (2.2) $D_-(t^\star_\delta)<\delta$ or (2.3) $D_+(t^\star_\delta)>\delta$, we have $\widetilde{I}^\star(\delta)=0$.  
    Thus, with $\gamma' = \gamma_\infty$ for the automatically fair and fair-boundary cases ($t^\star_\delta=0$), and $\gamma'= \gamma_\delta$ for the fairness impacted case ($t^\star_\delta>0$), we have
$\sup_{\P\in\mathcal{P}}\EN\lmb d_E\lsb\widehat{f}_{\delta,n},f^\star_\delta\rsb\rmb\le 
C n^{-(\gamma'+1)\beta/(2\beta+d)}$.
\end{itemize}
This implies that our FairBayes-DDP+ classifier is minimax optimal.

\subsection{Asymptotic Fairness}
For any $\delta\ge 0$,
at the population level, our fairness constraint enforces that 
$|\textup{DDP}(f)|\le \delta.$
However, based on a finite sample, 
in general one may slightly violate the constraint. 
When $\delta=0$, \cite{fukuchi2022minimax} defined 
a learning algorithm with output $\widehat{f}_{\delta,n}$ to be $(\alpha, \xi)$-consistently fair---for $\alpha>0$ and $\xi>0$---for an unfairness measure $U:\mathcal{F}\to\R$, if there
are constants $n_0 \ge 0$ and $C > 0$ independent of $n$ such that $\P(U(\widehat{f}_{\delta,n}) > Cn^{-\alpha}) \le\xi$ for all $n \ge n_0$,
over the randomness from the training data.
We adapt this definition
to fair classification.
\begin{definition}
 A sequence of classifiers $\widehat{f}_{\delta,n}$ depending on a sample of size $n$ is $(\delta,\alpha, \xi)$-consistently fair under demographic parity if
there
are constants $n_0 \ge 0$ and $C > 0$ independent of $n$, such that $\P(|\textup{DDP}(\widehat{f}_{\delta,n})| > \delta+Cn^{-\alpha}) \le\xi$ for all $n \ge n_0$, over the randomness from the training data.
\end{definition}
The following theorem demonstrates that the FairBayes-DDP+ algorithm is consistently fair. 
\begin{theorem}\label{thm:asymptotic fairness}
For any $\delta\ge 0$ and $\xi>0$, there is $C_\xi>0$ such that,  with $\Delta_n\asymp (\log\log n)^{-1}$, $r_n\asymp (\log n)^{-1}$ and offsets satisfying $ C_\xi (\phi_{n,1}\vee \phi_{n,0})<\lo\wedge \lz <r_n$,
the FairBayes-DDP+ estimate   of the $\delta$-fair Bayes-optimal classifier   is $(\delta,1/2,\xi)$-consistently fair. 
 In particular,
  there exist constants $c_{D,i}$, $i\in[6]$ and $L_{\ep}$ such that
 for $L_{\ep} (\lo\vee\lz)^\gamma<\ep\le \sqrt{8(p_1 \wedge p_0)}$,  we have
\begin{align}\label{ddpbd}
\nonumber\PN\lsb\left|{\textup{DDP}}(\widehat{f}_{\delta,n})\right|> \delta+\ep\rsb&\le \psi_{n,1,D}\lsb{\lo\wedge\lz}\rsb +\sum_{j\in\{-,+\}}{I}(\delta=D_j(t_\delta^\star)) \psi_{n,2,D}\lsb g_{\delta,j}(\omega(\lo\wedge\lz,r_n))\rsb\\
&+ c_{5,D}\exp(-c_{6,D}n\ep^2).
\end{align}
\end{theorem}
Theorem~\ref{thm:asymptotic fairness} demonstrates that if $\lo\wedge\lz >C_\xi(\phi_{n,1}\vee\phi_{n,0})$,
then the disparity level of $\widehat{f}_{\delta,n}$ will be no more than the pre-specified level $\delta$, up to a small term of order $n^{-1/2}$. 
This lower bound for offsets is necessary to ensure that the boundary sets and their probability measures are consistently estimated.
Smaller offsets could lead to inconsistent estimators of $(\taudso,\taudsz)$, which would increase the risk of violating the fairness constraint.
Moreover, the level of offsets also determines the tradeoff between fairness and accuracy. Larger offsets lead to slower convergence rates for the measure $d_E$,
but also to a smaller probability of disparity.

\section{Simulation Studies}
\label{sim}

\begin{figure}[t]
    \centering
    \begin{subfigure}[b]{0.48\textwidth}
        \includegraphics[width=\textwidth]{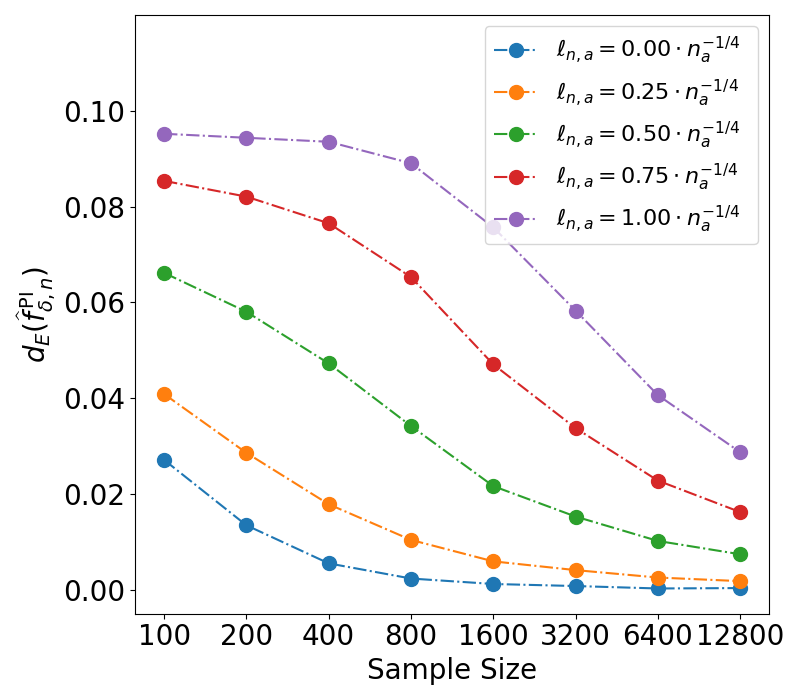}
        \caption{Estimated fairness-aware excess risk}
    \end{subfigure}%
              \begin{subfigure}[b]{0.48\textwidth}
         \centering
         \includegraphics[width=\textwidth]{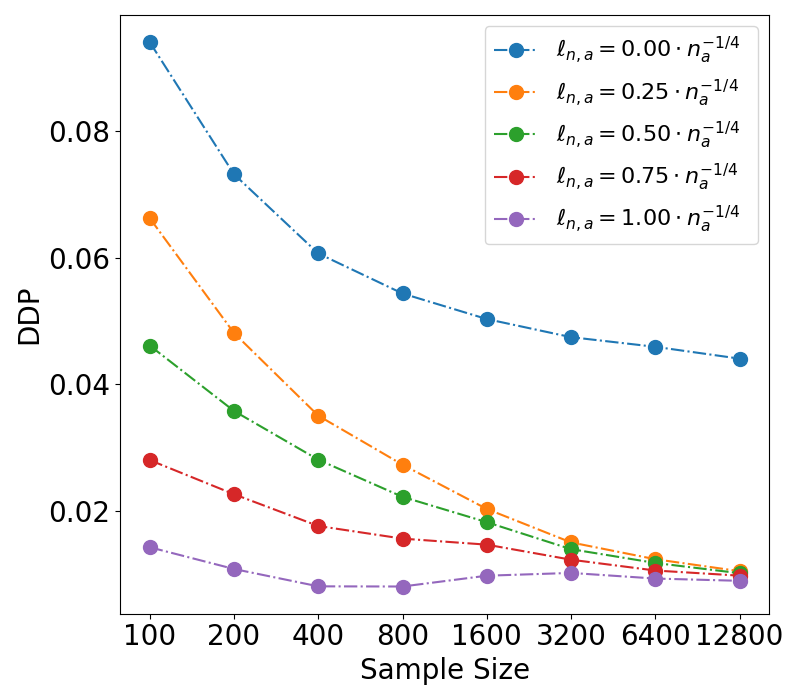}
         \caption{Estimated demographic disparity}
     \end{subfigure}
     \caption{Estimated fairness-aware excess risk and DDP of 
    our FairBayes-DDP+ classifier 
    $\widehat{f}_{\delta,n}$ in the setting from \Cref{sec:simu-syn}, for various sample sizes.} 
    \label{fig:eff_of_sample_size}
\end{figure}

\begin{table}[t]
    \centering
    \caption{Estimated fairness-aware excess risk and DDP of 
    our FairBayes-DDP+ classifier 
    $\widehat{f}_{\delta,n}$ in the setting from \Cref{sec:simu-syn},
    with $\ela = 0.25\cdot n_a^{-1/(2\beta+d)}$ and various sample sizes. }
    \label{tab:eff_of_sample_size}
    \begin{tabular}{c|c c c c c c c c c c c}\hline
Sample size  & 100 & 200 & 400 & 800 & 1600 & 3200 & 6400 & 12800\\\hline
$d_{E}(\widehat{f}_{\delta,n})$  &  0.041  &  0.029  &  0.018  &  0.010  &  0.006  &  0.004  &  0.003  &  0.002  \\
    (SD)                   &  (0.024) &  (0.019) &  (0.012) &  (0.008) &  (0.005) &  (0.005) &  (0.005) &  (0.005) \\\hline
DDP                         &  0.066  &  0.048  &  0.035  &  0.027  &  0.020  &  0.015  &  0.012  &  0.010  \\
(SD)                       &  (0.051) &  (0.036) &  (0.027) &  (0.022) &  (0.016) &  (0.012) &  (0.010) &  (0.008) \\\hline
    \end{tabular}
\end{table}

\subsection{Simulation Studies}\label{sec:simu-syn}
In this section, we conduct simulation studies to illustrate the numerical performance of our method. We consider a data-generating progress with standard components, similar  to e.g.,  \cite{CaiandWei}:
\begin{itemize}[]
    \item(1) Protected attribute: The protected attribute $A$ follows the Bernoulli distribution with parameter $1/2$.
   
    \item(2) Common feature: The common features $X=(X_1,X_2)$ are two-dimensional. For $a=\{0,1\}$, the conditional distribution of $(X_1,X_2)$ given the protected feature $A=a$ follows the uniform distribution on $[-1,1]^2$.
    \item(3) Regression functions: The conditional probability of $Y=1$ given $(X_1,X_2,A)= (x_1,x_2,a)$ is 
    $$\eta_a(x_1,x_2)= \frac{1+ (2a-1)s_1}{2}+\frac{ s_2\cdot \text{sign}(x_1)}{2}\lsb|x_1|(1-|x_2|)\rsb^\beta,$$
    for all $(x_1,x_2)\in [-1,1]^2$ and $a\in \A$. Here  $s_1$, $s_2$ and $\beta$ are hyperparameters that determine the group-wise thresholds, the margin condition, and the smoothness of the regression function. 
    we set $s_2>s_1>0$ so that $s_1+s_2\le 1$. 
\end{itemize}
It is clear that, for $a=\{0,1\}$, $\eta_a\in [0,1]$ is $\beta$-smooth, and it also satisfies the $\gamma_\delta$-margin condition with $\gamma_\delta=1/\beta$  when $\delta=0$ and $\gamma_\delta=1$ otherwise.
As shown in Section \ref{sec:detail_about_syn_dis}, the $\delta$-fair Bayes-optimal classifier takes values
$$f^\star_\delta(x_1,x_2,a) = I\lsb\eta_a(x_1,x_2)>\frac12+(2a-1)t_\delta^\star\rsb$$
for $x_1,x_2,a$, and this choice is unique almost surely with respect to the distribution of the data.
Here $t^\star_\delta$ satisfies $(s_1-s_2)/2\le t^\star_\delta\le s_1/2$ and solves the equation
$$\lsb\frac{s_1-2t_\delta^\star}{s_2}\rsb^{\frac1\beta}\lsb1-\frac1\beta\ln \lsb\frac{s_1-2t_\delta^\star}{s_2}\rsb\rsb =\delta\wedge \lmb \lsb\frac{s_1}{s_2}\rsb^{\frac1\beta}-
 \frac1{\beta}\lsb\frac{s_1}{s_2}\rsb^{\frac1\beta} \ln \lsb \frac{s_1}{s_2}\rsb\rmb.$$
Moreover, 
with  $q_\delta^\star=((s_1-2t_\delta^\star)/{s_2})^{1/\beta}$,
the misclassification rate of $f^\star_\delta$ is given by
\begin{equation*}
R(f^\star_{\delta})=  \frac12  - \frac{ s_1 q_\delta^\star}2 \lsb 1 - \ln\lsb q_\delta^\star\rsb  \rsb -\frac{s_2}{2(\beta+1)}\lsb \frac{1-\lsb q_\delta^\star \rsb^{\beta+1} }{\beta+1} + \lsb q_\delta^\star \rsb^{\beta+1}   \ln \lsb q_\delta^\star \rsb\rsb.
\end{equation*}
In our experiments, we set $s_1= 0.2$, $s_2=0.8$, $\beta=1$
 and generate samples  
of size  $n_{\mathrm{train}} = 2^j\cdot 50$, $j\in[6]$
 from the source distribution. 
 For each sample size, we estimate the regression functions by local polynomial estimators with a Gaussian kernel. 
 Additionally, we vary the bandwidth $h_{n,a}$ from  $0.5 \cdot n_a^{-1/4}$ to $5\cdot n_a^{-1/4}$, where $n_a$ is the sample size associated with group $A=a$, 
 and select the bandwidth that yields the best performance on a validation set of size $n_{\mathrm{val}} = 1000$. 
 For estimating the thresholds, 
 we let $\Delta_n=0.1\cdot (\log \log n)^{-1}$, $r_n = 0.1\cdot (\log n)^{-1}$ and
 consider offsets with levels $\ela = \{0,0.25,0.5,0.75,1\}\cdot n_a^{-1/4}$ to evaluate the effect of offsets. 

 For the resulting 
FairBayesDDP+ classifier
 $\widehat{f}_{\delta,n}$,
 we estimate the fairness-aware excess risk $d_{E}(\widehat{f}_{\delta,n})$
 and disparity $\textup{DDP}(\widehat{f}_{\delta,n})$
 on a test set with size $n_{\mathrm{test}}=1000$. 
 We 
repeat the experiments 1000 times. 
The results with $\delta=0$ are summarized in Figure \ref{fig:eff_of_sample_size} and Table \ref{tab:eff_of_sample_size}. 
As we can see, both the fairness-aware excess risk and disparity converge to zero as the sample size increases, 
lending support to the asymptotic consistency and fairness of our method.
For a given sample size, larger offsets lead to a a slower convergence of the fairness-aware excess risk $d_E$ and a faster convergence of the $\textup{DDP}$, which is consistent with our theoretical results from Theorem \ref{thm:convergence_rate} and Theorem \ref{thm:asymptotic fairness}.

\begin{table}[t]
    \centering
    \caption{Estimated fairness-aware excess risk and DDP of 
    our FairBayes-DDP+ classifier 
    $\widehat{f}_{\delta,n}$ in the setting from \Cref{sec:simu-syn},
   for various pre-specified disparity levels. }
    \label{tab:eff_of_delta}
    \begin{tabular}{c|c c c c c c c c c c}\hline
                  $\delta$  &   0.00   &   0.05   &   0.10   &   0.15   &  0.20  &  0.25   &   0.30   \\\hline
$d_{E}(\widehat{f}_{\delta,n})$  &  0.002  &  0.002  &  0.002   &  0.003  &  0.003  &  0.004 & 0.005  \\
    (SD)                   &  (0.005) &  (0.005) &  (0.005) &  (0.005) &  (0.005) &  (0.005) & (0.005)\\\hline
DDP                         &  0.010  &  0.050  &  0.100  &  0.150  &  0.200  &  0.250  &  0.300  \\
(SD)                       &  (0.008) &  (0.013) &  (0.013) &  (0.013) &  (0.013) &  (0.013)  & (0.014)\\\hline
    \end{tabular}
\end{table}

\begin{figure}[t]
    \centering
    \begin{subfigure}[b]{0.48\textwidth}
        \includegraphics[width=\textwidth]{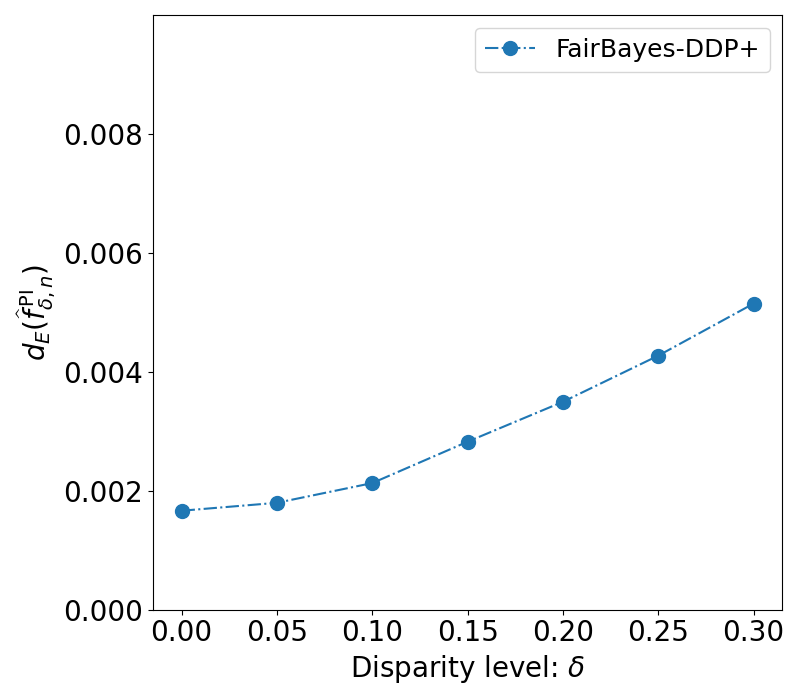}
        \caption{Estimated and population values of the fairness-aware excess risk.}
    \end{subfigure}%
    ~~
              \begin{subfigure}[b]{0.48\textwidth}
         \centering
         \includegraphics[width=\textwidth]{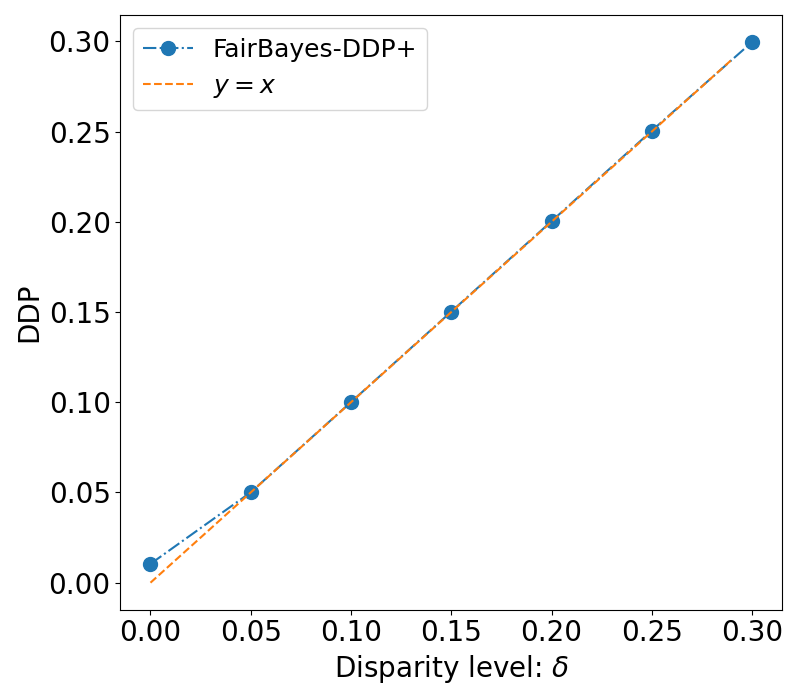}
         \caption{Estimated and population values of the demographic disparity.}
     \end{subfigure}
     \caption{Estimated and population values of the fairness-aware excess risk and DDP of 
    our FairBayes-DDP+ classifier 
    $\widehat{f}_{\delta,n}$ in the setting from \Cref{sec:simu-syn}, for various desired disparity levels.} \label{fig:eff_of_delta}
\end{figure}

Next, we set $n_{\mathrm{train}}=12800$, $\ela=0.25\cdot n_a^{-1/4}$ and consider different pre-specified levels of $\delta$. 
Again, we set $\beta=1$, and the bandwidth is chosen on a grid from $0.5 \cdot n_a^{-1/4}$ to $5\cdot n_a^{-1/4}$ to optimize performance on a validation set.
Figure \ref{fig:eff_of_delta} and Table \ref{tab:eff_of_delta} present the fairness-aware excess risks and disparity levels of our estimator under various pre-specified levels of disparity, based on 1000 simulations.  
As we can see, FairBayes-DDP+ effectively controls the disparity and achieves a vanishing fairness-aware excess risk.

\subsection{Empirical Data Analysis}\label{sec:eda}

To further support our theory and our proposed method, we conduct experiments on the benchmark ``Adult'' 
dataset \citep{misc_adult_2}, 
and compare our method with strong baseline methods.

{\bf Data Description.}  The  ``Adult'' dataset is a commonly considered dataset in fair statistical learning. It contains data on a sample of individuals.
The target variable $y$ measures if the income of an individual is more than \$50,000.  Age, marriage status, education level, and other related variables are included in $x$, and the protected attribute $a$ refers to gender.  To support our asymptotic theory from Section \ref{sec:Asymp_analysis}, 
we select three continuous features---``age", ``year of education", and ``working hours per week"---as predictors; 
these features have the largest empirical marginal correlation with the label. 
We adopt a standard data processing approach as in e.g., \cite{CHS2020}. 
In addition, we split the usual training set into a training
part (70\%) and a validation part (30\%) for model selection.

{\bf Baselines.} We consider several strong baselines proposed recently for fair classification:  (1) Adversarial training (ADV, \cite{ZLM2018}),
(2) KDE-based constrained optimization (KDE, \cite{CHS2020}),
(3) Post-processing through optimal transport (PPOT, \cite{xian2023fair}),
and (4) Post-processing through flipping (PPF, \cite{chen2023post}).

\textbf{Training details.} For our Fair Bayes-DDP+ method, we estimate the regression functions over three features
using local polynomial estimators. 
We use the Gaussian kernel and set the smoothness hyperparameter as $\beta=3$; which influences the choices below. 
We select the bandwidth $h_{n,a}$ 
with the best performance on the validation set,
ranging from $0.5\times n_a^{-1/(2\beta+d)}$ to $5 \times n_a^{-1/(2\beta+d)}$. To estimate the group-wise thresholds, we let $\Delta_n  = 0.1 \cdot (\log \log n)^{-1}$ and $r_n = 0.1\cdot (\log n)^{-1}$.
The offsets are set as $\ela = 0.1\cdot n_a^{-\beta/(2\beta+d)}$, for all $a$.

For other methods, we follow the training settings from \cite{CHS2020}. 
 A three-layer fully connected neural network with $32$ hidden neurons is trained with the Adam optimizer with the default hyperparameters  $(\beta_1,\beta_2)= (0.9,0.999)$. 
 The batch size, training epochs, and learning rate are set to be $512$, $200$ and $0.1$, respectively. 
For adversarial training \citep{ZLM2018}, we further use a two-layer fully connected neural network  with $16$ hidden neurons as the discriminator. In all cases, we train the model on the training set and perform early stopping based on the validation set. 
All  experiments use PyTorch; 
we repeat them 50 times.\footnote{The randomness of the experiments comes from the 
stochasticity of the batch selection in the optimization algorithm.}

\textbf{Simulation Results.} 
We  first evaluate the FairBayes-DDP+ algorithm with various pre-determined levels of disparity. We present the simulation results in Table \ref{tab:real_data_two_dimen}. We observe that FairBayes-DDP+ controls the disparity level at the pre-determined values, as desired.

We then compare FairBayes-DDP+ with baseline methods in Table \ref{tab:real_data_two_dimen1}. We observe that FairBayes-DDP+, PPOT, and PPF demonstrate comparable performance in terms of both accuracy and disparity control. This similarity arises because they are all post-processing methods that aim to estimate the fair Bayes-optimal classifier and are able to control the disparity directly. In contrast, KDE and ADV are in-processing methods where the disparity is controlled implicitly by tuning hyperparameters controlling the training process. Consequently, they exhibit inferior performance in disparity control compared to the post-processing methods.

To  further support FairBayes-DDP+, we compare its fairness-accuracy tradeoff with that of other baseline methods. 
For FairBayes-DDP+, PPOT and PPF, the level of unfairness is directly controlled, ranging from zero to the empirical DDP 
of the unconstrained classifier.
In KDE-based constrained optimization, fairness and accuracy are balanced through a tuning parameter that controls the ratio between the loss and the fairness regularization term. 
We let this tuning parameter  $\lambda$ vary from $0.05$ to $0.95$ to explore a wide range of the tradeoff. 
In adversarial training, the tradeoff is controlled by changing the hyperparameter $\alpha$ that handles the gradient of the discriminator. 
We vary this parameter from zero to five.
We empirically
find that in this range, the performance is representative and suffices for comparison. 
More details about the effects of $\lambda$ and $\alpha$ can be found in \cite{CHS2020} and \cite{ZLM2018}, respectively.

Figure \ref{fig::real_data_two_dimen} presents the empirical fairness-accuracy tradeoff, where each point represents a particular tuning parameter. Our FairBayes-DDP+ algorithm demonstrates the best tradeoff, followed by PPOT and PPF. For a given disparity level, FairBayes-DDP+ achieves the highest accuracy.
The KDE method performs satisfactorily in the high disparity regime; however, it may lose accuracy in the low disparity regime. This loss in accuracy could be attributed to its use of a Huber surrogate loss to handle the non-differentiability of the absolute value function at zero.
Here, adversarial training does not reduce the DDP to near zero, possibly due to the instability of minimax training.

\begin{table}[t]
 \caption{Classification accuracy and DDP on the ``Adult'' dataset}
 \label{tab:real_data_two_dimen1}
 \vskip -0.1in
 \begin{center}
 \begin{small}
 \begin{sc}
    \begin{tabular}{l|l|cc}\hline
 Methods &   Parameters &ACC & DDP\\\hline
 FairBayes-DDP+ (Proposed)     &    $\delta=0$   &    0.791 (0.001)    &   0.008 (0.003) \\
ADV\citep{ZLM2018}  &  $\alpha=5$    &   0.799 (0.004) &   0.055 (0.017)  \\
KDE\citep{CHS2020}     & $\lambda=0.95$  &    0.784 (0.002)   &   0.039 (0.008) \\
 PPOT     \citep{xian2023fair}      &       $\delta=0$           &    0.790  (0.001)  &   0.008 (0.004) \\
 PPF  \citep{chen2023post}          &     $\delta=0$             &    0.790  (0.001) &   0.007 (0.003)\\\hline
 \end{tabular}
    \vspace{-0.1in}
 \end{sc}
\end{small}
 \end{center}
 \vskip -0.1in
 \end{table}

\section{Summary and Discussion}
\label{sec:dis}
In this paper, 
we develop minimax optimal classifiers having a bounded demographic disparity.
Under appropriate smoothness and margin conditions, 
we show that there can be an additional term in the minimax lower bound, caused by the error in estimating the per-class thresholds. 
We also propose the FairBayes-DDP+ method for fair classification, prove its minimax optimality, and illustrate it in simulations and empirical data analysis.
In this work, our theory rests on the low-dimensional optimality of local polynomial methods, however,
empirically the plug-in method works well in higher-dimensional settings by leveraging neural nets (\cite{zeng2024bayesoptimal}). 
Formalizing this rigorously remains an intriguing direction for future work. 

\section*{Acknowledgements}
This work was 
partially supported by ARO W911NF-20-1-0080, ARO W911NF-23-1-0296,
NSF 2031895, NSF DMS 2046874, ONR N00014-21-1-2843, ONR N00014-18-2759, NSF – SCALE MoDL (2134209), a JP Morgan Faculty Award and the Sloan Foundation.

{\small
\setlength{\bibsep}{0.2pt plus 0.3ex}
\bibliographystyle{plainnat-abbrev}
\bibliography{main}
}



\appendix

\section*{Appendix}

\section*{Additional Notation and Definitions}
In this appendix, we use some additional notation.
For a real-valued function $f$ defined on $[a,b)$ for some $a<b$, we denote by $\lim_{x\to a^+} f(x)$ the limit from the right of $f$ at $a$, if it exists. Similarly, if $f$ is defined on $(b,a]$ for $b<a$,  we denote by $\lim_{x\to a^-} f(x)$ the limit from the left of $f$ at $a$, if it exists. 
For an interval $[a,b]$, and scalars $c\in \R$, $d>0$, we denote $c+ d[a,b] =[c+ da,c+ db] $.
For an integer $p\ge 1$, we let $ e_j$, $j\in[p]$ be the $j$-th standard basis vector, with $e_{jj}=1$ and $e_{jk}=0$ for $k\neq j$.
For an integer $p\ge 1$ and $ x=(x_{1},\ldots,x_{d})^\top\in \R^d$, 
denote by $B_{d,p}({x},r)$ the $d$-dimensional $\ell_p$ ball with center $ x$ and radius  $r\ge 0$, i.e.,
$B_{d,p}( x,r) = \{ y=(y_1,\ldots,y_d)^\top: \sum_{j=1}^d|y_j-x_{j}|^p\le r^p\}.$
Moreover, let $V_{d,p}$ be the volume of $B_{d,p} (0,1)$. 
For $q>0$ and $z = (z_1,\ldots,z_d)^\top\in [0,1]^d$, we define $\mathcal{C}_{z,q} = \{x=(x_1,\ldots,x_d)^\top: |x_i-z_i|\le 4q^{-1},i=1,2,\ldots,d\}$ 
as the cube of side length $8/q$ centered at $z$,
and $\mathcal{D}_{z,q} = 
B_{d,2}(z,2q^{-1})\setminus 
B_{d,2}(z,q^{-1})$ as a hyperspherical shell. 
The interior of a set $S$ is denoted by $\inter S$.
For a classifier $f$, we also define $  R(f):= \P\lsb Y\neq \widehat{Y}_f\rsb$.

Without a fairness constraint, a \emph{Bayes-optimal classifier}, which minimizes the misclassification rate, is defined as
$f^\star\in{\text{argmin}}_f\, [\P(Y\neq \widehat{Y}_f)]$.
Bayes-optimal classifiers are the ``best possible" method when fairness is not a concern.
Denoting the indicator function by $I(\cdot)$,
a classical result \citep[see e.g.,][etc]{DevroyeGL96}
is that 
all Bayes-optimal classifiers $f^\star:\X\times \{0,1\}\to[0,1]$ have the form
\begin{equation}\label{prop:ba-op}
f^\star(x,a)=I\left(\eta_a(x)>1/2\right)+\tau_aI\left(\eta_a(x)=1/2\right),
\end{equation}
for all $(x,a)\in \X\times \{0,1\}$, where and $\tau_0,\tau_1\in[0,1]$ are any two constants. 

\section{Fair Bayes-optimal Classifier with a Nonzero Disparity}
\label{ha}

\begin{figure}
    \centering
    \includegraphics[width = 0.9\textwidth]{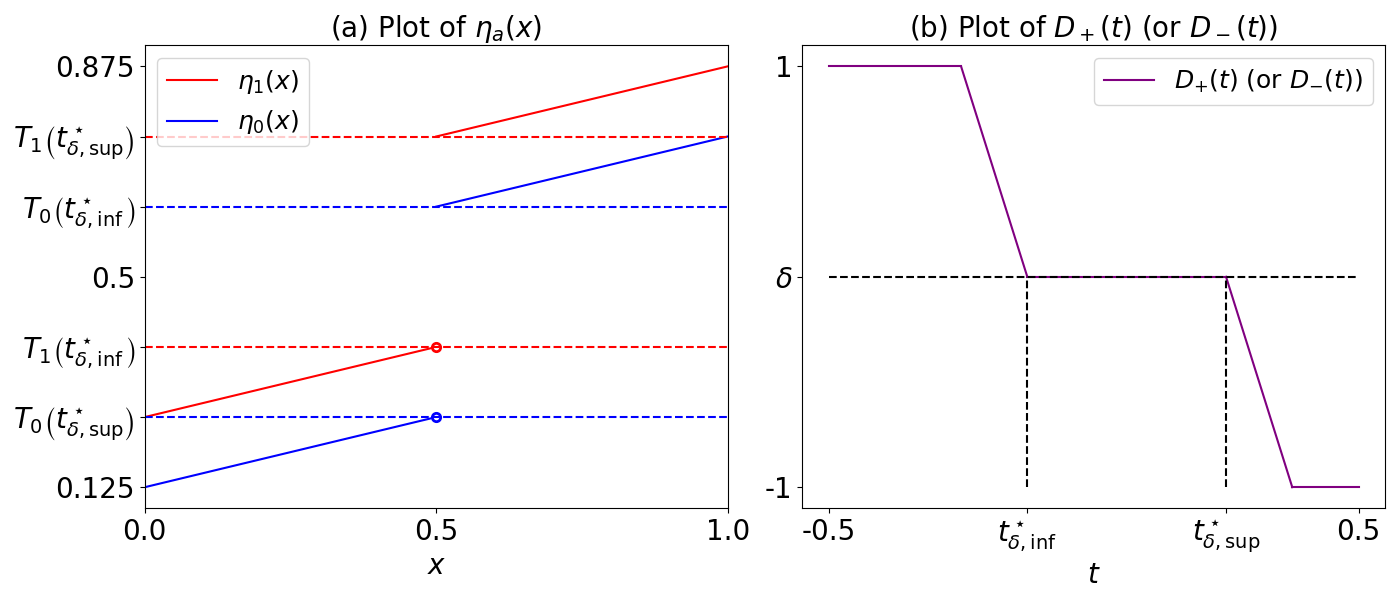}
    \caption{When both $D_+(t)$ and $D_-(t)$ are flat at $\delta$ (or $-\delta$), there exists an interval $[t^\star_{\delta,\inf}, t^\star_{\delta,\sup}]$ within which the conditions $D_+(t) = D_-(t) = \delta$ (or $-\delta$) always hold. In other words, the corresponding classifiers satisfy the hard constraint. Here, we set $\P(A=1)=1/2$, $X|A=a \sim U(0,1)$ and $\delta = 0$.}
    \label{fig:ill_t_sup_inf}
\end{figure}

Here we give the general form of Bayes-optimal classifiers for the case where the group-wise decision thresholds are not unique, following \cite{zeng2024bayesoptimal}.
For any $\delta>0$,
 define
 the following quantities, which can be viewed as  ``inverses" of the functions $D_-, D_+$ in the
 various cases:
\begin{equation}\label{eq:t-star-dp_inf}
  t_{\delta,\inf}^\star= \left\{\begin{array}{ll}
    \inf\left\{t: D_{-} (t)\le \delta\right\} = \inf\left\{t: D_{+} (t)\le \delta\right\}, & 
    \textnormal{in the fairness-impacted case }
    D_{-} (0)>\delta;\\
    0, &  \textnormal{otherwise},\\
  \end{array}  \right.
\end{equation}
\begin{equation}\label{eq:t-star-dp_sup}
  t_{\delta,\sup}^\star= \left\{\begin{array}{ll}
    \inf\left\{t: D_{-} (t)< \delta\right\} = \inf\left\{t: D_{+} (t) <\delta\right\}, &\textnormal{in the fairness-impacted case } D_{-} (0)>\delta;\\
    0, &  \textnormal{otherwise}.\\
  \end{array}  \right.
\end{equation}
We need both $ t_{\delta,\inf}^\star, t_{\delta,\sup}^\star$ to account for the case where $D_-,D_+$ are  ``flat" at $\delta$ or $-\delta$, so that they can take any value over a nonempty interval,  see Figure \ref{fig:ill_t_sup_inf} for an illustration.
\begin{proposition}[Fair Bayes-optimal classifiers]\label{thm-opt-dp}
For any $\delta\ge 0$, all 
\emph{$\delta$-fair Bayes-optimal classifiers} $f^\star_{\delta}$ have the following form:   for any $T_{\delta,1} \in [1/2+t^\star_{\delta,\inf}/(2p_1),1/2+t^\star_{\delta,\sup}/(2p_1)]$ and
$T_{\delta,0} \in [1/2-t^\star_{\delta,\sup}/(2p_0),1/2-t^\star_{\delta,\inf}/(2p_0)]$,
there are $(\tau_{\delta,1},\tau_{\delta,0})\in[0,1]^2$,
such that  
for all $x,a$,
\begin{equation}\label{eq:op-de-rule-dp}
 f^\star_{\delta}(x,a)=
I\left(\eta_{a}(x)>T_{\delta,a}\right)+ \tau_{\delta,a} I\left(\eta_a(x)=T_{\delta,a}\right).
\end{equation}
Further, 
$(T_{\delta,1},T_{\delta,0})$ and
$(\tau_{\delta,1},\tau_{\delta,0})$
are determined by the following constraints: 
\begin{itemize}[]
 \item (1).
 When $D_{-} (0)\le\delta$,
       \begin{equation}\label{eq:soft_constraint}
          \left| \Po\lsb\widehat{Y}_{f_\delta^\star}=1\rsb-\Pz\lsb\widehat{Y}_{f_\delta^\star}=1\rsb\right|\le \delta.
       \end{equation}
    \item (2). 
    When $D_{-} (0)>\delta$,
      \begin{equation}\label{eq:hard_constraint1}
            \Po\lsb\widehat{Y}_{f_\delta^\star}=1\rsb-\Pz\lsb\widehat{Y}_{f_\delta^\star}=1\rsb=\delta.
       \end{equation}

\end{itemize}
The form of $(\tau_{\delta,1},\tau_{\delta,0})$ is provided below.  
For scalars $a,b,c,d$ with $b\le c$ and $d>0$,
let $\rho((a+[b,c])/d)$ represent $[\rho((a+b/d)),\rho((a+c)/d)]$ with $\rho$ from \eqref{rho}, let $\widetilde{\delta}$ and $\widetilde{\delta}$ be defined in \eqref{dt}, 
and let, for $a\in\{0,1\}$, $\pisap$ and $\pisae$ be defined in \eqref{eq:pistar}.
We have the following four cases:
\begin{itemize}
\item Case (1). When $\pisoe=\pisze=0$,  
$(\tau_{\delta,1},\tau_{\delta,0})\in[0,1]^2$ can be arbitrary.

\item Case (2). When $\pisoe>\pisze=0$, 
$\tau_{\delta,0}\in[0,1]$ can be arbitrary, and 
we can take
$$\tau_{\delta,1}\in  \rho\lsb \frac{ \piszp-\pisop +\lmb  \widetilde{\delta},\widetilde{\delta} \rmb}{ \pisoe}\rsb.$$

\item Case (3). Similarly, when $\pisze>\pisoe=0$, 
$\tau_{\delta,1}\in[0,1]$ can be  arbitrary,  and we can take
$$\tau_{\delta,0}\in  \rho\lsb \frac{ \pisop-\piszp +\lmb -\widetilde{\delta},-\widetilde{\delta}\rmb }{ \Pz\lsb\eta_0(X)=T_{\delta,0}\rsb}\rsb  .$$
\item Case (4). Finally, when $\pisop>0$ and $\piszp>0$,
 we can take
\begin{equation*} 
\tau_{\delta,1} \in \rho\left( \frac{\piszp-\pisop +\lmb \widetilde{\delta},\widetilde{\delta}+\pisze\rmb}{ \pisoe}\rsb,\,\,
\text{ and }\,\,
\tau_{\delta,0} \in \rho\left(
\frac{ \pisop + \tau_{\delta,1}\pisoe -\piszp+\left[-\widetilde{\delta},-\widetilde{\delta} \right]
}{ \pisze}\right). 
 \end{equation*}
\end{itemize}

\begin{Remark}
When $\eta_1(X)$ and $\eta_0(X)$ have density functions on $[0,1]$, we have for $a\in\{0,1\}$, $\Pa(\eta_a(X)=1/2+(2a-1)t_\delta^\star/(2p_a))=0$ and the optimal classifier is deterministic.
With $t_\delta^\star\in[ t_{\delta,\inf}^\star, t_{\delta,\sup}^\star]$, and for all $x,a$, it takes values
\begin{equation}\label{eq:op-de-rule-dp_continuous}
 f^\star_\delta(x,a) = I\lsb\eta_a(x)>\frac12+\frac{(2a-1)t^\star_\delta}{2p_a}\rsb.
\end{equation}
\end{Remark}

\end{proposition}

\section*{Proofs}

In Sections \ref{sec:lemmas} to \ref{sec:pf-seceight}, we present the proofs of our theoretical results from the main text. 
We first introduce several technical lemmas that are essential for proving our theoretical results (Section \ref{sec:lemmas}), and defer their proofs to Section \ref{sec:tec_lems}.

\section{Additional Lemmas}\label{sec:lemmas}

\begin{lemma}\label{lem:interd1}
For any $ z=(z_{1},\ldots,z_{d})^\top\in \R^d$, $R>0$ and $  x=(x_{1},\ldots,x_{d})^\top\in B_{d,1}( z,R)$, we have, 
 for $0\le r\le 2R/(d+2)$,
$$\lambda[ B_{d,1}( z,R)\cap B_{d,2}( x,r)] \ge \frac{V_{d,1}}{V_{d,2}2^d}\cdot\lambda[ B_{d,2}( x,r)].$$
\end{lemma}

\begin{lemma}\label{lem:interd2}
For any $ z=(z_{1},\ldots,z_{d})^\top\in \R^d$, $R>0$ and $  x=(x_{1},\ldots,x_{d})^\top\in B_{d,2}(z,R)$, we have, for $0\le r\le R$,
\begin{equation*}
\lambda[ B_{d,2}( z,R)\cap B_{d,2}( x,r)] \ge \frac{3^{\frac{d+1}{2}}V_{d-1,2}}{2^{d+1}(d+1)V_{d,2}}\cdot\lambda [B_{d,2}( x,r)].
\end{equation*}

\end{lemma}

\begin{lemma}\label{lem:inter3}
Let $z$ be  a point such that $\mathcal{C}_{z,q}\subset [0,1]^d$.
Then, 
for any $r>0$,
there is $0<C_r\le 1$ also depending on $d$,
such that, 
for any $x=(x_1,\ldots,x_d)^\top\in [0,1]^d\setminus \mathcal{D}_{z,q}$, 
$$\lambda[B_{d,2}(x,r)\cap \lsb\mathcal{C}_{z,q}\setminus \mathcal{D}_{z,q}\rsb]\ge C_r\lambda[B_{d,2}(x,r)\cap \mathcal{C}_{z,q}].$$

\end{lemma}

\begin{lemma}\label{lem:misclassification}
For any classifier $f: \X\times \A\to[0,1]$, 
we have 
$$R(f)=\sum_{a\in\{0,1\}}p_a\int  \left[\lsb 1-2\eta_a(x)\rsb f(x,a) +\eta_a(x)\right] d\Pa(x),$$
and
\beq\label{ddpf} \textup{DDP}(f)=\sum_{a\in\{0,1\}}\int (2a-1)f(x,a)d\Pa(x).
\eeq
\end{lemma}

\begin{lemma}\label{lem:tail}
Let $\mathcal{S}_n=\mathcal{S}_{n,1} \cup \mathcal{S}_{n,0}$ with  $\mathcal{S}_n=\{(x_{i},a_i,y_{i})\}_{i=1}^{n}$ and, for $a\in\{0,1\}$, $S_{n,a}=\lbb (\xaj,a,y_{a,j})\rbb_{j=1}^{n_{a}}$ being an i.i.d.~sample. We have, for $a\in\{0,1\}$,
\begin{equation}\label{eq:nabound1}
    \PN\lsb\left|\frac{n_{a}}{n}-p_a\right|\ge \ep\rsb\le   2\exp\lsb-2n\ep^2\rsb.
\end{equation}
Moreover,  if $p_1 \wedge p_0>0$, we have, for $a\in\{0,1\}$ and $\ep \le  1/p_a$,
\begin{equation}\label{eq:nabound2}
    \PN\lsb\left|\frac{n}{n_{a}}-\frac1{p_a}\right|\ge \ep\rsb\le 2\exp\lsb-\frac{np_a^4\ep^2}{2}\rsb.
\end{equation}
\end{lemma}

\begin{lemma}\label{lem:uniformDDPbound}
For $a\in\{0,1\}$, we have that
 if $
\ep\le \sqrt{p_a/2}$, 
$$\PN\lsb\sup_{T\in\R}\left|\frac{1}{n_{a}}\sum_{j=1}^{n_{a}} I\lsb \eta_a\lsb\xaj\rsb> T\rsb- \Pa\lsb \eta_a(X)>T\rsb\right|>\ep \rsb\le 4\exp\lsb - {np_a\ep^2}\rsb.$$
\end{lemma}

\begin{lemma}\label{lem:diffbetweenDNandDR}
Let $D_{-}$ and $D_{+}$ be defined as in \eqref{eq:Disfun1} and \eqref{eq:Disfun2},  respectively. 
For $t \in \R$, 
we denote
\begin{equation*}
\begin{array}{l}
     D_{n,-}(t) = \frac{1}{n_1}\sum_{j=1}^{n_1} I\lsb \eta_1\lsb\xoj\rsb>\frac12+\frac{t}{2p_1}\rsb -\frac{1}{n_0}\sum_{j=1}^{n_0} I\lsb \eta_0\lsb\xzj\rsb\ge \frac12-\frac{t}{2p_0}\rsb,\\ 
      D_{n,+}(t) = \frac{1}{n_1}\sum_{j=1}^{n_1} I\lsb \eta_1\lsb\xoj\rsb\ge\frac12+\frac{t}{2p_1}\rsb -\frac{1}{n_0}\sum_{j=1}^{n_0} I\lsb \eta_0\lsb\xzj\rsb>\frac12-\frac{t}{2p_0}\rsb.
\end{array}
\end{equation*} 
Then, for $\ep \le \sqrt{(p_1 \wedge p_0) /2}$,
\begin{equation}\label{eq:diffdrc}
\max\left\{\PN \lsb\left| D_{n,+}(t)-D_{+}(t) \right|>\ep \rsb,
\PN \lsb\left| D_{n,-}(t)-D_{-}(t) \right|>\ep \rsb\right\}
\le 8\exp\lsb - \frac{n(p_1\wedge p_0)\ep^2}{4}\rsb.
\end{equation}
\end{lemma}

\begin{lemma}\label{lem:merge}
For a set $\{\iota_0,\iota_1,...,\iota_K\}\subset \R$ and $K>0$,
and for $\iota\in\{\iota_0,\iota_1,...,\iota_K\}$, let $\psi_{n,1,\iota}$ and $\psi_{n,2,\iota}$ be defined for all $\ep>0$ as in \eqref{psi},  with $c_{i,\iota}>0, i\in[4]$:
\begin{equation*}
\psi_{n,1,\iota}(\ep)=c_{1,\iota} \exp\lsb-c_{2,\iota} \lsb\ep/[\phi_{n,1}\vee\phi_{n,0}]\rsb^2 \rsb\text{ and }\psi_{n,2,\iota}(\ep) = c_{3,\iota}\exp\lsb-c_{4,\iota}n \ep^2\rsb.
\end{equation*}
Under the margin condition \eqref{ma2}, 
for any  $C_k>0, k\in[K]$, 
there exists $U_\ep>0$ such that, for $j\in\{-,+\}$ and $\ep<U_\ep$, with
$c_{1,\iota_0}=\sum_{k=1}^Kc_{1,\iota_k}$,
$c_{2,\iota_0}=\min_{k\in[K]} \lsb c_{2,\iota_k}C_k^2\rsb$, $c_{3,\iota_0}= \sum_{k=1}^K c_{3,\iota_k}$ and $c_{4,\iota_0}=
 \lsb \min_{k\in[K]} c_{4,\iota_k}\cdot U_\gamma^{-2}C_k^\gamma\rsb \wedge 1$, we have for all $\ep>0$ that
\begin{equation}\label{eq:summation1}
  \sum_{k=1}^K\psi_{n,1,\iota_k}(C_k\ep)  \le \psi_{n,1,\iota_0}(\ep),
\end{equation}
and
\begin{equation}\label{eq:summation2}
 I\lsb \delta=D_j(t^\star_\delta)\rsb \lsb \sum_{k=1}^K \psi_{n,2,\iota_k}\lsb g_{\delta,j}(\omega(C_k\ep,r_n))\rsb   \rsb\le  I\lsb \delta=D_j(t^\star_\delta)\rsb \psi_{n,2,\iota_0}\lsb g_{\delta,j}(\omega(\ep,r_n))\rsb.
\end{equation}
\end{lemma}

\begin{lemma}\label{lem:impacted}
Under the conditions of Theorem \ref{thm:estimate_of_t}, there are constants $L_{t_1}$, $U_{t_1}$ and $c_{i,t_1}$, $i\in[4]$ determining functions $\psi_{n,i,t_1}$ in \eqref{psi}, such that, 
for $\Delta_n\ge 0$ and $L_{t_1}(\phi_{n,1}\vee\phi_{n,0})< \ep< U_{t_1}$,
with  
$t_\delta^\star$ from \eqref{eq:t-star-dp}, 
\begin{equation}\label{D1}
 \PN\lsb\widehat{D}_{n}\lsb t_\delta^\star+\ep,0,0\rsb\ge \delta+\Delta_n\rsb\\
\le \psi_{n,1,t_1}(\ep) + I\left(\delta  = D_{-}(t_\delta^\star)\right) \psi_{n,2,t_1}\lsb \Delta_n+g_{\delta,-}\lsb\frac\ep2\rsb\rsb,
\end{equation}
and
\begin{equation}\label{D2}
 \PN\lsb\widehat{D}_{n}\lsb t_\delta^\star-\ep,0,0\rsb\le{\delta}-\Delta_n\rsb\\
\le \psi_{n,1,t_1}(\ep) + I\left(\delta = D_{+}(t_\delta^\star)\right) \psi_{n,2,t_1}\lsb \Delta_n+ g_{\delta,+}\lsb\frac\ep2\rsb\rsb.
\end{equation}
\end{lemma}
\begin{lemma}\label{lem:impacted1}
Under the conditions of Theorem \ref{thm:estimate_of_t}, 
there are constants $L_{t_2}$, $U_{t_2}$ and $c_{i,t_2}$, $i\in[4]$ determining functions $\psi_{n,i,t_2}$ in \eqref{psi}, such that, 
for  $L_{t_2}(\phi_{n,1}\vee\phi_{n,0})< \ep< U_{t_2}$,
\begin{itemize}
\item[(1)] if
$D_-(t^\star_\delta)=\delta$, for $g_{\delta,-}(2\ep)<\Delta_n\le g_{\delta,-}(2\ep)+\sqrt{(p_1\wedge p_0)/2}$,
\begin{equation}\label{D3}
 \PN\lsb\widehat{D}_{n}\lsb t_\delta^\star+\ep,0,0\rsb\le {\delta}-\Delta_n\rsb
\le\psi_{n,1,t_2}(\ep) + \psi_{n,2,t_2}\lsb \Delta_n -g_{\delta,-}\lsb2\ep\rsb\rsb;
\end{equation}
\item[(2)] if
$D_-(t^\star_\delta)<\delta$, for  $ \Delta_n<(\delta-D_-(t^\star_\delta))/2$,
\begin{equation}\label{D4}
 \PN\lsb\widehat{D}_{n}\lsb t_\delta^\star+\ep,0,0\rsb\ge {\delta}-\Delta_n\rsb
\le\psi_{n,1,t_2}(\ep);
\end{equation}
\item[(3)] if
$D_+(t^\star_\delta)=\delta$, for  $g_{\delta,+}(2\ep)<\Delta_n\le g_{\delta,+}(2\ep)+\sqrt{(p_1\wedge p_0)/2}$,
\begin{equation}\label{D5}
 \PN\lsb\widehat{D}_{n}\lsb t_\delta^\star-\ep,0,0\rsb\ge {\delta}+\Delta_n\rsb
\le\psi_{n,1,t_2}(\ep) + \psi_{n,2,t_2}\lsb \Delta_n -g_{\delta,+}\lsb2\ep\rsb\rsb;
\end{equation}
\item[(4)] if
$D_+(t^\star_\delta)<\delta$, for   $ \Delta_n<(D_+(t^\star_\delta)-\delta)/2$,
\begin{equation}\label{D6}
 \PN\lsb\widehat{D}_{n}\lsb t_\delta^\star-\ep,0,0\rsb\le {\delta}+\Delta_n\rsb
\le\psi_{n,1,t_2}(\ep).
\end{equation}
\end{itemize}

\end{lemma}

\begin{lemma}\label{lem:impacted2}
Under the conditions of Theorem \ref{thm:estimate_of_t}, 
there are constants $L_{r}$, $U_{r}$, $U_{\Delta,r}$ and  $c_{i,r}$, $i\in[4]$ determining functions $\psi_{n,i,r}$ in \eqref{psi} , such that, for $L_r(\phi_{n,1}\vee \phi_{n,0})<r_n<U_r$, and $2(g_{\delta,-}(4r_n)\vee g_{\delta,+}(4r_n))<\Delta_n<U_{\Delta,r}$,
\begin{itemize}
\item[(1)] if $t^\star_\delta=0$, 
\begin{equation}\label{Db1}
    \PN\lsb \htmid-\htmin>r_n \rsb  \le  \psi_{n,1,r}(r_n)+I(\delta =D_-(t^\star_\delta))\psi_{n,2,r}\lsb g_{\delta,-}\lsb\frac{r_n}{2}\rsb\rsb;
    \end{equation}
\item[(2)] if $t_\delta^\star>0$ and $D_+(t^\star_\delta)>\delta$,  
\begin{equation}\label{Db2}
    \PN\lsb \htmid-\htmin>r_n \rsb  \le  \psi_{n,1,r}\lsb\frac{r_n}{2}\rsb+I(\delta =D_-(t^\star_\delta))\psi_{n,2,r}\lsb g_{\delta,-}\lsb\frac{r_n}{4}\rsb\rsb;
    \end{equation}
\item[(3)] if $t_\delta^\star>0$ and $D_+(t^\star_\delta)=\delta>D_-(t^\star_\delta)$,  
\begin{equation}\label{Db3}
    \PN\lsb \htmid-\htmin\le r_n \rsb\le  \psi_{n,1,r}(r_n) + \psi_{n,2,r}\lsb g_{\delta,+}\lsb\frac{r_n}2\rsb\rsb,
    \end{equation}
    and
\begin{equation}\label{Db4}
    \PN\lsb \htmax-\htmid>r_n \rsb  \le  \psi_{n,1,r}\lsb\frac{r_n}{2}\rsb+\psi_{n,2,r}\lsb g_{\delta,+}\lsb\frac{r_n}{4}\rsb\rsb;
    \end{equation}
\item[(4)] if $t_\delta^\star>0$ and $D_-(t^\star_\delta)=D_+(t^\star_\delta)=\delta$,  
\begin{equation}\label{Db5}
    \PN\lsb \htmid-\htmin\le  r_n \rsb\le  \psi_{n,1,r}(r_n) + \psi_{n,2,r}\lsb  g_{\delta,+}\lsb\frac{r_n}2\rsb\rsb,
    \end{equation}
    and
\begin{equation}\label{Db6}
    \PN\lsb \htmid-\htmin\le r_n \rsb\le  \psi_{n,1,r}(r_n) + \psi_{n,2,r}\lsb g_{\delta,-}\lsb\frac{r_n}2\rsb\rsb.
    \end{equation}
\end{itemize}

\end{lemma}

\begin{lemma}\label{lem:expectationbound}
Under the conditions of Theorem \ref{thm:estimate_of_t}, 
let $\eta_1$ and $\eta_0$ satisfy the $\gamma$-exponent condition in the upper bound from Definition \ref{m} at 
level $\Tso$ with respect to $\Po$
and at level $\Tsz$ with respect to $\Pz$, respectively.
Then, for $a\in\{0,1\}$, the plug-in estimator with offset $\lo,\lz>0$, $r_n\asymp (\log\log n)^{-1}$ and $\Delta_n\asymp (\log n)^{-1}$ satisfies,
for some positive constant $C$,
\begin{align*}
\nonumber&\EN \int\, I{\{\eta_a(x)>\Tsda,\, \widehat\eta_a(x)\le \Thda+\ela
\}}\left|\eta_a(x)-T^\star_{\delta, a}\right|d\Pa(x)\\
&\qquad\qquad\qquad\le    C\lsb\lsb\phi_{n,1}\vee\phi_{n,0}\vee\ell_{n,a}\rsb^{\gamma+1} + I\left(0<D_{-}\lsb t_\delta^\star\rsb=\delta =D_{+}\lsb t_\delta^\star\rsb\right)\cdot n^{-\frac{\gamma+1}{2\gamma}}\rsb.
\end{align*}
An analogous bound holds for $\EN \int\, I{\{\eta_a(x)<\Tsda,\, \widehat\eta_a(x)\ge \Thda-\ela\}
}\left|\eta_a(x)-T^\star_{\delta, a}\right|d\Pa(x)$.
\end{lemma}

\begin{lemma}\label{lem:deltaneq}
    Under the condition of Theorem~\ref{thm:estimate_of_t}, there are constants $U_{\delta}$, $L_{\delta}$, $U_{\Delta,\delta}$  $c_{i,\delta}$, $i\in [4]$ determining functions $\psi_{n,i,\delta}$ in \eqref{psi} 
such that, for $L_{\delta}(\phi_{n,1}\vee\phi_{n,0})<r_n<U_{\delta}$ and $0<\Delta_n< U_{\Delta,\delta}$,
with $\widetilde{\delta}$ from \eqref{dt} and $\widehat{\delta}$ from \eqref{hdel}, 
    \begin{equation}\label{eq:deltares}
\PN\lsb \widehat{\delta}\neq \widetilde{\delta}\rsb\le \psi_{n,1,\delta}(r_n)+\psi_{n,2,\delta}(\Delta_n).
\end{equation}
\end{lemma}

In the following lemmas, we denote, for $a\in\{0,1\}$,
\begin{equation}\label{pihata}
  \pihap = \Pa\left(\widehat\eta_a(X)>\Thda+\ela\right);   \qquad\pihae= \Pa\left(|\widehat\eta_a(X)-\Thda|\le\ela\right).
\end{equation}

\begin{lemma}\label{lem:piset1}
 There is $L_{\pi_1}$, $U_{\pi_1}$, $U_{\Delta,\pi_1}$ and $c_{i,\pi}$, $i\in[4]$ determining functions $\psi_{n,i,\pi_1}$, $i\in\{1,2\}$ in \eqref{psi}, 
 such that, for
$L_{\pi_1}(\phi_{n,1}\vee \phi_{n,0})<r_n<U_{\pi_1}$, $2\lsb g_{\delta,-}(4r_n)\vee g_{\delta,+}(4r_n)\rsb<\Delta_n<U_{\Delta,\pi_1}$, $\ep>4U_\gamma(4(p_1\lo\vee p_0\lz))^\gamma$
 and $ 2L_{\pi_1}(\phi_{n,1}\vee \phi_{n,0})< \lo,\lz< 2r_n$,    
 with $\omega(\ep,r_n)$ from \eqref{xi}, we have
\begin{align}\label{pi11}
&\max\left\{
\PN(\pihap > \pisap +\ep),\,
\PN(\pihap < \pisap-\ep),\,
\PN(\pihae >\pisae+\ep),\,
\PN(\pihae < \pisae - \ep)
\right\}\\
&\qquad\le \psi_{n,1,\pi}\lsb\ela\rsb + \sum_{j=\{-,+\}}I(\delta=D_j(t^\star_\delta))
\psi_{n,2,\pi}\lsb g_{\delta,j}(\omega(\ela,r_n))\rsb.\nonumber
\end{align}
\end{lemma}

\begin{lemma}\label{lem:piset2}
With the same  $L_{\pi_1}$, $U_{\pi_1}$, $U_{\Delta,\pi_1}$ and $c_{i,\pi}$, $i\in[4]$ as in Lemma \ref{lem:piset1}, we have, for
$L_T(\phi_{n,1}\vee \phi_{n,0})<r_n<U_T$, $2\lsb g_{\delta,-}(4r_n)\vee g_{\delta,+}(4r_n)\rsb<\Delta_n<U_\Delta$, $4U_\gamma(4(p_1\lo\vee p_0\lz))^\gamma<\ep \le \sqrt{(p_1 \wedge p_0) /2}$
 and $ 2(L_\eta \vee L_T)(\phi_{n,1}\vee \phi_{n,0})< \lo,\lz< 2r_n$,   
with $\omega(\ep,r_n)$ from \eqref{xi}, $\pihnap,\pihnae$ from \eqref{hpin}
and $\pisap,\pisae$ from \eqref{eq:pistar},
\begin{align}\label{pi21}
&\max\left\{
\PN(\pihnap > \pisap +\ep),\,
\PN(\pihnap < \pisap-\ep),\,
\PN(\pihnae >\pisae+\ep),\,
\PN(\pihnae < \pisae - \ep)
\right\}\nonumber\\
&\qquad\le 
 \psi_{n,1,\pi}\lsb \ela\rsb +\sum_{j=\{-,+\}}I(\delta=D_j(t^\star_\delta))
\psi_{n,2,\pi}\lsb g_{\delta,j}(\omega(\ela,r_n))\rsb+4\exp\lsb-
\frac{np_a\ep^2}4\rsb.
\end{align}
\end{lemma}

\begin{lemma}\label{propRfunc}
Let $b>0$. For $0<\ep<b/2$, we have, with $\rho$ from \eqref{rho},
\beq\label{ri}
\rho\lsb\frac{a+2\ep}{b-\ep}\rsb -\rho\lsb\frac{a}{b}\rsb \le\frac{6\ep}{b}\text{ and } \ \ \rho\lsb\frac{a-2\ep}{b+\ep}\rsb -\rho\lsb\frac{a}{b}\rsb \ge\frac{-6\ep}{b} .
\eeq
\end{lemma}

\begin{lemma}\label{lem:pirbound}
 There exist constants $L_{\pi}$, $U_{\pi}$, $L_{\ep,\pi}$, $U_{\Delta,\pi}$, $c_{5,\pi}$,  $c_{6,\pi}$ and  with the same  $c_{i,\pi}$, $i\in[4]$ as in Lemma \ref{lem:piset1}
 such that, for $a\in\{0,1\}$,
 $L_\pi(\phi_{n,1}\vee \phi_{n,0})<r_n<U_\pi$, $2\lsb g_{\delta,-}(4r_n)\vee g_{\delta,+}(4r_n)\rsb<\Delta_n<U_{\Delta,\pi}$, $L_{\ep,\pi}(\lo\vee\lz)^\gamma<\ep \le \sqrt{(p_1 \wedge p_0) /2}$ and $ (L_\eta \vee L_T)(\phi_{n,1}\vee \phi_{n,0})< \ela/2 < r_n$,  
with 
$\pihap,\pihae$ from \eqref{pihata},
$\taudha$ from  \eqref{htau},
$\pisap,\pisae$ from \eqref{eq:pistar}  and $\taudsa$ from \eqref{eq:Taustar},
\begin{align}\label{eq:pibounds}
\nonumber&\max\left\{
\PN\lsb\pihae \taudha> \pisae  \taudsa+\ep\rsb,\,
\PN\lsb\pihae \taudha < \pisae   \taudsa-\ep\rsb
\right\}
\\
&\qquad\qquad\le4\psi_{n,1,\pi}\lsb \ela\rsb+4 \sum_{j=\{-,+\}}I(\delta=D_j(t^\star_\delta))
\psi_{n,2,\pi}\lsb g_{\delta,j}(\omega(\ela,r_n))\rsb+c_{5,\pi}\exp\lsb-c_{6,\pi}n\ep^2\rsb.
\end{align}    
\end{lemma}

The following proposition from \cite{tsy2007} demonstrates the point-wise convergence of the local polynomial estimator.
\begin{proposition}\label{prop:convergence_of_eta}
 Let $\mathcal{P}$ be a class of probability distributions for $(X,Y)$, such
that the regression function $\eta(x)=\P(Y=1\mid X=x)$ belongs to the H\"older class $\Sigma(\beta,L_\beta,\R^d)$ 
and the
marginal law of $X$ satisfies the strong density condition. 
Let $(X_i,Y_i)_{i=1}^n$ an i.i.d.~sample from $\P$,
and $\widehat\eta$ be the local polynomial estimator with kernel $K$ satisfying 
\eqref{Kernel} and $h_n\asymp n^{
-1/{(2\beta+d)}}$. 
Then there exist
constants $C_1,C_2>0$  such
that for any $\delta > 0$, $n \ge 1$ we have
$$\sup_{P \in\mathcal{P}} \PN\lsb\left|\widehat\eta(x) -\eta(x)\right| > \ep\rsb \le C_1 \exp\lsb-C_2 n^{\frac{2\beta}{2\beta+d}}\ep^2\rsb,$$
for almost all $x$ with respect to $\P_X$. The constants $C_1$, $C_2$  depend only on $\beta$, $d$,
$L$, $c_0$, $r_0$, $\mu_{\min}$, $\mu_{\max}$, and on the kernel $K$.
\end{proposition}
Proposition \ref{prop:convergence_of_eta} shows that the local polynomial estimators $(\widehat\eta_1,\widehat\eta_0)$ are
$(\phi_{n,1},\phi_{n,0})_{n\ge 1}$-pointwise convergent with $\phi_{n,1}=\phi_{n,0} = n^{-\beta/(2\beta+d)}$.

\section{Proofs of Results in  Section \ref{sec:prepare}}
\subsection{Proof of Proposition \ref{measure2}}
By \eqref{eq:t-star-dp} and \eqref{eq:hard_constraint1}, we have $t^\star_\delta=0$ when $\delta\ge  D_{-} (0)$ and
\begin{equation*}
t_\delta^\star\lmb\Po\lsb \widehat{Y}_{f_\delta^\star}=1\rsb- \Pz\lsb \widehat{Y}_{f^\star_\delta}=1\rsb\rmb=
\left\{\begin{array}{lcr}
    0, &  D_{-} (0)\le\delta;\\
    t_\delta^\star\delta, &  D_{-} (0)>\delta.
\end{array}\right.
\end{equation*}
Using Lemma \ref{lem:misclassification}, it follows that
\begin{align*}
&d_{E}\lsb f,f_\delta^\star\rsb
=2\sum_{a\in\A}p_a\int_\X \lsb \frac12 +\frac{(2a-1)t_\delta^\star}{2p_a}-\eta_a(x)\rsb\lsb f(x,a)-f^\star_\delta(x,a)\rsb d\Pa(x)\\
&= d_R(f,f^\star_\delta)+2\sum_{a\in\A}p_a\lsb \frac{(2a-1)t_\delta^\star}{2p_a}\rsb\lmb\int \lsb{f}(x,a)-{f}_\delta^\star(x,a)\rsb d\Pa(x)\rmb\\
&= d_R(f,f^\star_\delta)+\sum_{a\in\A}(2a-1)t_\delta^\star\lmb \Pa\lsb \widehat{Y}_f=1\rsb- \Pa\lsb \widehat{Y}_{f^\star_\delta}=1\rsb\rmb\\
&= d_R(f,f^\star_\delta)+t_\delta^\star\lmb\Po\lsb \widehat{Y}_f=1\rsb- \Po\lsb \widehat{Y}_{f^\star_\delta}=1\rsb  -\Pz\lsb \widehat{Y}_{f}=1\rsb + \Pz\lsb \widehat{Y}_{f_\delta^\star}=1\rsb\rmb\\
&= \left\{\begin{array}{lcl}
 d_R(f,f^\star_\delta),    &  D_{-} (0)\le\delta;\\
 d_R(f,f^\star_\delta)+t_\delta^\star\lmb\textup{DDP}(f)  -\delta\rmb,      &  D_{-} (0)>\delta.
\end{array}\right.
\end{align*}

Now note that $t_\delta^\star>0$ when $D_{-} (0)>\delta$ 
and $t_\delta^\star<0$ when $\delta<-D_{+} (0)$. 
This implies
that if $|\textup{DDP}(f)|\le \delta$, then
$d_R \lsb {f},f_\delta^\star\rsb\ge d_E({f},f^\star_\delta)$.

\subsection{Proof of Theorem \ref{thm:lower_bound}}
\label{pf:thm:lower_bound}

In the automatically fair and fair-boundary cases
when $\delta\ge \max\lbb |D_{-} (0)|,|D_{+} (0)|\rbb$, 
all unconstrained Bayes-optimal classifiers are $\delta$-fair Bayes-optimal classifiers. 
In this scenario, the fair classification problem is simply a standard unconstrained classification problem, and the minimax lower bound is the same as the lower bound \eqref{eq:lowbnd:auto} from  \cite{tsy2007}.

Next, we consider the fairness-impacted case. 
In what follows, we assume $\delta=0$ and write $\gamma:=\gamma_0$ without loss of generality. 
We will generally omit mentioning $\delta$ further in this proof.
In addition to the usual lower bound for classification problems, 
in the fairness-impacted case, 
the minimax lower bound may contain 
a second term
due to the estimation of thresholds. Accordingly, the proof of the theorem also contains two parts.

In the first part, we start from the strategy of \cite{tsy2007}; with some modifications, 
either in order to streamline the proof,
or as required by the fairness constraint. 
For $\vsig \in \{0, 1\}^m$, 
we construct a family of distributions $\P_\vsig$ on $[0,1]^d\times\{0,1\}\times\{0,1\}$  such that $D_{\vsig,r}(t^\star_\vsig)<0<D_{\vsig,l}(t^\star_\vsig)$,  and apply Assouad’s lemma adapted to the fair classification problem.
In the second part, we construct two distributions, $\P_{1}$ and $\P_{-1}$, on $\MR^d\times\{0,1\}\times\{0,1\}$ such that
 $D_{\pm1,r}(t^\star_{\pm1})=0=D_{\pm1,l}(t^\star_{\pm1})$,
and then apply Le Cam's lemma to show
that 
the second term appears in the lower bound for the fairness-impacted case.

\textbf{Part I}:
For an integer $q\ge 1$ divisible by eight,
 we consider the following regular grid in the unit cube:
\beq\label{gq}
\mathcal{G}_q =
\left\{\left(\frac{8k_1+4}{q},\frac{8k_2+4}{q},\ldots,\frac{8k_d+4}{q}\right): k_i \in \{0, 1, ..., q/8-1\},  i \in [d]\right\}.
\eeq
Observe that the cardinality of  $\mathcal{G}_q$ is $M=8^{-d}{q}^{d}$,
and denote by
$x_1, x_2, ..., x_M$ the points in $\mathcal{G}_q$. Let $m\le M$ be a positive integer to be specified later.
Writing $ B_0=[0,1]^d\setminus \cup_{j=1}^{m} B_{d,2}(x_j,2q^{-1})$, we have that
$ B_0,B_{d,2}(x_1,2q^{-1}),\ldots,B_{d,2}(x_{m},2q^{-1})$ forms a partition of $[0,1]^d$;
note in particular any two distinct points $x_i,x_j$ are at distance at least $8/q$, and so the balls do not intersect. 
We next define 
a collection $\mathcal{H} = \{\P_\vsig :\vsig \ \in \{0,1\}^m\}$
 of probability distributions $\P_\vsig$ on $\mathcal{Z} = \MR^d \times \{0, 1\} \times \{0, 1\}$, indexed by the vertices of the hypercube,
by specifying the marginal distributions of $X$ and $A$, and the conditional distribution $\P_{Y|X,A=a}$.

\begin{itemize}
\item {\it Construction of marginal distributions of $X$ and $A$}:

We construct $A$ and $X$ to be independently distributed with the marginal distributions of $X$  and $A$ not depending on $\vsig$. 
For any $\P_\vsig \in\mathcal{H}$, we set $\P_\vsig(A=1)=1/2$. 
For a certain $w$ with $0<w<m^{-1}$, to be chosen later, $X$ has a density $\mu$ with respect to the Lebesgue measure $\lambda$ on $\MR^d$, defined in the following way: 
\begin{align}\label{mu}
\mu(x)=\left\{\begin{array}{lcl}
w/\lambda[B_{d,2}(x_j,q^{-1})],    &  x\in B_{d,2}(x_j,q^{-1}), \ \ j\in[m];\\
(1-mw)/\lambda[ B_0],    &  x\in B_0;\\
0,    & \textnormal{otherwise}.
\end{array}\right.
\end{align}
\begin{figure}
    \centering
    \includegraphics[width = 0.5\textwidth]{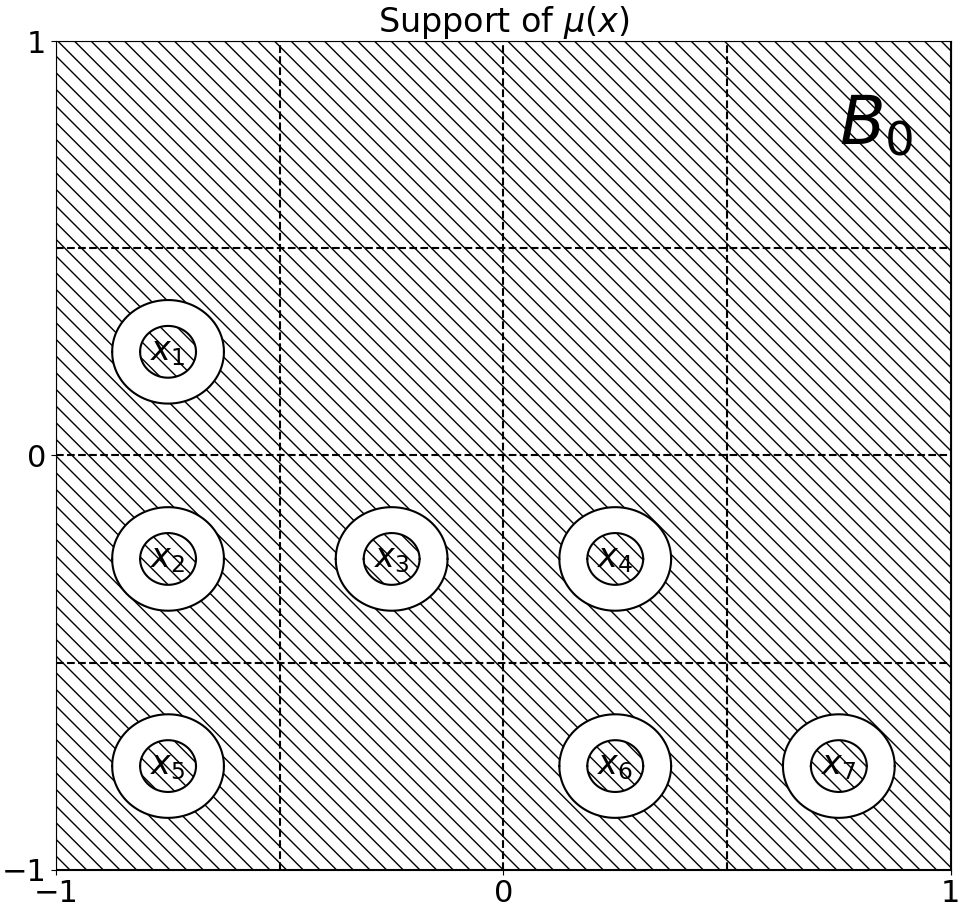}
    \caption{The shaded areas illustrate the support of $\mu$ in a two-dimensional setting.}
    \label{fig:illcase1}
\end{figure}

Figure \ref{fig:illcase1} provides an illustration of the support of the function $\mu(x)$. Note that $\mu$ has a constant density over the ball
$B_{d,2}(x_j,q^{-1})$ \footnote{Here, we use $q^{-1}$ rather than $2q^{-1}$.}, for all $j$;
as well as on $B_0$.
Clearly, $\mu$ is a probability density function on $[0,1]^d$. 

\item {\it Construction of conditional distribution of $Y$ given $X$ and $A$}:

Let $u$ be an infinitely differentiable and non-increasing function on $[0,\infty)$, with bounded derivatives of all orders,
such that
$u(x)=1$ when $x\le 1$,
$u(x)\in (0,1)$ 
when $x\in (1, 2)$,
and $u(x)=0$ when $x\ge  2$.   
For $0< c_\beta \le q^\beta/2$,
let $\phi:\MR^d \to [0,\infty)$ be the function defined, for $x\in \R^d$, as
\beq\label{phi}
\phi(x)=c_\beta q^{-\beta} u(q\|x\|),
\eeq
and note that by the choice of $c_\beta$ and the properties of $u$, 
$\phi(x)\le 1/2$ for all $x$.
For any $\P_\vsig \in\mathcal{H}$, denote
$\eta_{\vsig,a}(x,a)=\P_{\vsig}(Y=1\mid X=x,A=a)$  for all $x,a$. 
We set
\begin{equation}\label{eta}
\eta_{\vsig,1}(x) =\left\{\begin{array}{lcl}
 (1+\sigma_j\phi(x-x_j))/2, 
 & x\in B_{d,2}(x_j,2q^{-1}),\, j\in[m];\\
 1/2, & \mathrm{otherwise};
\end{array}\right .
\end{equation}
and
\begin{equation}\label{eta0}
\eta_{\vsig,0}(x)= 1/2,  \ x\in[0,1]^d.
\end{equation}

\end{itemize}
Next, we consider the fair Bayes-optimal classifiers under $\P_{\vsig}\in\mathcal{H}$. 
Define $D_{\vsig,-}$ to be $D_{-}$ from \eqref{eq:Disfun1} for the distribution $\P_\vsig$ and define $D_{\vsig,+}$, $t_\vsig^\star$, $T_{\vsig,a}^\star$, $\tau_{\vsig,a}^\star$, $g_{\vsig,a}$
similarly. 
By the definition of the distribution of $X$ from \eqref{mu}, by \eqref{eta}, \eqref{eta0},
and due to our choice $w<m^{-1}$,
\begin{align*}
&-1\le D_{\vsig,-} (0)=\P_{\vsig,X|A=1}\lsb\eta_{\vsig,1}(X)>\frac12\rsb-\P_{\vsig,X|A=0}\lsb\eta_{\vsig,0}(X)\ge\frac12\rsb\\
&=\sum_{j=1}^{m}I(\sigma_j=1)\int_{B_{d,2}(x_j,q^{-1})}\mu(x)dx-1
=w  \sum_{j=1}^{m}I(\sigma_j=1)-1<0.
\end{align*}
Similarly, 
\begin{align*}
D_{\vsig,+} (0)=\P_{\vsig,X|A=1}\lsb\eta_{\vsig,1}(X)\ge\frac12\rsb-\P_{\vsig,X|A=0}\lsb\eta_{\vsig,0}(X)>\frac12\rsb
=1>0.
\end{align*}
This implies $D_+(0)\ge -D_-(0)$. In addition, one can verify that for any $t>0$, one has
$D_{\vsig,+}(t)<0$.
Thus, we have
$t_\vsig^\star=\sup_t\lbb D_{\vsig,+}(t)>0\rbb=0$ and $D_{\vsig,-}(t_\vsig^\star)<0< D_{\vsig,+}(t_\vsig^\star)$.
Further, from 
\eqref{Tdel}, 
$T_{\vsig,a}^\star=1/2$ for $a\in\{0,1\}$.
Moreover, 
due to \eqref{eq:op-de-rule-dp},
and
since $\delta=0$  and $D_{\vsig,l}(t)<0$,
\eqref{eq:hard_constraint1} becomes
\begin{align*}
&\P_{\vsig,X|A=1}\lsb\eta_{\vsig,1}(X)>\frac12\rsb
+\tau_{\vsig,1}^\star \P_{\vsig,X|A=1}\lsb\eta_{\vsig,1}(X)=\frac12\rsb\\
&-
\P_{\vsig,X|A=0}\lsb\eta_{\vsig,0}(X)>\frac12\rsb
-\tau_{\vsig,0}^\star \P_{\vsig,X|A=0}\lsb\eta_{\vsig,0}(X)=\frac12\rsb = 0.
\end{align*}
Since by \eqref{mu}, \eqref{eta}, and \eqref{eta0}, 
$\P_{\vsig,X|A=1}\lsb\eta_{\vsig,1}(X)>{1}/{2}\rsb=w\sum_{j=1}^{m} I(\sigma_j=1)$, while 
$\P_{\vsig,X|A=0}\lsb\eta_{\vsig,0}(X)>{1}/{2}\rsb=0$,  and
$\P_{\vsig,X|A=0}\lsb\eta_{\vsig,0}(X)={1}/{2}\rsb=1$, this is equivalent to
\begin{align*}
&w  \sum_{j=1}^{m} I(\sigma_j=1)
+\tau_{\vsig,1}^\star \P_{\vsig,X|A=1}\lsb\eta_{\vsig,1}(X)={1}/{2}\rsb
=\tau_{\vsig,0}^\star.
\end{align*}
Hence, to ensure that 
a classifier is Bayes-optimal, due to
\eqref{eq:hard_constraint1} in 
Theorem \ref{thm-opt-dp}, 
it suffices to take $\tau_{\vsig,1}^\star=0$ and $\tau_{\vsig,0}^\star=w\sum_{j=1}^{m} I(\sigma_j=1)\in [0,1]$. Thus, 
based on \eqref{eq:op-de-rule-dp},
using that $T_{\vsig,a}^\star=1/2$,
a fair Bayes-optimal classifier is given by
\begin{align}\label{fstarvsig}
\left\{\begin{array}{ll}
         f^\star_{\vsig}(x,1)=I\left(x\in\bigcup_{j\in[m]:\sigma_j=1}B_{d,2}(x_j,2q^{-1})\right) ;&\\
     f^\star_{\vsig}(x,0)=w\sum_{j=1}^{m} I(\sigma_j=1).&
     \end{array}\right.
\end{align}
Now, we verify the  distributional conditions:
\begin{itemize}

\item {\it Smoothness Condition from Definition \ref{h}}:
For any $b \in \mathbb{N}^d$
such that $|b| \le \lbeta_+$, the partial derivative $D^b\eta_{\vsig,1}$ at $x\in  B_{d,2}(x_j,2q^{-1})$ exists and
$D^b\eta_{\vsig,1}(x) = c_\beta q^{|b|-\beta}\phi_j/2  \cdot D^b u(q\|x- x_j\|)$. 
Since $\|\cdot\|$ is infinitely differentiable on $\R^d/\{0\}$ and $u(\cdot)$ is infinitely differentiable on $[0,\infty)$ with $u(t)= 1$ for $0<t\le 1$, we have that $u(\|\cdot\|)$
is infinitely differentiable on $\R^d$ with bounded derivatives of all order. Thus,
there is a constant $M$ such that $|D^b u(q\|x- x_j\|)|<M$. 
Therefore, for any $j \in \{1,\ldots,m\}$ and any
$x,x' \in B_{d,2}(x_j,2q^{-1})$, we have, when $c_\beta$ is small enough, that
$|\eta_{\vsig,1}(x') - \eta_{\vsig,1}(x)| \le L_\beta \|x'-x\|^\beta$.
This implies that $\eta_{\vsig,a}$ belongs to the H\"older
class $\Sigma(\beta,L_\beta,\MR^d
)$.

 \item {\it Margin Condition from Definition \ref{m}}: 
 Since $D_{\vsig,r}(t_\vsig^\star)<0< D_{\vsig,l}(t_\vsig^\star)$, we only need to verify condition \eqref{ma1}. We have from \eqref{eta} that, for $\vsig\in\{0,1\}^m$,
 \begin{equation*}
g_{\vsig,+} (\ep)=\P_{\vsig,X|A=1}\lsb \frac12-\ep\le \eta_{\vsig,1}(X)< \frac12\rsb+\P_{\vsig,X|A=0}\lsb \frac12<\eta_{\vsig,0}(X)\le \frac12+\ep\rsb=0
\end{equation*}
and
\begin{align*}
&g_{\vsig,-} (\ep)=\P_{\vsig,X|A=1}\lsb \frac12< \eta_{\vsig,1}(X)\le \frac12+\ep\rsb+\P_{\vsig,X|A=0}\lsb \frac12-\ep\le\eta_{\vsig,0}(X)< \frac12\rsb\\
&=\sum_{j=1}^{m} I\lsb\sigma_j=1\rsb\cdot \P_{\vsig,X|A=1}\lsb 0<\phi(X-x_j)<2\ep\rsb\le\sum_{j=1}^{m}\int_{B_{d,2}(x_j,2q^{-1})} I (0<\phi(x-x_j)<2\ep)\mu(x)dx\\
&=\sum_{j=1}^{m}\int_{B_{d,2}(x_j,q^{-1})} I (0<\phi(x-x_j)<2\ep)\frac{w}{\lambda[B_{d,2}(x_j,q^{-1})]}dx
= {mw}\cdot I\lsb\epsilon > c_\beta q^{-\beta}/2\rsb,
\end{align*}
where in the last step we have used that due to \eqref{phi},
we have $\phi(x-x_j)\le c_\beta q^{-\beta}$ for all $x$.
Therefore, the $\gamma$-margin condition is satisfied if $mw = \Theta(q^{-\gamma\beta}).$ 
\item {\it Strong density condition  from Definition \ref{sd}}: 
Let $m=\lfloor q^{d-\gamma\beta}\rfloor$, $w=c_w q^{-d}$ for some $c_w>0$ and take $q\to\infty$ as $n\to\infty$.  
The condition $\gamma\beta \le d$ ensures that $m\ge 1$. 
Next, note that since by definition $X$ is independent of $A$, we have $\mu_a = \mu$ for all $a$. 
Now, let $\Omega_{\mu} = B_0\cup_{j=1}^{m}  B_{d,2}(x_j,q^{-1}) $ be the support of $\mu$. 
Recall
that 
$\mathcal{C}_{z,q} = \{y: |y_i-z_i|\le 4q^{-1}, i=1,2,\ldots,d\}$ and $\mathcal{D}_{z,q}=B_{d,2}(z,2q^{-1})\setminus B_{d,2}(z,q^{-1})$. 
Recalling $\mathcal{G}_q$ from \eqref{gq},
we have that
$$[0,1]^d = \cup_{z\in\mathcal{G}_q} \mathcal{C}_{z,q} \ \ \ \text{and} \ \ \ \Omega_\mu = [0,1]^d \setminus \lsb\cup_{i=1}^m \mathcal{D}_{x_i,q}\rsb.$$ 
For $x\in\Omega_\mu$, we have, for all $i$ that $x\in [0,1]^d\setminus \mathcal{C}_{x_i,q}$. Then, by Lemma \ref{lem:inter3},  for any $r>0$, there exists $0<C_r\le 1$ such that,
\begin{align*}   
&\lambda[B_{d,2}(x,r)\cap \Omega_\mu]  = \lambda\bigl[B_{d,2}(x,r)\cap [0,1]^d\bigr] - \sum_{i=1}^m\lambda[B_{d,2}(x,r)\cap \mathcal{D}_{x_i,q}]\\
&=\sum_{z\in\mathcal{G}_q}\lambda[B_{d,2}(x,r)\cap \mathcal{C}_{z,q}] - \sum_{i=1}^m\lambda[B_{d,2}(x,r)\cap \mathcal{D}_{x_i,q}]\\
&=\sum_{z\in\mathcal{G}_q\setminus \lsb\cup_{i=1}^m\{x_i\}\rsb}\lambda[B_{d,2}(x,r)\cap \mathcal{C}_{z,q}]+ \sum_{i=1}^m\lambda[B_{d,2}(x,r)\cap \mathcal{C}_{x_i,q}]-\sum_{i=1}^m \lambda[B_{d,2}(x,r)\cap \mathcal{D}_{x_i,q}]\\
&=\sum_{z\in\mathcal{G}_q\setminus \lsb\cup_{i=1}^m\{x_i\}\rsb}\lambda[B_{d,2}(x,r)\cap \mathcal{C}_{z,q}]+ \sum_{i=1}^m\lambda[B_{d,2}(x,r)\cap \lsb\mathcal{C}_{x_i,q}\setminus\mathcal{D}_{x_i,q}\rsb]\\
&\ge \sum_{z\in\mathcal{G}_q\setminus \lsb\cup_{i=1}^m\{x_i\}\rsb}\lambda[B_{d,2}(x,r)\cap \mathcal{C}_{z,q}]+ C_r\sum_{i=1}^m\lambda[B_{d,2}(x,r)\cap \mathcal{C}_{x_i,q} ]\\
&\ge C_r\sum_{z\in\mathcal{G}_q}\lambda[B_{d,2}(x,r)\cap \mathcal{C}_{z,q} ]= C_r\lambda[B_{d,2}(x,r)\cap [0,1]^d].
\end{align*}
On the other hand,
if $x\in \cup_{j=1}^m  B_{d,2}(x_j,q^{-1}) $, 
due to \eqref{mu},
 we have 
 due to the choice of $w$ that
 $\mu(x)=c_w q^{-d}/{\lambda[ B_{d,2}(x_j,q^{-1}) ]}=c_wV_{d,2}^{-1}$.
On the other hand, if $x\in  B_0$, due to \eqref{mu}, we have $\mu(x)=(
1-mc_w q^{-d})/({1-m2^dq^{-d}}V_{d,2})$.
Thus, the conditional distribution of $X$ given $A=a$ satisfies the strong density condition
with $c_\mu = C_r$,
$r_\mu = 1$,
and
$\mu_{\min} \le \mu_{\max}$
if 
$\mu_{\min}\le  (
1-mc_w q^{-d})/({1-m2^dq^{-d}}V_{d,2}) \le \mu_{\max}$
and
$\mu_{\min}\le  c_w/V_{d,2}\le \mu_{\max}$.
\end{itemize}

Finally, we derive the first term of the minimax lower bound. 
For $r\in\{0,1\}$
and $\vsig \in \{-1, 1\}^m$,
denote $\vsig_{j,r}=(\sigma_{1}, \ldots , \sigma_{j-1}, r, \sigma_{j+1}, \ldots , \sigma_m)$,
Clearly, $\P^{\otimes n}_{\vsig_{j,1}}$ is absolutely continuous with respect to $\P^{\otimes n}_{\vsig_{j,0}}$. 
Moreover, recall that the total variation distance between $\P^{\otimes n}_{\vsig_{j,1}}$ and $\P^{\otimes n}_{\vsig_{j,0}}$ can be expressed as
$$\textup{TV}\left(\P^{\otimes n}_{\vsig_{j,1}},\P^{\otimes n}_{\vsig_{j,0}}\right)
=1-\int\lsb \frac{\P^{\otimes n}_{\vsig_{j,1}}(z)}{\P^{\otimes n}_{\vsig_{j,0}}(z)} \wedge1\rsb d\P^{\otimes n}_{\vsig_{j,0}}(z).$$

Now, we provide an upper bound on the 
Kullback–Leibler divergence between  
$\P^{\otimes n}_{\vsig_{j,1}}$ and $\P^{\otimes n}_{\vsig_{j,0}}$, taking $j=1$ without loss of generality. 
As $\P_{\vsig_{1,1}}(A=a)=\P_{\vsig_{1,0}}(A=a)=1/2$ for $a\in\{0,1\}$,
recalling $\eta_{\vsig,a}$ from \eqref{eta} and \eqref{eta0},
we have
\begin{align*}
&\textup{KL}\lsb \P_{\vsig_{1,1}},\P_{\vsig_{1,0}}\rsb 
=\sum_{a\in\A}\P_{\vsig_{1,1}}(A=a)\int \eta_{\vsig_{1,1},a}(x)\log\frac {\eta_{\vsig_{1,1},a}(x)}{\eta_{\vsig_{1,0},a}(x)}\mu(x)dx\\
&=\frac12\lmb\int\eta_{\vsig_{1,1},1}(x)\log\frac {\eta_{\vsig_{1,1},1}(x)}{\eta_{\vsig_{1,0},1}(x)}\mu(x)dx+\int (1-\eta_{\vsig_{1,1},1}(x))\log\frac {1-\eta_{\vsig_{1,1},1}(x)}{1-\eta_{\vsig_{1,0},1}(x)}\mu(x)dx\rmb\\
&=\frac12\int_{ B_{d,2}(x_1,2q^{-1})}\lmb\frac{1+\phi(x-x_1)}{2}\log\lsb 1+\phi(x-x_1)\rsb+\frac{1-\phi(x-x_1)}{2}\log\lsb{1-\phi(x-x_1)}\rsb\rmb\mu(x)dx\\
&\le\frac12\int_{ B_{d,2}(x_1,2q^{-1})}\lmb\frac{1+\phi(x-x_1)}{2}\phi(x-x_1)-\frac{1-\phi(x-x_1)}{2}\phi(x-x_1)\rmb\mu(x)dx\\
&\le\frac12\int_{ B_{d,2}(x_1,2q^{-1})}\lsb \phi^2(x-x_1)\frac{w}{\lambda\lmb B_{d,2}(x_1,q^{-1})\rmb}\rsb dx\le\frac{w}{2}c_\beta^2q^{-2\beta}.
\end{align*}
Here, the first inequality holds since, for $0<x<1$, $\log(1+x)<x$ and $\log(1-x)<-x$;
and the last inequality holds 
due to \eqref{phi}.
By Pinsker's inequality \citep{tsybakov2009nonparametric}, we therefore have
\begin{align}\label{tvbd}
&\textup{TV}\lsb \P^{\otimes n}_{\vsig_{1,1}},\P^{\otimes n}_{\vsig_{1,-1}}\rsb\le\frac12\sqrt{\textup{KL}\lsb \P^{\otimes n}_{\vsig_{1,1}},\P^{\otimes n}_{\vsig_{1,-1}}\rsb}
=\frac12\sqrt{n\textup{KL}\lsb \P_ {\vsig_{1,1}},\P_{\vsig_{1,-1}}\rsb}\le\sqrt{\frac{1}{8}}c_\beta\sqrt{nw} q^{-\beta}.
\end{align}
Recalling that $w=c_w q^{-d}$ and taking $q=n^{{1}/({2\beta+d})}$ with $c_w={2}c_\beta^{-2}$, we have $\textup{TV}\lsb \P^{\otimes n}_{\vsig_{1,1}},\P^{\otimes n}_{\vsig_{1,-1}}\rsb\le 1/2$. 

To complete the proof for the first term in minimax lower bound, we apply Assouad's lemma \citep{assouad1983deux,tsybakov2009nonparametric} 
to the class $\mathcal{H}$. 
Let  $\nu$ denote the distribution of a Bernoulli variable with parameter $1/2$, so that
for $\sigma\sim \nu$,
$\nu(\sigma = 1) = \nu(\sigma = 0) =\frac12$.
For data-dependent sets $\widehat{G}_{a,1}$ and $\widehat{G}_{a,\tau}$ for $a\in\{0,1\}$,
let $\widehat{f}_{\delta,n}$ be the classifier with, for all $x,a$,
$$\widehat{f}_{\delta,n}(x,a)= I(x\in \widehat{G}_{a,1})+\widehat\tau I(x\in \widehat{G}_{a,\tau}).$$
We use $\E_\vsig$ to denote expectation under the distribution $\P_\vsig$. Then, by the definition of $d_E$ from \eqref{eq:measure}, 
using that $T_{\vsig,a}^\star=1/2$,
and by \eqref{fstarvsig}, 
\eqref{eta}
\begin{align*}
&\sup_{\P\in\mathcal{P}_\Sigma}\EN d_{E}(\widehat{f}_{\delta,n}, f^\star)\ge  \sup_{\vsig\in\{0,1\}^m} \E^{\otimes n}_\vsig d_E(\widehat{f}_{\delta,n}, f_\vsig^\star)\\
&= \sup_{\vsig\in\{0,1\}^m} \E^{\otimes n}_\vsig \lbb\int\lsb \widehat{f}_{\delta,n}(x,1)-f_\vsig^\star(x,1)\rsb\lsb \frac12-\eta_{\vsig,1}(x)\rsb\mu(x) dx\rbb\\
&=\sup_{\vsig\in \{0,1\}^m}\E^{\otimes n}_\vsig \lbb\sum_{j=1}^{m}\int_{ B_{d,2}(x_j,2q^{-1})}\left|\widehat{f}_{\delta,n}(x,1)-\sigma_j\right|\frac{\phi(x-x_j)}{2}\mu(x)dx\rbb\\
&\ge \frac12\E^{\otimes{m}}_{\nu}\lbb\E^{\otimes n}_\vsig \lbb\sum_{j=1}^m\int_{ B_{d,2}(x_j,2q^{-1})}\left|\widehat{f}_{\delta,n}(x,1)-\sigma_j\right|\phi(x-x_j)\mu(x)dx\rbb\rbb.
\end{align*}
In the last line, we have written 
$\E^{\otimes{m}}_{\nu}$ for the expectation over $\vsig = (\sigma_1,\ldots, \sigma_m)$ with i.i.d.~$\sigma_i\sim \nu$ for all $i\in [m]$.
Recalling the definition $\vsig_{j,0}$, the last term equals
\begin{align*}
& \frac12\E^{\otimes{m}}_{\nu}\lbb\sum_{j=1}^m\E^{\otimes n}_{\vsig_{j,0}}\lbb\frac{\P^{\otimes n}_{\vsig}}{\P^{\otimes n}_{\vsig_{j,0}}} \int_{ B_{d,2}(x_j,2q^{-1})}\left|\widehat{f}_{\delta,n}(x,1)-\sigma_j\right|\phi(x-x_j)\mu(x)dx\rbb\rbb\\
&=\frac12\E^{\otimes\{m-1\}}_{\nu}\lbb\sum_{j=1}^m
\E_{\sigma_j\sim \nu}\E^{\otimes n}_{\vsig_{j,0}}\lbb\frac{\P^{\otimes n}_{\vsig}}{\P^{\otimes n}_{\vsig_{j,0}}} \int_{ B_{d,2}(x_j,2q^{-1})}\left|\widehat{f}_{\delta,n}(x,1)-\sigma_j\right|\phi(x-x_j)\mu(x)dx\rbb\rbb\\
&\ge\frac12\E^{\otimes\{m-1\}}_{\nu}\lbb\sum_{j=1}^m\E^{\otimes n}_{\vsig_{j,0}}\lbb\lsb\frac{\P^{\otimes n}_{\vsig_{j,1}}}{\P^{\otimes n}_{\vsig_{j,0}}}\wedge1\rsb \E_{\sigma_j\sim \nu}\int_{ B_{d,2}(x_j,2q^{-1})}\left|\widehat{f}_{\delta,n}(x,1)-\sigma_j\right|\phi(x-x_j)\mu(x)dx\rbb\rbb.
\end{align*}
Since for all $z\in[0,1]$, $\E_{\sigma_j\sim \nu}|z-\sigma_j| = (|z-1|+|z|)/2=1/2$,
this can be further written as
\begin{align*}
& \frac14\E^{\otimes\{m-1\}}_{\nu}\lbb\sum_{j=1}^m\E^{\otimes n}_{\vsig_{j,0}}\lbb\lsb1-\textup{TV}\lsb \P^{\otimes n}_{\vsig_{j,1}},\P^{\otimes n}_{\vsig_{j,-1}}\rsb\rsb \int_{ B_{d,2}(x_j,2q^{-1})}\phi(x-x_j)\mu(x)dx\rbb\rbb.
\end{align*}
Using \eqref{tvbd} with $m = \lfloor q^{d-\gamma\beta} \rfloor$, $w=c_wq^{-d}$ and $q=n^{{1}/({2\beta+d})}$,
as well as by
$$
\int_{ B_{d,2}(x_1,2q^{-1})}\phi(x-x_1)\mu(x)dx
=
\int_{ B_{d,2}(x_1,q^{-1})}\phi(x-x_j)\frac{w}{\lambda\lmb B_{d,2}(x_1,q^{-1})\rmb}dx
=wc_\beta q^{-\beta},
$$
this is lower bounded by
$
\frac{mwc_\beta q^{-\beta}}{8}
\ge C'n^{-\frac{\beta(\gamma+1)}{2\beta+d}}$, as desired.
This finishes the argument of the first part.

\begin{figure}
    \centering
    \includegraphics[width = 0.8\textwidth]{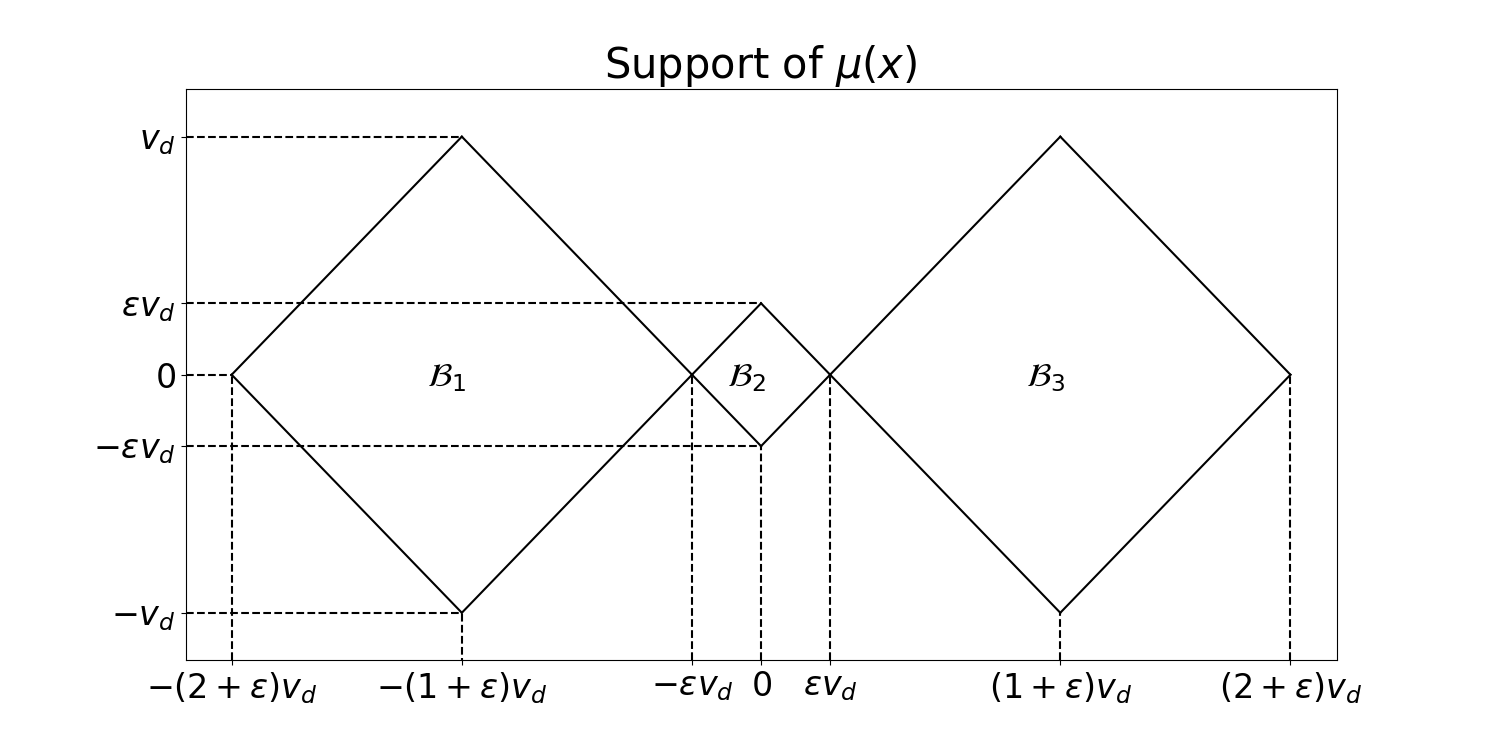}
    \includegraphics[width = 0.8\textwidth]{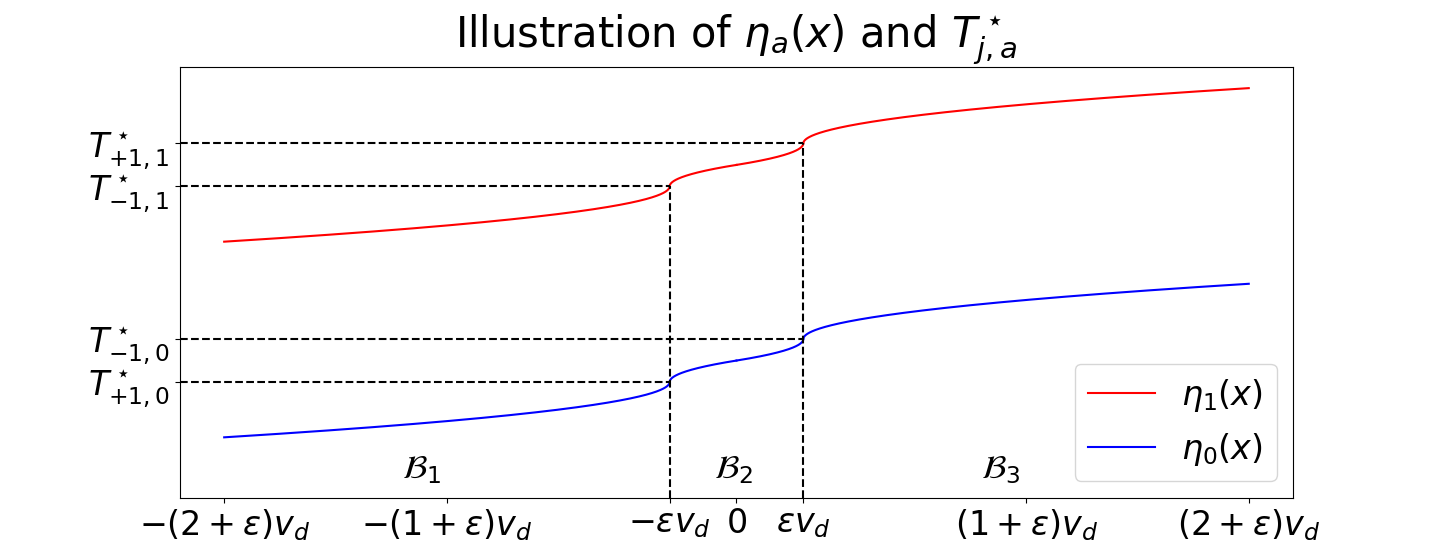}
    \caption{Illustration of constructions in the lower bound based on a two-dimensional setting. In the upper panel, we plot the support of $\mu(x)$, i.e., $\mathcal{B}_1$, $\mathcal{B}_2$ and $\mathcal{B}_3$.
    In the lower panel, we plot the regression functions for two groups (Here, we only plot $\eta_a(x_1)$ as $\eta_a(x)$ only depends on the first coordinate of $x$), and the fair thresholds under $\P_{+1}$ and $\P_{-1}$.}
    \label{fig:lb}
\end{figure}

\textbf{Part 2.} In this part, we apply Le Cam's method to prove that the second term
on the right hand side of \eqref{eq:lowbnd:impacted2}
appears in the lower bound; see Figure \ref{fig:lb} for an illustration of the construction. 
Recall that $V_{d,p}$ is the volume of a $d$-dimensional unit $\ell_p$ ball and let $v_d = V_{d,1}^{-1/d}$.
We will construct  two distributions $\P_{1}$ and $\P_{-1}$  on 
$\mathcal{X}\times \{0,1\}$ with $
\mathcal{X} = [-3,3]\times [-1,1]^{d-1}$.  For  $0<s\le  1$ specified later, let
$
 \mathcal{B}_1 = \{{x} = (x_1,\ldots,x_d)^\top\in \MR^d: |x_1+(1+s)v_d|+|x_2|+\ldots |x_d|\le v_d\},
$
$ \mathcal{B}_2 = \{{x} = (x_1,\ldots,x_d)^\top\in \MR^d: |x_1|+|x_2|+\ldots |x_d|\le sv_d\},$
and
$
 \mathcal{B}_3 = \{{x} = (x_1,\ldots,x_d)^\top\in \MR^d: |x_1-(1+s)v_d|+|x_2|+\ldots |x_d|\le v_d\}.
$
By construction, we have $\lambda[ \mathcal{B}_1] = \lambda[ \mathcal{B}_3] = 1$ and $\lambda[ \mathcal{B}_2] = s^d$.
We will the use subscripts $+1$ and $-1$ to denote quantities corresponding to $\P_{1}$ and $\P_{-1}$, respectively.
\begin{itemize}
\item {\it Construction of marginal distributions of $X$ and $A$}:
We set $\P_{1}(A=1)=\P_{-1}(A=1)=0.5$ and 
for $j\in\{-1,1\}$ and $a\in\{0,1\}$,
denote by $\mu_{j,a}$ the conditional density function of $X$ given $A=a$ under $\P_{j}$, defined as, for $x=(x_1,...,x_d)^\top$,
$$
\left\{\begin{array}{lcl}
  \mu_{1,1}(x)=\mu_{-1,0}(x)=\left\{\begin{array}{lcl}
     1/(2+2s^d) ,   &  x_1\in  \mathcal{B}_1\cup  \mathcal{B}_2;\\
    1/2 ,   &  x_1\in    \mathcal{B}_3;\\
    0,    &  \mathrm{otherwise};
  \end{array} \right.   &  \\
\mu_{1,0}(x)=\mu_{-1,1}(x)=\left\{\begin{array}{lcl}
     1/2,   &  x_1\in   \mathcal{B}_1;\\
       1/(2+2s^d),   &  x_1\in  \mathcal{B}_2 \cup \mathcal{B}_3;\\
    0,    &  \mathrm{otherwise}.
  \end{array} \right. \label{mupm}
\end{array}\right.$$

\item {\it Construction of conditional distribution of $Y$ given $X$ and $A$}:
For $j\in\{-1,1\}$ and all $x,a$, 
consider
the regression functions
$\eta_{j,a}(x)=\P_{j}(Y=1\mid A=a,X=x)$,
defined as
 \begin{align}\label{etaj}
\eta_{j,a}(x)=\eta_{a}(x) =\left\{\begin{array}{lll}
   1/2 +(2a-1)/4 - c_\beta(sv_d)^{d/\gamma} - c_\beta (-x_1-sv_d)^{d/\gamma}, & -(2+s)v_d\le x_1 < - sv_d;\\
    1/2 +(2a-1)/4 -c_\beta (sv_d)^{d/\gamma} +c_\beta (x_1+sv_d)^{d/\gamma}, & -sv_d\le x_1 < 0;\\
   1/2 +(2a-1)/4 + c_\beta(sv_d)^{d/\gamma} -c_\beta (-x_1+sv_d)^{d/\gamma}, & 0\le x_1 <  sv_d;\\
   1/2 +(2a-1)/4 +c_\beta (sv_d)^{d/\gamma} + c_\beta(x_1-sv_d)^{d/\gamma}, & sv_d\le x_1 \le (2 + s)v_d.
    \end{array}\right.
\end{align}
Here $c_\beta\in (0,1)$ is chosen small enough that $\eta_{j,a} \in\Sigma\lsb\beta,L_\beta,\R^d\rsb$ for all $j,a$, which can be done since $\gamma\beta \le d$.
\end{itemize}
Next, we consider the fair Bayes-optimal classifiers under $\P_{1}$ and $\P_{-1}$.
Define, for $j\in\{-1,1\}$, $D_{j,r}$ to be $D_{-}$ from \eqref{eq:Disfun1} for the distribution $\P_{j}$ and define $D_{j,\ell}$, $t_j^\star$, $T_{j,a}^\star$, $\tau_{j,a}^\star$, $g_{j,a}$ similarly.
It can be readily verified that 
\begin{align}\label{eq:probcase2}
\nonumber&\P_{1,X|A=1}(\eta_{1,1}(X)>3/4+ c_\beta(sv_d)^{d/\gamma})
=\P_{1,X|A=0}(\eta_{1,0}(X)\ge 1/4- c_\beta(sv_d)^{d/\gamma})\\
&=\P_{-1,X|A=1}(\eta_{-1,1}(X)>3/4-c_\beta(sv_d)^{d/\gamma})
=\P_{-1,X|A=0}(\eta_{-1,0}(X)\ge 1/4+c_\beta(sv_d)^{d/\gamma})
=\frac12.
\end{align}
For instance, $\eta_{1,1}(x)>3/4+c_\beta(sv_d)^{d/\gamma}$ holds if and only if $sv_d\le x_1 \le (2 + s)v_d$. 
Noting that $\mu_{j,a}(x)\equiv 0$ when $x\notin \cup_{j=1}^3\mathcal{B}_j$, we have
\begin{align*}
&\P_{1,X|A=1}(sv_d\le X_1 \le (2 + s)v_d)=\int_{\lsb\cup_{j=1}^3\mathcal{B}_j\rsb \cap \{sv_d\le x_1 \le (2 + s)v_d\}}\mu_{1,1}(x)dx
=\int_{ \mathcal{B}_3}\mu_{1,1}(x)dx
=\frac{1}{2}\lambda( \mathcal{B}_3) = 1/2.
\end{align*}
Recalling  \eqref{eq:Disfun1},
using that $p_0=p_1=1/2$,
\eqref{eq:probcase2} further implies that
for $j\in\{-1,1\}$, 
\begin{align*}
&D_{j,r}\lsb \frac14+j\cdot c_\beta(sv_d)^{d/\gamma}\rsb\\
&=
\P_{j,X|A=1}\lsb\eta_{j,1}(X)>\frac34+j\cdot c_\beta(sv_d)^{d/\gamma}\rsb-\P_{j,X|A=0}\lsb\eta_{j,0}(X)\ge\frac14-j\cdot c_\beta(sv_d)^{d/\gamma}\rsb
=\frac12-\frac12 = 0.
\end{align*}
As both $\eta_1(X)$ and $\eta_0(X)$ are continuous random variable on $\cup_{j=1}^3\mathcal{B}_j$,
We thus deduce that
$t_{j}^\star=1/4+ j\cdot c_\beta(sv_d)^{d/\gamma}$. In fact, for any $\ep>0$, we have
\begin{align*}
&D_{j,r}(t_j^\star-\ep) = D_{j,r}(t_j^\star-\ep) - D_{j,r}(t_j^\star) + D_{j,r}(t_j^\star)\\
&=\P_{j,X|A=1}\lsb\frac12 +t^\star -\ep<\eta_{j,1}(X)\le \frac12 +t^\star\rsb  +\P_{j,X|A=0}\lsb\frac12 -t^\star \le \eta_{j,0}(X)< \frac12 -t^\star+\ep \rsb>0,
\end{align*}
and
\begin{align*}
&D_{j,r}(t_j^\star+\ep) = D_{j,r}(t_j^\star+\ep) - D_{j,r}(t_j^\star) + D_{j,r}(t_j^\star)\\
&=-\P_{j,X|A=1}\lsb\frac12 +t^\star <\eta_{j,1}(X)\le \frac12 +t^\star+\ep\rsb - \P_{j,X|A=0}\lsb\frac12 -t^\star -\ep\le \eta_{j,0}(X)< \frac12 -t^\star \rsb<0.
\end{align*}
Hence based on \eqref{eq:op-de-rule-dp_continuous}, if we  set $\tau_{j,1}^\star=\tau_{j,0}^\star=0$,
a fair Bayes-optimal classifier is given by
\begin{align}\label{fpmstar}
&f_{\pm 1}^\star(x,a)=I\lsb\eta_{\pm1,a}(x)>1/2+(2a-1)\lsb1/4\pm c_\beta(sv_d)^{d/\gamma}\rsb\rsb.
\end{align}
Now, we verify the distributional conditions:
\begin{itemize}
\item {\it Smoothness Condition from Definition \ref{h}}:
Since $\gamma\beta\le d$, 
for any $b \in \mathbb{N}^d$
such that $|b| \le \lbeta_+$, we have
that $|b|\le d/\gamma$. Thus, for $a\in\{0,1\}$, the partial derivative $D^b \eta_{j,a}$ exists and
is bounded by $\Pi_{j=1}^{|b|}\lsb d/\gamma-j+1\rsb \lsb |x_1|+s v_d\rsb^{d/\gamma-|b|}$.
As $|x_1|$ and $s$ are bounded, when  $c_\beta>0$ is small enough, for $j\in\{-1,1\}$ and $a\in\{0,1\}$, $\eta_{j,a}$ belongs to the H\"older
class $\Sigma(\beta,L_\beta,\MR^d
)$.

\item {\it Margin Condition  from Definition \ref{m}}: In this case, we have $D_{j,-}(t_j^\star)=0=D_{j,+}(t_j^\star)$ for $j=\{-1,1\}$.
To verify the margin condition, we need to provide both lower and upper bounds for 
$$g_{\delta,j}(t_j^\star,\ep)=\sum_a\P_{j,X|A=a}\lsb 0<|\eta_{j,a}-T_{j,a}^\star|< \ep\rsb \ \ \text{ for }  j=\{-1,1\}, \ep\in  (0,\ep_0],$$
with some $\ep_0>0$. 
We set $\ep_0= c_\beta\lsb  v_d\rsb^{{d}/{\gamma}}$.
By construction, we have for $j\in\{-1,1\}$ and all $x \in \X$ that $\mu_{j,1}(x)=\mu_{-j,0}(x)$ and $\eta_{-j,1}(x)=\eta_{-j,0}(x)-1/2$. Moreover,  $T^\star_{j,1}-T^\star_{-j,0}= 1/2+1/4+j\cdot c_\beta(sv_d)^{d/\gamma} -(1/2-(1/4 -j\cdot c_\beta(sv_d)^{d/\gamma})) =1/2$.
Thus,
\begin{align}\label{eq:equivalencej}
\nonumber&\P_{j,X|A=1}\lsb T^\star_{j,1}<\eta_{j,1}(X)<T^\star_{j,1}+\ep\rsb =\int_{\X} I\lsb T^\star_{j,1}<\eta_{j,1}(x)<T^\star_{j,1}+\ep \rsb \mu_{j,1}(x)dx\\
\nonumber&=\int_{\X} I\lsb T^\star_{j,1}-\frac12<\eta_{j,1}(x)-\frac12<T^\star_{j,1}+\ep-\frac12 \rsb \mu_{j,1}(x)dx\\
\nonumber&=\int_{\X} I\lsb T^\star_{-j,0}<\eta_{-j,0}(x) <T^\star_{-j,0}+\ep \rsb \mu_{-j,0}(x)dx\\
&=\P_{-j,X|A=0}\lsb T^\star_{-j,0}<\eta_{-j,0}(X)<T^\star_{-j,0}+\ep\rsb.
\end{align}
Again, by construction, we have for $a\in\{0,1\}$ and $x\in \mX$ that 
$\mu_{1,a}(x)=\mu_{1,1-a}(-x)$ and $\eta_{1,1-a}(-x)=1-\eta_{1,a}(x)$. Moreover,
$T_{1,a}^\star = 1/2 + (2a-1)t^\star_{1} =1 - (1/2 + (1-2a)t_1^\star) = 1-T_{1,1-a}^\star$.
Thus,
\begin{align}\label{eq:equivalencea}
\nonumber&\P_{1,X|A=a}\lsb T^\star_{1,a}<\eta_{1,a}(X)<T^\star_{1,a}+\ep\rsb =\int_{\X} I\lsb 1-T^\star_{1,a}-\ep<1-\eta_{1,a}(x)<1-T^\star_{1,a} \rsb \mu_{1,a}(x)dx\\
\nonumber&=\int_{\X} I\lsb T^\star_{1,1-a}-\ep<\eta_{1,1-a}(-x)<T^\star_{1,1-a} \rsb \mu_{1,1-a}(-x)dx\\
\nonumber&=\int_{\X} I\lsb T^\star_{1,1-a}-\ep<\eta_{1,1-a}(x)<T^\star_{1,1-a} \rsb \mu_{1,1-a}(x)dx\\
&=\P_{1,X|A=1-a}\lsb T^\star_{1,1-a}-\ep<\eta_{1,1-a}(X)<T^\star_{1,1-a}\rsb.
\end{align}
Thus,  for $j\in\{-1,1\}$,
\begin{align*}
&g_{\delta,j}(t_j^\star,\ep)=\sum_a\P_{j,X|A=a}\lsb 0<|\eta_{j,a}-T_{j,a}^\star|< \ep\rsb\\
&= \sum_a\lsb\P_{j,X|A=a}\lsb T_{j,a}^\star -\ep<\eta_{j,a}<T_{j,a}^\star\rsb +\P_{j,X|A=a}\lsb T_{j,a}^\star <\eta_{j,a}<T_{j,a}^\star+ \ep\rsb\rsb\\
&= 2\lsb\P_{1,X|A=1}\lsb T_{1,1}^\star -\ep<\eta_{1,1}<T_{1,1}^\star\rsb +\P_{1,X|A=1}\lsb T_{1,1}^\star <\eta_{1,1}<T_{1,1}^\star+ \ep\rsb\rsb,\end{align*}
where the last equality follows \eqref{eq:equivalencej} and \eqref{eq:equivalencea}. 
Next, we provide upper bounds for $\P_{1,X|A=1}\lsb T_{1,1}^\star -\ep<\eta_{1,1}<T_{1,1}^\star\rsb$ and  $\P_{1,X|A=1}\lsb T_{1,1}^\star <\eta_{1,1}<T_{1,1}^\star+ \ep\rsb$ when $0<\ep\le  c_\beta\lsb  v_d\rsb^{{d}/{\gamma}}$.
 
We first observe the  following two facts:
\begin{itemize}
\item Fact 1. By construction, we have that
\begin{align}\label{eq:subsets}
\nonumber&\mathcal{B}_1 \subset \{x: -(2+s)v_d\le x_1\le -v_d\}, \ \ \mathcal{B}_2 \subset \{x: -sv_d\le x_1\le sv_d\},\\
&\text{and }  \mathcal{B}_3 \subset \{x:  v_d\le x_1\le (2+s)v_d\}.
&\end{align}
\item Fact 2. Recalling that
for any $d\ge 1$,
$V_{d,1}$ is the volume of the  unit $\ell_1$ ball, we have
\begin{align}\label{eq:Vod1}
  \nonumber &V_{d,1} =\int_{\sum_{j=1}^d|x_j|\le 1}dx =
\int_{-1}^1\lsb \int_{\sum_{j=2}^d|x_j|\le 1-|x_1|} dx_2\ldots dx_d\rsb dx_1\\
  \nonumber&= 2\int_{0}^1\lsb \int_{\sum_{j=2}^d|x_j|\le 1-x_1} dx_2\ldots dx_d\rsb dx_1
= 2\int_{0}^1\lsb \int_{\sum_{j=2}^d|x_j|\le x_1} dx_2\ldots dx_d\rsb dx_1\\
&=2V_{d-1,1}\int_{0}^1 x_1^{d-1} dx_1 = \frac{2V_{d-1,1}}{d}.
\end{align}
This further implies $V_{d,1} = 2^d/(d!)$ and $v_d=(d!)^{1/d}/2 $.
\end{itemize}

For $\P_{1,X|A=1}\lsb T_{1,1}^\star<\eta_{1,1}<T_{1,1}^\star+\ep\rsb$,
    since  $\eta_{1,1}(x)$ only depends on $x_1$, is continuous, and is strictly increasing as a function of $x_1$, its inverse $\eta^{-1}_{1,1}$ 
    as a function of $x_1$
    exists with $\eta^{-1}_{1,1}(T_{1,1}^\star)=sv_d$.
    Moreover, for $0<\ep\le\ep_0$,
    $\eta^{-1}_{1,1}(T_{1,1}^\star +\ep)=sv_d+ \lsb{\ep}/{c_\beta}\rsb^{\gamma/d}$.  
    Thus,  $ T^\star_{1,1}<\eta_{1,1}(x)<T^\star_{1,1}+\ep$ is equivalent to $sv_d< x_1< sv_d+\lsb{\ep}/{c_\beta}\rsb^{\gamma/d}$. 
    Further, by \eqref{eq:subsets}, we have $\mathcal{B}_j\cap\{sv_d< x_1< sv_d+\lsb{\ep}/{c_\beta}\rsb^{\gamma/d}\}=\emptyset$ for $j=1,2$. Thus,
 for  $0<\ep \le \ep_0$,
  \begin{align}\label{eq:ualb0}
  &\P_{1,X|A=1}\lsb T^\star_{1,1}<\eta_{1,1}(X)<T^\star_{1,1}+\ep\rsb=  \P_{1,X|A=1}\lsb sv_d< X_1<  sv_d+\lsb\frac{\ep}{c_\beta}\rsb^{\gamma/d}\rsb\\
  \nonumber  &=  \int I\lsb sv_d< x_1< sv_d+\lsb \frac{\ep}{c_\beta}\rsb^{\gamma/d}\rsb \mu_{1,1}(x)dx 
  = \sum_{j=1}^3 \int_{\mathcal{B}_j} I\lsb sv_d< x_1< sv_d+\lsb\frac{\ep}{c_\beta}\rsb^{\gamma/d}\rsb \mu_{1,1}(x)dx\\
     \nonumber   &=  \frac12 \int_{\mathcal{B}_3} I\lsb sv_d< x_1<sv_d+ \lsb\frac{\ep}{c_\beta}\rsb^{\gamma/d} \rsb  dx.
\end{align}  
Since $0<\ep\le  c_\beta\lsb  v_d\rsb^{{d}/{\gamma}}$,
and recalling \eqref{eq:Vod1},
this further equals
  \begin{align}\label{eq:ualb1}
\nonumber& \frac12 \int_{sv_d}^{sv_d+\lsb\frac{\ep}{c_\beta}\rsb^{\gamma/d}} \lsb \int_{\sum_{j=2}^d|x_j|\le v_d -| x_1-(1+s)v_d|} dx_2\ldots dx_d\rsb dx_1\\
 \nonumber&= \frac12 \int_{sv_d}^{sv_d+\lsb\frac{\ep}{c_\beta}\rsb^{\gamma/d}} \lsb \int_{\sum_{j=2}^d|x_j|\le   x_1- sv_d} dx_2\ldots dx_d\rsb dx_1
 = \frac12 \int_{0}^{\lsb\frac{\ep}{c_\beta}\rsb^{\gamma/d} } \lsb \int_{\sum_{j=2}^d|x_j|\le x_1} dx_2\ldots dx_d\rsb dx_1\\
&= \frac12 \int_{0}^{\lsb\frac{\ep}{c_\beta}\rsb^{\gamma/d} }V_{d-1,1}x_1^{d-1}dx_1=\frac{V_{d-1,1}}{2d}\lsb\frac{\ep}{c_\beta}\rsb^\gamma= \frac{V_{d,1}}{4}\lsb\frac{\ep}{c_\beta}\rsb^\gamma.
\end{align}  

To bound $\P_{1,X|A=1}\lsb T_{1,1}^\star -\ep<\eta_{1,1}<T_{1,1}^\star\rsb$, 
we
first study the inverse $\eta_{1,1}^{-1}$ of $\eta_{1,1}$, viewed as a function of $x_1$.
We have $\eta^{-1}_{1,1}(T_{1,1}^\star)=sv_d$ and
$$\eta^{-1}_{1,1}(T_{1,1}^\star-\ep)=
\left\{\begin{array}{ll}
    sv_d -\lsb\frac{\ep}{c_\beta}\rsb^{\gamma/d}, & 0<\ep\le c_\beta(sv_d)^{d/\gamma};\\
\lsb 2\lsb sv_d\rsb^{d/\gamma} - \frac{\ep}{c_\beta}\rsb^{\gamma/d} -sv_d, &c_\beta(sv_d)^{d/\gamma}< \ep\le 2c_\beta(sv_d)^{d/\gamma};\\
 -\lsb \frac{\ep}{c_\beta}- 2\lsb sv_d\rsb^{d/\gamma} \rsb^{\gamma/d} -sv_d,     & 2c_\beta(sv_d)^{d/\gamma}< \ep \le 2c_\beta(sv_d)^{d/\gamma} + c_\beta(2v_d)^{d/\gamma}.
\end{array}\right.$$
Thus, for $0<\ep\le 2c_\beta(sv_d)^{d/\gamma} + c_\beta(2v_d)^{d/\gamma}$, $T^\star_{1,1}-\ep<\eta_{1,1}(x)<T^\star_{1,1}$ is equivalent to
$$\left\{\begin{array}{ll}
     sv_d -\lsb\frac{\ep}{c_\beta}\rsb^{\gamma/d}<x_1<sv_d, &  \text{ when } 0<\ep\le c_\beta(sv_d)^{d/\gamma};\\
\lsb 2\lsb sv_d\rsb^{d/\gamma} - \frac{\ep}{c_\beta}\rsb^{\gamma/d} -sv_d<x_1<sv_d, &   \text{ when } c_\beta(sv_d)^{d/\gamma}< \ep\le 2c_\beta(sv_d)^{d/\gamma};\\
 -\lsb \frac{\ep}{c_\beta}- 2\lsb sv_d\rsb^{d/\gamma} \rsb^{\gamma/d} -sv_d<x_1<sv_d,     & \text{ when } 2c_\beta(sv_d)^{d/\gamma}< \ep \le 2c_\beta(sv_d)^{d/\gamma} + c_\beta(2v_d)^{d/\gamma}.
\end{array}\right.
$$
Now we consider several cases.
\begin{itemize}
\item (1)  $0< \ep\le c_\beta(sv_d)^{d/\gamma}$. We can write
\begin{align*}
    &\P_{1,X|A=1}\lsb T^\star_{1,1}-\ep<\eta_{1,1}(X)<T^\star_{1,1}\rsb =  \P_{1,X|A=1}\lsb sv_d -\lsb\frac\ep{c_\beta}\rsb^{\gamma/d}< X_1<   sv_d\rsb\\
    &=  \int I\lsb sv_d-\lsb\frac{\ep}{c_\beta}\rsb^{\gamma/d}< x_1<  sv_d\rsb \mu_{1,1}(x)dx\\
    & = \sum_{j=1}^3 \int_{\mathcal{B}_j} I\lsb sv_d-\lsb\frac{\ep}{c_\beta}\rsb^{\gamma/d}< x_1<  sv_d\rsb \mu_{1,1}(x)dx.
\end{align*}
In this case, we have
$sv_d -\lsb{\ep}/{c_\beta}\rsb^{{\gamma}/{d}}\ge 0$. Using \eqref{eq:subsets}, we have $\mathcal{B}_j \cap \{ x: sv_d -\lsb{\ep}/{c_\beta}\rsb^{{\gamma}/{d}}<x_1<sv_d\}=\emptyset$ for $j=1,3$. Thus,
this further equals
\begin{align*}    
       &  \int_{\mathcal{B}_2} I\lsb sv_d-\lsb\frac{\ep}{c_\beta}\rsb^{\gamma/d}< x_1<  sv_d\rsb  \frac{1}{2+2s^d}dx\\
& = \frac{1}{2+2s^d}\int_{sv_d-\lsb\frac{\ep}{c_\beta}\rsb^{\gamma/d}}^{ sv_d} \lsb \int_{\sum_{j=2}^d|x_j|\le sv_d-|x_1|} dx_2\ldots dx_d\rsb dx_1\\
& = \frac{1}{2+2s^d} \int_{0}^{\lsb\frac{\ep}{c_\beta}\rsb^{\gamma/d} } \lsb \int_{\sum_{j=2}^d|x_j|\le x_1} dx_2\ldots dx_d\rsb dx_1\\
&=\frac{V_{d-1,1}}{2+2s^d}\int_{0}^{\lsb\frac{\ep}{c_\beta}\rsb^{\gamma/d} }x_1^{d-1} dx_1=\frac{V_{d,1}}{4+4s^d}\lsb\frac{\ep}{c_\beta}\rsb^\gamma.
\end{align*}
As $1\le 1+s^d\le 2$ when $0\le s\le 1$, we thus have, for $0< \ep\le c_\beta(sv_d)^{d/\gamma}$,
$$\frac{V_{d,1}}{8}\lsb\frac{\ep}{c_\beta}\rsb^\gamma\le \P_{1,X|A=1}\lsb T^\star_{1,1}-\ep<\eta_{1,1}(X)<T^\star_{1,1}\rsb \le \frac{V_{d,1}}{4}\lsb\frac{\ep}{c_\beta}\rsb^\gamma.$$

\item  (2)  $c_\beta(sv_d)^{d/\gamma}< \ep\le  2c_\beta(sv_d)^{d/\gamma}$. In this case, we have
$-sv_d\le \lsb 2\lsb sv_d\rsb^{d/\gamma} - \frac{\ep}{c_\beta}\rsb^{\gamma/d} -sv_d< 0$. 
Again, by \eqref{eq:subsets}, we have $\mathcal{B}_j \cap \{ x: sv_d -\lsb\ep/c_\beta\rsb^{\gamma/d}<x_1<sv_d\}=\emptyset$ for $j=1,3$. Thus,
\begin{align}
    &\P_{1,X|A=1}\lsb T^\star_{1,1}-\ep<\eta_{1,1}(X)<T^\star_{1,1}\rsb =  \P_{1,X|A=1}\lsb \lsb 2\lsb sv_d\rsb^{d/\gamma} - \frac{\ep}{c_\beta}\rsb^{\gamma/d} -sv_d< X_1<   sv_d\rsb\nonumber\\
    &=  \int I\lsb \lsb 2\lsb sv_d\rsb^{d/\gamma} - \frac{\ep}{c_\beta}\rsb^{\gamma/d} -sv_d< x_1<   sv_d\rsb \mu_{1,1}(x)dx\nonumber\\
    & = \sum_{j=1}^3 \int_{\mathcal{B}_j} I\lsb \lsb 2\lsb sv_d\rsb^{d/\gamma} - \frac{\ep}{c_\beta}\rsb^{\gamma/d} -sv_d< x_1<   sv_d\rsb \mu_{1,1}(x)dx\nonumber\\
       & =   \int_{\mathcal{B}_2} I\lsb \lsb 2\lsb sv_d\rsb^{d/\gamma} - \frac{\ep}{c_\beta}\rsb^{\gamma/d} -sv_d< x_1<   sv_d\rsb  \frac{1}{2+2s^d}dx.\label{tminepbd}
\end{align}
On one hand, since  $I\lsb \lsb 2\lsb sv_d\rsb^{d/\gamma} - \ep/c_\beta\rsb^{\gamma/d} -sv_d< x_1<   sv_d\rsb\le 1$,
this is upper bounded by
\begin{align*}
       & \int_{\mathcal{B}_2}   \frac{1}{2+2s^d}dx =    \frac{1}{2+2s^d}\lambda ({\mathcal{B}_2} )  = \frac{V_{d,1}}{2+2s^d} (sv_d)^d <  \frac{V_{d,1}}{2} \lsb\frac{\ep}{c_\beta}\rsb^\gamma,
\end{align*}
where  the
last inequality holds since $1+s^d>1$ and $(\ep/c_\beta)^\gamma>(sv_d)^{d}$ when $ \ep>c_\beta(sv_d)^{d/\gamma}$.

On the other hand, since $I\lsb \lsb 2\lsb sv_d\rsb^{d/\gamma} - \ep/c_\beta\rsb^{\gamma/d} -sv_d< x_1<   sv_d\rsb\ge I\lsb  0< x_1<   sv_d\rsb$ when  $\lsb 2\lsb sv_d\rsb^{{d}/{\gamma}} - {\ep}/{c_\beta}\rsb^{{\gamma}/{d}} -sv_d<0$, 
the \eqref{tminepbd} is lower bounded by 
\begin{align*}
        &\int_{\mathcal{B}_2} I\lsb  0< x_1<   sv_d\rsb  \frac{1}{2+2s^d}dx= \frac{1}{2}  \int_{\mathcal{B}_2}   \frac{1}{2+2s^d}dx =    \frac{1}{4+4s^d}\lambda ({\mathcal{B}_2} ) \\
        &= \frac{V_{d,1}}{4+4s^d} (sv_d)^d \ge  \frac{V_{d,1}}{2^{3+\gamma}} \lsb\frac{\ep}{c_\beta}\rsb^\gamma,
\end{align*}
where the  
last inequality holds since $1+s^d<2$ and $(\ep/c_\beta)^\gamma\le 2^\gamma(sv_d)^{d}$ when $ \ep \le 2c_\beta(sv_d)^{d/\gamma}$.

As a result, we have, for  $c_\beta(sv_d)^{d/\gamma}< \ep\le  2c_\beta(sv_d)^{d/\gamma}$,
$$\frac{V_{d,1}}{2^{3+\gamma}}\lsb\frac{\ep}{c_\beta}\rsb^\gamma\le \P_{1,X|A=1}\lsb T^\star_{1,1}-\ep<\eta_{1,1}(X)<T^\star_{1,1}\rsb \le \frac{V_{d,1}}{2}\lsb\frac{\ep}{c_\beta}\rsb^\gamma.$$

\item  (3)  $2c_\beta(sv_d)^{d/\gamma}< \ep\le  2c_\beta(sv_d)^{d/\gamma} + c_\beta(2v_d)^{d/\gamma}$. 
In this case, we have
$ -\lsb {\ep}/{c_\beta}- 2\lsb sv_d\rsb^{{d}/{\gamma}} \rsb^{{\gamma}/ {d}} -sv_d< -sv_d$. 
By \eqref{eq:subsets},  
$\mathcal{B}_2 \cap \biggl\{ x: -\lsb {\ep}/{c_\beta}- 2\lsb sv_d\rsb^{{d}/{\gamma}} \rsb^{{\gamma}/ {d}} -sv_d
<x_1<sv_d\biggr\}=\textnormal{int}\mathcal{B}_2$,
where $\textnormal{int}$ denotes the interior of a set,
and 
$\mathcal{B}_3 \cap \biggl\{ x: -\lsb {\ep}/{c_\beta}- 2\lsb sv_d\rsb^{{d}/{\gamma}} \rsb^{{\gamma}/ {d}} -sv_d<x_1<sv_d\biggr\}=\emptyset$. 
 Thus,
\begin{align*}
    &\P_{1,X|A=1}\lsb T^\star_{1,1}-\ep<\eta_{1,1}(X)<T^\star_{1,1}\rsb =  \P_{1,X|A=1}\lsb -\lsb \ep/c_\beta- 2\lsb sv_d\rsb^{d/\gamma} \rsb^{\gamma/d} -sv_d< X_1<   sv_d\rsb\\
    &=  \int I\lsb -\lsb \ep/c_\beta- 2\lsb sv_d\rsb^{d/\gamma} \rsb^{\gamma/d} -sv_d< x_1<  sv_d\rsb \mu_{1,1}(x)dx\\
    & = \sum_{j=1}^3 \int_{\mathcal{B}_j} I\lsb -\lsb \ep/c_\beta- 2\lsb sv_d\rsb^{d/\gamma} \rsb^{\gamma/d} -sv_d< x_1<  sv_d\rsb \mu_{1,1}(x)dx\\
       & = \int_{\mathcal{B}_1} I\lsb -\lsb \ep/c_\beta- 2\lsb sv_d\rsb^{d/\gamma} \rsb^{\gamma/d} -sv_d< x_1<  sv_d\rsb  \frac{1}{2+2s^d}dx+ \int_{\textnormal{int}\mathcal{B}_2}  \frac{1}{2+2s^d}dx.
\end{align*}
The first term further equals
\begin{align*}
    &\frac{1}{2+2s^d}\int_{-\lsb \ep/c_\beta- 2\lsb sv_d\rsb^{d/\gamma} \rsb^{\gamma/d} -sv_d}^{ -sv_d} \lsb \int_{\sum_{j=2}^d|x_j|\le v_d-|x_1+(1+s)v_d|} dx_2\ldots dx_d\rsb dx_1\\
    & = \frac{1}{2+2s^d}\int_{-\lsb \ep/c_\beta- 2\lsb sv_d\rsb^{d/\gamma} \rsb^{\gamma/d} -sv_d}^{ -sv_d} \lsb \int_{\sum_{j=2}^d|x_j|\le -x_1-sv_d} dx_2\ldots dx_d\rsb dx_1\\
   & = \frac{1}{2+2s^d}\int^{\lsb \ep/c_\beta- 2\lsb sv_d\rsb^{d/\gamma} \rsb^{\gamma/d} }_{0} \lsb \int_{\sum_{j=2}^d|x_j|\le x_1} dx_2\ldots dx_d\rsb dx_1\\
   & = \frac{V_{d,1}}{4+4s^d}\lsb \ep/c_\beta- 2\lsb sv_d\rsb^{d/\gamma} \rsb^{\gamma}.
\end{align*}
We have on one hand, 
\begin{equation*}
 \frac{s^d}{2+2s^d}+ \frac{V_{d,1}}{4+4s^d}\lsb \frac{\ep}{c_\beta}- 2\lsb sv_d\rsb^{d/\gamma} \rsb^{\gamma}  <\frac{V_{d,1}}{2+2s^d} {(sv_d)^d} +\frac{V_{d,1}}{4+4s^d}\lsb \frac\ep{c_\beta}\rsb^\gamma \le \frac{3V_{d,1}}{4}\lsb \frac\ep{c_\beta}\rsb^\gamma,
\end{equation*}
where the last inequality holds since $1+s^d>1$ and $\lsb \ep/c_\beta\rsb^{\gamma}>(sv_d)^d $ when $ \ep>2c_\beta(sv_d)^{d/\gamma}>c_\beta(sv_d)^{d/\gamma}$.
On the other hand, we have, when $2c_\beta(sv_d)^{d/\gamma}<\ep\le 4c_\beta(sv_d)^{d/\gamma}$,
\begin{equation*}
 \frac{s^d}{2+2s^d}+ \frac{V_{d,1}}{4+4s^d}\lsb \frac{\ep}{c_\beta}- 2\lsb sv_d\rsb^{d/\gamma} \rsb^{\gamma} \ge \frac{s^d}{2+2s^d} = \frac{V_{d,1}}{2+2s^d} (sv_d)^d \ge  \frac{V_{d,1}}{2^{2+2\gamma}}\lsb\frac\ep{c_\beta}\rsb^\gamma,
\end{equation*}
and when $4c_\beta(sv_d)^{d/\gamma}<\ep$,
\begin{equation*}
 \frac{s^d}{2+2s^d}+ \frac{V_{d,1}}{4+4s^d}\lsb \frac{\ep}{c_\beta}- 2\lsb sv_d\rsb^{d/\gamma} \rsb^{\gamma} \ge \frac{V_{d,1}}{4+4s^d}\lsb \frac{\ep}{2c_\beta} \rsb^{\gamma} \ge \frac{V_{d,1}}{2^{3+\gamma}}\lsb\frac\ep{c_\beta}\rsb^\gamma,
\end{equation*}
As a result,  we have, for  $2c_\beta(sv_d)^{d/\gamma}< \ep\le  2c_\beta(sv_d)^{d/\gamma} + c_\beta(2v_d)^{d/\gamma}$,
$$\frac{V_{d,1}}{2^{3+2\gamma}}\lsb\frac{\ep}{c_\beta}\rsb^\gamma\le \P_{1,X|A=1}\lsb T^\star_{1,1}-\ep<\eta_{1,1}(X)<T^\star_{1,1}\rsb \le \frac{3V_{d,1}}{4}\lsb\frac{\ep}{c_\beta}\rsb^\gamma.$$
\end{itemize}
In particular, 
 we have, for  $0< \ep\le  c_\beta v_d^{d/\gamma}$,
\begin{equation}\label{eq:ualb2}
\frac{V_{d,1}}{2^{3+2\gamma}}\lsb\frac{\ep}{c_\beta}\rsb^\gamma\le \P_{1,X|A=1}\lsb T^\star_{1,1}-\ep
<\eta_{1,1}(X)<T^\star_{1,1}\rsb \le \frac{3V_{d,1}}{4}\lsb\frac{\ep}{c_\beta}\rsb^\gamma.
\end{equation}

Combining \eqref{eq:ualb0},  \eqref{eq:ualb1} and \eqref{eq:ualb2}, we get,
for $j\in\{-1,+1\}$  and $0<\ep\le  c_\beta v_d^{d/\gamma}$,
\begin{align*}
    V_{d,1}(2^{-1}+2^{-(2+2\gamma)})
    \lsb\frac{\ep}{c_\beta}\rsb^\gamma \le g_{\delta,j}(t_j^\star,\ep)=2 \P_{1,X|A=1}\lsb 0<|\eta_{1,1}-T_{1,1}^\star|< \ep\rsb \le 2V_{d,1}\lsb\frac{\ep}{c_\beta}\rsb^\gamma.
\end{align*} 
Thus, for $0<x\le 1/2$,
$g_{\delta,j}^{-1}(t_{j}^\star,x)\le c_\beta\lsb 2x/V_{d,1}\rsb ^{1/\gamma}$, 
and the $\gamma$-margin condition from Definition \ref{m} is satisfied.

\item {\it Strong density condition from Definition \ref{sd}}:
Denote $C_d=[-3,3]^d$ and by $\Omega_\mu = \cup_{j=1}^3 \mathcal{B}_j$ the support of $\mu_{\pm 1,a}$. By construction, we have, for $a\in\{0,1\}$ and $j\in\{-1,1\}$, since $s\le 1$,
$$\frac{1}{4}\le \frac{1}{2+2s^d}\le \mu_{j,a}(x)\le \frac{1}{2}.$$
Thus, we can take $\mu_{\min}=1/4$ and $\mu_{\max}=1/2$.
We then show that  $\Omega_\mu$ is a regular set by considering the following two cases: (1) $x\in\mathcal{B}_1 \cup \mathcal{B}_3$
and (2) $x\in\mathcal{B}_2$.
Let $\Xi_d=V_{d,1}/(V_{d,2}2^d)$.
\begin{itemize}
\item (1) When $x\in\mathcal{B}_1 \cup \mathcal{B}_3$: 
Without loss of generality, we assume $x\in\mathcal{B}_1$.
By Lemma \ref{lem:interd1}
where $\mathcal{B}_1= B_{d,1}( z,R)$
with $z = ((1+s)v_d,\ldots,0)^\top$ and $R = v_d$, 
we have, when $r\le 2v_d/(d+2)$, 
$$
\lambda[B_{d,2}(x,r)\cap \Omega_\mu]\ge \lambda[B_{d,2}(x,r)\cap \mathcal{B}_1]\ge \Xi_d\lambda[B_{d,2}(x,r)] 
\ge\Xi_d\lambda[B_{d,2}(x,r)\cap C_d].
$$

\item (2) When $x\in\mathcal{B}_2$:
Without loss of generality, we assume $x_1>0$. 
\begin{itemize}
\item  When $r\le 2sv_d/(d+2)$, by Lemma \ref{lem:interd1}
where $\mathcal{B}_2= B_{d,1}( z,R)$
with $z = 0$ and $R = sv_d$,  
we have
$$\lambda[B_{d,2}(x,r)\cap \Omega_\mu]
\ge \lambda[B_{d,2}(x,r)\cap \mathcal{B}_2]\ge \Xi_d\lambda[B_{d,2}(x,r)]\ge \Xi_d\lambda[B_{d,2}(x,r)\cap C_d].
$$

\item When $2sv_d/(d+2)<r\le 2\sqrt2sv_d$, we have, by Lemma \ref{lem:interd1}
where $\mathcal{B}_2= B_{d,1}( z,R)$
with $z = 0$ and $R = sv_d$,
\begin{align*}
   &\lambda[B_{d,2}(x,r)\cap \Omega_\mu]
\ge \lambda[B_{d,2}(x,2sv_d/(d+2))\cap \mathcal{B}_2]\\&\ge  \Xi_d\lambda[B_{d,2}(x,2sv_d/(d+2))]
=
\frac{\Xi_d}{2^{d/2}(d+2)^d}\lambda[B_{d,2}(x,2\sqrt2sv_d)]\\
&\ge \frac{\Xi_d}{2^{d/2}(d+2)^d}\lambda[B_{d,2}(x,r)]
\ge\frac{\Xi_d}{2^{d/2}(d+2)^d}\lambda[B_{d,2}(x,r)\cap C_d].
\end{align*}

\item When $\sqrt2{s}v_d<r/2<v_d/(d+2)$, we denote $z=(sv_d,0,\ldots,0)^\top\in \mathcal{B}_3$. As $x_1>0$, we have $z\in B_{2,d}(x,\sqrt{2}sv_d)$. 
Noting that when $r>2\sqrt{2}sv_d$, $r-\sqrt{2}sv_d>r/2$, 
this further implies $B_{2,d}(z, {r}/{2}) \subset B_{2,d}(z, r-\sqrt{2}sv_d) \subset B_{2,d}(x,r)$. 
Now, since $z \in \mathcal{B}_1$, by Lemma \ref{lem:interd1} again,
as $\mathcal{B}_3= B_{d,1}( z,R)$
with $z = (-(1+s)v_d,\ldots,0)^\top$ and $R = v_d$, 
we have
$$\lambda [B_{2,d}(x, r)\cap \Omega_\mu]\ge \lambda [B_{2,d}(z, r/2)\cap \mathcal{B}_3] \ge \Xi_d \lambda[\mathcal{B}_1]\ge \Xi_d \lambda[\mathcal{B}_1\cap C_d].$$ 

\end{itemize}

\end{itemize}
The strong density condition is thus satisfied.

\end{itemize}

Using \eqref{etaj}, 
and in particular that $\eta_{1,1} = \eta_{-1,1}$ or $\P_{1,Y|X,A=a}(y)=\P_{-1,Y|X,A=a}(y)$, 
the 
Kullback–Leibler divergence  between $\P_{1}$ and $\P_{-1}$ can be expressed
as
\begin{align*}
&\textup{KL}(\P_{1},\P_{-1})=\int_{\X\times \A\times\mathcal{Y}}\log \frac{d\P_{1}(x,a,y)}{d\P_{-1}(x,a,y)}d\P_{1}(x,a,y)\\
&=\int_{\X \times\A\times\mathcal{Y}}\log \frac{d\P_{1,A}(a)d\P_{1,X|A=a}(x)d\P_{1,Y|X,A=a}(y)}{d\P_{-1,A}(a)d\P_{-1,X|A=a}(x)d\P_{-1,Y|X,A=a}(y)}d\P_{1}(x,a,y)\\
&=\sum_{a\in\{0,1\}}\P_{1}(A=a)\int_{\X}\log \frac{d\P_{1,X|A=a}(x)}{d\P_{-1,X|A=a}(x)}d\P_{1,X|A=a}(x).
\end{align*}
Using the definition of $\mu_{j,a}$,
and in particular that
for all $a$,
$\mu_{1,a} \neq \mu_{-1,a}$ only for $x\in  \mathcal{B}_1 \cup  \mathcal{B}_3$,
as well as that $\lambda[ \mathcal{B}_1] = \lambda[ \mathcal{B}_3] = 1$,
this further equals
\begin{align*}
&\sum_{a\in\{0,1\}}\frac12\int \log \frac{\mu_{1,a}(x)}{\mu_{-1,a}(x)}\mu_{1,a}(x)dx\\
&=\frac12\int_{  \mathcal{B}_1 \cup  \mathcal{B}_3}\lmb \frac{1}{2}\log(1+s^d) - \frac{1}{2+2s^d} \log(1+s^d)\rmb dx
=\frac{s^d}{2+2s^d}\log(1+s^d)\le  Cs^{2d}.
\end{align*}
By Pinsker's inequality, we have
\begin{align}\label{pin}
&\textup{TV}\lsb \P^{\otimes n}_{1},\P^{\otimes n}_{-1}\rsb\le\frac12\sqrt{\textup{KL}\lsb \P^{\otimes n}_{1},\P^{\otimes n}_{-1}\rsb}
=\frac12\sqrt{n\textup{KL}\lsb \P_{1},\P_{-1}\rsb}\le C'\sqrt{n}s^d.
\end{align}
Recall that  $\nu$ denotes the distribution of a Rademacher variable.
We have, by \eqref{eq:measure}, 
     \begin{align}
&\sup_{\P\in\mathcal{P}_\Sigma}\E_\P d_{E}(\widehat{f}_{\delta,n},f^\star) \ge
\sup_{j\in\{-1,1\}} \E^{\otimes n}_{\P_j}d_E \lsb\widehat{f}_{\delta,n}, f_{j}^\star\rsb\nonumber\\
&= \sup_{j\in\{-1,1\}}\E^{\otimes n}_{\P_j} \sum_{a\in\{0,1\}} \lbb\int_{\cup_{j=1}^3 \mathcal{B}_j} \lsb \widehat{f}_{\delta,n}(x,a)-f_{j}^\star(x,a)\rsb\lsb T_{j,a}^\star-\eta_{j,a}(x) \rsb  \mu_{j,a}(x)dx\rbb\nonumber\\
&\ge \sup_{j\in\{-1,1\}}\E^{\otimes n}_{\P_j} \sum_{a\in\{0,1\}} \lbb\int_{ \mathcal{B}_2}\lsb \widehat{f}_{\delta,n}(x,a)-f_{j}^\star(x,a)\rsb\lsb T_{j,a}^\star-\eta_{j,a}(x) \rsb  \mu_{j,a}(x)dx\rbb\nonumber\\
&\ge \E_{\nu}\E^{\otimes n}_{\P_\nu} \sum_{a\in\{0,1\}} \lbb\int_{ \mathcal{B}_2}\lsb \widehat{f}_{\delta,n}(x,a)-f_{\nu}^\star(x,a)\rsb\lsb T_{\nu,a}^\star-\eta_{\nu,a}(x) \rsb  \mu_{\nu,a}(x)dx\rbb.\label{delb}
\end{align}
Here, the second to last inequality holds since $\lsb \widehat{f}_{\delta,n}(x,a)-f_{j}^\star(x,a)\rsb\lsb T_{j,a}^\star-\eta_{j,a}(x) \rsb \ge 0$, which follows the fact that $\widehat{f}_{\delta,n}(x,a)\in[0,1]$, 
while $f_{j}^\star(x,a) =1$ when $\eta_{j,a}(x)>  T_{j,a}^\star$
and $f_{j}^\star(x,a) =0$ when $\eta_{j,a}(x)<  T_{j,a}^\star$.

Now, define the distribution $\P_0=(\P_{1}+\P_{-1})/2$. We have  that $\P^{\otimes n}_{1}$ and $\P^{\otimes n}_{-1}$ are absolutely continuous with respect to $\P^{\otimes n}_0$. Moreover, 
by \eqref{eq:subsets}, 
 \begin{align*} 
\eta_{j,a}(x)=\left\{\begin{array}{lll}
   1/2 +(2a-1)/4 - c_\beta(sv_d)^{d/\gamma} - c_\beta (-x_1-sv_d)^{d/\gamma}, & x\in \mathcal{B}_1;\\
    1/2 +(2a-1)/4 -c_\beta (sv_d)^{d/\gamma} +c_\beta (x_1+sv_d)^{d/\gamma}, & x\in \mathcal{B}_2 \cap \{x_1 < 0\};\\
   1/2 +(2a-1)/4 + c_\beta(sv_d)^{d/\gamma} -c_\beta (-x_1+sv_d)^{d/\gamma}, &x\in \mathcal{B}_2 \cap \{x_1 \ge 0\};\\
   1/2 +(2a-1)/4 +c_\beta (sv_d)^{d/\gamma} + c_\beta(x_1-sv_d)^{d/\gamma}, & x\in \mathcal{B}_3.
    \end{array}\right.
\end{align*}
In particular, we have, for $x\in\mathcal{B}_2$,
 \begin{equation}\label{boundcase22} 
\eta_{j,a}(x)=
    1/2 +(2a-1)/4 +\sign(x_1)c_\beta (sv_d)^{d/\gamma} -\sign(x_1)c_\beta (-|x_1|+sv_d)^{d/\gamma}.
  \end{equation}
  and
 \begin{equation*} 
  1/2 +(2a-1)/4 -c_\beta (sv_d)^{d/\gamma} \le \eta_{j,a}(x)\le    1/2 +(2a-1)/4 + c_\beta (sv_d)^{d/\gamma} .
  \end{equation*}
By \eqref{fpmstar}, for $x\in\textnormal{int}(\mathcal{B}_2)$, we thus have
\begin{equation*}
f_{+1}^\star(x,1) =f_{-1}^\star(x,0)= 0 \ \ \text{and} 
   \ \  f_{-1}^\star(x,1) =f_{+1}^\star(x,0) =1.
\end{equation*}
In other words, we have $f_{j}^\star(x,a) = I(j+1\neq 2a)$
on $\textnormal{int}(\mathcal{B}_2)$.
Moreover, since the boundary of $\mathcal{B}_2$ has zero measure, we can set $f_{j}^\star(x,a)$ arbitrarily on this boundary.
Thus,
\eqref{delb} can be written and bounded as
\begin{align}\label{eq:measureequ}
\nonumber&\E_{\nu} \sum_{a\in\{0,1\}}\E^{\otimes n}_{\P_0} \lsb\frac{d\P^{\otimes n}_{\nu}}{d\P^{\otimes n}_{0}}\rsb 
\lbb\int_{ \mathcal{B}_2}\lsb \widehat{f}_{\delta,n}(x,a)-I(\nu+1\neq 2a)\rsb \lsb T^\star_{\nu,a}-\eta_{\nu,a}(x)\rsb\mu_{\nu,a}(x)dx\rbb\\
\nonumber&\ge \sum_{a\in\{0,1\}}\E^{\otimes n}_{\P_0} \lsb\frac{d\P^{\otimes n}_{1}}{d\P^{\otimes n}_{0}}\wedge \frac{d\P^{\otimes n}_{-1}}{d\P^{\otimes n}_{0}}\rsb 
\E_{\nu} \lbb\int_{ \mathcal{B}_2}\lsb \widehat{f}_{\delta,n}(x,a)-I(\nu+1\neq 2a)\rsb \lsb T^\star_{\nu,a}-\eta_{\nu,a}(x)\rsb\mu_{\nu,a}(x)dx\rbb\\
\nonumber&= \E_{\P_0}(1-\textup{TV}\lsb\P^{\otimes n}_{1},\P^{\otimes n}_{-1}\rsb 
\sum_{a\in\{0,1\}}\E_{\nu} \lbb\int_{ B_2}\lsb \widehat{f}_{\delta,n}(x,a)-I(\nu+1\neq 2a)\rsb \lsb T^\star_{\nu,a}-\eta_{\nu,a}(x)\rsb\mu_{\nu,a}(x)dx\rbb.\\
&~
\end{align}
Then, if we take $s \asymp n^{-1/(2d)}$ 
with $\textup{TV}\lsb \P^{\otimes n}_{1},\P^{\otimes n}_{-1}\rsb\le C'\sqrt{n}s^d<1/2$,  by \eqref{pin},
\eqref{eq:measureequ} is further lower bounded by 
\begin{align*}
&\frac12\sum_{a\in\{0,1\}}\E_{\nu} \lbb\int_{ B_2}\lsb \widehat{f}_{\delta,n}(x,a)-I(\nu+1\neq 2a)\rsb \lsb T^\star_{\nu,a}-\eta_{\nu,a}(x)\rsb\mu_{\nu,a}(x)dx\rbb\\
&=\frac14 \int_{ B_2}\lsb \widehat{f}_{\delta,n}(x,1)-I(1+1\neq 2)\rsb \lsb T^\star_{1,1}-\eta_{1,1}(x)\rsb\mu_{1,1}(x)dx\\
&+\frac14 \int_{ B_2}\lsb \widehat{f}_{\delta,n}(x,1)-I(-1+1\neq 2)\rsb \lsb T^\star_{-1,1}-\eta_{-1,1}(x)\rsb\mu_{-1,1}(x)dx\\
&+\frac14\int_{ B_2}\lsb \widehat{f}_{\delta,n}(x,0)-I(1+1\neq 0)\rsb \lsb T^\star_{1,0}-\eta_{1,0}(x)\rsb\mu_{1,0}(x)dx\\
&+\frac14\int_{ B_2}\lsb \widehat{f}_{\delta,n}(x,0)-I(-1+1\neq 0)\rsb \lsb T^\star_{-1,0}-\eta_{-1,0}(x)\rsb\mu_{-1,0}(x)dx.
\end{align*}
By \eqref{boundcase22} and  the fact that $T_{j,a}^\star=1/2+(2a-1)(1/4+j\cdot c_\beta(sv_d)^{d/\gamma})$, 
\begin{align*}
    \eta_{j,a}-T_{j,a}^\star
   &=- c_\beta(2a-1)j (sv_d)^{d/\gamma}+c_\beta\sign(x_1)\lsb  (sv_d)^{d/\gamma} - (sv_d-|x_1|)^{d/\gamma}\rsb.
\end{align*} 
Moreover, we have $\mu_{j,a}(x)=1/(2+2s^d)$ for $x\in\mathcal{B}_2$.
Thus, denoting $W_d =(sv_d)^{d/\gamma}$ and $h(x)=  \sign(x_1)\lmb (sv_d)^{d/\gamma} - (sv_d-|x|)^{d/\gamma}\rmb$,
this further equals 
\begin{align}\label{eq:measureequ2}
\nonumber&\frac{c_\beta}{8+8s^d} \lbb\int_{ B_{2}}  \widehat{f}_{\delta,n}(x,1)\lsb  W_d - h(x)\rsb dx\rbb
+\frac{c_\beta}{8+8s^d} \lbb\int_{ B_{2}}\lsb  \widehat{f}_{\delta,n}(x,1)-1\rsb \lsb  -W_d - h(x)\rsb dx\rbb\\
\nonumber&+\frac{c_\beta}{8+8s^d} \lbb\int_{ \mathcal{B}_{2}}\lsb  \widehat{f}_{\delta,n}(x,0)-1\rsb \lsb  -W_d -  h(x)\rsb dx\rbb+\frac{c_\beta}{8+8s^d} \lbb\int_{ \mathcal{B}_{2}}  \widehat{f}_{\delta,n}(x,0)\lsb  W_d -  h(x)\rsb dx\rbb\\
\nonumber&=\frac{c_\beta}{8+8s^d} \lbb\int_{ B_{2}} \lsb
\widehat{f}_{\delta,n}(x,1) W_d - \widehat{f}_{\delta,n}(x,1) h(x) 
- \widehat{f}_{\delta,n}(x,1)W_d - \widehat{f}_{\delta,n}(x,1)h(x) +W_d +  h(x)\rsb dx\rbb\\
\nonumber&+\frac{c_\beta}{8+8s^d} \lbb\int_{ \mathcal{B}_{2}}\lsb  -\widehat{f}_{\delta,n}(x,0)W_d- \widehat{f}_{\delta,n}(x,0)h(x)+W_d+  h(x) +\widehat{f}_{\delta,n}(x,0)W_d -\widehat{f}_{\delta,n}(x,0)h(x) \rsb dx\rbb.
\end{align}
Cancelling terms, this equals
\begin{align}
\nonumber&\frac{c_\beta}{8+8s^d} \lbb\int_{ B_{2}} \lsb
 - 2\widehat{f}_{\delta,n}(x,1) h(x) 
   +W_d +  h(x)\rsb dx\rbb\\
\nonumber&+\frac{c_\beta}{8+8s^d} \lbb\int_{ \mathcal{B}_{2}}\lsb - 2\widehat{f}_{\delta,n}(x,0)h(x)+W_d+  h(x)     \rsb dx\rbb\\
&=\frac{c_\beta}{4+4s^d} \lbb\int_{ B_{2}} \lsb
 - \lsb\widehat{f}_{\delta,n}(x,1)+\widehat{f}_{\delta,n}(x,0)\rsb h(x) 
   +W_d +  h(x)\rsb dx\rbb.
\end{align}
Note that $0\le \widehat{f}_{\delta,n}(x,a)\le 1$ for $a\in\{0,1\}$. By definition of $W_d$ and $h(x)$, we have, for $a\in\{0,1\}$,
\begin{align*}
&W_d+h(x) -2\widehat{f}_{\delta,n}(x,a)h(x) =W_d +(1-2\widehat{f}_{\delta,n}(x,a))h(x)
\ge W_d -|h(x)|\\
&= (sv_d)^{d/\gamma}-\lsb (sv_d)^{d/\gamma} - (sv_d-|x_1|)^{d/\gamma}\rsb = (sv_d-|x_1|)^{d/\gamma}.
\end{align*}
Then, \eqref{eq:measureequ2} is lower bounded by
\begin{align*}
&\frac{c_\beta}{4+4s^d} \int_{ \mathcal{B}_{2}}   (sv_d-|x_1|)^{d/\gamma} dx=\frac{c_\beta}{2+2s^d} \int_{ \mathcal{B}_{2}\cap \{x_1>0\}}   (sv_d-x_1)^{d/\gamma} dx\\
&= \frac{c_\beta}{2+2s^d} \int_{0}^{sv_d} \lsb\int_{\sum_{j=2}^d|x_j|\le sv_d-x_1}  (sv_d-x_1)^{d/\gamma}   dx_2\ldots dx_d \rsb dx_1 \\
&= \frac{c_\beta}{2+2s^d} \int_{0}^{sv_d}  (sv_d-x_1)^{d/\gamma} \lsb\int_{\sum_{j=2}^d|x_j|\le sv_d-x_1}  dx_2\ldots dx_d \rsb dx_1 \\
&=\frac{c_\beta V_{d-1,1}}{2+2s^d} \int_{0}^{sv_d}   (sv_d-x_1)^{d/\gamma} \lsb{sv_d-x_1}\rsb^{d-1} dx_1\\
&= \frac{c_\beta \gamma V_{d-1,1}}{(2+2s^d)(1+\gamma)d} (sv_d)^{d/\gamma+d} \ge C_{\beta,\gamma,d} s^{\frac{d}\gamma+d} 
\ge C'n^{-\frac{1+\gamma}{2\gamma}}.
\end{align*}
In the last inequality, we have used that  $s \asymp n^{-1/(2d)}$.
This finishes the proof.

\section{Proofs of Theorems in Section \ref{sec:FairBayes}}\label{sec:pf-FairBayes}

The following counterexample from \cite{Vert2009} demonstrates the advantage of offsets. Assume that
$\X \subset \R$, that the density $g$ is such that $g(x) = 1/2$ for $x \in [0, 1]$ and that $g(x) < 1/2$
elsewhere. Assume $\widehat{g}$ is an consistent estimator of $g$ such that $\|\widehat{g}-g\|_\infty\le \ep$ for some small  $\ep>0$. 
If $\widehat{g}(x)=g(x)+\ep$ for $x\in[0,1]$, we have  $\Lambda_f\lsb1/2\rsb=\emptyset$ and $\widehat{\Lambda}_f\lsb1/2\rsb \supset [0, 1]$. Thus, the standard plug-in estimate fails to  estimate
$\Lambda_f\lsb1/2\rsb$ consistently even as $\ep$ tends to 0. However, $\widetilde{\Lambda}_{g,\ell_n}\lsb1/2\rsb$ with a positive offset $\ell_n>\ep$ can become consistent.

\subsection{Proof of Proposition \ref{prop:checkcondition}}
 \begin{itemize}
 \item In the fairness-impacted case, we have $D_-(0)>0$ and $\widetilde{\delta}=\delta$. Recalling \eqref{eq:Disfun1}, \eqref{eq:Disfun2}, \eqref{eq:t-star-dp} and \eqref{eq:pistar},
we have, by the left continuity of $D_+$ and the right continuity of $D_-$,
{\begin{align*}
D_{+}(t_\delta^\star) &= \Po\lsb\eta_1(X)\ge\frac12 + \frac{t^\star_\delta}{2p_1} \rsb- 
\Pz\lsb\eta_0(X)>\frac12 - \frac{t^\star_\delta}{2p_0}\rsb \\
&= \Po\lsb\eta_1(X)>\Tsdo \rsb+ \Po\lsb\eta_1(X)=\Tsdo\rsb- 
\Pz\lsb\eta_0(X)>\Tsdz\rsb   \\
&=\pisop+\pisoe -\piszp \ge \delta,
\end{align*}
and
\begin{align*}
D_{-}(t_\delta^\star) &= \Po\lsb\eta_1(X)>\frac12 + \frac{t^\star_\delta}{2p_1} \rsb- 
\Pz\lsb\eta_0(X)\ge\frac12 - \frac{t^\star_\delta}{2p_0}\rsb \\
&= \Po\lsb\eta_1(X)>\Tsdo \rsb- 
\Pz\lsb\eta_0(X)>\Tsdz\rsb - \Pz\lsb\eta_0(X)=\Tsdz\rsb  \\
&=\pisop -\piszp -\pisze\le \delta.
\end{align*}}
It follows that
$$\delta - \pisop +\piszp =\delta-D_{+}(t_\delta^\star) +\pisoe\le \pisoe,$$ and
thus,
and interpreting $x/0 = 0$ for all $x\in \R$ in what follows,
$\frac{\piszp- \pisop +\delta }{\pisoe}\le 1$.
Similarly, $\pisop -\piszp-\delta  =D_{-}(t_\delta^\star) -\delta+\pisze\le \pisze$
and $\frac{ \pisop -\piszp-\delta}{\pisze}\le 1$.
As a result, with $x\mapsto \sigma(x) = \max(x,0)$,
\begin{align*}
\textup{DDP}(f^\star_\delta)&=\sum_{a\in\{0,1\}}\lmb (2a-1)\lsb \Pa\lsb\eta_a(X)>\Tsda\rsb +\taudsa\Pa\lsb\eta_a(X)=\Tsda\rsb\rsb\rmb\\
&=\pisop+\taudso\cdot \pisoe -\piszp-\taudsz\cdot\pisze\\
&=\pisop+\pisoe \rho\lsb \frac{\piszp- \pisop +\delta }{\pisoe} \rsb -\piszp-\pisze \rho\lsb \frac{ \pisop -\piszp-\delta}{\pisze} \rsb\\
&=\pisop+ \sigma\lsb {\piszp- \pisop +\delta } \rsb -\piszp-\sigma\lsb { \pisop -\piszp-\delta} \rsb=\delta,
\end{align*}
were the last equation follows by checking the cases $\piszp- \pisop +\delta \ge0$ and $\piszp- \pisop +\delta < 0$.

 \item In the automatically-fair and fair-boundary cases, we have $D_-(0)\le0$ and $\widetilde{\delta}=0$. 
  \begin{itemize}
      \item If $D_-(0)\le 0$, 
we have,
$$-\pisoe\le D_{+}(0) -\pisoe = \pisop -\piszp =D_{-}(0) +\pisze\le \pisze.$$
It holds that
$(\piszp- \pisop)/{\pisoe}\le 1$ and $(\pisop -\piszp)/{\pisze}\le 1$.
As a result, with $x\mapsto \sigma(x) = \max(x,0)$,
\begin{align*}
\textup{DDP}(f^\star_\delta)&=\pisop+\pisoe \rho\lsb \frac{\piszp- \pisop  }{\pisoe} \rsb -\piszp-\pisze \rho\lsb \frac{ \pisop -\piszp}{\pisze} \rsb\\
&=\pisop+ \sigma\lsb {\piszp- \pisop  } \rsb -\piszp-\pisze \sigma\lsb { \pisop -\piszp} \rsb=0,
\end{align*}
were the last equation follows by checking the cases $\piszp- \pisop  \ge0$ and $\piszp- \pisop < 0$.
\item   If $D_-(0)> 0$, 
we have,
$$ \pisop -\piszp =D_{-}(0) +\pisze>\pisze.$$
It holds that
$(\piszp- \pisop)/{\pisoe}\le 0$ and $(\pisop -\piszp)/{\pisze}> 1$.
As a result, with $x\mapsto \sigma(x) = \max(x,0)$,
\begin{align*}
\textup{DDP}(f^\star_\delta)&=\pisop+\pisoe \rho\lsb \frac{\piszp- \pisop  }{\pisoe} \rsb -\piszp-\pisze \rho\lsb \frac{ \pisop -\piszp}{\pisze} \rsb\\
&=\pisop+   -\piszp-\pisze =D_-(0).
\end{align*}  
\end{itemize}
\end{itemize}

\section{Proofs of Theorems in Section \ref{sec:Asymp_analysis}}\label{sec:pf-seceight}

\subsection{Proof of Theorem \ref{thm:estimate_of_t}}
Recall the definition of $\htmid$, $\htmin$ and $\htmax$ from \eqref{hts}.
By construction, we have for any $\ep>0$ that
\begin{align*}
\begin{array}{ll}
     \lbb\htmid > t^\star_\delta +\ep\rbb \subset \lbb \widehat{D}_n(t^\star_\delta +\ep,0,0)\ge \delta\rbb; & \lbb\htmid < t^\star_\delta +\ep\rbb \subset \lbb \widehat{D}_n(t^\star_\delta +\ep,0,0)\le \delta\rbb;\\
\lbb\htmin > t^\star_\delta +\ep\rbb \subset \lbb \widehat{D}_n(t^\star_\delta +\ep,0,0)\ge \delta+\Delta_n\rbb; &\lbb\htmin < t^\star_\delta +\ep\rbb \subset \lbb \widehat{D}_n(t^\star_\delta +\ep,0,0)\le \delta+\Delta_n\rbb;\\
\lbb\htmax > t^\star_\delta +\ep\rbb \subset \lbb \widehat{D}_n(t^\star_\delta +\ep,0,0)\ge \delta-\Delta_n\rbb; &\lbb\htmax < t^\star_\delta +\ep\rbb \subset \lbb \widehat{D}_n(t^\star_\delta +\ep,0,0)\le \delta-\Delta_n\rbb.
\end{array}
\end{align*}
Let $L_{t_1}$, $U_{t_1}$ and $c_{i,t_1}$ be defined in Lemma \ref{lem:impacted}; let $L_{t_2}$, $U_{t_2}$ and $c_{i,t_2}$ be defined in Lemma \ref{lem:impacted1}; and let $L_r$, $U_r$, $U_{\Delta,r}$, and $c_{i,r}$ be defined in Lemma \ref{lem:impacted2}. During this proof, we take $L_t= L_{t_1}\vee L_{t,2}\vee L_r$ and  $U_t= U_{t_1}\wedge U_{t,2}\wedge U_r$. We further denote $$\mathcal{E}_t=\psi_{n,1,t}(\ep) +\sum_{j=\{-1,1\}}I(\delta = D_j(t^\star_\delta))\psi_{n,2,t}\lsb g_{\delta,j}\lsb\omega(\ep,r_n)\rsb\rsb.$$
Next, we consider the following four cases: (1) $t_\delta^\star=0$, (2)
$t^\star_\delta>0$ and $D_+(t^\star_\delta)>\delta$, (3) $t^\star_\delta>0$ and  $D_+(t^\star_\delta)=\delta>D_-(t^\star_\delta)$ and (4) $t^\star_\delta>0$ and  $D_+(t^\star_\delta)=\delta=D_-(t^\star_\delta)$.

\begin{itemize}
    \item[Case (1) $t_\delta^\star=0$.]
~

In this case, we have $\Istar=0$ and $\omega(\ep,r_n)=r_n$. By construction, we have $\htd=\htmin$ when $\htmid-\htmin\le r_n$.
By \eqref{D1} of Lemma \ref{lem:impacted}, \eqref{Db1} of Lemma \ref{lem:impacted2} and Lemma \ref{lem:merge}, when $L_t(\phi_{n,1}\vee \phi_{n,0})<r_n<U_t$, $2(g_{\delta,-}(4r_n)\vee g_{\delta,+}(4r_n))<\Delta_n<U_{\Delta,t}:=U_{\Delta,r}$ and $L_t(\phi_{n,1}\vee \phi_{n,0})<\ep<r_n$, we have, with $c_{i,t}>0,i\in[4]$ that
\begin{align*}
    &\PN\lsb \htd>t^\star_\delta+\ep\rsb=\PN\lsb \htd>\ep\rsb =\PN\lsb \htd>\ep, \htd=\htmin\rsb+ \PN\lsb \htd>\ep, \htd\neq\htmin\rsb\\
    &\le \PN\lsb \htmin>\ep\rsb+ \PN\lsb  \htd\neq\htmin\rsb\\
    &\le \PN\lsb \widehat{D}_n(t^\star_\delta +\ep,0,0)\ge \delta+\Delta_n\rsb+ \PN\lsb  \htmid-\htmin >r_n\rsb\\
    &\le  \psi_{n,1,t_1}(\ep)+I(\delta = D_-(t^\star_\delta))\psi_{n,2,t_1}\lsb \Delta_n+g_{\delta,-}\lsb \frac{\ep}{2}\rsb\rsb+\psi_{n,1,r}(r_n)+I(\delta = D_-(t^\star_\delta))\psi_{n,2,r}\lsb g_{\delta,-}\lsb\frac{r_n}{2}\rsb\rsb\\
       &\le  \psi_{n,1,t_1}(\ep)+I(\delta = D_-(t^\star_\delta))\psi_{n,2,t_1}\lsb g_{\delta,-}\lsb \frac{r_n}{4}\rsb\rsb+\psi_{n,1,r}(r_n)+I(\delta = D_-(t^\star_\delta))\psi_{n,2,r}\lsb g_{\delta,-}\lsb\frac{r_n}{4}\rsb\rsb\\
       &\le  \psi_{n,1,t}(\ep)+I(\delta = D_-(t^\star_\delta))\psi_{n,2,t}\lsb g_{\delta,-}\lsb {r_n}\rsb\rsb \le \mathcal{E}_t.
\end{align*}
Here, the third last inequality holds since $\Delta_n+g_{\delta,-}(\ep/2)>\Delta_n>4g_{\delta,-}(4r_n)>g_{\delta,-}(r_n/4)$.

Moreover, by definition,
$\PN\lsb\htd<t^\star_\delta-\ep\rsb\le \PN\lsb\htd<0\rsb=0\le \mathcal{E}_t$.
  \item[Case (2)  $t_\delta^\star>0$ and $\delta < D_+(t^\star_\delta)$.]
    ~\\
In this case, we have $\Istar=0$ and $\omega(\ep,r_n)=r_n$. By construction, we have $\htd=\htmin$ when $\htmid-\htmin\le r_n$. By \eqref{D1} of Lemma \ref{lem:impacted}, \eqref{Db2} of Lemma \ref{lem:impacted2} and Lemma \ref{lem:merge}, when $L_t(\phi_{n,1}\vee \phi_{n,0})<r_n<U_t$, $2(g_{\delta,-}(4r_n)\vee g_{\delta,+}(4r_n))<\Delta_n<U_{\Delta,t}:=U_{\Delta,r}$ and $L_t(\phi_{n,1}\vee \phi_{n,0})<\ep<r_n$, we have,  with $c_{i,t}>0,i\in[4]$ that
\begin{align*}
    &\PN\lsb \htd>t^\star_\delta+\ep\rsb=\PN\lsb \htd>t^\star_\delta+\ep, \htd=\htmin\rsb+ \PN\lsb \htd>t^\star_\delta+\ep, \htd\neq\htmin\rsb\\
    &\le \PN\lsb \htmin>t^\star_\delta+\ep\rsb+ \PN\lsb  \htd\neq\htmin\rsb\\
    &\le \PN\lsb \widehat{D}_n(t^\star_\delta +\ep,0,0)\ge \delta+\Delta_n\rsb+ \PN\lsb  \htmid-\htmin >r_n\rsb\\
    &\le  \psi_{n,1,t_1}(\ep)+I(\delta=D_-(t^\star_\delta))\psi_{n,2,t_1}\lsb \Delta_n+g_{\delta,-}\lsb \frac{\ep}{2}\rsb\rsb+\psi_{n,1,r}\lsb\frac{r_n}{2}\rsb+I(\delta=D_-(t^\star_\delta))\psi_{n,2,r}\lsb g_{\delta,-}\lsb\frac{r_n}{4}\rsb\rsb\\
       &\le  \psi_{n,1,t_1}(\ep)+I(\delta=D_-(t^\star_\delta))\psi_{n,2,t_1}\lsb g_{\delta,-}\lsb \frac{r_n}{4}\rsb\rsb+\psi_{n,1,r}\lsb \ep\rsb+I(\delta=D_-(t^\star_\delta))\psi_{n,2,r}\lsb g_{\delta,-}\lsb\frac{r_n}{4}\rsb\rsb\\
       &\le   \psi_{n,1,t}(\ep) +I(\delta=D_-(t^\star_\delta))\psi_{n,2,t}\lsb g_{\delta,-}\lsb {r_n}\rsb\rsb\le \mathcal{E}_t.
\end{align*}
Here, the third last inequality holds since $\Delta_n-g_{\delta,-}(\ep/2)>\Delta_n-g_{\delta,-}(4r_n)>g_{\delta,-}(4r_n)\ge g_{\delta,-}(r_n/4)$.

On the other hand, by \eqref{D6} of Lemma \ref{lem:impacted1}, \eqref{Db4} of Lemma \ref{lem:impacted2} and Lemma \ref{lem:merge}, when $L_t(\phi_{n,1}\vee \phi_{n,0})<r_n<U_t$, $2(g_{\delta,-}(4r_n)\vee g_{\delta,+}(4r_n))<\Delta_n<U_{\Delta,t}:=U_{\Delta,r}\wedge ((D_+(t^\star_\delta)-\delta)/2)$ and $L_t(\phi_{n,1}\vee \phi_{n,0})<\ep<r_n$, we have,  with $c_{i,t}>0,i\in[4]$ that
\begin{align*}
    &\PN\lsb \htd<t_\delta^\star-\ep\rsb =\PN\lsb \htd<t_\delta^\star-\ep, \htd=\htmin\rsb+ \PN\lsb \htd<t^\star_\delta-\ep, \htd\neq\htmin\rsb\\
    &\le \PN\lsb \htmin<t^\star_\delta-\ep\rsb+ \PN\lsb  \htd\neq\htmin\rsb\\
    &\le \PN\lsb \widehat{D}_n(t^\star_\delta -\ep,0,0)\le \delta+\Delta_n\rsb+ \PN\lsb\htmid-\htmin> r_n\rsb\\
    &\le  \psi_{n,1,t_2}(\ep)+\psi_{n,1,r}\lsb\frac{r_n}{2}\rsb+I(\delta=D_-(t^\star_\delta))\psi_{n,2,r}\lsb g_{\delta,-}\lsb\frac{r_n}{4}\rsb\rsb\\
       &\le  \psi_{n,1,t_1}(\ep)+\psi_{n,1,r}\lsb \ep\rsb+I(\delta=D_-(t^\star_\delta))\psi_{n,2,r}\lsb g_{\delta,-}\lsb\frac{r_n}{4}\rsb\rsb\\
       &\le   \psi_{n,1,t}(\ep) +I(\delta=D_-(t^\star_\delta))\psi_{n,2,t}\lsb g_{\delta,-}\lsb{r_n}\rsb\rsb\le \mathcal{E}_t.
\end{align*}
   \item[Case (3)  $t_\delta^\star>0$ and $D_-(t^\star_\delta)<\delta = D_+(t^\star_\delta)$.]
    ~\\
Again, we have  $\Istar=0$ and $\omega(\ep,r_n)=r_n$.   By construction, we have $\htd=\htmax$ when $\htmid-\htmin> r_n$ and $\htmax-\htmid< r_n$.  By \eqref{D4} of Lemma \ref{lem:impacted1}, \eqref{Db3}, \eqref{Db4} of Lemma \ref{lem:impacted2} and Lemma \ref{lem:merge}, when $L_t(\phi_{n,1}\vee \phi_{n,0})<r_n<U_t$, $2(g_{\delta,-}(4r_n)\vee g_{\delta,+}(4r_n))<\Delta_n<U_{\Delta,t}:=U_{\Delta,r}\wedge ((\delta-D_-(t^\star_\delta)/2)$ and $L_t(\phi_{n,1}\vee \phi_{n,0})<\ep<r_n$,  we have, with $c_{i,t}>0,i\in[4]$ that
\begin{align*}
    &\PN\lsb \htd>t_\delta^\star+\ep\rsb =\PN\lsb \htd>t_\delta^\star+\ep, \htd=\htmax\rsb+ \PN\lsb \htd>t^\star_\delta+\ep, \htd\neq\htmax\rsb\\
    &\le \PN\lsb \htmax>t^\star_\delta+\ep\rsb+ \PN\lsb  \htd\neq\htmax\rsb\\
    &\le \PN\lsb \widehat{D}_n(t^\star_\delta +\ep,0,0)\ge \delta-\Delta_n\rsb+ \PN\lsb\htmid-\htmin\le r_n\rsb+\PN\lsb\htmax-\htmid> r_n\rsb\\
    &\le   \psi_{n,1,t_2}(\ep) + \psi_{n,1,r}(r_n)
    + \psi_{n,2,r}\lsb   g_{\delta,+}\lsb\frac{r_n}2\rsb\rsb +\psi_{n,1,r}\lsb\frac{r_n}{2}\rsb+\psi_{n,2,r}\lsb  g_{\delta,+}\lsb\frac{r_n}4\rsb\rsb\\
    &\le \psi_{n,1,t_2}(\ep)  + 2\psi_{n,1,r}\lsb\frac{r_n}2\rsb
    + 2\psi_{n,2,r}\lsb  g_{\delta,+}\lsb\frac{r_n}{4}\rsb\rsb
    \le  \psi_{n,1,t}(\ep) + \psi_{n,2,t} \lsb   g_{\delta,+}\lsb{r_n}\rsb\rsb\le \mathcal{E}_t.
\end{align*}
Here, the third last inequality holds since $\Delta_n-g_{\delta,-}(2\ep)>\Delta_n-g_{\delta,-}(4r_n)>g_{\delta,+}(4r_n)\ge g_{\delta,+}(r_n/4)$.

On the other hand, by \eqref{D2} of Lemma \ref{lem:impacted}, \eqref{Db3}, \eqref{Db4} of Lemma \ref{lem:impacted2} and Lemma \ref{lem:merge}, when $L_t(\phi_{n,1}\vee \phi_{n,0})<r_n<U_t$, $2(g_{\delta,-}(4r_n)\vee g_{\delta,+}(4r_n))<\Delta_n<U_{\Delta,t}:=U_{\Delta,r}$ and $L_t(\phi_{n,1}\vee \phi_{n,0})<\ep<r_n$, we have, with $c_{i,t}>0,i\in[4]$ that
\begin{align*}
    &\PN\lsb \htd<t_\delta^\star-\ep\rsb =\PN\lsb \htd<t_\delta^\star-\ep, \htd=\htmax\rsb+ \PN\lsb \htd<t^\star_\delta-\ep, \htd\neq\htmax\rsb\\
    &\le \PN\lsb \htmax<t^\star_\delta-\ep\rsb+ \PN\lsb  \htd\neq\htmax\rsb\\
    &\le \PN\lsb \widehat{D}_n(t^\star_\delta -\ep,0,0)\le \delta-\Delta_n\rsb+ \PN\lsb\htmid-\htmin\le r_n\rsb+\PN\lsb\htmax-\htmid> r_n\rsb\\
    &\le   \psi_{n,1,t_1}(\ep) +  \psi_{n,2,t_1}\lsb \Delta_n+g_{\delta,+}\lsb\frac{\ep}{2}\rsb\rsb + \psi_{n,1,r}(r_n)
    + \psi_{n,2,r}\lsb   g_{\delta,+}\lsb\frac{r_n}2\rsb\rsb +\psi_{n,1,r}\lsb\frac{r_n}{2}\rsb \\
    &+\psi_{n,2,r}\lsb  g_{\delta,+}\lsb\frac{r_n}4\rsb\rsb\\
    &\le \psi_{n,1,t_1}(\ep) +  \psi_{n,2,t_1}\lsb g_{\delta,+}\lsb\frac{r_n}4\rsb\rsb + 2\psi_{n,1,r}\lsb\frac{r_n}2\rsb
    + 2\psi_{n,2,r}\lsb  g_{\delta,+}\lsb\frac{r_n}{4}\rsb\rsb
    \le  \psi_{n,1,t}(\ep) + \psi_{n,2,t} \lsb   g_{\delta,-}\lsb{r_n}\rsb\rsb\le \mathcal{E}_t.
\end{align*}
    \item[Case (4)  $t_\delta^\star>0$ and $D_-(t^\star_\delta)=\delta = D_+(t^\star_\delta)$.]
    ~\\
 In this case, we have $\Istar=1$ and $\omega(\ep,r_n)=\ep$.   By construction, we have $\htd=\htmid$ when $\min(\htmid-\htmin,\htmax-\htmid)> r_n$.  By \eqref{D1}, \eqref{D2} of Lemma  \ref{lem:impacted}, \eqref{Db5}, \eqref{Db6} of Lemma \ref{lem:impacted2} and Lemma \ref{lem:merge},  when $L_t(\phi_{n,1}\vee \phi_{n,0})<r_n<U_t$, $2(g_{\delta,-}(4r_n)\vee g_{\delta,+}(4r_n))<\Delta_n<U_{\Delta,t}:=U_{\Delta,r}$ and $L_t(\phi_{n,1}\vee \phi_{n,0})<\ep<r_n$, we have,  with $c_{i,t}>0,i\in[4]$ that,
\begin{align*}
    &\PN\lsb \htd>t_\delta^\star+\ep\rsb =\PN\lsb \htd>t_\delta^\star+\ep, \htd=\htmid\rsb+ \PN\lsb \htd>t^\star_\delta+\ep, \htd\neq\htmid\rsb\\
    &\le \PN\lsb \htmid>t^\star_\delta+\ep\rsb+ \PN\lsb  \htd\neq\htmid\rsb\\
    &\le \PN\lsb \widehat{D}_n(t^\star_\delta +\ep,0,0)\ge \delta\rsb+ \PN\lsb\htmid-\htmin\le r_n\rsb+\PN\lsb\htmax-\htmid\le r_n\rsb\\
    &\le   \psi_{n,1,t_1}(\ep) +  \psi_{n,2,t_1}\lsb g_{\delta,-}\lsb\frac\ep2\rsb\rsb + 2\psi_{n,1,r}(r_n)
    + \psi_{n,2,r}\lsb   g_{\delta,+}\lsb\frac{r_n}2\rsb\rsb + \psi_{n,2,r}\lsb  g_{\delta,-}\lsb\frac{r_n}2\rsb\rsb\\
    &\le  \psi_{n,1,t}(\ep) + \psi_{n,2,t} \lsb   g_{\delta,-}\lsb {\ep}\rsb\rsb +\psi_{n,2,t} \lsb   g_{\delta,+}\lsb {\ep}\rsb\rsb\le \mathcal{E}_t.
\end{align*}
Similarly,
\begin{align*}
    &\PN\lsb \htd<t_\delta^\star-\ep\rsb =\PN\lsb \htd<t_\delta^\star-\ep, \htd=\htmid\rsb+ \PN\lsb \htd<t^\star_\delta-\ep, \htd\neq\htmid\rsb\\
    &\le \PN\lsb \htmid<t^\star_\delta-\ep\rsb+ \PN\lsb  \htd\neq\htmid\rsb\\
    &\le \PN\lsb \widehat{D}_n(t^\star_\delta -\ep,0,0)\le \delta\rsb+ \PN\lsb\htmid-\htmin\le r_n\rsb+\PN\lsb\htmax-\htmid\le r_n\rsb\\
    &\le   \psi_{n,1,t_1}(\ep) +  \psi_{n,2,t_1}\lsb g_{\delta,+}\lsb\frac\ep2\rsb\rsb + 2\psi_{n,1,r}(r_n)
    + \psi_{n,2,r}\lsb   g_{\delta,+}\lsb\frac{r_n}2\rsb\rsb + \psi_{n,2,r}\lsb  g_{\delta,-}\lsb\frac{r_n}2\rsb\rsb\\
    &\le  \psi_{n,1,t}(\ep) + \psi_{n,2,t} \lsb   g_{\delta,-}\lsb {\ep}\rsb\rsb +\psi_{n,2,t} \lsb   g_{\delta,+}\lsb {\ep}\rsb\rsb\le \mathcal{E}_t.
\end{align*}

\end{itemize}

\subsection{Proof of Theorem \ref{thm:convergence_rate}}
By definition, we have
$$\widehat{f}_{\delta,n}(x,a) = \left\{ 
\begin{array}{lr}
 1 ,   &  \lbb (x,a): \widehat{\eta}_a(x)>\Thda+\ela\rbb,\\
 \taudha,    & \lbb (x,a): \left|\widehat{\eta}_a(x)-\Thda\right|<\ela\rbb,\\
0,  &  \lbb (x,a): \widehat{\eta}_a(x)<\Thda-\ela\rbb;
\end{array}
\right.
\text{ and } {f}^{\star}_{\delta}(x,a) = \left\{
\begin{array}{lr}
 1 ,   &   \lbb   (x,a): {\eta}_a(x)>\Tsda\rbb,\\
 \taudsa,    & \lbb (x,a):   {\eta}_a(x)=\Tsda\rbb,\\
0,  &    \lbb  (x,a):{\eta}_a(x)<\Tsda\rbb.
\end{array}
\right.$$
It follows that
$$\widehat{f}_{\delta,n}(x,a)- {f}^{\star}_{\delta}(x,a) = \left\{
\begin{array}{lr}
 0 ,   &   \lbb   (x,a): \widehat{\eta}_a(x)>\Thda+\ela, {\eta}_a(x)>\Tsda\rbb,\\
1- \taudsa,    & \lbb (x,a): \widehat{\eta}_a(x)>\Thda+\ela,  {\eta}_a(x)=\Tsda\rbb,\\
1,  &    \lbb \widehat{\eta}_a(x)>\Thda+\ela, (x,a):{\eta}_a(x)<\Tsda\rbb,\\
 \taudha-1 ,   &   \lbb   (x,a): \left|\widehat{\eta}_a(x)-\Thda\right|\le \ela, {\eta}_a(x)>\Tsda\rbb,\\
\taudha- \taudsa,    & \lbb (x,a):\left| \widehat{\eta}_a(x)-\Thda\right|\le\ela,  {\eta}_a(x)=\Tsda\rbb,\\
\taudha,  &    \lbb  \left|\widehat{\eta}_a(x)-\Thda\right|\le\ela , (x,a):{\eta}_a(x)<\Tsda\rbb,\\
 -1 ,   &   \lbb   (x,a): \widehat{\eta}_a(x)\le \Thda- \ela , {\eta}_a(x)>\Tsda\rbb,\\
- \taudsa,    & \lbb (x,a): \widehat{\eta}_a(x)<\Thda-\ela,  {\eta}_a(x)=\Tsda\rbb,\\
0,  &    \lbb \widehat{\eta}_a(x)<\Thda-\ela, (x,a):{\eta}_a(x)<\Tsda\rbb.\\
\end{array}
\right.$$
Since $\Tsda - \eta_a(x) = 0$ on $\lbb (x,a): {\eta}_a(x)=\Tsda\rbb$, we have
\begin{align*}
&(\widehat{f}_{\delta,n}(x,a)-f_{\delta}^\star(x,a))(\Tsda-f_{\delta}^\star(x,a)) \\
&= \left\{\begin{array}{lcl}
      \Tsda-\eta_a(x),  &  \{(x,a): \widehat{\eta}_a(x)>\Thda+\ela ,\eta_a(x)<\Tsda \};\\
   (\taudha-1)(\Tsda-\eta_a(x)),  &  \{(x,a): |\widehat{\eta}_a(x)-\Thda|\le\ela ,\eta_a(x)>\Tsda \};\\
   \taudha(\Tsda-\eta_a(x)),  &  \{(x,a): |\widehat{\eta}_a(x)-\Thda|\le\ela ,\eta_a(x)<\Tsda \};\\
      -(\Tsda-\eta_a(x)),  &  \{(x,a): \widehat{\eta}_a(x)<\Thda-\ela ,\eta_a(x)>\Tsda \};\\
0, & \textnormal{otherwise},
\end{array}\right.\\
&\le |\eta_a(x)-\Tsda| \lmb I(\eta_a(x)>\Tsda,\widehat\eta_a(x)\le\Thda +\ela) + I(\eta_a(x)<\Tsda,\widehat\eta_a(x)\ge\Thda-\ela)\rmb.
\end{align*}
Then, by \eqref{eq:measure} and Lemma \ref{lem:expectationbound} for $a\in\{0,1\}$, we conclude that 
\begin{align*}
&\EN\lmb d_E\lsb\widehat{f}_{\delta,n},f^\star\rsb\rmb=2\sum_{a\in\{0,1\}}p_a\EN\lmb\int (\widehat{f}_{\delta,n}(x,a)-f^\star(x,a))(\Tsda-\eta_a(x))  d\P_{X|A=a}(x)\rmb\\
&=2\sum_{a\in\{0,1\}}p_a\EN \int\, I{\{\eta_a(x)>\Tsda,\, \widehat\eta_a(x)\le \Thda+\ela
\}}\left|\eta_a(x)-T^\star_{\delta, a}\right|d\Pa(x)\\
&+2\sum_{a\in\{0,1\}}p_a\EN \int\, I{\{\eta_a(x)<\Tsda,\, \widehat\eta_a(x)\ge \Thda-\ela\}}\left|\eta_a(x)-T^\star_{\delta, a}\right|d\Pa(x)\\
&\le  C \lsb (\phi_{n,1}\vee \phi_{n,0}\vee \lo\vee \lz) +\widetilde{I}^\star(\delta)n^{-1/(2\gamma_\delta)}\rsb^{\gamma_\delta+1}.
\end{align*}

\subsection{Proof of Theorem~\ref{thm:asymptotic fairness}}
Recalling \eqref{eq:t-star-dp} and \eqref{eq:pistar},
we have by definition 
{\begin{align*}
D_{+}(t_\delta^\star) &= \Po\lsb\eta_1(X)\ge\frac12 + \frac{t^\star_\delta}{2p_1} \rsb- 
\Pz\lsb\eta_0(X)>\frac12 - \frac{t^\star_\delta}{2p_0}\rsb \\
&= \Po\lsb\eta_1(X)>\Tsdo \rsb+ \Po\lsb\eta_1(X)=\Tsdo\rsb- 
\Pz\lsb\eta_0(X)>\Tsdz\rsb   \\
&=\pisop+\pisoe -\piszp \ge \delta,
\end{align*}
and
\begin{align*}
D_{-}(t_\delta^\star) &= \Po\lsb\eta_1(X)>\frac12 + \frac{t^\star_\delta}{2p_1} \rsb- 
\Pz\lsb\eta_0(X)\ge\frac12 - \frac{t^\star_\delta}{2p_0}\rsb \\
&= \Po\lsb\eta_1(X)>\Tsdo \rsb- 
\Pz\lsb\eta_0(X)>\Tsdz\rsb - \Pz\lsb\eta_0(X)=\Tsdz\rsb  \\
&=\pisop -\piszp -\pisze\le \delta.
\end{align*}}
Recall that for all $x,a$,
$$ \widehat{f}_{\delta,n}(x,a) = I\lsb\widehat{\eta}_a(X)>\Thda+\ela\rsb +\taudha I\lsb\left|\widehat{\eta}_a(x)-\Thda\right|\le\ela\rsb.$$
The disparity level of our method $\widehat{f}_{\delta,n}$ can be expressed as 
\begin{align*}
\textup{DDP}(\widehat{f}_{\delta,n})&=\Po\lsb \widehat{Y}_{\widehat{f}_{\delta,n}} =1 \rsb-\Pz\lsb \widehat{Y}_{\widehat{f}_{\delta,n}} =1 \rsb\\
&= \Po\lsb\widehat{\eta}_1(X)>\Thdo+\lo\rsb +\taudho \Po\lsb\left|\widehat{\eta}_1(x)-\Thdo\right|\le\lo\rsb\\
&-\Pz\lsb\widehat{\eta}_0(X)>\Thdz+\lz\rsb +\taudhz \Pz\lsb\left|\widehat{\eta}_0(x)-\Thdz\right|\le\lz\rsb\\
&=\pihop+\pihoe\taudho-\pihzp-\pihze\taudhz.
\end{align*}
From Proposition \ref{prop:checkcondition}, we have
$|\pisop+\pisoe \taudso -\piszp-\pisze \taudsz|\le\delta.  $
Then,
by Lemmas \ref{lem:piset1} and \ref{lem:pirbound}, we have,
for $(L_\pi\vee L_{\pi_1})(\phi_{n,1}\vee \phi_{n,0})<r_n<(U_\pi\wedge U_{\pi_1})$, $2\lsb g_{\delta,-}(4r_n)\vee g_{\delta,+}(4r_n)\rsb<\Delta_n<(U_{\Delta,\pi}\wedge U_{\Delta,\pi_1})$, 
 $L_{\ep,\pi}<\ep/4 \le \sqrt{(p_1 \wedge p_0) /2}$ and $ (L_\eta \vee L_T)(\phi_{n,1}\vee \phi_{n,0})< \ela/2 < r_n$, 
\begin{align*}
\begin{split}
&\PN\left(\textup{DDP}(\widehat{f}_{\delta,n})>\delta + \ep\right)=\PN\left(\textup{DDP}(\widehat{f}_{\delta,n})>\pisop+\pisoe \taudso-\piszp-\pisze\taudsz+\ep\right)\\
&\le \PN\lsb\pihop - \pisop >\frac\ep4\rsb
+ \PN\lsb\taudho\pihoe - \pisoe \taudso>\frac\ep4\rsb\\
&\qquad\qquad\qquad\qquad+ \PN\lsb \pihzp - \piszp <-\frac\ep4\rsb
+\PN\lsb \pihze \taudhz- \pisze\taudsz<-\frac\ep4\rsb\\
&\le  10\psi_{n,1,\pi}\lsb\lo \wedge \lz\rsb +10\sum_{j\in\{-,+\}} I\lsb\delta=D_j(t^\star_\delta)\rsb\psi_{n,2,\pi}\lsb g\lsb \xi\lsb\lo \wedge \lz,r_n\rsb\rsb\rsb + 2c_{5,\pi}\exp\lsb - \frac{c_{6,\pi}n\ep^2}{16}\rsb.
\end{split}
\end{align*}
For $c_{1,D}=20c_{1,\pi}$, $c_{2,D}=c_{2,\pi}/16$, $c_{3,D}=20c_{3,\pi}$, $c_{4,D}=c_{4,\pi}/(4^{2\gamma})$, $c_{5,D}=c_{5,\pi}$, and  $c_{6,D}=c_{6,\pi}/16$, 
this is upper bounded by
\begin{align*}
\frac12\psi_{n,1,D}\lsb{\lo \wedge \lz}\rsb +\frac12\psi_{n,2,D}\lsb g\lsb \xi\lsb\lo \wedge \lz,r_n\rsb\rsb\rsb  + \frac12c_{5,D}\exp\lsb - {c_{6,D}n\ep^2}\rsb.
\end{align*}
Similarly, the same upper bound holds for $\PN\left(\textup{DDP}(\widehat{f}_{\delta,n})<-\delta - \ep\right)$.
As a result, there exist constants $c_{D,i}$, $i\in[6]$ such that \eqref{ddpbd} holds.

\section{Proofs of Lemmas}\label{sec:tec_lems}

\subsection{Proof of Lemma \ref{lem:interd1}}
\label{plem:interd1}
  Without loss of generality, we assume $ z = {0}$ by translation,
 and that $R=1$ by scaling.
 The vertices of the polyhedron $B_{d,1} (0,1)$  are $\{
\sigma \cdot e_j:
 j\in [d],\sigma\in\{-1,1\}\}$. 
 Without loss of generality, we assume
 that the vertex of $B_{d,1}({0},1)$  closest to $ x$ is $e_1$. Then, we have
 $\sum_{j=1}^d|x_j|\le 1 $ and 
 $x_{1} > \max_{j=2,\ldots,d} |x_{j}|$.
When $r\le 1/(d+2)$,
we will show that 
\beq\label{int} B_{d,1}\lsb  x-\frac{r e_1}{2},\frac{r}2\rsb\subset B_{d,2}( x,r)\cap B_{d,1} (0,1).
\eeq
First, as $B_{d,1}( x,r )\subset B_{d,2}( x,r ) $, 
we have  $B_{d,1}(  x-{ e_1r/2},{r}/2) 
\subset 
B_{d,1}(  x,r) 
\subset   B_{d,2}( x ,r ).$
Next, let $ y=(y_1,\ldots,y_d)^\top\in B_{d,1}(  x-{e_1}r/{2},r/2)$. We have
$$\left| y_1-x_{1}+\frac{r}2\right|+\sum_{j=2}^d|y_j-x_j|\le \frac{r}2.$$
We consider the following two cases: (1)
$\sum_{j=1}^d |x_j|\le 1-r$ and (2) $ 1-r <\sum_{j=1}^d|x_j|\le 1$.
\begin{itemize}[]
\item (1) When $\sum_{j=1}^d |x_j|\le 1-r$, we have 
\begin{align*}
    \sum_{j=1}^d|y_j|&\le \left| y_1-x_{1}+\frac{r}2\right|+\sum_{j=2}^d|y_j-x_j|+ \left |x_1-\frac{r}2\right| +\sum_{j=2}^d|x_j| \\
    &\le \frac{r}2 + \left|x_1-\frac{r}2\right| + \sum_{j=2}^d|x_j|\le  \frac{r}{2} + \frac{r}{2}+\sum_{j=1}^d|x_j|\le 1.
\end{align*}
\item (2) When $1-r<\sum_{j=1}^d |x_j|\le 1$, we have $x_1>1-r-\sum_{j=2}^d|x_j|
\ge 1-r-(d-1)x_1.$ 
It follows that $x_1>(1-r)/d\ge r/2$ as $r\le 2/(d+2)$. Thus, starting with the same argument as above,
\begin{equation*}
    \sum_{j=1}^d|y_j| 
    \le \frac{r}2 + \left|x_1-\frac{r}2\right| + \sum_{j=2}^d|x_j|= \frac{r}{2} +x_1- \frac{r}{2}+\sum_{j=1}^d|x_j|\le 1.
\end{equation*}
\end{itemize}
This implies \eqref{int}, and we then have
\begin{align*}
  \lambda\lmb B_{d,1}( 0,1)\cap B_{d,2}( x,r)\rmb  \ge   \lambda\lmb B_{d,1}\lsb  x-\frac{r e_1}{2},\frac{r}2\rsb\rmb =\frac{V_{d,1}r^d}{2^d} =\frac{V_{d,1}}{V_{d,2}2^d}\cdot V_{d,2}r^d = \frac{V_{d,1}}{V_{d,2}2^d}\lambda[ B_{d,2}( x,r)].
\end{align*}

\subsection{Proof of Lemma \ref{lem:interd2}}
\label{plem:interd2}
As in the proof of Lemma \ref{lem:interd1},
without loss of generality, we can assume $ z = {0}$
 and $R=1$.
Further, since  the $\ell_2$-ball is rotation-invariant, we can assume without loss of generality that $ x= x_1 e_1$ with $x_1\in[0,1]$.
For $r\le 1$, and for
$ y=(y_1,\ldots,y_d)^\top\in B_{d,2}( x-e_1, 1)\cap B_{d,2}( x, r)$, we have
$$(y_1-x_1+1)^2+\sum_{j=2}^dy_j^2\le 1 \ \text{ and } \ (y_1-x_1)^2+\sum_{j=2}^dy_j^2\le r^2.$$
It follows that
$$  -\sqrt{1-\sum_{j=2}^dy_j^2}\le x_1-\sqrt{r^2-\sum_{j=2}^dy_j^2}  \le                 y_1\le\sqrt{1-\sum_{j=2}^dy_j^2} +x_1-1    \le\sqrt{1-\sum_{j=2}^dy_j^2}.$$
It follows that  $ y \in B_{d,2}( 0,1)$. This shows that 
$B_{d,2}( x-e_1, 1)\cap B_{d,2}( x, r) \subset  B_{d,2}( 0,1)\cap B_{d,2}( x, r) $.
Moreover,
$$\{ y: \| y -  x +e_1\|_2^2 = 1\} \cap \{ y: \| y -  x\|_2^2 = r^2\} =\lbb  y: y_1 = x_1-\frac{r^2}{2}, \sum_{j=2}^d y_j^2 = r^2-\frac{r^4}{4}\rbb. $$
It follows that
\begin{align*}
& \lambda[B_{d,2}( 0,1)\cap B_{d,2}( x, r)]   \ge\lambda[B_{d,2}( x-e_1, 1)\cap B_{d,2}( x, r) ]\\
&=
\int  I\lsb{x_1-r}\le y_1\le {x_1-r^2/2}, \sum_{j=2}^d y_j^2 \le r^2-(y_1-x_1)^2\rsb d y \\
&\qquad\qquad\qquad\qquad\qquad\qquad+\int I\lsb {x_1-r^2/2}\le y_1\le x_1,  \sum_{j=2}^d y_j^2 \le 1-(y_1-x_1+1)^2\rsb d y\\
&=\int_{x_1-r}^{x_1-\frac{r^2}{2}} \lsb r^2-(y_1-x_1)^2 \rsb^{\frac{d-1}{2}}V_{d-1,2}dy_1 + \int_{x_1-\frac{r^2}{2}}^{x_1} \lsb 1-(y_1-x_1+1)^2 \rsb^{\frac{d-1}{2}}V_{d-1,2}dy_1.
\end{align*}
By 
substituting $t=\sqrt{r^2-(y_1-x_1)^2}$ for the first integral, 
and using $r\le 1$,
this is lower bounded by
\begin{align*}
&V_{d-1,2}\int_{0}^{\sqrt{r^2-\frac{r^4}{4}}} \frac{t^{d}}{\sqrt{r^2-t^2}}dt   \ge V_{d-1,2} \int_{0}^{\frac{\sqrt{3}r}{2}} \frac{t^{d}}{r}dt\\
&\ge  r^d V_{d,2} \frac{3^{\frac{d+1}{2}}V_{d-1,2}}{2^{d+1}(d+1)V_{d,2}} = \frac{3^{\frac{d+1}{2}}V_{d-1,2}}{2^{d+1}(d+1)V_{d,2}}\cdot\lambda [B_{d,2}( x,r)].
\end{align*}

\subsection{Proof of Lemma \ref{lem:inter3}}
\label{plem:inter3}

We consider two  cases: (1) $x\in[0,1]^d\setminus \mathcal{C}_{z,q}$ and (2)
$x\in \mathcal{C}_{z,q}\setminus \mathcal{D}_{z,q}$.
\begin{itemize}
    \item Case (1): $x\in[0,1]^d\setminus \mathcal{C}_{z,q}$.
    
    Since the result holds for any  $C_r\le 1$  when $B_{d,2}(x,r)\cap \mathcal{D}_{z,r}=\emptyset$, 
    we only need to consider the case that $B_{d,2}(x,r)\cap \mathcal{D}_{z,r}\neq\emptyset$. 
    In this case, we have $\|x-z\|< 2q^{-1}+r$. 
    Letting $z_x= z+3q^{-1}(x-z)/\|x-z\|$,   
    we can verify that
    $B_{d,2}\lsb z_x,q^{-1}\rsb \subset B_{d,2}(x,r)\cap \lsb\mathcal{C}_{z,q}\setminus \inter\mathcal{D}_{z,q}\rsb.$
    In fact,  for any point $y\in B_{d,2}\lsb z_x,q^{-1}\rsb$, we have 
$$\|y-x\|\le \|y-z_x\|+\|z_x-x\|\le q^{-1} +\|x-z\| -3q^{-1} = \|x-z\|-2q^{-1}< r.$$
Moreover,
$$\|y-z\|\le \|y-z_x\|+\|z_x-z\|= q^{-1}  +3q^{-1} = 4q^{-1}, $$
and
$\|y-z\|\ge\|z_x-z\|- \|y-z_x\|= 3q^{-1} -q^{-1} = 2q^{-1}$.
It follows that
$$y\in B_{d,2}(x,r)\cap \lsb B_{d,2}(z,4q^{-1})\setminus \inter B_{d,2}(z,2q^{-1}) \rsb \subset B_{d,2}(x,r)\cap\lsb \mathcal{C}_{z,q}\setminus \inter\mathcal{D}_{z,q}\rsb.
$$
Since $\lambda[\mathcal{C}_{z,q}]=8^dq^{-d}$,
and the boundary of $\mathcal{D}_{z,q}$ has zero Lebesgue measure,
we thus have
\begin{align*}
    \lambda[B_{d,2}(x,r)\cap \lsb \mathcal{C}_{z,q}\setminus\mathcal{D}_{z,q} \rsb]&\ge  \lambda[B_{d,2}(z_x,q^{-1})] = V_{d,2}q^{-d}=
    8^{-d}V_{d,2}\lambda[\mathcal{C}_{z,q}]\\
    &\ge  8^{-d}V_{d,2}\lambda[B_{d,2}(x,r)\cap\mathcal{C}_{z,q}].
\end{align*}

\item (2) Case (2):
$x\in \mathcal{C}_{z,q}\setminus \mathcal{D}_{z,q}$. In this case, we consider two sub-cases: (2.1) $x\in   B_{d,2}(z,q^{-1})$; and (2.2) $x\in \mathcal{C}_{z,q}\setminus B_{d,2}(z,2q^{-1})$.
\begin{itemize}
    \item Case (2.1): $x\in  B_{d,2}(z,q^{-1})$.
\begin{itemize}
    \item    When $r\le q^{-1}$, we have
    $B_{d,2}(x,r) \subset B_{d,2}(z,2q^{-1})\subset \mathcal{C}_{z,q}$ since, for any point $y \in B_{d,2}(x,r)$, $\|y-z\|\le \|y-x\|+\|x-z\|\le r+q^{-1}\le 2q^{-1}$. 
    Thus, by Lemma \ref{lem:interd2} with $R=q^{-1}$, 
    denoting $\Psi_d= 3^\frac{d+1}{2}V_{d-1,2}/(2^{d+1}(d+1)V_{d,2})$,
    we have
\begin{align*}
   & \lambda[B_{d,2}(x,r)\cap \lsb \mathcal{C}_{z,q}\setminus\mathcal{D}_{z,q}\rsb] =
  \lambda[B_{d,2}(x,r)\cap B_{d,2}(z,q^{-1})]\\
  &\ge \Psi_d   
    \lambda[B_{d,2}(x,r)]  =\Psi_d   
    \lambda[B_{d,2}(x,r)\cap \mathcal{C}_{z,q}].
\end{align*}
\item When, $q^{-1}< r< (4\sqrt{d}+1)q^{-1}$, we have
 \begin{align*}
   &\lambda[B_{d,2}(x,r)\cap \lsb \mathcal{C}_{z,q}\setminus\mathcal{D}_{z,q}\rsb]\ge
   \lambda[B_{d,2}(x,q^{-1})\cap \lsb \mathcal{C}_{z,q}\setminus\mathcal{D}_{z,q}\rsb]\\
   &=
  \lambda[B_{d,2}(x,q^{-1})\cap B_{d,2}(z,q^{-1})]\ge\Psi_d   
    \lambda[B_{d,2}(x,q^{-1})]
    =\frac{\Psi_{d}}{(4\sqrt{d}+1)^d}   
    \lambda[B_{d,2}(x,(4\sqrt{d}+1)q^{-1})]\\
    &\ge \frac{\Psi_{d}}{(4\sqrt{d}+1)^d}   
    \lambda[B_{d,2}(x,(4\sqrt{d}+1)q^{-1})\cap \mathcal{C}_{z,q}]
 \ge\frac{\Psi_{d}}{(4\sqrt{d}+1)^d}   
    \lambda[B_{d,2}(x,r)\cap \mathcal{C}_{z,q}].
    \end{align*}

    \item When $r\ge(4\sqrt{d}+1)q^{-1}$, we have $\mathcal{C}_{z,q}\subset B_{d,2}(x,r)$ since for any point $y \in \mathcal{C}_{z,q}$, $\|y-x\|\le \|y-z\|+\|z-x\|\le 4\sqrt{d}q^{-1}+q^{-1}\le (4 \sqrt{d}+1)q^{-1}$. We thus have,
 \begin{align}
    &\lambda[B_{d,2}(x,r)\cap\lsb \mathcal{C}_{z,q}\setminus \mathcal{D}_{z,q}\rsb] =
  \lambda[\mathcal{C}_{z,q}\setminus \mathcal{D}_{z,q}] = (8^d - (2^d-1)V_{d,2})q^{-d}\nonumber
  \\   &= \frac{8^d -(2^d-1)V_{d,2}}{8^d}\lambda[\mathcal{C}_{z,q}] = \frac{8^d -(2^d-1)V_{d,2}}{8^d}\lambda[B_{d,2}(x,r)\cap\mathcal{C}_{z,q}].\label{lr}
\end{align}

    \end{itemize}
\item Case (2.2): $x\in \mathcal{C}_{z,q}\setminus B_{d,2}(z,2q^{-1})$.
\begin{itemize}
    \item  When $r\le q^{-1}$, the result holds when $\|x-z\|>3q^{-1}$ since in this case, $B_{d,2}(x,r)\cap \mathcal{D}_{z,q}=\emptyset$. 
    When $\|x-z\| \le 3q^{-1}$, we consider the set
$\mathcal{M} = \{\tilde{y} = 2x-y : y \in B_{d,2}(x,r)\cap \mathcal{D}_{z,q}\}$.
Since $\mathcal{M}$ is a translation and reflection of $B_{d,2}(x,r)\cap \mathcal{D}_{z,q} $,
we clearly have 
$\lambda[\mathcal{M}] = \lambda[ B_{d,2}(x,r)\cap \mathcal{D}_{z,q} ].$
Moreover, we can verify that
$\mathcal{M}\subset B_{d,2}(x,r)\cap \lsb B_{d,2}(z,4q^{-1})\setminus B_{d,2}(z,3q^{-1})\rsb \subset B_{d,2}(x,r)\cap \lsb \mathcal{C}_{z,q}\setminus \mathcal{D}_{z,q}\rsb.$
In fact, for any $\tilde{y}\in\mathcal{M}$ with $y = 2x-\tilde{y} \in B_{d,2}(x,r)\cap \lsb\mathcal{C}_{z,q}\setminus\mathcal{D}_{z,q}\rsb$, we have $\|\tilde{y}-x\| = \|x-y\|\le r \le q^{-1}$
and
 $$\|\tilde{y}-z\| \le \|\tilde{y}-x\| + \|x-z\|\le q^{-1}+3q^{-1} =4q^{-1}.$$
 Moreover, since $x = (y+\tilde{y})/2$, it holds that
 $\|z-x\| \le \max(\|z-\tilde{y}\|,\|z-y\|).$
 As $\|z-y\|\le 2q^{-1} <3q^{-1}\le \|z-x\|$, we have $\|\tilde{y}-z\|\ge 3q^{-1}.$
As a result,  
$$ \tilde{y} \in B_{d,2}(x,r)\cap \lsb B_{d,2}(z,4q^{-1})\setminus B_{d,2}(z,3q^{-1})\rsb.$$
Therefore,
$$\lambda[B_{d,2}(x,r)\cap \lsb\mathcal{C}_{z,q}\setminus\mathcal{D}_{z,q}\rsb]\ge \lambda[\mathcal{M}] =\lambda[B_{d,2}(x,r)\cap \mathcal{D}_{z,q}],
$$
which, 
using $\mathcal{C}_{z,q} = \bigl(\mathcal{C}_{z,q}\setminus\mathcal{D}_{z,q}\bigr)\cup \mathcal{D}_{z,q}$,
implies that 
\begin{equation}\label{lr2}\lambda[B_{d,2}(x,r)\cap \lsb\mathcal{C}_{z,q}\setminus\mathcal{D}_{z,q}\rsb]\ge \frac12\lambda[B_{d,2}(x,r)\cap \mathcal{C}_{z,q}].
\end{equation}
\item When $q^{-1}< r< 8\sqrt{d}q^{-1}$.  We
 consider the set
$\mathcal{N} = \{\tilde{y} = x+(rq)^{-1} (y-x) : y \in B_{d,2}(x,r)\cap \mathcal{C}_{z,q}\}$.
Since $\mathcal{N}$ is a   scaling  of $B_{d,2}(x,r)\cap \mathcal{C}_{z,q} $ with scaling coefficient $(rq)^{-1}$,
we clearly have 
$\lambda[\mathcal{N}] = (rq)^{-d}\lambda[ B_{d,2}(x,r)\cap \mathcal{C}_{z,q} ].$
Moreover, we can verify that
$\mathcal{N}\subset B_{d,2}(x,q^{-1})\cap \mathcal{C}_{z,q}.$
In fact, for any $\tilde{y}\in\mathcal{N}$ with $y = x+ rq(y-x)\in B_{d,2}(x,r)\cap \mathcal{C}_{z,q}$, we have $\|\tilde{y}-x\| = \|(rq)^{-1}(y-x)\|\le q^{-1}$,
and
\begin{align*}
    |\tilde{y}_i-z_i| &=|\tilde{y}_i-x_i+x_i-z_i| \le 
|\tilde{y}_i-x_i|+|x_i-z_i|I\lsb(\tilde{y}_i-x_i)\cdot(x_i-z_i)\ge 0 \rsb\\
&=(rq)^{-1}|{y}_i-x_i| + |x_i-z_i|I\lsb({y}_i-x_i)\cdot(x_i-z_i)\ge 0 \rsb\\
&= |{y}_i-x_i| + |x_i-z_i|I\lsb({y}_i-x_i)\cdot(x_i-z_i)\ge 0 \rsb\\
&= \max\lsb |y_i-x_i|,|y_i-z_i|\rsb \le 4q^{-1},
\end{align*}
which uses that $(rq)^{-1}\le 1$ when $r>q^{-1}$ and $I\lsb(\tilde{y}_i-x_i)\cdot(x_i-z_i)\ge 0\rsb =I\lsb({y}_i-x_i)\cdot(x_i-z_i)\ge 0 \rsb.$ As a result,
$\tilde{y}\in  B_{d,2}(x,q^{-1})\cap \mathcal{C}_{z,q}.$

Now, using \eqref{lr2}, we have, for $q^{-1}<r<8\sqrt{d}q^{-1}$,
 \begin{align*}
   &\lambda[B_{d,2}(x,r)\cap \lsb \mathcal{C}_{z,q}\setminus\mathcal{D}_{z,q}\rsb]\ge
   \lambda[B_{d,2}(x,q^{-1})\cap \lsb \mathcal{C}_{z,q}\setminus\mathcal{D}_{z,q}\rsb]\\
   &=
  \frac12\lambda[B_{d,2}(x,q^{-1})\cap\mathcal{C}_{z,q}]\ge\frac12\lambda[\mathcal{N}] = \frac{1}{2r^dq^d} \lambda [B_{d,2}(x,r)\cap \mathcal{C}_{z,q}] 
  \ge \frac{1}{2^{3d+1}d^{\frac{d}{2}}}  \lambda [B_{d,2}(x,r)\cap \mathcal{C}_{z,q}].
\end{align*}
 \item When $r\ge 8\sqrt{d}q^{-1}$, we have $\mathcal{C}_{z,q}\subset B_{d,2}(x,r)$, since for any point $y \in \mathcal{C}_{z,q}$, $\|y-x\|\le \|y-z\|+\|z-x\|\le 4\sqrt{d}q^{-1}+4\sqrt{d}q^{-1}\le 8\sqrt{d}q^{-1}$. Thus, the conclusion follows from the same reasoning as \eqref{lr}.
\end{itemize}
\end{itemize}
    \end{itemize}

\subsection{Proof of Lemma \ref{lem:misclassification}}
\label{plem:misclassification}

By definition, $\widehat{Y}$ is conditionally independent of $Y$ given $X$ and $A$. Thus,
 \begin{align*}
 &\P(\widehat{Y}_f=1,Y=0\mid X=x,A=a)=f(x,a)(1-\eta_a(x)), \textnormal{ and }\\
 &\P(\widehat{Y}_f=1,Y=0\mid X=x,A=a)=\eta_a(x)(1-f(x,a)).
 \end{align*}
 This implies that
 \begin{align*}
R(f)&=\P(Y\neq \widehat{Y}_{f})=\sum_{a\in\A}p_a \P(Y\neq \widehat{Y}_{f}|A=a)\\
 &=\sum_{a\in\{0,1\}}p_a\int_\X \lsb \P(\widehat{Y}_f=1,Y=0\mid X=x,A=a)+\P(\widehat{Y}_f=1,Y=0\mid X=x,A=a)\rsb d\Pa(x)\\ 
 &=\sum_{a\in\{0,1\}}p_a\int_\X \lsb f(x,a)(1-\eta_a(x))+\eta_a(x)(1-f(x,a)) \rsb d\Pa(x)\\
 &=\sum_{a\in\{0,1\}}p_a\int_\X \lsb (1-2\eta_a(x))f(x,a)+\eta_a(x)\rsb d\Pa(x).
\end{align*}
For the second result,
 \begin{align*}
\textup{DDP}(f)&=\P\lsb \widehat{Y}_{f}=1\mid A=1\rsb-\P\lsb \widehat{Y}_{f}=1\mid A=0\rsb\\
 &=\sum_{a\in\{0,1\}}(2a-1)\P\lsb \widehat{Y}_{f}=1\mid A=a\rsb=\sum_{a\in\{0,1\}}\int (2a-1)f(x,a)d\Pa(x).
\end{align*}
This finishes the proof.

\subsection{Proof of Lemma \ref{lem:tail}}
\label{plem:tail}

For $a\in \{0,1\}$ and $(x_j,a_j,y_j)\in\mathcal{S}_a$, denote $I_{a,j}=I(a_j=a)$. We have that $I_{a,j}$ are i.i.d.~copies of $I(A=a)$ with $n_{a}=\sum_{j=1}^{n} I_{a,j}$, 
$ I_{a,j}\in\{0,1\}$, and  $\E(I_{a,j})=p_a$. Then, by Hoeffding's inequality, 
$\PN\lsb\left|\frac{n_{a}}{n}-p_a\right|\ge \ep\rsb\le   2\exp\lsb-2n\ep^2\rsb$.
Next,  when $\left|
{n_{a}}/{n}-p_a\right|\le \delta \le {p_a}/2$, we have
\begin{equation*}
\left|\frac{n}{n_{a}}-\frac1{p_a}\right|= \frac{n}{n_{a}p_a}\left|\frac{n_{a}}{n}-p_a\right|\le \frac{2\delta}{p_a^2}.
\end{equation*}
Thus, by taking $\ep = \frac{2\delta}{p_a^2} \le 1/p_a$,
\begin{equation*}
    \PN\lsb\left|\frac{n}{n_{a}}-\frac1{p_a}\right|\ge \ep\rsb\le   \PN\lsb\left|\frac{n_{a}}{n}-p_a\right| \ge \frac{p_a^2\ep}{2}\rsb\le 2\exp\lsb-\frac{np_a^4\ep^2}{2}\rsb.
\end{equation*}

\subsection{Proof of Lemma \ref{lem:uniformDDPbound}}
\label{plem:uniformDDPbound}
When $\epsilon\le \sqrt{p_a/2}$ and $|{n_{a}}/{n}-p_a|\le \ep\sqrt{{p_a}/2}$, we have, 
$n_a\ge n\lsb p_a- \ep\sqrt{{p_a}/{2}}\rsb \ge {n}p_a/{2}.$
By the Dvoretzky–Kiefer–Wolfowitz inequality and \eqref{eq:nabound1} of Lemma \ref{lem:tail}, when $\ep\le\sqrt{{p_a}/{2}}$,
\begin{align*}
&\PN\lsb\sup_{T\in\R}\left|\frac{1}{n_{a}}\sum_{j=1}^{n_{a}} I\lsb \eta_a\lsb\xaj\rsb> T\rsb- \Pa\lsb \eta_a(X)>T\rsb\right|>\ep \rsb\\
&=\PN\lsb\sup_{T\in\R}\left|\frac{1}{n_{a}}\sum_{j=1}^{n_{a}} I\lsb \eta_a\lsb\xaj\rsb> T\rsb- \Pa\lsb \eta_a(X)>T\rsb\right|>\ep, 
\left|\frac{n_{a}}{n}-p_a\right|\le \sqrt{\frac{p_a}{2}}\ep \rsb\\
&+\PN\lsb\sup_{T\in\R}\left|\frac{1}{n_{a}}\sum_{j=1}^{n_{a}} I\lsb \eta_a\lsb\xaj\rsb> T\rsb- \Pa\lsb\eta_a(X)>T\rsb\right|>\ep,\left|\frac{n_{a}}{n}-p_a\right|> \sqrt{\frac{p_a}{2}}\ep \rsb\\
&\le 2\exp\lsb-2n\lsb p_a-\sqrt{\frac{p_a}{2}}\ep\rsb \ep^2\rsb +  \PN\lsb 
\left|\frac{n_{a}}{n}-p_a\right|\ge \sqrt{\frac{p_a}{2}}\ep \rsb\\
&\le2\exp\lsb -np_a\ep^2\rsb+2\exp \lsb -{np_a\ep^2}\rsb = 4\exp \lsb -{np_a\ep^2}\rsb.
\end{align*}

\subsection{Proof of Lemma \ref{lem:diffbetweenDNandDR}}
\label{plem:diffbetweenDNandDR}

Here, we only study the first term on the left hand side of \eqref{eq:diffdrc}; the same argument applies to the second one. 
For 
$t \in \R$ and 
all $j \in [n_1]$,
let $I_{1,j}(t)=I(\eta_1(\xoj)>1/2+t/({2p_1}))$,
and for all $j\in [n_0]$, let $I_{0,j}(t)=I(\eta_0(\xzj)\ge 1/2-t/({2p_0)})$. We have, for $a\in\{0,1\}$ and $j\in[n_a]$,
\begin{equation*}
\begin{array}{lcl}
 \E_{X|A=1}\lsb I_{1,j}(t)\rsb=\Po\lsb\eta_1(X)>\frac12+\frac{t}{2p_1}\rsb; &\E_{X|A=0}\lsb I_{0,j}(t)\rsb=\Pz\lsb\eta_0(X) \ge \frac12-\frac{t}{2p_0}\rsb;\\
D_{n,-}\lsb t\rsb=\frac{1}{n_{1}}\sum\limits_{j=1}^{n_{1}}I_{1,j}(t)-\frac1{n_{0}}\sum\limits_{j=1}^{n_{0}}I_{0,j}(t);  &
D_{-}\lsb t \rsb =\frac{1}{n_{1}}\sum\limits_{j=1}^{n_{1}}\E[I_{1,j}(t)]-\frac1{n_{0}}\sum\limits_{j=1}^{n_{0}}\E[I_{0,j}(t)].
\end{array}
\end{equation*}
By Lemma \ref{lem:uniformDDPbound}, we have, for $\ep \le \sqrt{(p_1\wedge p_0)/2}$ and $a\in \{0,1\}$
$$
\PN\lsb\sup_{T\in\R}\left|\frac1{n_{a}}\sum_{j=1}^{n_{a}}\left( I_{a,j}(t) -\E_{X|A=a}[I_{a,j}(t)]\right)\right|>\ep\rsb\le 4\exp\lsb-{np_a\ep^2}\rsb.$$ It follows that, for $\ep\le \sqrt{(p_1\wedge p_0)/2}$,
\begin{align*}
&\PN\lsb|{D}_{n,r}(t)-D_{-}(t)|> \ep\rsb\\
&\le\sum_{a=0}^1 
\PN\lsb\left|\frac1{n_a}\sum_{j=1}^{n_{a}}\left( I_{a,j}(t)-\E_{X|A=a}[I_{a,j}(t)]\right)\right| >\frac\ep2 \rsb 
\le8\exp\lsb-\frac{n (p_1\wedge p_0)\ep^2}{4} \rsb.
\end{align*}

\subsection{Proof of Lemma \ref{lem:merge}}
\label{plem:{lem:merge}}

\begin{proof}
For \eqref{eq:summation1}, by construction, we have, with $c_{1,\iota_0}=\sum_{k=1}^Kc_{1,\iota_k}$ and
$c_{2,\iota_0}=\min_{k\in[K]} \lsb c_{2,\iota_k}C_k^2\rsb$,
\begin{align*}
&\sum_{k=1}^K\psi_{n,1,\iota_k}(C_k\ep)=\sum_{k=1}^K\lsb c_{1,\iota_k} \exp\lsb-c_{2,\iota_k} \lsb\frac{C_k\ep}{\phi_{n,1}\vee\phi_{n,0}}\rsb^2\rsb\rsb\\
&\le\lsb\sum_{k=1}^Kc_{1,\iota_k}\rsb\exp\lsb\lsb -\min_{k\in[K]} \lsb c_{2,\iota_k}C_k^2\rsb\rsb \lsb\frac{\ep}{\phi_{n,1}\vee\phi_{n,0}}\rsb^2\rsb=
\psi_{n,1,\iota_0}(\ep).
\end{align*}

For  \eqref{eq:summation2},  under the margin condition \eqref{ma2}, we have, for $j\in\{-,+\}$, $k\in[K]$ and $\lsb\max_{k\in[K]}c_k \vee 1\rsb \ep<\ep_0$,  if $\delta=D_j\lsb t_\delta^\star \rsb$, that
$$ g_{\delta,j}(C_k)>U^{-1}_\gamma(C_k\ep)^\gamma> U_\gamma^{-2}C_k^\gamma(U_\gamma \ep^\gamma)>U_\gamma^{-2}C_k^\gamma g_{\delta,j}(\ep).$$
Thus, with $\Istar$ from \eqref{istar},
\begin{align*}
&g_{\delta,j}(\omega(C_k\ep,r_n))= \Istar \cdot g_{\delta,j}(C_k\ep)+(1-\Istar)\cdot g_{\delta,j}(r_n)>\Istar \cdot U_\gamma^{-2}C_k^\gamma \cdot g_{\delta,j}(\ep)+(1-\Istar)\cdot g_{\delta,j}(r_n) \\
&>(U_\gamma^{-2}C_k^\gamma \wedge 1)\lsb \Istar \cdot g_{\delta,j}(\ep)+(1-\Istar)\cdot g_{\delta,j}(r_n)\rsb=
(U_\gamma^{-2}C_k^\gamma \wedge 1) g_{\delta,j}(\omega(\ep,r_n)).
\end{align*}
It then follows that, with $c_{3,\iota_0}= \sum_{k=1}^K c_{3,\iota_k}$ and $c_{4,\iota_0}=
 \lsb \min_{k\in[K]} c_{4,\iota_k}\cdot U_\gamma^{-2}C_k^\gamma\rsb \wedge 1$,
\begin{align*}
&I\lsb \delta=D_j(t^\star_\delta)\rsb\lsb\sum_{k=1}^K\psi_{n,2,\iota_k}\lsb g_{\delta,j}\lsb \omega(C_1\ep,r_n)\rsb\rsb  \rsb
=I\lsb \delta=D_j(t^\star_\delta)\rsb\lsb \sum_{k=1}^K c_{3,\iota_k}\exp\lsb-c_{4,\iota_k}n  g_{\delta,j}^2\lsb \omega(C_1\ep,r_n)\rsb\rsb  \rsb\\
&\le I\lsb \delta=D_j(t^\star_\delta)\rsb \sum_{k=1}^K c_{3,\iota_k}\exp\lsb-\lsb \lsb c_{4,\iota_k}\cdot U_\gamma^{-2}C_1^\gamma\rsb\wedge 1 \rsb n  g_{\delta,j}^2\lsb \omega( \ep,r_n)\rsb  \rsb\\
&\le I\lsb \delta=D_j(t^\star_\delta)\rsb\lsb \lsb \sum_{k=1}^K c_{3,\iota_k}\rsb\exp\lsb-\lsb \lsb \min_{k\in[K]}c_{4,\iota_k}\cdot U_\gamma^{-2}C_k^\gamma\rsb\wedge 1\rsb n  g_{\delta,j}^2\lsb \omega( \ep,r_n)\rsb\rsb\rsb\\
&= I\lsb \delta=D_j(t^\star_\delta)\rsb \psi_{n,2,\iota_0}\lsb g_{\delta,j}(\omega( \ep,r_n))\rsb.
\end{align*}

\end{proof}

\subsection{Proof of Lemma \ref{lem:impacted}}
\label{plem:{lem:impacted}}

We only prove \eqref{D1}, since \eqref{D2} can be verified in the same vein. 
We define the following events: \begin{equation*}
\begin{array}{lcl}
  E_{1}=\lbb\widehat{D}_{n}\lsb t_{\delta}^\star+\ep,0,0\rsb\ge\delta +\Delta_n \rbb;    &  
  E_{2}=\lbb \max_j\left|\widehat{\eta}_1\lsb\xoj\rsb-\eta_1\lsb \xoj\rsb\right|\le \frac{\ep}{8p_1}\rbb; \\
  E_{3}=\lbb \max_j\left|\widehat{\eta}_0\lsb\xzj\rsb-\eta_0\lsb \xzj\rsb\right|\le \frac{\ep}{8p_0}\rbb; &  E_{4}=\lbb\left|\frac{n}{n_{1}}-\frac1{p_1}\right|\le \frac{\ep}{5p_1}\rbb;   \\
  E_{5}=\lbb\left|\frac{n}{n_{0}}-\frac1{p_0}\right|\le \frac{\ep}{5p_0}\rbb; &  E_{6}=\lbb {D}_{n,-}\lsb t_{\delta}^\star+\frac{\ep}{2} \rsb<\delta +\Delta_n\rbb.
\end{array}
\end{equation*}
When $E_{2}$ and $E_{4}$ hold with $\ep \le 1/4$,
we have, 
with $t_\delta^\star$ and $T_{\delta,a}^\star$  from \eqref{eq:t-star-dp},
\begin{align*}
&\frac1{n_{1}}\sum_{j=1}^{n_{1}}I\lsb\widehat\eta_1\lsb \xoj\rsb>1/2+\frac{n\lsb t_{\delta}^\star+\ep\rsb}{2n_{1}}\rsb
=
\frac1{n_{1}}\sum_{j=1}^{n_{1}}I\lsb\widehat\eta_1\lsb \xoj\rsb>T^\star_{\delta,1}+\frac{\ep}{2p_1}+ \frac{t_{\delta}^\star+\ep}2\lsb\frac{n}{n_{1}}-\frac{1}{p_1}\rsb\rsb\\
&\le \frac1{n_{1}}\sum_{j=1}^{n_{1}}I\lsb\eta_1\lsb \xoj\rsb>T^\star_{\delta,1}+\frac{\ep}{2p_1}-\frac{(1+\ep)\ep}{10p_1}-\left|\widehat\eta_1\lsb \xoj\rsb-\eta_1\lsb\xoj\rsb\right|\rsb\\
&\le \frac1{n_{1}}\sum_{j=1}^{n_{1}} I\lsb\eta_1\lsb \xoj\rsb>T^\star_{\delta,1}+\frac{\ep}{4p_1}\rsb.
\end{align*}
Similarly, when $E_{a,3}$ and $E_{a,5}$ hold with $\ep\le 1/4$,
\begin{equation*}
\frac1{n_{0}}\sum_{j=1}^{n_{0}}I\lsb\widehat\eta_0\lsb \xzj\rsb >1/2-\frac{n\lsb t_{\delta}^\star+\ep\rsb}{2n_{0}}\rsb\ge \frac1{n_{0}}\sum_{j=1}^{n_{0}}I\lsb\eta_0\lsb \xzj\rsb\ge T^\star_{\delta,0}-\frac{\ep}{4p_0}\rsb.
\end{equation*}
This implies that, when $E_2, \ldots, E_6$ hold with $\ep \le 1/4$, we have
\begin{align*}
&\widehat{D}_{n}\lsb t_{\delta}^\star+\ep,0,0\rsb
=\frac1{n_{1}}\sum_{j=1}^{n_{1}}I\lsb\widehat\eta_1\lsb \xoj\rsb>1/2+\frac{n\lsb t_{\delta}^\star+\ep\rsb}{2n_{1}}\rsb-\sum_{j=1}^{n_{0}}I\lsb\widehat\eta_0\lsb \xzj\rsb >1/2-\frac{n\lsb t_{\delta}^\star+\ep\rsb}{2n_{0}}\rsb\\
&\le\frac1{n_{1}}\sum_{j=1}^{n_{1}} I\lsb\eta_1\lsb \xoj\rsb> T^\star_{\delta,1}+\frac{\ep}{4p_1}\rsb-\frac1{n_{0}}\sum_{j=1}^{n_{0}}I\lsb\eta_0\lsb \xzj\rsb\ge T^\star_{\delta,0}-\frac{\ep}{4p_0}\rsb
= {D}_{n,-}\lsb t_{\delta}^\star+\frac{\ep}{2} ,0,0\rsb<\delta+\Delta_n.
\end{align*}
It follows that $\PN(\cap_{j=1}^6 E_j)=0$ and
\begin{equation*}
\PN\lsb\widehat{D}_{n}\lsb  t_{\delta}^\star+\ep,0,0\rsb>\delta \rsb\le \PN\lsb\cap_{j=1}^6 E_j\rsb+\sum_{j=2}^6\PN\lsb E_j^c\rsb=\sum_{j=2}^6\PN\lsb E_j^c\rsb.
\end{equation*}
Next, we bound $\PN(E_2^c), \ldots, \PN(E_6^c)$ in order.
First, as  $(\widehat\eta_1,\widehat\eta_0)$ are
$(\phi_{n,1},\phi_{n,0})_{n\ge 1}$-pointwise convergent, 
by \eqref{eq:regression_function}, we have,
for 
$ 8L_\eta (p_1\phi_{n,1}\vee p_0\phi_{n,0}) < \ep < 8U_{\eta}(p_1\wedge p_0)$, i.e., when  $L_\eta \phi_{n,a} < \ep/(8p_a) < U_{\eta}$ for $a\in \{0,1\}$, that
\begin{equation*}
\PN(E_2^c)\le c_{1,\eta} \exp\lsb{-\frac{c_{2,\eta}}{64p_1^2}\lsb\frac{\ep}{\phi_{n,1}}\rsb^2 }\rsb\text{ and }\PN(E_3^c)\le c_{1,\eta} \exp\lsb{-\frac{c_{2,\eta}}{64p_0^2}\lsb\frac{\ep}{\phi_{n,0}}\rsb^2}\rsb.
\end{equation*}
Second, using \eqref{eq:nabound2} of Lemma~\ref{lem:tail} and that $\phi_{n,a}\ge c_\mu n^{-1/2}$, 
for $0<\ep \le 5$, so that $\ep/(5p_a)\le 1/p_a$,
we have
\begin{equation*}
\PN\lsb E_4^c\rsb\le 2\exp\lsb-\frac{np_1^2\ep^2}{50}\rsb\le 2\exp\lsb-\frac{c_\mu^2p_1^2}{50}\lsb\frac{\ep}{\phi_{n,1}}\rsb^2\rsb, \text{ and } \PN\lsb E_5^c\rsb\le 2\exp\lsb-\frac{c_\mu^2p_0^2}{50}\lsb\frac{\ep}{\phi_{n,0}}\rsb^2\rsb.
\end{equation*}
Finally, to bound $\PN(E_6^c)$, denote $c_\delta=\delta -D_{-}(t_{\delta}^\star)$ and note that
\begin{align*}
&\PN\lsb E_6^c\rsb
=\PN\lsb{D}_{n,-}\lsb t_{\delta}^\star+\frac{\ep}{2}\rsb\ge\delta + \Delta_n\rsb
=
\PN\lsb({D}_{n,-}-D_{-})\lsb t_{\delta}^\star+ \frac\ep{2}\rsb\ge\delta+ \Delta_n -D_{-}\lsb t_{\delta}^\star+ \frac{\ep}{2}\rsb\rsb\\
&\le8\exp\lsb-\frac{n (p_1\wedge p_0) \lsb\delta+ \Delta_n -D_{-}\lsb t_{\delta}^\star+ \frac{\ep}{2}\rsb\rsb^2}{4} \rsb =8\exp\lsb-\frac{n (p_1\wedge p_0) \lsb c_\delta+ \Delta_n+g_{\delta,-}\lsb \frac\ep2\rsb\rsb^2}{4} \rsb \\
&\le8 I\left(\delta =D_{-}(t_{\delta}^\star)\right)\exp\lsb-\frac{n (p_1\wedge p_0) \lsb  \Delta_n+g_{\delta,-}\lsb \frac\ep2\rsb\rsb^2}{4} \rsb+ 8\exp\lsb-\frac{n (p_1\wedge p_0)  c_\delta^2}{4} \rsb.
\end{align*}
For $\ep \le 5c_\delta/{\sqrt{2(p_1\vee p_0)}}$,
this can be upper bounded by
\begin{align*}
& 
8 I\left(\delta =D_{-}(t_{\delta}^\star)\right)\exp\lsb-\frac{n (p_1\wedge p_0) \lsb \Delta_n+ g_{\delta,-}\lsb \frac\ep2\rsb\rsb^2}{4} \rsb+ 8 \exp\lsb-\frac{n(p_1\wedge p_0)^2\ep^2}{50}\rsb\\
&\le 8 I\left(\delta =D_{-}(t_{\delta}^\star)\right)\exp\lsb-\frac{n (p_1\wedge p_0) \lsb \Delta_n+ g_{\delta,-}\lsb \frac\ep2\rsb\rsb^2}{4} \rsb+ 8 \exp\lsb-\frac{c_\mu^2(p_1\wedge p_0)^2}{50}\lsb\frac{\ep}{\phi_{n,1}\vee\phi_{n,0}}\rsb^2\rsb.
\end{align*}

In conclusion, by taking $L_{t_1} = 8L_\eta(p_1\vee p_0)$ and $U_{t_1}=(8p_1U_\eta)\wedge (8p_0U_\eta)\wedge (5c_\delta/{\sqrt{2(p_1\vee p_0)}})
\wedge (1/4)$,
we have that, for $L_{t_1}(\phi_{n,1}\vee \phi_{n,0})< \ep< U_{t_1}$,
\eqref{D1} holds with 
$\psi_{n,1,t_1}(\ep)=c_{1,t_1} \exp(-c_{2,t_1} (\ep/[\phi_{n,1}\vee\phi_{n,0}])^2 )$ and $\psi_{n,2,t_1}(\ep) = c_{3,t_1}\exp\lsb-c_{4,t_1}n \ep^2 \rsb$, where $c_{1,t_1}=2c_{1,\eta}+12$, $c_{2,t_1}=(c_{2,\eta}\wedge c_\mu^2(p_1\wedge p_0)^2)/({64(p_1\vee p_0)^2}\vee 50)$, $c_{3,t_1}=8$, and $c_{4,t_1}=(p_1\wedge p_0)/4$.

\subsection{Proof of Lemma \ref{lem:impacted1}}
\label{plem:{lem:impacted1}}
Here,
we only prove \eqref{D3} and \eqref{D4}, since \eqref{D5} and \eqref{D6} can be verified similarly. To prove \eqref{D3}, 
we define the following events:  \begin{equation*}
\begin{array}{lcl}
  \overline{E}_{1}=\lbb\widehat{D}_{n}\lsb t_{\delta}^\star+\ep,0,0\rsb\le\delta -\Delta_n\rbb;    &  
  \overline{E}_{2}=\lbb \max_j\left|\widehat{\eta}_1\lsb\xoj\rsb-\eta_1\lsb \xoj\rsb\right|\le \frac{\ep}{4p_1}\rbb; \\
  \overline{E}_{3}=\lbb \max_j\left|\widehat{\eta}_0\lsb\xzj\rsb-\eta_0\lsb \xzj\rsb\right|\le \frac{\ep}{4p_0}\rbb; &  \overline{E}_{4}=\lbb\left|\frac{n}{n_{1}}-\frac1{p_1}\right|\le \frac{\ep}{4p_1}\rbb;   \\
  \overline{E}_{5}=\lbb\left|\frac{n}{n_{0}}-\frac1{p_0}\right|\le \frac{\ep}{4p_0}\rbb; &  \overline{E}_{6}=\lbb {D}_{n,l}\lsb t_{\delta}^\star+2\ep \rsb>\delta -\Delta_n \rbb.
\end{array}
\end{equation*}
Following the arguments used for proving \eqref{D1}, we have the following facts:
\begin{itemize}
\item[(1)] With $\ep\le 1$,
$\PN\lsb\widehat{D}_{n}\lsb  t_{\delta}^\star+\ep,0,0\rsb\le \delta-\Delta_n \rsb\le \sum_{j=2}^6\PN\lsb \overline{E}_j^c\rsb$.
\item[(2)] For 
$ 4L_\eta (p_1\phi_{n,1}\vee p_0\phi_{n,0}) < \ep < 4U_{\eta}(p_1\wedge p_0)$, i.e., when  $L_\eta \phi_{n,a} < \ep/(4p_a) < U_{\eta}$ for $a\in \{0,1\}$,
\begin{equation*}
\PN(\overline{E}_2^c)\le c_{1,\eta} \exp\lsb{-\frac{c_{2,\eta}}{16p_1^2}\lsb\frac{\ep}{\phi_{n,1}}\rsb^2 }\rsb\text{ and }\PN(\overline{E}_3^c)\le c_{1,\eta} \exp\lsb{-\frac{c_{2,\eta}}{16p_0^2}\lsb\frac{\ep}{\phi_{n,0}}\rsb^2}\rsb.
\end{equation*}

\item[(3)] 
For $0<\ep \le 4$, so that $\ep/(4p_a)\le 1/p_a$,
\begin{equation*}
\PN\lsb \overline{E}_4^c\rsb\le 2\exp\lsb-\frac{c_\mu^2p_1^2}{32}\lsb\frac{\ep}{\phi_{n,1}}\rsb^2\rsb \text{ and } \PN\lsb \overline{E}_5^c\rsb\le 2\exp\lsb-\frac{c_\mu^2p_0^2}{32}\lsb\frac{\ep}{\phi_{n,0}}\rsb^2\rsb.
\end{equation*}
\end{itemize}
Now, to bound $\PN(\overline{E}_6^c)$, since $D_-(t^\star_\delta)=\delta$, we have $$\delta -\Delta_n-D_{-}\lsb t_\delta^\star+2\ep\rsb=D_{-}(t_\delta^\star)-D_{-}\lsb t_\delta^\star+2\ep\rsb -\Delta_n = g_{\delta,-}\lsb2\ep\rsb-\Delta_n.$$ 
Since $\Delta_n\asymp (\log\log n)^{-1}$ and $\ep<r_n\asymp (\log n)^{-1}$, we have  from Condition \ref{ma1} that
when $n$ is large enough,  
$$ (\log\log n)^{-1} \asymp \Delta_n - U_\gamma 2^\gamma\ep^\gamma <\Delta_n -g_{\delta,-}(2\ep)< \sqrt{(p_1\wedge p_0)/2}.$$
Then, 
by Lemma \ref{lem:diffbetweenDNandDR},
\begin{align*}
\PN\lsb \overline{E}_6^c\rsb
&=\PN\lsb{D}_{n,-}\lsb t_{\delta}^\star+2{\ep}\rsb\le\delta-\Delta_n \rsb
=
\PN\lsb({D}_{n,-}-D_{-})\lsb t_{\delta}^\star+ 2\ep\rsb\le\delta -\Delta_n-D_{-}\lsb t_{\delta}^\star+ 2{\ep}\rsb\rsb\\
&=\PN\lsb({D}_{n,-}-D_{-})\lsb t_{\delta}^\star+ 2\ep\rsb\le -\lsb \Delta_n-g_{\delta,-}(2{\ep})\rsb\rsb\le8\exp\lsb-\frac{n (p_1\wedge p_0)  \lsb \Delta_n-g_{\delta,-}(2\ep)\rsb^2}{4} \rsb.
\end{align*}

In conclusion, by taking $L_{t_2} = 4L_\eta(p_1\vee p_0)$ and $U_{t_2}=(4p_1U_\eta)\wedge (4p_0U_\eta)\wedge 1$,
we have that, for $L_{t_2}(\phi_{n,1}\vee \phi_{n,0})< \ep< U_{t_2}$,
\eqref{D3} holds with
$\psi_{n,1,t_2}(\ep)=c_{1,t_2} \exp(-c_{2,t_2} (\ep/[\phi_{n,1}\vee\phi_{n,0}])^2 )$ and $\psi_{n,2,t_2}(\ep) = c_{3,t_2}\exp\lsb-c_{4,t_2}n \ep^2 \rsb$, where $c_{1,t_2}=2c_{1,\eta}+4$, $c_{2,t_2}=(c_{2,\eta}\wedge c_\mu^2(p_1\wedge p_0)^2)/({16(p_1\vee p_0)^2}\vee 32)$, $c_{3,t_2}=8$, and $c_{4,t_2}=(p_1\wedge p_0)/4$.

Next, to prove \eqref{D4}, 
we define the following events:  \begin{equation*}
\begin{array}{lcl}
  \widetilde{E}_{1}=\lbb\widehat{D}_{n}\lsb t_{\delta}^\star+\ep,0,0\rsb\ge\delta -\Delta_n\rbb;    &  
  \widetilde{E}_{2}=\lbb \max_j\left|\widehat{\eta}_1\lsb\xoj\rsb-\eta_1\lsb \xoj\rsb\right|\le \frac{\ep}{4p_1}\rbb; \\
  \widetilde{E}_{3}=\lbb \max_j\left|\widehat{\eta}_0\lsb\xzj\rsb-\eta_0\lsb \xzj\rsb\right|\le \frac{\ep}{4p_0}\rbb; &  \widetilde{E}_{4}=\lbb\left|\frac{n}{n_{1}}-\frac1{p_1}\right|\le \frac{\ep}{4p_1}\rbb;   \\
  \widetilde{E}_{5}=\lbb\left|\frac{n}{n_{0}}-\frac1{p_0}\right|\le \frac{\ep}{4p_0}\rbb; &  \widetilde{E}_{6}=\lbb {D}_{n,l}\lsb t_{\delta}^\star \rsb<\delta -\Delta_n \rbb.
\end{array}
\end{equation*}
Again, by the arguments used in proving \eqref{D1}, we obtain the following facts:
\begin{itemize}
\item[(1)] With $\ep\le 1$,
$\PN\lsb\widehat{D}_{n}\lsb  t_{\delta}^\star+\ep,0,0\rsb\ge \delta-\Delta_n \rsb\le \sum_{j=2}^6\PN\lsb \widetilde{E}_j^c\rsb$.
\item[(2)] For 
$ 4L_\eta (p_1\phi_{n,1}\vee p_0\phi_{n,0}) < \ep < 4U_{\eta}(p_1\wedge p_0)$, i.e., when  $L_\eta \phi_{n,a} < \ep/(4p_a) < U_{\eta}$ for $a\in \{0,1\}$,
\begin{equation*}
\PN(\widetilde{E}_2^c)\le c_{1,\eta} \exp\lsb{-\frac{c_{2,\eta}}{16p_1^2}\lsb\frac{\ep}{\phi_{n,1}}\rsb^2 }\rsb\text{ and }\PN(\widetilde{E}_3^c)\le c_{1,\eta} \exp\lsb{-\frac{c_{2,\eta}}{16p_0^2}\lsb\frac{\ep}{\phi_{n,0}}\rsb^2}\rsb.
\end{equation*}

\item[(3)] 
For $0<\ep \le 4$, so that $\ep/(4p_a)\le 1/p_a$,
\begin{equation*}
\PN\lsb \widetilde{E}_4^c\rsb\le 2\exp\lsb-\frac{c_\mu^2p_1^2}{32}\lsb\frac{\ep}{\phi_{n,1}}\rsb^2\rsb \text{ and } \PN\lsb \widetilde{E}_5^c\rsb\le 2\exp\lsb-\frac{c_\mu^2p_0^2}{32}\lsb\frac{\ep}{\phi_{n,0}}\rsb^2\rsb.
\end{equation*}
\end{itemize}
Now, to bound $\PN(\widetilde{E}_6^c)$, let $c_\delta=\delta-D_{-}(t^\star_\delta)$. For $\Delta_n<c_\delta/2$ and  $\ep \le c_\delta{\sqrt{2/(p_1\vee p_0)}}$,
by Lemma \ref{lem:diffbetweenDNandDR},
\begin{align*}
\PN\lsb \widetilde{E}_6^c\rsb
&=\PN\lsb{D}_{n,-}\lsb t_{\delta}^\star\rsb\ge\delta-\Delta_n \rsb
=
\PN\lsb({D}_{n,-}-D_{-})\lsb t_{\delta}^\star\rsb\ge\delta -\Delta_n-D_{-}\lsb t_{\delta}^\star\rsb\rsb\\
&=\PN\lsb({D}_{n,-}-D_{-})\lsb t_{\delta}^\star\rsb\ge \frac{c_\delta}{2}\rsb\le 8 \exp\lsb \frac{n(p_1\wedge p_0)c_\delta^2}{16}\rsb\\
&\le 
8 \exp\lsb-\frac{n(p_1\wedge p_0)^2\ep^2}{32}\rsb
\le 8 \exp\lsb-\frac{c_\mu^2(p_1\wedge p_0)^2}{32}\lsb\frac{\ep}{\phi_{n,1}\vee\phi_{n,0}}\rsb^2\rsb.
\end{align*}

In conclusion, by taking $L_{t_2} = 4L_\eta(p_1\vee p_0)$ and $U_{t_2}=(4p_1U_\eta)\wedge (4p_0U_\eta)\wedge c_\delta{\sqrt{2/(p_1\vee p_0)}}\wedge 1$,
we have that, for $L_{t_2}(\phi_{n,1}\vee \phi_{n,0})< \ep< U_{t_2}$,
\eqref{D4} holds with
$\psi_{n,1,t_2}(\ep)=c_{1,t_2} \exp(-c_{2,t_2} (\ep/[\phi_{n,1}\vee\phi_{n,0}])^2 )$,
where $c_{1,t_2}=2c_{1,\eta}+12$ and $c_{2,t_2}=(c_{2,\eta}\wedge c_\mu^2(p_1\wedge p_0)^2)/({16(p_1\vee p_0)^2}\vee 32)$.

\subsection{Proof of Lemma \ref{lem:impacted2}}
\label{plem:impacted2}

\begin{proof}
Here, we only  prove \eqref{Db1}, \eqref{Db2} and \eqref{Db3} as the other three claims can be verified similarly: 
\eqref{Db4} is analogous to \eqref{Db2}, \eqref{Db5} and \eqref{Db6} are analogous to \eqref{Db3}.

For \eqref{Db1},  recalling the definition of $\htmin$ and $\htmid$ in \eqref{hts}, we have $0\le \htmin\le \htmid$ and
$\lbb\htmid> r_n\rbb \subset \{\widehat{D}_n(r_n,0,0)\ge \delta\}$.
Now, $t^\star_\delta=0$ if and only if $D_-(0)\le \delta$. 
Further, take $L_r=L_{t_1}$ and $U_r=U_{t,1}$. By Lemma \ref{lem:impacted}, when $L_r<r_n<U_r$, we have, with $c_{i,r}=c_{i,t_1}, i\in[4]$, that
\begin{align*}
    &\PN\lsb \htmid-\htmin>r_n \rsb\le   \PN\lsb \htmid>r_n\rsb\\
    &\le   \PN\lsb \widehat{D}_n(r_n,0,0)\ge \delta\rsb 
        \le  \psi_{n,1,r}(r_n)+I(\delta = D_-(t^\star_\delta))\psi_{n,2,r}\lsb g_{\delta,-}\lsb\frac{r_n}{2}\rsb\rsb.
    \end{align*}

For \eqref{Db2},  by the definition of $\htmin$ and $\htmid$, 
 we have $0\le \htmin\le \htmid$, as well as
\begin{align*}
       &\lbb\htmin < t_\delta^\star -\frac{r_n}2\rbb \subset \lbb \widehat{D}_n\lsb t^\star_\delta-\frac{r_n}{2},0,0\rsb\le \delta+\Delta_n\rbb
\end{align*}
and
$\lbb\htmid> t^\star_\delta +\frac{r_n}2\rbb \subset \lbb\widehat{D}_n\lsb t^\star_\delta + \frac{r_n}2,0,0\rsb\ge \delta\rbb$.

Take $L_r=2(L_{t_1}\vee L_{t_2})$, $U_r=2(U_{t,1}\wedge U_{t,2})$ and
$U_{\Delta_1}=  (D_+(t^\star_\delta)-\delta)/2$. By Lemmas \ref{lem:impacted} and \ref{lem:impacted1}, when $L_{b}(\phi_{n,1}\vee\phi_{n,0})< r_n < U_r$ and
$ g_{\delta,+}(4r_n)<  \Delta_n < U_{\Delta,r}$, we have, with $c_{1,r}=c_{1,t_1}+c_{1,t_2}$, $c_{2,r}=c_{2,t_1}\wedge c_{2,t_2}$, $c_{3,r}=c_{3,t_2}$ and $c_{4,r}=c_{4,t_1}$ that
\begin{align*}
    &\PN\lsb \htmid-\htmin>r_n \rsb\le  \PN\lsb \htmin<t^\star_\delta-\frac{r_n}2\rsb +  \PN\lsb \htmid>t^\star_\delta+\frac{r_n}{2}\rsb\\
    &\le   \PN\lsb \widehat{D}_n\lsb t^\star_\delta-\frac{r_n}{2},0,0\rsb\le  \delta+\Delta_n\rsb +\PN\lsb \widehat{D}_n\lsb t^\star_\delta+\frac{r_n}2,0,0\rsb\ge  \delta\rsb \\
        &\le  \psi_{n,1,t_2}\lsb\frac{r_n}{2}\rsb+\psi_{n,1,t_1}\lsb\frac{r_n}{2}\rsb +I(\delta =D_-(t^\star_\delta))\psi_{n,2,t_1}\lsb g_{\delta,-}\lsb\frac{r_n}{4}\rsb\rsb\\
        &\le \psi_{n,1,r}\lsb\frac{r_n}2\rsb+I(\delta =D_-(t^\star_\delta))\psi_{n,2,r}\lsb g_{\delta,-}\lsb\frac{r_n}{4}\rsb\rsb .
    \end{align*}
    
For \eqref{Db3},  again by the definition of $\htmin$ and $\htmid$, 
 we have $0\le \htmin\le \htmid$, as well as
\begin{align*}
       &\lbb\htmin > t_\delta^\star -2r_n\rbb \subset \lbb \widehat{D}_n\lsb t^\star_\delta-2{r_n},0,0\rsb\ge \delta+\Delta_n\rbb
\end{align*}
and
$\lbb\htmid<t^\star_\delta -{r_n}\rbb \subset \lbb\widehat{D}_n\lsb t^\star_\delta-r_n,0,0\rsb\le \delta\rbb$.

Take $L_r=L_{t_1}\vee (L_{t_2}/2)$, $U_r=(U_{t,1}\wedge U_{t,2})/2$ and
$U_{\Delta,r}= \sqrt{(p_1\wedge p_0)/2} \wedge ((\delta-D_-(t^\star_\delta))/2)$. By Lemmas \ref{lem:impacted} and \ref{lem:impacted1},  when $L_{r}(\phi_{n,1}\vee\phi_{n,0})< r_n < U_r$ and
$ g_{\delta,+}(4r_n)<  \Delta_n < U_{\Delta_1}$, we have, with $c_{1,r}=c_{1,t_1}+c_{1,t_2}$, $c_{2,r}=c_{2,t_1}\wedge c_{2,t_2}$, $c_{3,r}=c_{3,t_1}+c_{3,t_2}$ and $c_{4,r}=c_{4,t_1}\wedge c_{4,t_2}$ that
\begin{align*}
    &\PN\lsb \htmid-\htmin<r_n \rsb\le  \PN\lsb \htmid>t^\star_\delta-2r_n\rsb +  \PN\lsb \htmid<t^\star_\delta-r_n\rsb\\
    &\le   \PN\lsb \widehat{D}_n\lsb t^\star_\delta-2r_n,0,0\rsb\ge  \delta+\Delta_n\rsb +\PN\lsb \widehat{D}_n\lsb t^\star_\delta-r_n,0,0\rsb\le  \delta\rsb \\
        &\le \psi_{n,1,t_2}(2r_n) + \psi_{n,2,t_2}\lsb \Delta_n -g_{\delta,+}\lsb4r_n\rsb\rsb + \psi_{n,1,t_1}(r_n) +   \psi_{n,2,t_1}\lsb g_{\delta,+}\lsb\frac{r_n}2\rsb\rsb\\
        &\le  \psi_{n,1,r}(r_n) + \psi_{n,2,r}\lsb \lsb \Delta_n -g_{\delta,+}\lsb4r_n\rsb\rsb \wedge g_{\delta,+}\lsb\frac{r_n}2\rsb\rsb \le\psi_{n,1,r}(r_n) + \psi_{n,2,r}\lsb   g_{\delta,+}\lsb\frac{r_n}2\rsb\rsb.
    \end{align*}
    The last inequality holds since, when $\Delta_n>2g_{\delta,+}(4r_n)$, $\Delta_n-g_{\delta,+}(4r_n)>g_{\delta,+}(4r_n)>g_{\delta,+}(r_n/2)$.
\end{proof}

\subsection{Proof of Lemma \ref{lem:expectationbound}}
\label{plem:expectationbound}

Here, we only prove the first claim with $a=1$, 
since the other results can be derived with the same argument. 
During the proof, we use $C,C_1,C_2$, etc.,  to represent constants that may vary from line to line. Let $0<C_J<1/4$ be a constant determined later. For 
$J_n=\lfloor C_J\cdot \log_2 n\rfloor$ and
$1\le j\le J_n$, 
with $L_\eta$, $U_\eta$ and $(\phi_{n,1},\phi_{n,0})_{n\ge 1}$  from Definition \ref{pc},
$L_T$ and  $U_T$ from Corollary \ref{constT},
$D_{-}$ from \eqref{eq:Disfun1}, $D_{+}$ from \eqref{eq:Disfun1},
$t_\delta^\star$ from \eqref{eq:t-star-dp},
$\delta $ from \eqref{dt}, 
and $\gamma$ from the $\gamma$-exponent condition in the upper bound from Definition \ref{m},
we denote
\begin{align}\label{abd}
a_{n,j}&=2^{j}(L_\eta\vee L_T)(\phi_{n,1}\vee \phi_{n,0}),\\ 
b_{n,j}&=\left\{
\begin{array}{ll}
    2^{2+j/\gamma}L_\gamma^{-1/\gamma}n^{-1/2\gamma}, & t^\star_\delta>0 \text{ and } D_-(t^\star_\delta)=D_+(t^\star_\delta)=\delta;\\
  0,   &  \text{otherwise},
\end{array}
\right.\\
d_{n,j}&=a_{n,j}+b_{n,j}+\ell_{n,1}.
\end{align}
Here, we set $C_J$ such that $d_{n,J_n}<\Delta_n \asymp (\log n)^{-1}$.
From Definition \ref{pc}, it follows that
   $$\PN\lsb|\widehat{\eta}_{1}-\eta^\star_{1}|>\frac12a_{n,j-1}\rsb\le c_{1,\eta}\exp\lsb-c_{2,\eta}(L_\eta\vee L_T)^22^{2{j-2}}\rsb\le C_{1}\exp\lsb-C_{2}2^{2j}\rsb,$$
where in the second inequality we have defined the constants $C_1, C_2$ appropriately.
 We first handle the case
 where $t^\star_\delta>0$ and $D_-(t^\star_\delta)=D_+(t^\star_\delta)=\delta$, and then consider the complement of this case.
 \begin{itemize}[]
\item \textbf{Scenario 1}: $t^\star_\delta>0$ and $D_-(t^\star_\delta)=D_+(t^\star_\delta)=\delta$.

In this case, we have
 $$d_{n,j}=a_{n,j}+b_{n,j}+\ell_{n,1}\le C\lsb2^j(\phi_{n,1}\vee\phi_{n,0})+2^{j/\gamma}n^{-{1}/{2\gamma}}+\ell_{n,1}\rsb.$$
 Moreover, by the margin condition \eqref{ma2}  from Definition \ref{m},, we have
 $$g_{\delta,-}\lsb\frac{b_{n,j}}{2}\rsb\wedge g_{\delta,+}\lsb\frac{b_{n,j}}{2}\rsb \ge L_\gamma 2^{-2\gamma} b_{n,j}^\gamma = 2^j n^{-\frac12}.$$
By Corollary \ref{constT},
 and using appropriate parts of $1/2\cdot a_{n,j-1}+b_{n,j-1}$ in the upper bound,
\begin{align*}
&\PN\lsb\Thdo-\Tsdo>1/2\cdot a_{n,j-1}+b_{n,j-1}\rsb\\
&\le c_{1,T}\exp\lsb-c_{2,T}\lsb\frac{a_{n,j-1}}{2(\phi_{n,1}\vee\phi_{n,0})}\rsb^2\rsb
+c_{3,T}\exp\lsb-c_{4,T}n \lsb g_{\delta,-}\lsb \frac{b_{n,j}}2 \rsb \wedge g_{\delta,+}\lsb \frac{b_{n,j}}2 \rsb\rsb^2\rsb\\
&\le c_{1,T}\exp\lsb-c_{2,T}2^{2(j-2)}(L_\eta\vee L_T)^2\rsb+c_{3,T}\exp\lsb-c_{4,T}2^{2j}\rsb
\le C_{1}\exp\lsb-C_{2} 2^{2j}\rsb.
\end{align*}
Thus, we have
 \begin{align}\label{htt}
\PN\left(\Thdo-\Tsdo> a_{n,j-1}/2+b_{n,j-1}\right)
\le  C_1\exp\lsb-C_22^{2j}\rsb.
\end{align}
\item \textbf{Scenario 2}: Other cases.

In this case, we have 
 $d_{n,j}=a_{n,j}+\ell_{n,1}\le C\lsb2^j(\phi_{n,1}\vee\phi_{n,0})+\lo\rsb.$
 By the margin condition \eqref{ma2}  from Definition \ref{m}, we have,  for $j\in\{-,+\}$ that $g_{\delta,j}({r_n}/{4})>L_\gamma 2^{-2\gamma}r_n^\gamma$. It follows that,
$$I(\delta= D_-(t^\star_\delta))\psi_{n,2,T}
\lsb g_{\delta,-}\lsb\frac{r_n}{4}\rsb\rsb + I(\delta= D_+(t^\star_\delta))\psi_{n,2,T}
\lsb g_{\delta,+}\lsb\frac{r_n}{4}\rsb\rsb \le 2\psi_{n,2,T}\lsb L_\gamma 2^{-2\gamma}r_n^\gamma \rsb.$$
When $r_n\asymp (\log\log n)^{-1}$, we have $nr_n^{2\gamma} \ge Cn^{\frac12}\ge  C2^{2J_n}$. Then, by Corollary~\ref{constT}, we have
 \begin{align*}
&\PN\left(\Thdo-\Tsdo> a_{n,j-1}/2+b_{n,j-1}\right)
=
\PN\left(\Thdo-\Tsdo> a_{n,j-1}/2\right)\\
&\le \psi_{n,1,T}(a_{n,j-1}/2) + 2\psi_{n,2,T}( L_\gamma 2^{-2\gamma}r_n^\gamma)\\
&=c_{1,T}\exp\lsb-c_{2,T} 2^{2(j-2)}(L_\eta\vee L_T)^2\rsb + 2c_{3,T}\exp\lsb -c_{3,T} n L^2_\gamma 2^{-4\gamma}r_n^{2\gamma}\rsb
\le  C_1\exp\lsb-C_22^{2j}\rsb.
\end{align*}
 \end{itemize}
Now, we 
consider the disjoint sets
\begin{align*}
\mathcal{C}_0&=\lbb\eta_1>\Tsdo+d_{n,J_n},\,\widehat{\eta}_1 \le \Thdo+\ell_{n,1}\rbb;\quad
\mathcal{C}_1=\lbb\Tsdo<\eta_1\le \Tsdo+d_{n,1}, \,\widehat{\eta}_1\le \Thdo+\ell_{n,1}\rbb;\\
\mathcal{C}_j&=\lbb\Tsdo+d_{n,j-1}<\eta_1\le \Tsdo+d_{n,j},\, \widehat{\eta}_1\le \Thdo+\ell_{n,1}\rbb, \ \ j\ge2.
\end{align*}
Clearly,
 $\lbb x: \eta_1(x)>\Tsdo,\, \widehat{\eta}_1(x)\le \Thdo+\lo\rbb \subset \cup_{j=0}^{J_n} \mathcal{C}_j$. 
 We bound the quantity of interest over the sets $\mathcal{C}_j$ individually.

\begin{itemize}[]
\item \textbf{On $\mathcal{C}_1$}:
Since $\mathcal{C}_1 \subset \{0 < |\eta_1 - \Tsdo| \le  d_{n,1}\}$, 
using the margin condition from Definition \ref{m},
we have
 \begin{align*}
\EN\lmb\int_{\mathcal{C}_1} \left|\eta_1(x)-\Tsdo\right|d\P_{X|A=1}(x)\rmb
\le d_{n,1}\P_{X|A=1}\lsb \Tsdo<\eta_1(X)\le \Tsdo +d_{n,1}\rsb\le U_\gamma d_{n,1}^{1+\gamma}.
  \end{align*}
 \textbf{On $\mathcal{C}_j$ for $2\le j\le J_n$}:
For $j\ge 2$, we have
due to the definitions from \eqref{abd},
$\mathcal{C}_j \subset \mathcal{C}_{j,1} \cup \mathcal{C}_{j,2}$
with 
\begin{equation*}
\mathcal{C}_{j,1}=\lbb|\widehat\eta_1-\eta_1|> \frac12a_{n,j-1}\rbb \cap\lbb \Tsdo<\eta_1(x)\le \Tsdo+d_{n,j}\rbb,
\end{equation*}
and
\begin{equation*}
\mathcal{C}_{j,2}=\lbb\Thdo-\Tsdo>\frac12a_{n,j-1}+ b_{n,j-1}\rbb \cap \lbb \Tsdo<\eta_1(x)\le \Tsdo+d_{n,j}\rbb.
\end{equation*}
Using Fubini's theorem,
the margin condition from Definition \ref{m},
and \eqref{htt},
\begin{align*}
&\EN\lmb\int_{\mathcal{C}_j}\left|\eta_1(x)-\Tsdo\right|d\P_{X|A=1}(x)\rmb\\
&\le\EN\lmb\int_{\mathcal{C}_{j,1}}\left|\eta_1(x)-\Tsdo\right|d\P_{X|A=1}(x)\rmb+\EN\lmb\int_{\mathcal{C}_{j,2}}\left|\eta_1(x)-\Tsdo\right|d\P_{X|A=1}(x)\rmb\\
&=d_{n,j}\int \lmb\PN\lsb\left|\widehat\eta_1(x)-\eta_1(x)\right|> \frac12a_{n,j-1}\rsb +\PN\lsb\Thdo-\Tsdo> \frac12a_{n,j-1}+b_{n,j-1}\rsb \rmb\\
&\qquad\qquad\qquad\qquad\qquad\qquad\qquad\cdot I\lsb \Tsdo<\eta_1(x)\le \Tsdo+d_{n,j}\rsb d\P_{X|A=1}(x)\\
&\le d_{n,j} C_{1}\exp\lsb-C_22^{2j} \rsb\P_{X|A=1}(\Tsdo<\Thdo\le \Tsdo+d_{n,j})
\le C_1U_\gamma d^{\gamma+1}_{n,j}\exp\lsb-C_22^{2j} \rsb.
\end{align*}

\item \textbf{On $\mathcal{C}_0$}:
Finally, 
$$\mathcal{C}_0\subset \lbb\left|\widehat\eta_1-\eta_1\right|> a_{n,J_n}/2\rbb\cup \lbb\Thdo-\Tsdo > a_{n,J_n}/2+b_{n,J_n}\rbb.$$
Using
that for all $x$, $|\eta_1(x)-\Tsdo|\le 1$,
by Fubini’s theorem, 
and by the margin condition from Definition \ref{m},
and \eqref{htt},
we obtain that
\begin{align*}
&\EN\lsb\int_{\mathcal{C}_0}\left|\eta_1(x)-\Tsdo\right|d\P_{X|A=1}\rsb\le\EN\P_{X|A=1}\lsb\mathcal{C}_0\rsb\\
&\le\PN\lsb\left|\widehat\eta_1-\eta_1\right|>\frac12a_{n,J_n}\rsb+\PN\lsb\Thdo-\Tsdo>\frac12a_{n,J_n}+b_{n,J_n}\rsb\\
&\le 2C_1\exp\lsb-C_22^{2J_n}\rsb = 2C_1\exp\lsb-C_2n^{2C_J}\rsb
\le C\lsb\phi_{n,1}\vee\phi_{n,0}\vee\ell_{n,1}\vee n^{-\frac{1}{2\gamma}}\rsb^\gamma.
\end{align*}
\end{itemize}
To conclude, we have
\begin{align*}
&\EN \int_{\eta_1(x)>\Tsdo,\widehat{\eta}_1(x)\le \Thdo+\lo}|\eta_1(x)-T^\star_{1}|d\Po(x)\le\sum_{j=0}^{J_n} \EN \int_{\mathcal{C}_j}|\eta_1(x)-T^\star_{1}|d\Po(x)\\
&\le  U_\gamma d_{n,1}^{1+\gamma}+\sum_{j=2}^{J_n} C_1U_\gamma d^{\gamma+1}_{n,j}\exp\lsb-C_22^{2j} \rsb\\
&\le  \left\{\begin{array}{ll}
C\sum_{j=1}^{J_n}\exp\lsb-C_22^{2j}\rsb \lsb2^j(\phi_{n,1}\vee\phi_{n,0})+2^{j/\gamma} n^{-\frac1{2\gamma}}+\lo\rsb^{\gamma+1},  & 0<D_-(t^\star_\delta)=\delta =D_{+}\lsb t_\delta^\star\rsb ; \\
C\sum_{j=1}^{J_n}\exp\lsb-C_22^{2j} \rsb\lsb2^j(\phi_{n,1}\vee\phi_{n,0})+\lo\rsb^{\gamma+1},&  \text{otherwise} ,
\end{array}\right.\\
&\le C\lsb\lsb\phi_{n,1}\vee\phi_{n,0}\vee\ell_{n,1}\rsb^{\gamma+1} + I\left(0<D_-(t^\star_\delta)=\delta =D_{+}\lsb t_\delta^\star\rsb\right)\cdot n^{-\frac{\gamma+1}{2\gamma}}\rsb.
\end{align*}
This finishes the proof.

\subsection{Proof of Lemma \ref{lem:deltaneq}}
\label{plem:deltaneq}

By the definition of $\widetilde\delta$ from and $\widehat{\delta}$, we have
$$
\PN\lsb \widehat{\delta}\neq \widetilde{\delta}\rsb =\left\{\begin{array}{ll}
  \PN\lsb\widehat{D}_n(r_n,0,0) \ge \delta+\Delta_n\rsb,  & D_-(0) \le \delta; \\
  \PN\lsb\widehat{D}_n(r_n,0,0) < \delta+\Delta_n\rsb,  & D_-(0) > \delta.
\end{array}\right.
$$
We consider the two cases in order:
\begin{itemize}
    \item Case (1): $D_-(0) \le \delta$.

In this case, we have $t_\delta^\star=0$. By \eqref{D1} of Lemma \ref{lem:impacted}, with $L_{t_1}$, $U_{t_1}$ and $c_{i,t_1}, i\in[4]$ from Lemma \ref{lem:impacted}, we have, for
$L_{t_1}\lsb \phi_{n,1}\vee \phi_{n,0}\rsb< r_n<U_{t_1}$ and $\Delta_n>0$,
\begin{align*}
    \PN\lsb\widehat{D}_n(r_n,0,0)\ge \delta+\Delta_n\rsb \le \psi_{n,1,t_1}(r_n) + \psi_{n,2,t_1}(\Delta_n).
\end{align*}
Thus,
\eqref{eq:deltares} holds with
 $c_{i,\delta}=c_{i,t_1}, i\in[4]$.
    \item Case (2): $D_-(0) > \delta$.

In this case, we define the following events:  \begin{equation*}
\begin{array}{lcl}
  \widehat{E}_{1}=\lbb\widehat{D}_{n}\lsb r_n,0,0\rsb<\delta +\Delta_n\rbb;    &  
  \widehat{E}_{2}=\lbb \max_j\left|\widehat{\eta}_1\lsb\xoj\rsb-\eta_1\lsb \xoj\rsb\right|\le \frac{r_n}{2p_1}\rbb; \\
  \widehat{E}_{3}=\lbb \max_j\left|\widehat{\eta}_0\lsb\xzj\rsb-\eta_0\lsb \xzj\rsb\right|\le  \frac{r_n}{2p_0}\rbb; &    \widehat{E}_{4}=\lbb {D}_{n,-}\lsb 2r_n \rsb\ge \delta+\Delta_n \rbb.
\end{array}
\end{equation*}

Follow the same arguments for proving \eqref{D1}, we have the following facts:
\begin{itemize}
\item[(1)] with $\Delta_n>0$,
\begin{equation*}
\PN\lsb\widehat{D}_{n}\lsb  r_n,0,0\rsb< \delta+\Delta_n \rsb\le \PN\lsb\widehat{D}_{n}\lsb  r_n,0,0\rsb< D_-(0)-\Delta_n \rsb\le \sum_{j=2}^4\PN\lsb \widehat{E}_j^c\rsb.
\end{equation*}
\item[(2)] for 
$ 2L_\eta (p_1\phi_{n,1}\vee p_0\phi_{n,0}) < r_n < 2U_{\eta}(p_1\wedge p_0)$, i.e., when  $L_\eta \phi_{n,a} < \ep/(2p_a) < U_{\eta}$ for $a\in \{0,1\}$,
\begin{equation*}
\PN(\widehat{E}_4^c)\le c_{1,\eta} \exp\lsb{-\frac{c_{2,\eta}}{4p_1^2}\lsb\frac{\ep}{\phi_{n,1}}\rsb^2 }\rsb\text{ and }\PN(\widehat{E}_3^c)\le c_{1,\eta} \exp\lsb{-\frac{c_{2,\eta}}{4p_0^2}\lsb\frac{\ep}{\phi_{n,0}}\rsb^2}\rsb.
\end{equation*}
Now, to bound $\PN(\widehat{E}_4^c)$. We have $t^\star_\delta>0$ when $D_-(0)>\delta$. For  $r_n<t^\star_\delta/4$,
$$\delta +\Delta_n-D_{-}\lsb 2r_n\rsb\le D_{+}(t_\delta^\star)+\Delta_n-D_{-}\lsb t_\delta^\star/2\rsb \le \Delta_n.$$ 
Then, for $\Delta_n \le \sqrt{(p_1 \wedge p_0) /2}$,
by Lemma \ref{lem:diffbetweenDNandDR},
\begin{align*}
\PN\lsb \widehat{E}_4^c\rsb
&=\PN\lsb{D}_{n,-}\lsb 2r_n\rsb\ge\delta+\Delta_n  \rsb
=
\PN\lsb({D}_{n,-}-D_{-})\lsb 2r_n\rsb\ge\delta+\Delta_n -D_{-}\lsb 2r_n\rsb\rsb\\
&=\PN\lsb({D}_{n,-}-D_{-})\lsb2r_n \rsb\ge  \Delta_n\rsb\le 8 \exp\lsb \frac{n(p_1\wedge p_0)\Delta_n^2}{4}\rsb.
\end{align*}

In conclusion, by taking $L_{\delta} = 2L_\eta(p_1\vee p_0)$, $U_{\delta}=(t^\star_\delta/4)\wedge (2U_\gamma(p_1\wedge p_0))$ and $U_{\Delta,\delta} = \sqrt{(p_1\wedge p_0)/2}$,
we have that, for $L_{\delta}(\phi_{n,1}\vee \phi_{n,0})< r_n< U_{\delta}$ and $0<\Delta_n<U_{\Delta,\delta}$,
\eqref{eq:deltares} holds with
with $c_{1,\delta}=2c_{1,\eta}$, $c_{2,\delta}=c_{2,\eta}/({4(p_1\vee p_0)^2})$, $c_{3,\delta}=8$ and $c_{4,\delta}=(p_1\wedge p_0)/4$.
\end{itemize}
    \end{itemize}

\subsection{Proof of Lemma \ref{lem:piset1}}
\label{plem:piset1}
During the proof, we denote
$\mathcal{E}_{\pi,a}=\psi_{n,1,\pi}\lsb {\ela} \rsb +\sum_{j\in\{-,+\}}I(\delta=D_j(t^\star_\delta))\psi_{n,2,\pi}\lsb \xi\lsb{\ela},r_n\rsb  \rsb.$
For the first two terms of  \eqref{pi11}, note that
\begin{align}
&I\left(\eta_a(x)>\Tsdo+\ela+|\widehat\eta_a(x)-\eta_a(x)|+|\Thda-\Tsda|\right)\nonumber\\
\label{s}
&\le I\left(\widehat\eta_a(x)>\Thda+\ela\right) = I\left(\eta_a(x)>\Tsda+\ela+\eta_a(x)-\widehat\eta_a(x) +\Thda-\Tsda\right)\\
\nonumber
&\le I\left(\eta_a(x)>\Tsda+\ela-|\widehat\eta_a(x)-\eta_a(x)|-|\Thda-\Tsda|\right).
\end{align}
Using the margin condition from Definition \ref{m},
it follows that, on the event that 
$\sup_{x\in\Omega}|\widehat{\eta}_a(x)-\eta_a(x)|\le \ela/2$ and $|\Thda-\Tsda|\le\ela/2$,  
   \begin{equation}\label{boundsofetas}
   \Pa({\eta}_a(X)> \Tsda )-U_\gamma  (4p_a\ela)^\gamma\le  \Pa(\widehat{\eta}_a(X)> \Thda+\ela)\le \Pa({\eta}_a(X)> \Tsda ).
   \end{equation} 
Thus, if we take $L_{\pi_1}=L_t$, $U_{\pi_1}=U_T$ and $U_{\Delta,\pi_1}=U_{\Delta,T}$,
by Corollary \ref{constT} and as $(\widehat\eta_1,\widehat\eta_0)$ are
$(\phi_{n,1},\phi_{n,0})$-pointwise convergent, 
there exist constants $c_{i,\pi}>0. i\in[4]$ such that,
for $L_{\pi_1}(\phi_{n,1}\vee \phi_{n,0})<r_n<U_{\pi_1}$, $2\lsb g_{\delta,-}(4r_n)\vee g_{\delta,+}(4r_n)\rsb<\Delta_n<U_{\Delta,\pi_1}$,
$\ep> U_\gamma  (4p_a\ela)^\gamma $ and $ L_{\pi_1}(\phi_{n,1}\vee \phi_{n,0})< \ela/2 < r_n$, with 
the functions 
$\psi_{n,j,\iota}$ from \eqref{psi},  
  \begin{align*}
&\PN(\pihap > \pisap+\ep)\le \PN(\pihap > \pisap)\\
&\le \PN\lsb\sup_{x\in\Omega}|\widehat{\eta}_a(x)-\eta_a(x)|> \frac\ela2\rsb +\PN\lsb|\Thda-\Tsda|>\frac\ela2\rsb\\
&\le\psi_{n,1,\eta}\lsb \frac\ela2\rsb+  \psi_{n,1,T}\lsb \frac\ela2\rsb +\sum_{j\in\{-,+\}}I(\delta=D_j(t^\star_\delta))\psi_{n,2,T}\lsb\xi\lsb \frac{\ela}{2} ,r_n\rsb \rsb
\\
&\le\psi_{n,1,\pi}\lsb {\ela} \rsb +\sum_{j\in\{-,+\}}I(\delta=D_j(t^\star_\delta))\psi_{n,2,\pi}\lsb \xi\lsb{\ela},r_n\rsb  \rsb=\mathcal{E}_{\pi,a}.
  \end{align*}
Similarly, 
\begin{align*}
\PN(\pihap <\pisap-\ep)&\le \PN(\pihap <\pisap-U_\gamma  (4p_a\ela)^\gamma )
\le\mathcal{E}_{\pi,a}.
  \end{align*}

For the last two terms of \eqref{pi11}, 
on the event that  $\sup_{x\in\Omega}|\widehat{\eta}_1(x)-\eta_1(x)|\le \lo/2$ and $|\Thdo-\Tsdo|<\lo/2$,
\begin{align*}
&I(\eta_1(x)=\Tsdo)
\le I(|\widehat\eta_1(x)-\Thdo|\le\lo)
\le I(|\eta_1(x)-\Tsdo|\le \lo+|\widehat\eta_1(x)-\eta_1(x)|+|\Thdo-\Tsdo|).
\end{align*}
Again, by the margin condition from Definition \ref{m},
on the event that    $\sup_{x\in\Omega}|\widehat{\eta}_1(x)-\eta_1(x)|\le \lo/2$ and $|\Thdo-\Tsdo|<\lo/2$,
   \begin{equation*}
  \Po({\eta}_1(x) =\Tsdo) +2U_\gamma  (4p_1\lo)^\gamma\ge 
  \Po(|\widehat{\eta}_1(x)- \Thdo|\le\lo)\ge       \Po({\eta}_1(x) =\Tsdo).
   \end{equation*}
 Thus, by Corollary \ref{constT} and the fact that $(\widehat\eta_1,\widehat\eta_0)$ are
$(\phi_{n,1},\phi_{n,0})$-pointwise convergent, we have, 
for $L_{\pi_1}(\phi_{n,1}\vee \phi_{n,0})<r_n<U_{\pi_1}$, $2\lsb g_{\delta,-}(4r_n)\vee g_{\delta,+}(4r_n)\rsb<\Delta_n<U_{\Delta,\pi_1}$,
$\ep> U_\gamma  (4p_a\ela)^\gamma $ and $ (L_\eta \vee L_T)(\phi_{n,1}\vee \phi_{n,0})< \ela/2 < r_n$ that
\begin{align*}
\PN(\pihoe >\pisoe + \ep )\le 
\PN(\pihoe  >\pisoe +2U_\gamma  (4p_1\lo)^\gamma ) \le \mathcal{E}_{\pi,a}
  \end{align*}
  and
$\PN(\pihoe <\pisoe -\ep)\le\PN(\pihoe <\pisoe )
\le \mathcal{E}_{\pi,a}$,
finishing the proof.

\subsection{Proof of Lemma \ref{lem:piset2}}
\label{plem:piset2}
During the proof, we denote, 
$$ \mathcal{E}'_{\pi,a}= \psi_{n,1,\pi}\lsb \ela \rsb +\sum_{j\in\{\-,+} I(\delta= D_j(t^\star_\delta))\cdot
\psi_{n,2,\pi}\lsb g_{\delta,j}(\ela)\rsb+ 4\exp\lsb-
\frac{np_a\ep^2}4\rsb  $$
For $a\in\{0,1\}$, define the following events:
\begin{align*}
&E_{a,1}=\lbb\max_{j\in [n_a]}|\widehat{\eta}_a(\xaj)-\eta_a(\xaj)|\le \ela/2\rbb, \qquad E_{a,2}= \lbb|\Thda-\Tsda|\le \ela/2\rbb,\\
&E_{a,3}=\lbb\frac{1}{n_a}\sum\limits_{j=1}^{n_a}I\left({\eta}_a(\xaj)> \Tsda  \right) -\Pa\left(\eta_a(X)>\Tsda \right)\le\ep\rbb,\\
&E_{a,4}=\lbb
\frac{1}{n_a}\sum\limits_{j=1}^{n_a}I({\eta}_a(\xaj)\ge \Tsda +2\ela ) -\Pa(\eta_a(X)>\Tsda +{2\ela}) \ge- \frac\ep2\rbb,\\
&E_{a,5}=\lbb
\frac{1}{n_a}\sum\limits_{j=1}^{n_a}I(|{\eta}_a(\xaj)- \Tsda |\le 2\ela ) -\Pa(|\eta_a(X)-\Tsda|\le{2\ela}) \le  \frac\ep2\rbb,\\
&E_{a,6}=\lbb\frac{1}{n_a}\sum\limits_{j=1}^{n_a}I({\eta}_a(\xaj)= \Tsda  ) -\Pa(\eta_a(X)=\Tsda )
\ge -\ep\rbb.
\end{align*}
We recall the functions 
$\psi_{n,j,\iota}$ from \eqref{psi}, for $j\in\{1,2\}$,
defined by constants $c_{i,\iota}$ with $\iota\in \{\eta,T,\pi\}$ and $i\in [4]$.  
For the first two terms in \eqref{pi21}, 
we have from
\eqref{s} with $x = x_{a,j}$
that, when $E_{a,i}$, $i\in [3]$ hold,  
   \begin{align*}
        \pihnap&=\frac{1}{n_a}\sum\limits_{j=1}^{n_a}I\lsb \widehat{\eta}_a(\xaj)> \Thda+\ela\rsb\le\frac{1}{n_a}\sum\limits_{j=1}^{n_a}I\lsb {\eta}_a(\xaj)> \Tsda\rsb\\
        &\le
        \Po\lsb{\eta}_a(X)> \Tsda \rsb +\ep
        =\pisap +\ep.
   \end{align*}
Moreover,  when $E_{a,1}$, $E_{a,2}$ and $E_{a,4}$ hold with $\ep\ge 4U_\gamma(4p_a\ela)^\gamma$, by the margin condition from Definition \ref{m}
and the definition of $\pisap$ from \eqref{pihata},
   \begin{align*}
       \pihnap&=\frac1{n_a}\sum\limits_{j=1}^{n_a}I\lsb \widehat{\eta}_a(\xaj)> \Thda+\ela\rsb\ge \frac1{n_a}\sum\limits_{j=1}^{n_a}I\lsb {\eta}_a(\xaj)> \Tsda+2\ela\rsb\\
       &\ge \Po\lsb{\eta}_a(X)> \Tsda +{2\ela} \rsb-\frac\ep2
       \ge\pisap -\frac\ep2- \Pa\lsb\Tsda<\eta_a(X)\le \Tsda +{2\ela}\rsb\\
            &\ge  \pisap -\frac\ep2 - U_\gamma(4p_a\ela)^\gamma \ge \pisap-\ep.
   \end{align*}
Thus, by Corollary \ref{constT}, Lemma \ref{lem:uniformDDPbound}, Lemma \ref{lem:merge} and the fact that $(\widehat\eta_1,\widehat\eta_0)$ are
$(\phi_{n,1},\phi_{n,0})$-pointwise convergent, 
we have, for  $L_{\pi_1}(\phi_{n,1}\vee \phi_{n,0})<r_n<U_{\pi_1}$, $2\lsb g_{\delta,-}(4r_n)\vee g_{\delta,+}(4r_n)\rsb<\Delta_n<U_{\Delta,\pi_1}$, $ 4U_\gamma  (4p_a\ela)^\gamma \le \ep\le \sqrt{p_a/2}$ and $ L_{\pi_1}(\phi_{n,1}\vee \phi_{n,0})\le \ela/2 \le r_n$, with $c_{i,\pi}>0, i\in[4]$, that
  \begin{align*}
&\PN(\pihnap >\pisap + \ep)\le \sum_{j=1}^3 \PN(E_{a,j}^c)\\
&\le\psi_{n,1,\eta}\lsb \frac\ela2\rsb+  \psi_{n,1,T}\lsb \frac\ela2\rsb +\sum_{j\in\{-,+\}}I(\delta=D_j(t^\star_\delta) \cdot
\psi_{n,2,T}\lsb g_{\delta,j}\xi\lsb \frac{\ela}2 ,r_n\rsb\rsb +4\exp\lsb-
{np_a\ep^2}\rsb\\
&\le\psi_{n,1,\pi}\lsb \ela\rsb+  \sum_{j\in\{-,+\}}I(\delta=D_j(t^\star_\delta) 
\psi_{n,2,\pi}\lsb g_{\delta,j}\lsb \xi\lsb {\ela} ,r_n\rsb\rsb\rsb +4\exp\lsb-
{np_a\ep^2}\rsb
=\mathcal{E}'_{\pi,a}.
  \end{align*}
Similarly, 
 \begin{align*}
\PN\lsb\pihnap <\pisap - \ep\rsb&\le
\sum_{j=1,2,4} 
\PN(E_{a,j}^c)
\le  \mathcal{E}'_{\pi,a}.
  \end{align*}

For the last two terms in \eqref{pi21}, note that
\begin{align*}
& I\left(|\eta_a(x)-\Tsda|\le \ela-|\widehat\eta_a(x)-\eta_a(x)|-|\Thda-\Tsda|\right)
\le I\left(|\widehat\eta_a(x)-\Thda|\le\ela\right) \\
&\le I\left(|\eta_a(x)-\Tsda|\le \ela+|\widehat\eta_a(x)-\eta_a(x)|+|\Thda-\Tsda|\right).
\end{align*}
When $E_{a,1}$, $E_{a,2}$ and $E_{a,5}$ hold with $\ep> 4U_\gamma  (4p_a\ela)^\gamma $,
 by the margin condition from Definition \ref{m},
   \begin{align*}
    \pihnae&=\frac{1}{n_a}\sum\limits_{j=1}^{n_a}I\lsb |\widehat{\eta}_a(\xaj)- \Thda|\le\ela\rsb
        \le\frac{1}{n_a}\sum\limits_{j=1}^{n_a}I\lsb |{\eta}_a(\xaj)- \Tsda|\le 2\ela\rsb\\
        &\le \Po\lsb|{\eta}_a(X)- \Tsda|\le 2\ela \rsb +\frac\ep2
        =\pisae +\frac\ep2 +\Po\lsb 0<|{\eta}_a(X)- \Tsda|\le 2\ela \rsb\\
        &\le \pisae +\frac\ep2 +2U_\gamma(4p_a\ela)^\gamma\le \pisae+\ep.
   \end{align*}
and when $E_{a,1}$, $E_{a,2}$ and $E_{a,6}$ hold,
   \begin{align*}
    \pihnae&=\frac{1}{n_a}\sum\limits_{j=1}^{n_a}I\lsb |\widehat{\eta}_a(\xaj)- \Thda|>\ela\rsb
        \ge\frac{1}{n_a}\sum\limits_{j=1}^{n_a}I\lsb {\eta}_a(\xaj)=\Tsda\rsb\\
        &\ge  \Po\lsb {\eta}_a(X)= \Tsda\rsb -\ep
        =\pisae -\ep .
   \end{align*}
Thus, we have, for   $L_{\pi_1}(\phi_{n,1}\vee \phi_{n,0})<r_n<U_{\pi_1}$, $2\lsb g_{\delta,-}(4r_n)\vee g_{\delta,+}(4r_n)\rsb<\Delta_n<U_{\Delta,\pi_1}$, $ 4U_\gamma  (4p_a\ela)^\gamma \le \ep\le \sqrt{p_a/2}$ and $ (L_\eta \vee L_T)(\phi_{n,1}\vee \phi_{n,0})\le \ela/2 \le r_n$, with $c_{i,\pi}>0, i\in[4]$ that
 \begin{align*}
\PN\lsb\pihnae >\pisae + \ep\rsb\le\PN(E_{a,1}^c)+ \PN(E_{a,2}^c)+ \PN(E_{a,5}^c)
\le  \mathcal{E}'_{\pi,a}.
  \end{align*}
 Similarly,
 \begin{align*}
\PN\lsb\pihnae <\pisae - \ep\rsb\le\PN(E_{a,1}^c)+ \PN(E_{a,2}^c)+ \PN(E_{a,6}^c)
\le \mathcal{E}'_{\pi,a}.
  \end{align*}

\subsection{Proof of Lemma \ref{propRfunc}}
\label{ppropRfunc}

Recalling the function $\rho$ from \eqref{rho},
we consider the following three cases in order:
(1) $a\ge b>0$; (2) $a\le 0<b$. and (3) $0<a< b$.

\begin{itemize}[]

\item Case 1: If $a\ge b>0$, we have $\rho(a/b)=1$. Moreover, with $0<\ep<b/2$,
$({a+2\ep})/({b-\ep})>1$
and
$({a-2\ep})/({b+\ep})-1 = ({a-b-3\ep})/({b+\ep})> -3\ep/b> -6\ep/b$. 
It follows that, with $0<\ep< b/2$:
$$\rho\lsb\frac{a+2\ep}{b-\ep}\rsb -\rho\lsb\frac{a}{b}\rsb =0< \frac{6\ep}{b}\text{ and } \ \ \rho\lsb\frac{a-2\ep}{b+\ep}\rsb -\rho\lsb\frac{a}{b}\rsb >-\frac{6\ep}{b}.$$

\item Case 2: If $a\le 0< b$, we have $\rho(a/b)=0$. 
Moreover, with $0<\ep< b/2$, 
$({a+2\ep})/({b-\ep})m 4\ep/b<  \frac{6\ep}{b}$
and
$({a-2\ep})/({b+\ep})<0$. It follows that, with $0<\ep< b/2$:
$$\rho\lsb\frac{a+2\ep}{b-\ep}\rsb -\rho\lsb\frac{a}{b}\rsb \le \frac{4\ep}{b}\text{ and } \ \ \rho\lsb\frac{a-2\ep}{b+\ep}\rsb -\rho\lsb\frac{a}{b}\rsb =0\ge  -\frac{6\ep}{b}.$$

\item Case 3: If $0<a< b$, we have $\rho(a/b)=a/b$. Moreover, with $0<\ep< b/2$, 
$$\frac{a+2\ep}{b-\ep}-\frac{a}{b} =\frac{(a+2b)\ep}{b(b-\ep)}\le \frac{3b\ep}{b(b-\ep)}\le \frac{6\ep}{b}$$
and
$$\frac{a-2\ep}{b+\ep}-\frac{a}{b} =\frac{(-a-2b)\ep}{b(b+\ep)}\ge \frac{-3b\ep}{b(b+\ep)}\ge -\frac{3\ep}{b}\ge -\frac{6\ep}{b}.$$
Hence \eqref{ri} follows.
\end{itemize}

\subsection{Proof of Lemma \ref{lem:pirbound}}
\label{plem:pirbound}

We consider the following two cases for $\pisae$ from \eqref{eq:pistar}: (1) $\pisae = 0$ and (2) $\pisae > 0$.

Case (1): When $\pisae = 0$, since by \eqref{htau}, $0\le \taudha\le 1$
and since  $\pihae\ge 0$ for $\pihae$ from \eqref{pihata}, for any $\ep >0$, we have
\begin{equation*}
\PN\lsb\pihae\taudha < \pisae \taudsa-\ep\rsb
\le\PN\lsb \pihae<0 \rsb=0.
\end{equation*}
Moreover, if we take $L_{\pi}=L_{\pi_1}$, $U_{\pi}=U_{\pi_1}$ $L_{\ep,\pi}=2^{2+2\gamma}U_\gamma(p_1\vee p_0)^\gamma$ and $U_{\Delta,\pi}=U_{\Delta,\pi_1}$,
by \eqref{pi11} from Lemma \ref{lem:piset1},  for  $L_{\pi_1}(\phi_{n,1}\vee \phi_{n,0})<r_n<U_{\pi_1}$, $2\lsb g_{\delta,-}(4r_n)\vee g_{\delta,+}(4r_n)\rsb<\Delta_n<{U_\Delta,\pi_1}$, $ L_{\ep,\pi} (\lo\vee\lz)^\gamma \le \ep\le \sqrt{p_a/2}$ and $ (L_\eta \vee L_T)(\phi_{n,1}\vee \phi_{n,0})\le \ela/2 \le r_n$, 
we have
\begin{align*}
&\PN\lsb\pihae\taudha > \pisae \taudsa+\ep \rsb
= \PN\lsb\pihae\taudha>\ep\rsb 
\le \PN\lsb\pihae>\pisae+\ep\rsb\\
&\le
\psi_{n,1,\pi}\lsb\ela\rsb+\sum_{j\in\{-,+\}}I(\delta=D_j(t^\star_\delta))
\psi_{n,2,\pi}\lsb g_{\delta,j}\omega(\ela,r_n)\rsb.
\end{align*}
Thus, \eqref{eq:pibounds} holds with $c_{5,\pi}=c_{6,\pi}=0$.

 Case (2): When $\pisae > 0$, we distinguish the following two cases:

(A). When $\widehat\delta=\widetilde\delta$, $\pihnap\ge \pisap - {\pisae}\ep/{12} $, $\widehat\pi_{n,1-a,+}\le \pi^\star_{1-a,+} + {\pisae}\ep/{12} $, $\pihnae\ge \pisae - {\pisae}\ep/{12} $ and $\pihae \le \pisae+\ep/2$, 
recalling $\widehat{\delta}_a$ from Algorithm \ref{alg:FBC1}, and using Lemma \ref{propRfunc},
we have
\begin{align*}
&\pihae\taudha =\pihae \rho\lsb\frac{\widehat\pi_{n,1-a,+}- \pihnap  +\widehat{\delta} }{\pihnae}\rsb \le \lsb\pisae+\frac\ep2\rsb \rho\lsb\frac{ \pi^\star_{1-a,+}- \pisap + \widetilde{\delta}+ \ep\pisae/{6}}{\pisae - \ep\pisae/{12}}\rsb\\
&\le
 \lsb\pisae+\frac\ep2\rsb  \lsb \rho\lsb\frac{ \pi^\star_{1-a,+}- \pisap+{\delta} }{\pisae }\rsb +\frac\ep2\rsb \le \pisae  \taudsa +\ep;
\end{align*}
Thus, if we take $L_{\pi}=L_{\pi_1}\vee L_{\delta}$, $U_{\pi}=U_{\pi_1}\wedge U_{\delta}$, $L_{\ep,\pi}=48U_\gamma\lsb ((4p_1)^\gamma/\pisoe)\vee ((4p_0)^\gamma/\pisze)\rsb$ and $U_{\Delta,\pi}=U_{\Delta,\pi_1}\wedge U_{\Delta,\delta}$,
by Lemmas \ref{lem:deltaneq}, \ref{lem:piset1} and  \ref{lem:piset2}, we have, with $L_{\pi}(\phi_{n,1}\vee \phi_{n,0})< r_n<U_{\pi}$, $\Delta_n<U_{\Delta,\pi}$, $2(L_{\pi_1}\vee L_{\delta})(\phi_{n,1}\vee \phi_{n,0})<
\lo,\lz< 2r_n$, and $  L_{\ep,\pi}(\lo\vee\lz)^\gamma<\ep \le \sqrt{p_a/2}$, that
\begin{align*}
&\PN\lsb\pihae \rho\lsb\frac{\widehat\pi_{n,1-a,+}  - \pihnap + \widehat{\delta}_a }{\pihnae}\rsb > \pisae   \rho\lsb\frac{ \pi^\star_{1-a,+}- \pisap +{\delta}}{\pisae }\rsb +\ep\rsb\\
&\le \PN\lsb\pihnap< \pisap - \frac{\pisae\ep}{12}\rsb + \PN\lsb\widehat\pi_{n,1-a,+}>\pi^\star_{1-a,+} + \frac{\pisae\ep}{12} \rsb\\
& + \PN\lsb\pihnae< \pisae - \frac{\pisae\ep}{12}\rsb +\PN\lsb\pihae > \pisae+\frac\ep2\rsb\\
&\le4\psi_{n,1,\pi}\lsb \ela\rsb+  4\sum_{j\in\{-,+\}}I(\delta=D_j(t^\star_\delta))
\psi_{n,2,\pi}\lsb g_{\delta,j}\omega(\ela,r_n)\rsb+12\exp\lsb-\frac
{np_a(\pisae)^2\ep^2}{576}\rsb.
\end{align*}
Choosing 
$c_{5,\pi}=12$ and $c_{6,\pi}={p_a{\pi}^{\star 2}_{a,=}}/{576}$ proves the first claimed inequality.

(B). When $\pihnap\le \pisap + {\pisae}\ep/{12} $, $\widehat\pi_{n,1-a,+}\ge \pi^\star_{1-a,+} - {\pisae}\ep/{12} $, $\pihnae\le \pisae + {\pisae}\ep/{12} $ and $\pihae \ge \pisae-\ep/2$, 
we have
\begin{align*}
&\pihae \rho\lsb\frac{ \widehat\pi_{n,1-a,+} - \pihnap + \widehat{\delta}_a}{\pihnae}\rsb \ge \lsb\pisae-\frac\ep2\rsb \rho\lsb\frac{ \pi^\star_{1-a,+}-\pisap -{\delta}- \ep\pisae/{6}}{\pisae + \ep\pisae/{12}}\rsb\\
&\ge
 \lsb\pisae-\frac\ep2\rsb  \lsb \rho\lsb\frac{\pi^\star_{1-a,+}- \pisap -{\delta} }{\pisae }\rsb -\frac\ep2\rsb \ge \pisae   \rho\lsb\frac{\pi^\star_{1-a,+}- \pisap {-\delta} }{\pisae }\rsb -\ep.
\end{align*}
Thus, similarly to case (A),
\begin{align*}
&\PN\lsb\pihae \rho\lsb\frac{\widehat\pi_{n,1-a,+}- \pihnap +\widehat{\delta}_a }{\pihnae}\rsb < \pisae   \rho\lsb\frac{\pi^\star_{1-a,+}- \pisap {-\delta} }{\pisae }\rsb -\ep\rsb\\
&\le \PN\lsb\pihnap> \pisap + \frac{\pisae\ep}{12}\rsb + \PN\lsb\widehat\pi_{n,1-a,+}<\pi^\star_{1-a,+} - \frac{\pisae\ep}{12} \rsb\\& + \PN\lsb\pihnae> \pisae + \frac{\pisae\ep}{12}\rsb +\PN\lsb\pihae < \pisae-\frac\ep2\rsb,
\end{align*}
and the second claimed inequality follows as in case (A).

\section{Bayes-optimal Classifier for Data Distribution from Section \ref{sec:simu-syn}}\label{sec:detail_about_syn_dis}
In this section, we derive the $\delta$-fair Bayes optimal classifier and its misclassification rate for the data distribution proposed in Section \ref{sec:simu-syn}. 
Let $p_a=P(A=1)$ be the probability of $A$ being $1$, let $\mu_a$ the conditional density function of $(X_1,X_2)$ given $A=a$, and let $\eta_a(X_1,X_2)$ be the probability of $Y=1$ given $A=a$ and $X_1,X_2$.
According to the construction, for $a\in\{0,1\}$ and $(x_1,x_2)\in[-1,1]^2$,
\begin{equation}\label{eq:syncons}
  (1) \ p_a  =\frac12, \
  (2) \ \mu_a(x_1,x_2)  \equiv \frac{1}4 \ \text{ and } \  (3) \ \eta_a(x_1,x_2)= \frac{1+ (2a-1)s_1}{2}+\frac{ s_2\cdot \text{sign}(x_1)}{2}\lsb|x_1|(1-|x_2|)\rsb^\beta.
\end{equation}
In the following, we define the half cubes  $\mathbb{B}^2_+ $ and $\mathbb{B}^2_- $ as 
$$\mathbb{B}^2_+ = \lbb x=(x_1,x_2)\in [-1,1]^2, x_1\ge 0 \rbb \text { and } \mathbb{B}^2_- = \lbb x=(x_1,x_2)\in [-1,1]^2, x_1< 0 \rbb,
$$
respectively.
As $\lsb|x_1|(1-|x_2|)\rsb^\beta\in [0,1]$ on $[-1,1]^2$, we have
 \begin{equation}\label{eq:rangeofeta}
\left\{\begin{array}{cc}
\eta_a(x_1,x_2)\in \lmb\frac{1+(2a-1)s_1}{2}, \frac{1+(2a-1)s_1+s_2}{2}\rmb  \text{ on }  \mathbb{B}^2_{+} ;\\
\\
 \eta_a(x_1,x_2)\in \lmb\frac{1+(2a-1)s_1-s_2}{2}, \frac{1+(2a-1)s_1}{2}\rmb \text{ on }  \mathbb{B}^2_{-}. 
\end{array}\right.\end{equation} 

In order to derive $D_{-}(t)$, we  first calculate $\Poto(\eta_1(X_1,X_2)>1/2+t)$ and $\Potz(\eta_0(X_1,X_2)>1/2-t)$. For $\Poto(\eta_1(X_1,X_2)>1/2+t)$, we 
consider two cases: (1) $ (s_1-s_2)/2\le t \le  s_1/2$ and (2) $s_1/2<t\le (s_1+s_2)/2$.
\begin{itemize}
  
\item Case (1): $ (s_1-s_2)/2\le t \le  s_1/2$.

    In this case, we have $ (1+s_1-s_2)/2\le 1/2+t \le  (1+s_1)/2$. By \eqref{eq:rangeofeta}, 
  \begin{equation}\label{eq:seteq1}
      \lbb\eta_1(x_1,x_2) >\frac12+t\rbb \cap \mathbb{B}^2_+  = \mathbb{B}^2_+,
  \end{equation}     and
  
\begin{align}\label{eq:seteq2}
\nonumber    &\lbb\eta_1(x_1,x_2) >\frac12+t\rbb \cap \mathbb{B}^2_-=
     \lbb
{ \text{sign}(x_1)}\lsb \lab x_1\rab \lsb 1-\lab x_2\rab\rsb\rsb^\beta >\frac{2t-s_1}{s_2}
\rbb \cap \mathbb{B}^2_-      \\
\nonumber &= 
     \lbb
-\lsb -x_1 \lsb 1-\lab x_2\rab\rsb\rsb^\beta >\frac{2t-s_1}{s_2}
\rbb \cap \mathbb{B}^2_-      =
     \lbb - x_1 \lsb 1-\lab x_2\rab\rsb <\lsb\frac{s_1-2t}{s_2}\rsb^{\frac1\beta}
\rbb \cap   \mathbb{B}^2_-  \\
\nonumber&=     \lbb \lsb 1-\lab x_2\rab\rsb <-\frac1{x_1}\lsb\frac{s_1-2t}{s_2}\rsb^{\frac1\beta}
\rbb \cap   \mathbb{B}^2_-=\lbb  -1\le x_1< 0,  1+\frac1{x_1}\lsb\frac{s_1-2t}{s_2}\rsb^{\frac1\beta}<|x_2|\le 1 
\rbb\\
\nonumber&=\lbb -1\le x_1\le -\lsb\frac{s_1-2t}{s_2}\rsb^{\frac1\beta},  1+\frac1{x_1}\lsb\frac{s_1-2t}{s_2}\rsb^{\frac1\beta}<|x_2|\le 1
\rbb\\
&\cup\lbb -\lsb\frac{s_1-2t}{s_2}\rsb^{\frac1\beta}< x_1 <0, |x_2|\le 1
\rbb.
\end{align}
Here, the last equality holds since the event
$\lbb|x_2|> 1+({(s_1-2t)}/{s_2})^{(1/\beta)}
/x_1\rbb$ is implied by  $-((s_1-2t)/{s_2})^{(1/\beta)}< x_1 <0$. 
Then, by \eqref{eq:syncons},
\begin{align*}
 &  \Poto\lsb\eta_1(X_1,X_2)>\frac12+t\rsb = \int_{\{\eta_1(x_1,x_2)>\frac12+t\}}
\mu_{1}(x_1,x_2) dx_1dx_2\\
&= \frac14\lsb \int_{\{\eta_1(x_1,x_2)>\frac12+t\}\cap \mathbb{B}^2_+}
 dx_1dx_2 +\int_{\{\eta_1(x_1,x_2)>\frac12+t\}\cap \mathbb{B}^2_-}
 dx_1dx_2   \rsb\\
 &=\frac12+\frac14\int_{-\lsb\frac{s_1-2t}{s_2}\rsb^{\frac1\beta}}^{0}\lsb \int^{ 1}_{ -1} dx_2\rsb dx_1+ \frac14\int^{-\lsb\frac{s_1-2t}{s_2}\rsb^{\frac1\beta}}_{-1}\lsb \int_{1+\frac1{x_1}\lsb\frac{2t-s_1}{s_2}\rsb^{\frac1\beta}< |x_2|\le 1}dx_2\rsb dx_1\\
 &=\frac12+\frac12\int_{-\lsb\frac{s_1-2t}{s_2}\rsb^{\frac1\beta}}^{0} dx_1+  \frac12\int^{-\lsb\frac{s_1-2t}{s_2}\rsb^{\frac1\beta}}_{-1}\lsb   -\frac1{x_1}\lsb\frac{s_1-2t}{s_2}\rsb^{\frac1\beta} \rsb dx_1 \\
 &=\frac12+ \frac{1}{2}  \lsb\frac{s_1-2t}{s_2}\rsb^{\frac1\beta}
- \frac12\lsb\frac{s_1-2t}{s_2}\rsb^{\frac1\beta} \lsb\ln (-x_1)\Bigg|^{-\lsb\frac{s_1-2t}{s_2}\rsb^{\frac1\beta}}_{-1}\rsb\\
 &=\frac12+ \frac{1}{2} \lsb\frac{s_1-2t}{s_2}\rsb^{\frac1\beta}
\lsb 1 -  \frac1{\beta}\ln \lsb\frac{s_1-2t}{s_2}\rsb\rsb.
\end{align*}

\item Case (2): $ s_1/2\le t \le  (s_1+s_2)/2$.
 
    In this case, we have $(1+s_1)/2<1/2+t<(1+s_1+s_2)/2$. Again,    
    by \eqref{eq:rangeofeta}, 
\begin{equation}\label{eq:seteq3}
    \lbb\eta_1(x_1,x_2) >\frac12+t\rbb \cap \mathbb{B}^2_-   = \emptyset,
\end{equation} 
    and
\begin{align}\label{eq:seteq4}
 \nonumber   &\lbb\eta_1(x_1,x_2) >\frac12+t\rbb \cap \mathbb{B}^2_+ = 
     \lbb
{  \text{sign}(x_1)}\lsb \lab x_1\rab \lsb 1-\lab x_2\rab\rsb\rsb^\beta >\frac{2t-s_1}{s_2}
\rbb \cap \mathbb{B}^2_+      \\
\nonumber  &= 
     \lbb  x_1 \lsb 1-\lab x_2\rab\rsb >\lsb\frac{2t-s_1}{s_2}\rsb^{\frac1\beta}
\rbb \cap   \mathbb{B}^2_+ = \lbb \lsb 1-\lab x_2\rab\rsb >\frac1{x_1}\lsb\frac{2t-s_1}{s_2}\rsb^{\frac1\beta}
\rbb \cap   \mathbb{B}^2_+\\
\nonumber  &=\lbb  0\le x_1\le 1, |x_2|< 1-\frac1{x_1}\lsb\frac{2t-s_1}{s_2}\rsb^{\frac1\beta}
\rbb\\
&=\lbb \lsb\frac{2t-s_1}{s_2}\rsb^{\frac1\beta}\le x_1 \le 1, |x_2|< 1-\frac1{x_1}\lsb\frac{2t-s_1}{s_2}\rsb^{\frac1\beta}
\rbb.
\end{align}
Here, the last equality holds since the set
$\lbb|x_2|< 1-({(2t-s_1)}/{s_2})^{(1/\beta)}
/x_1\rbb$ is empty set when  $0\le x_1 <(2t-s_1)/s_2)^{(1/\beta)}$.
Thus, 
\begin{align*}
 &  \Poto\lsb\eta_1(X_1,X_2)>\frac12+t\rsb = \int_{\{\eta_1(x_1,x_2)>t\}}
\mu_{1}(x_1,x_2) dx_1dx_2\\
&= \frac14\lsb \int_{\{\eta_1(x_1,x_2)>t\}\cap \mathbb{B}^2_+}
 dx_1dx_2 +\int_{\{\eta_1(x_1,x_2)>t\}\cap \mathbb{B}^2_-}
 dx_1dx_2   \rsb\\
 &=\frac{1}{4}\int_{\{\eta_1(x_1,x_2)>t\}\cap \mathbb{B}^2_+}
 dx_1dx_2 =\frac14 \int_{\lsb\frac{2t-s_1}{s_2}\rsb^{\frac1\beta}}^1\lsb \int_{  \frac1{x_1}\lsb\frac{2t-s_1}{s_2}\rsb^{\frac1\beta}-1}^{ 1-\frac1{x_1}\lsb\frac{2t-s_1}{s_2}\rsb^{\frac1\beta}} dx_2\rsb dx_1\\
 &=\frac12\int_{\lsb\frac{2t-s_1}{s_2}\rsb^{\frac1\beta}}^1\lsb   1-\frac1{x_1}\lsb\frac{2t-s_1}{s_2}\rsb^{\frac1\beta} \rsb dx_1 \\
 &=\frac{1}{2}\lsb 1- \lsb\frac{2t-s_1}{s_2}\rsb^{\frac1\beta}\rsb
 -  \frac12\lsb\frac{2t-s_1}{s_2}\rsb^{\frac1\beta} \lsb\ln x\Bigg|_{\lsb\frac{2t-s_1}{s_2}\rsb^{\frac1\beta}}^{1}\rsb\\
 &=\frac{1}{2}- \frac12\lsb\frac{2t-s_1}{s_2}\rsb^{\frac1\beta}
\lsb 1- \frac1{\beta} \ln \lsb \frac{2t-s_1}{s_2}\rsb\rsb.\\
\end{align*}
\end{itemize}
In summary, we have,
\begin{equation}\label{eq:expofsimu1}
\Poto\lsb\eta_1(X)>\frac12+t\rsb=\left\{\begin{array}{ll}
1,  &  t \le\frac{s_1-s_2}{2};\\
  \frac12+ \frac{1}{2} \lsb\frac{s_1-2t}{s_2}\rsb^{\frac1\beta}
 \lsb 1-  \frac1{\beta} \ln \lsb\frac{s_1-2t}{s_2}\rsb\rsb,  & \frac{s_1-s_2}{2}< t \le \frac{s_1}{2};\\
 \frac{1}{2}- \frac12\lsb\frac{2t-s_1}{s_2}\rsb^{\frac1\beta}
\lsb 1- \frac1{\beta} \ln \lsb \frac{2t-s_1}{s_2}\rsb\rsb,   &   \frac{s_1}{2}\le t \le \frac{s_1+s_2}{2};\\
0,   &   t > \frac{s_1+s_2}{2}.
\end{array}\right.
\end{equation}
To find $\Potz\lsb\eta_0(X_1,X_2)\ge\frac12-t\rsb$,
we note that $\eta_0(x_1,x_2) = \eta_1(x_1,x_2)-s_1$ and $\mu_0(x_1,x_2)\equiv \mu_1(x_1,x_2)$. Thus,
\begin{align}\label{eq:expofsimu2}
   \nonumber  & \Potz\lsb\eta_0(X_1,X_2)\ge\frac12-t\rsb =\int_{\{\eta_0(x_1,x_2)>\frac12-t\}}\mu_0(x_1,x_2)dx_1dx_2\\
   \nonumber &=\int_{\{\eta_1(x_1,x_2)>\frac12+s_1-t\}}\mu_1(x_1,x_2)dx_1dx_2 =\Poto\lsb\eta_1(X)>\frac12+s_1-t\rsb\\
  \nonumber&=\left\{\begin{array}{ll}
1 ,  &  s_1-t\le \frac{s_1-s_2}{2};\\ 
  \frac12+ \frac{1}{2} \lsb\frac{2t-s_1}{s_2}\rsb^{\frac1\beta}
 \lsb 1-  \frac1{\beta}  \ln \lsb\frac{2t-s_1}{s_2}\rsb\rsb,  & \frac{s_1-s_2}{2}< s_1-t \le \frac{s_1}{2};\\
  \frac{1}{2}- \frac12\lsb\frac{s_1-2t}{s_2}\rsb^{\frac1\beta}\lsb 1-
 \frac1{\beta}\ln \lsb \frac{s_1-2t}{s_2}\rsb\rsb,   &   \frac{s_1}{2}< s_1-t \le \frac{s_1+s_2}{2};\\
0,  &   s_1-t > \frac{s_1+s_2}{2},
\end{array}\right.\\
&=\left\{\begin{array}{ll}
  0,  & t\le \frac{s_1-s_2}{2};\\
  \frac{1}{2}- \frac12\lsb\frac{s_1-2t}{s_2}\rsb^{\frac1\beta}\lsb 1-
 \frac1{\beta}\ln \lsb \frac{s_1-2t}{s_2}\rsb\rsb,   &   \frac{s_1-s_2}{2}< t \le \frac{s_1}{2};\\ 
  \frac12+ \frac{1}{2} \lsb\frac{2t-s_1}{s_2}\rsb^{\frac1\beta}
 \lsb 1-  \frac1{\beta}  \ln \lsb\frac{2t-s_1}{s_2}\rsb\rsb,  & \frac{s_1}{2}< t \le \frac{s_1+s_2}{2};\\
 1,  & t>\frac{s_1+s_2}{2}.
\end{array}\right.
\end{align}
Recalling \eqref{eq:Disfun1}, by
\eqref{eq:expofsimu1} and \eqref{eq:expofsimu2}, 
\begin{align*}
&  D_{-}\lsb t\rsb=
\P_{X_1,X_2|A=1}\lsb\eta_{1}(X_1,X_2)>\frac12+t\rsb - \P_{X_1,X_2|A=0}\lsb\eta_{0}(X_1,X_2)>\frac12-t\rsb\\
&=\left\{\begin{array}{ll}
  1,  & t\le \frac{s_1-s_2}{2};\\
\lsb\frac{s_1-2t}{s_2}\rsb^{\frac1\beta}\lsb 1-
 \frac1{\beta}\ln \lsb \frac{s_1-2t}{s_2}\rsb\rsb,   &   \frac{s_1-s_2}{2}< t \le \frac{s_1}{2};\\ 
  - \lsb\frac{2t-s_1}{s_2}\rsb^{\frac1\beta}
 \lsb 1-  \frac1{\beta}  \ln \lsb\frac{2t-s_1}{s_2}\rsb\rsb,  & \frac{s_1}{2}< t \le \frac{s_1+s_2}{2};\\
 -1,  & t>\frac{s_1+s_2}{2}.
\end{array}\right.
\end{align*}
Since $s_1<s_2$, we have
$D_{-} (0) = 
   \lsb\frac{s_1}{s_2}\rsb^{\frac1\beta}\lsb 1-
 \frac1{\beta} \ln\lsb \frac{s_1}{s_2}\rsb\rsb>0.$
Thus, based on \eqref{eq:op-de-rule-dp_continuous},
the $\delta$-fair Bayes-optimal classifier is given, for all $x,a$, by
\begin{equation*}
f_\delta^\star(x,a)=I\lsb\eta_{a}(x)>\frac12+(2a-1)t^\star_\delta\rsb +\tau_{\delta,a}^\star I\lsb\eta_{a}(x)=\frac12+(2a-1)t^\star_\delta\rsb,
\end{equation*}
with $t^\star_\delta = I(D_{-} (0)>\delta)\cdot\sup_{t}\{D_{-}(t)>\delta\}$ and $(\tau_{\delta,1}^\star,\tau_{\delta,0}^\star)\in[0,1]^2$ being arbitrary. Specifically,  $t^\star_\delta$ satisfies $0\le t^\star_\delta<s_1/2$ and
$$ \lsb\frac{s_1-2t^\star_\delta}{s_2}\rsb^{\frac1\beta}\lsb 1-
 \frac1{\beta}\ln \lsb \frac{s_1-2t}{s_2}\rsb\rsb=\delta\wedge \lmb \lsb\frac{s_1}{s_2}\rsb^{\frac1\beta}\lsb 1-
 \frac1{\beta} \ln\lsb \frac{s_1}{s_2}\rsb\rsb\rmb.$$
 For the misclassification rate of $ f^\star_\delta$, we have
\begin{align*}
\P(Y\neq\widehat{Y}_f)&=\sum_{a\in\{0,1\}}p_a\int  \left[\lsb 1-2\eta_a(x_1,x_2)\rsb f^\star_\delta(x_1,x_2,a)  
 +\eta_a(x_1,x_2)\right] d\P_{X_1,X_2|A=a}(x)\\
&=\frac18\sum_{a\in\{0,1\}}\int_{-1}^1\int_{-1}^1  \left[\lsb 1-2\eta_a(x_1,x_2)\rsb f^\star_\delta(x_1,x_2,a)  +\eta_a(x_1,x_2)\right] dx_1dx_2 \\
&=\frac18\int_{-1}^1\int_{-1}^1   \lmb f^\star_\delta(x_1,x_2,1)  +f^\star_\delta(x_1,x_2,1)  \rmb dx_1dx_2 
+\frac18\int_0^1\int_0^1 \lmb\eta_1(x_1,x_2)+\eta_0(x_1,x_2)\rmb dx_1dx_2\\
&-\frac14\int_{\{\eta_1(x_1,x_2)>\frac12+t^\star_\delta\}}  \eta_1(x_1,x_2) dx_1dx_2  -\frac14\int_{\{\eta_0(x_1,x_2)>\frac12-t^\star_\delta\}} \eta_0(x_1,x_2) dx_1dx_2\\
=:& \mathrm{(I)} + \mathrm{(II)} + \mathrm{(III)} +\mathrm{(IV)} .
\end{align*}
 We now calculate the above four terms in turn. 

For (I), denote $q^\star_\delta = ((s_1-2t^\star_\delta)/s_2)^{(1/\beta)}$.
Since $0\le t_\delta^\star\le s_1/2$, by \eqref{eq:expofsimu1} and \eqref{eq:expofsimu2},  
\begin{align*}\mathrm{(I)} &=\frac12\lmb\Poto\lsb \eta_1(X_1,X_2)>\frac12+ t^\star_\delta\rsb+\Potz\lsb \eta_0(X_1,X_2)>\frac12- t^\star_\delta\rsb\rmb\\
&=
\frac14+ \frac{1}{4}q^\star_\delta 
 \lsb 1-    \ln \lsb q^\star_\delta \rsb\rsb
 +\frac14- \frac{1}{4}q^\star_\delta 
 \lsb 1-    \ln \lsb q^\star_\delta \rsb\rsb
 =\frac12. \end{align*}
Further,
\begin{align*}\mathrm{(II)} &=\frac18 \int_{-1}^1\int_{-1}^1\lmb 1+ s_2\cdot \textup{sign}(x_1) \lsb|x_1|(1-|x|_2)\rsb^\beta\rmb dx_1dx_2\\
&=\frac12 +
s_2\lsb \int_{-1}^0 (-1)^{\beta+1}x_1^\beta dx_1 + \int_{0}^1 x_1^\beta dx_1\rsb \cdot\int_{-1}^1 (1-|x_2|)^\beta dx_2 \\
&=\frac12 +s_2\cdot\lsb -\int_{0}^1 x_1^\beta dx_1+\int_{0}^1 x_1^\beta dx_1\rsb \cdot  \int_{-1}^1 (1-|x_2|)^\beta dx_2 =\frac12.
 \end{align*}

For (III),  by \eqref{eq:seteq1} and \eqref{eq:seteq2}, we have that
\begin{align*}
 \mathrm{(III)} &=-\frac14\int_{\{\eta_1(x_1,x_2)>\frac12+t^\star_\delta\}}    \eta_1(x_1,x_2) dx_1dx_2 \\
&=-\frac14\lsb\int_{\{\eta_1(x_1,x_2)>\frac12+t^\star_\delta\}\cap \mathbb{B}_+^2}  \eta_1(x_1,x_2) dx_1dx_2 +\int_{\{\eta_1(x_1,x_2)>\frac12+t^\star_\delta\}\cap \mathbb{B}_-^2}  \eta_1(x_1,x_2) dx_1dx_2 \rsb\\
&= -\frac14\int_{0}^1\int_{-1}^1  \eta_1(x_1,x_2) dx_1dx_2 -\frac14\int_{-\lsb\frac{s_1-2t^\star_\delta}{s_2}\rsb^{\frac1\beta}}^{0}\lsb \int^{ 1}_{ -1} \eta_1(x_1,x_2)  dx_2\rsb dx_1\\
 &- \frac14\int^{-q_\delta^\star}_{-1}\lsb \int_{ 1+\frac{q_\delta^\star}{x_1}<|x_2|\le 1} \eta_1(x_1,x_2) dx_2\rsb dx_1\\
 &=  -\frac14\int_{0}^1\int_{-1}^1 \frac{1+ s_1+ s_2\lsb|x_1|(1-|x_2|)\rsb^\beta }{2} dx_1dx_2\\
 &-\frac14\int_{-q_\delta^\star}^{0}\lsb \int^{ 1}_{ -1} \frac{1+ s_1- s_2\lsb|x_1|(1-|x_2|)\rsb^\beta }{2} dx_2\rsb dx_1\\
 &- \frac14\int^{-q_\delta^\star}_{-1}\lsb \int_{ 1+\frac{q_\delta^\star}{x_1}<|x_2|\le 1} \frac{1+ s_1- s_2\lsb|x_1|(1-|x_2|)\rsb^\beta }{2}dx_2\rsb dx_1.
 \end{align*}
 This further equals
 \begin{align*}
 &- \frac{1+s_1}4 - \frac{s_2}8\int_{0}^1 x_1^\beta dx_1 \cdot \int_{-1}^1 {(1-|x_2|)^\beta } dx_1dx_2-\frac{1+s_1}4 q_\delta^\star\\
  &+\frac{s_2}8
  \int_{- q_\delta^\star}^{0} (-x_1)^\beta dx_1 \cdot \int^{ 1}_{ -1} (1-|x_2|)^\beta  dx_2 dx_1-\frac{1+s_1}4 q_\delta^\star\int^{- q_\delta^\star}_{-1}-\frac1{x_1} dx_1 \\
    & + \frac{s_2}8\int^{- q_\delta^\star}_{-1} (-x_1)^{\beta}\lsb \int_{ 1+\frac{q_\delta^\star}{x_1} <|x_2|\le 1} (1-|x_2|)^\beta dx_2\rsb dx_1\\
  &=- \frac{1+s_1}4 - \frac{s_2}{4(\beta+1)^2}-\frac{1+s_1}4 q_\delta^\star+\frac{s_2}{4(\beta+1)^2} \lsb q^\star_\delta\rsb^{{\beta+1}} +\frac{1+s_1}{4} q_\delta^\star\ln \lsb q^\star_\delta\rsb  \\
    & +\frac{s_2}{4(\beta+1)}\int^{- q_\delta^\star}_{-1}     -\frac{ (q_\delta^\star)^{\beta+1}}{x_1}    dx_1\\ 
      &=- \frac{1+s_1}4 \lsb 1+ q_\delta^\star-q_\delta^\star \ln\lsb q_\delta^\star\rsb \rsb - \frac{s_2}{4(\beta+1) }\lsb\frac{1- \lsb q_\delta^\star\rsb^{\beta+1}}{\beta+1} + \lsb q^\star_\delta\rsb^{\beta+1} \ln \lsb q^\star_\delta \rsb \rsb  .
     \end{align*}

For (IV),  by \eqref{eq:seteq3}, \eqref{eq:seteq4}, $s_1/2\le s_1-t^\star\le s_1$ and as $\eta_0(x_1,x_2)=\eta_1(x_1,x_2)-s_1$,
we have that
\begin{align*}
\mathrm{(IV)} &=-\frac14\int_{\{\eta_0(x_1,x_2)>\frac12-t^\star_\delta\}}    \eta_0(x_1,x_2) dx_1dx_2=-\frac14\int_{\{\eta_1(x_1,x_2)>\frac12+s_1-t^\star_\delta\}}   \lsb \eta_0(x_1,x_2)\rsb dx_1dx_2 \\ 
&=-\frac14\int_{\{\eta_1(x_1,x_2)>\frac12+s_1-t^\star_\delta\}\cap \mathbb{B}_+^2 }  \eta_0(x_1,x_2) dx_1dx_2 -\frac14\int_{ \{\eta_1(x_1,x_2)>\frac12+s_1-t^\star_\delta\}\cap \mathbb{B}_-^2}  \eta_0(x_1,x_2) dx_1dx_2 \\
&= -\frac14 \int_{\lsb\frac{s_1-2t^\star_\delta}{s_2}\rsb^{\frac1\beta}}^1\lsb \int_{  \frac1{x_1} \lsb\frac{s_1-2t^\star_\delta}{s_2}\rsb^{\frac1\beta}-1}^{ 1-\frac1{x_1} q_\delta^\star} \eta_0(x_1,x_2)dx_2\rsb dx_1\\ 
 &=   -\frac14 \int_{ q_\delta^\star}^1\lsb \int_{  \frac1{x_1} q_\delta^\star-1}^{ 1-\frac1{x_1} q_\delta^\star}  \frac{1- s_1+ s_2\lsb|x_1|(1-|x_2|)\rsb^\beta }{2} dx_2\rsb dx_1\\
 &= -\frac{1-s_1}4 \int_{ q_\delta^\star}^1\lsb  1-\frac1{x_1} q_\delta^\star\rsb    dx_1 -\frac{s_2}8\int_{ q_\delta^\star}^1x_1^\beta\lsb \int_{  \frac1{x_1} q_\delta^\star-1}^{ 1-\frac1{x_1} q_\delta^\star}(1-|x_2|)^\beta dx_2\rsb dx_1\\
  &= -\frac{1-s_1}4\lsb1- q_\delta^\star\rsb-\frac{1-s_1}{4} q_\delta^\star \ln \lsb q_\delta^\star \rsb-\frac{s_2}{4(\beta+1 )}\int_{ q_\delta^\star}^1 \lsb x_1^\beta-\frac1{x_1} \lsb q_\delta^\star \rsb^{{\beta+1}} \rsb dx_1\\
    &= -\frac{1-s_1}4\lsb1- q_\delta^\star +q_\delta^\star \ln\lsb q_\delta^\star\rsb\rsb -\frac{s_2}{4(\beta+1)}\lsb \frac{1-\lsb q_\delta^\star \rsb^{\beta+1} }{\beta+1} + \lsb q_\delta^\star \rsb^{\beta+1}   \ln \lsb q_\delta^\star \rsb\rsb.
  \end{align*}
Summing up, we have
 \begin{align*} R(f^\star_\delta) &= \mathrm{(I)}+\mathrm{(II)}+\mathrm{(III)} + \mathrm{(IV)} \\
 &=\frac12 +\frac12 - \frac{1+s_1}4 \lsb 1+ q_\delta^\star-q_\delta^\star \ln\lsb q_\delta^\star\rsb \rsb   -\frac{1-s_1}4\lsb1- q_\delta^\star +q_\delta^\star \ln\lsb q_\delta^\star\rsb\rsb\\
 &-\frac{s_2}{2(\beta+1)}\lsb \frac{1-\lsb q_\delta^\star \rsb^{\beta+1} }{\beta+1} +\lsb q_\delta^\star \rsb^{\beta+1}   \ln \lsb q_\delta^\star \rsb\rsb\\
  &=\frac12  - \frac{ s_1 q_\delta^\star}2 \lsb 1 - \ln\lsb q_\delta^\star\rsb  \rsb -\frac{s_2}{2(\beta+1)}\lsb \frac{1-\lsb q_\delta^\star \rsb^{\beta+1} }{\beta+1} + \lsb q_\delta^\star \rsb^{\beta+1}   \ln \lsb q_\delta^\star \rsb\rsb.
 \end{align*}

\end{sloppypar}
\end{document}